\documentclass[11pt,a4paper]{article}
\usepackage[margin=2.0cm]{geometry}
\usepackage[utf8]{inputenc}
\usepackage[T1]{fontenc}
\usepackage{lmodern}
\usepackage{amsmath,amssymb,amsfonts,amsthm,mathtools,bm,graphicx,longtable,enumitem,empheq,array,xcolor}
\usepackage{float}
\floatstyle{ruled}
\newfloat{algorithm}{tbp}{loa}
\floatname{algorithm}{Algorithm}
\floatstyle{plain}
\usepackage{pgfplots}
\pgfplotsset{compat=1.17}
\usepackage[most]{tcolorbox}
\newtcolorbox{glance}{enhanced,breakable,colback=blue!3!white,colframe=blue!45!black,
 boxrule=0.5pt,arc=2pt,left=6pt,right=6pt,top=4pt,bottom=4pt,fonttitle=\bfseries\small,
 title={This section at a glance},fontupper=\small}
\usepackage[hidelinks]{hyperref}
\newcommand{\rev}[1]{#1}

\newcounter{algline}
\newlength{\algindent}
\newenvironment{algorithmic}[1][1]{%
  \setcounter{algline}{0}\setlength{\algindent}{0pt}%
  \begin{list}{}{\setlength{\leftmargin}{2.25em}\setlength{\rightmargin}{0pt}%
    \setlength{\topsep}{2pt}\setlength{\itemsep}{0pt}\setlength{\parsep}{0pt}}%
}{\end{list}}
\newcommand{\State}{\stepcounter{algline}\item[\scriptsize\arabic{algline}:]\leavevmode\hspace*{\algindent}\hangindent=\algindent\ignorespaces}
\newcommand{\Require}{\item[]\textbf{Input:} }
\newcommand{\Ensure}{\item[]\textbf{Output:} }
\newcommand{\Guarantee}{\item[]\textbf{Guarantee:} }
\newcommand{\For}[1]{\State\textbf{for} #1 \textbf{do}\par\addtolength{\algindent}{1em}}
\newcommand{\EndFor}{\par\addtolength{\algindent}{-1em}}
\newcommand{\Loop}{\State\textbf{loop}\par\addtolength{\algindent}{1em}}
\newcommand{\EndLoop}{\EndFor}
\newcommand{\If}[1]{\State\textbf{if} #1 \textbf{then}\par\addtolength{\algindent}{1em}}
\newcommand{\EndIf}{\EndFor}
\newcommand{\Else}{\EndIf\State\textbf{else}\par\addtolength{\algindent}{1em}}
\newcommand{\ElsIf}[1]{\EndIf\State\textbf{else if} #1 \textbf{then}\par\addtolength{\algindent}{1em}}
\newcommand{\Return}{\textbf{return} }
\newcommand{\Comment}[1]{\hfill\textit{#1}}

\newcolumntype{L}[1]{>{\raggedright\arraybackslash}p{#1}}
\newtheorem{theorem}{Theorem}
\newtheorem{proposition}{Proposition}
\newtheorem{lemma}{Lemma}
\newtheorem{corollary}{Corollary}
\newtheorem{assumption}{Assumption}
\theoremstyle{definition}
\newtheorem{definition}{Definition}
\newtheorem{remark}{Remark}
\allowdisplaybreaks[1]

\allowdisplaybreaks

\newcommand{\ba}{\mathbf{a}}
\newcommand{\bb}{\mathbf{b}}
\newcommand{\bc}{\mathbf{c}}
\newcommand{\bd}{\mathbf{d}}
\newcommand{\bolde}{\mathbf{e}}

\newcommand{\bg}{\mathbf{g}}
\newcommand{\bh}{\mathbf{h}}

\newcommand{\bn}{\mathbf{n}}

\newcommand{\bp}{\mathbf{p}}
\newcommand{\bq}{\mathbf{q}}
\newcommand{\br}{\mathbf{r}}
\newcommand{\bs}{\mathbf{s}}

\newcommand{\bu}{\mathbf{u}}
\newcommand{\bv}{\mathbf{v}}
\newcommand{\bw}{\mathbf{w}}
\newcommand{\bx}{\mathbf{x}}
\newcommand{\by}{\mathbf{y}}
\newcommand{\bz}{\mathbf{z}}

\newcommand{\bA}{\mathbf{A}}
\newcommand{\bB}{\mathbf{B}}
\newcommand{\bC}{\mathbf{C}}
\newcommand{\bD}{\mathbf{D}}

\newcommand{\bG}{\mathbf{G}}
\newcommand{\bH}{\mathbf{H}}
\newcommand{\bI}{\mathbf{I}}
\newcommand{\bJ}{\mathbf{J}}

\newcommand{\bL}{\mathbf{L}}
\newcommand{\bM}{\mathbf{M}}

\newcommand{\bP}{\mathbf{P}}

\newcommand{\bU}{\mathbf{U}}
\newcommand{\bV}{\mathbf{V}}

\newcommand{\btheta}{\mbox{\boldmath $\theta$}}

\newcommand{\blambda}{\mbox{\boldmath $\lambda$}}

\newcommand{\1}{\mbox{\boldmath $1$}}
\newcommand{\0}{\mbox{\boldmath $0$}}

\newcommand{\be}{\begin{equation}}
\newcommand{\ee}{\end{equation}}

\newcommand{\matrixb}{\left[ \begin{array}}
\newcommand{\matrixe}{\end{array} \right]}   

\newcommand{\argmax}{\operatornamewithlimits{\arg \max}}
\newcommand{\argmin}{\operatornamewithlimits{\arg \min}}

\newcommand{\diag}{\mathop{\rm diag}\nolimits}

\def\*{\circledast}

\newcommand{\bvec}[1]{\boldsymbol{#1}}

\newcommand{\vm}{\bvec{m}}

\begin{document}
\title{The Euler Logarithm and Its Inverse in Machine Learning: \\ Loss Design, Cross Entropy, Mirror Geometry, and Adaptive Learning Algorithms}
\author{Andrzej Cichocki\\[3pt]
\small Systems Research Institute, Polish Academy of Sciences, Warsaw, Poland\\
\small Warsaw University of Technology, Warsaw, Poland, and
\small RIKEN AIP, Tokyo, Japan\\
\small\texttt{cichockiand@gmail.com}}
\date{}
\maketitle

\begin{abstract}

A probability close to zero has very different consequences across machine learning: it can let a single mislabeled example dominate training, leave a promising action almost never sampled, or send a token to an irrelevant expert. Follow-the-regularized-leader (FTRL) and mirror descent (MD) govern such behavior through the choice of a regularizer and the Bregman divergence it induces. This paper studies a compact and interpretable family of losses, probability maps and optimization geometries generated by the normalized, freely scaled Euler logarithm. Two exponents independently set how quickly its inverse decays for very negative scores and how quickly it grows for very positive ones, while a scale parameter moves the transition between the two regimes without changing the value and slope of the link at one.

For classification, Euler cross-entropy reweights low-confidence examples through an explicit gradient weight, preserves class rankings, recovers the class posterior at the population optimum, and contains a bounded member with guarantees under symmetric label noise; paired with the matching Euler softmax, it becomes convex in the logits. For optimization, the Euler potential yields exact mirror descent and follow-the-regularized-leader updates with explicit strong-convexity and radius constants and with polynomial rather than exponential floors on small weights. With dimension-dependent exponents and an optimized scale, the leading full-information regret constant comes within about two percent of the entropic one. 
 We also derive the exact self-concordance constant of the barrier hybrids and a closed-form Hessian-matching scale with an exact robustness bound for relatively smooth problems.

Tables of metrics, benchmarks for up to one million coordinates and a scenario guide indicate when this geometry should be preferred to entropy, \rev{Tsallis} or barrier regularizers, and when it should not. Certified routines for the inverse and for simplex normalization support sixteen algorithms for online learning, sparse and adaptive optimization, reinforcement learning, games and mixture-of-experts routing. Finally, six  claimed benefits (tail control, penalty bounds, robustness to noisy labels and outliers, exploration in reinforcement learning, heavy tails, and sparsity) of generalized logarithms are examined one by one, separating proved results from open hypotheses.

\end{abstract}
\begin{center}\textit{Dedicated to Professor Shun-ichi Amari, my close scientific collaborator and mentor, on the occasion of his 90th birthday.
}\end{center}
\tableofcontents
\clearpage
\section{Introduction}

Follow-the-Regularized-Leader (FTRL) and Mirror Descent (MD) are foundational frameworks in online optimization that sit at the frontier of modern machine learning because they unify optimization geometry with regularization, directly inspiring state-of-the-art adaptive algorithms (like Adam, AdaGrad, and Muon) used to train today's largest AI models. Traditional gradient descent assumes that parameter space is Euclidean. FTRL and MD generalize gradient descent to non-Euclidean geometries: they use a Bregman divergence to match the update step to the natural geometry of the feasible set. Moreover, FTRL solves an explicit regularized optimization problem over all past gradients at every step, whereas standard subgradient methods rely on local Euclidean projections. Adaptive methods such as AdaGrad, RMSProp and Adam can be viewed as diagonal approximations of MD and FTRL in which the local metric adapts to past gradient magnitudes.

Such learning algorithms are shaped by two closely connected choices: how an objective measures error and how an update moves through the space of possible solutions. Cross-entropy, exponentiated gradient, and natural-gradient methods illustrate the power of logarithmic and information-geometric constructions. A natural next step is to make these constructions more adaptable: can a small, interpretable family of functions connect loss sensitivity, optimization geometry, and implementable algorithms with explicit theoretical guarantees?

This perspective owes much to Professor Shun-ichi Amari, to whom this paper is dedicated on the occasion of his 90th birthday. His development of information geometry established a powerful language for studying statistical models through metrics, dual affine connections, and divergence functions \cite{Amaribook,Amari2016}. His natural-gradient formulation brought this geometric viewpoint directly into learning: the size of a parameter change is measured through its effect on the predictive distribution \cite{Amari1998}. This insight continues to guide the search for learning algorithms whose geometry reflects the structure of the problem.

For a regular statistical model $p_{\boldsymbol\theta}$, the Fisher information matrix (FIM) and the associated natural-gradient direction are
\begin{align}
\bI(\boldsymbol\theta)
&=\mathbb E_{X\sim p_{\boldsymbol\theta}}\!\left[\nabla_{\boldsymbol\theta}\ln p_{\boldsymbol\theta}(X)\,
\nabla_{\boldsymbol\theta}\ln p_{\boldsymbol\theta}(X)^T\right],
\label{eq:amari-intro-fisher}\\
\operatorname{grad}_{\mathcal I}L(\boldsymbol\theta)
&=\bI(\boldsymbol\theta)^{-1}\nabla_{\boldsymbol\theta}L(\boldsymbol\theta),
\label{eq:amari-intro-natural}
\end{align}
where the second expression assumes a positive-definite FIM in identifiable local coordinates. {For a tangent displacement $\bv$, the Fisher norm is $\|\bv\|_{\bI}=\sqrt{\bv^T\bI(\boldsymbol\theta)\bv}$. When the gradient is nonzero, $-\operatorname{grad}_{\mathcal I}L/\|\operatorname{grad}_{\mathcal I}L\|_{\bI}$ minimizes the first-order change in $L$ among Fisher-unit displacements. The natural-gradient vector field is invariant under smooth invertible changes of coordinates.}

Amari's work also illuminates the geometric meaning of divergence functions \cite{Amari2009,Amari-PAN,Amari2016}. For the Euler potential studied here, the primal coordinates $\bw$ and dual coordinates {$\nabla F_{a,b,\lambda}(\bw)$} are linked by a convex potential, and its Bregman divergence induces a Hessian metric $\nabla^2F_{a,b,\lambda}(\bw)=\diag\bigl(\log_{a,b,\lambda}'(w_i)\bigr)$ on the positive orthant. In the logarithmic limit $(a,b)\to(0,0)$ this metric becomes $\diag(1/w_i)$, whose restriction to the simplex tangent space is the categorical Fisher--Rao metric. General Euler parameters then provide a controlled deformation of this familiar geometry. This is the central link to Amari's legacy in the present work: statistical geometry supplies the conceptual foundation, while the Euler family offers a concrete way to investigate tunable losses and update geometries. A general Euler Hessian is, however, a chosen optimization metric; it need not equal the Fisher information matrix of a given predictive model unless that model's log-likelihood induces the same curvature.

The classical two-parameter Euler logarithm is the unscaled starting point,
\begin{equation}
\log_{a,b}(x)=\frac{x^b-x^a}{b-a},\qquad x>0,\quad a\le0<b,
\end{equation}
with the ordinary logarithm as the continuous limit $(a,b)\to(0,0)$. Its historical and mathematical foundations include Euler's work on Lambert's series and the Borges--Roditi and Sharma--Taneja--Mittal constructions \cite{lambert1758,Euler1779,Borges1998,sharma1975,mittal1975,kaniadakis2005}. Mirror-descent and exponentiated-gradient updates built on the two-parameter Euler logarithm and on related trace-form entropies were introduced in our earlier work \cite{Cichocki2025arXiv,Cichocki2025,Cichocki2024}. Our contribution is to develop this established kernel into a connected machine-learning framework: from loss derivatives and inverse evaluation to constrained updates, convergence guarantees, and online learning.

Sections~2--13 extend this link to the normalized three-parameter family $\log_{a,b,\lambda}(x)=[\log_{a,b}(\lambda x)-\log_{a,b}(\lambda)]/[\lambda \log_{a,b}'(\lambda)]$, with $\lambda>0$. Here $\log_{a,b}$ denotes the unscaled link. The appeal of the two exponents is their interpretable control over asymptotic behavior; the third parameter changes where those behaviors meet. Because exchanging $a$ and $b$ leaves the function unchanged, we use ordered exponents $a\le0<b$. These exponents govern the tails, while symmetric combinations determine local curvature. In the regime $a<0<b$, the inverse accepts every real input, making it particularly useful for unconstrained dual-coordinate updates. On the boundary $a=0<b$, the associated one-hot classification loss is bounded. The domain characterization in Lemma~\ref{lem:audit-domains} therefore acts as a design guide: it identifies which parameter regimes support the computational and statistical behavior sought in an application.

For classification, the framework makes the influence of individual examples explicit. Building the loss on the Euler logarithm changes the softmax gradient through a computable multiplier. The bounded boundary family, which includes generalized cross-entropy \cite{ZhangSabuncu2018}, attenuates low-confidence examples; opposite-sign deformations can instead emphasize sufficiently hard examples. This gives a principled way to formulate different learning priorities. Proposition~\ref{thm:EulerCE-population} further shows how the bounded family transforms population-optimal probabilities: for its interior parameter range, class rankings are preserved even though the optimal probability values change. This distinction helps align loss selection with the intended task, whether accurate class decisions or calibrated probability estimates. Noise robustness remains a statistical question to evaluate under a specified noise model.

For optimization, integrating the Euler logarithm generates a convex potential whose curvature determines how strongly each coordinate responds to a gradient. We develop this mirror geometry alongside categorical Fisher-based calculations \cite{Amari1998,Amaribook,Martens2020}. The two metrics have complementary roles: the Fisher metric reflects a predictive model, whereas the Euler Hessian supplies a tunable optimization geometry. Making their roles explicit provides a sound basis for designing preconditioners and for connecting learning objectives with update rules.

\paragraph{Purpose of the paper.}
{We ask which properties of a learning algorithm can be designed by choosing its logarithmic link. The Euler exponents govern the tails, and the scale sets their relative contribution at the probabilities encountered in training. The ordinary exponential identity $\exp(u+v)=\exp(u)\exp(v)$ usually fails after deformation. Consequently, dividing Euler exponentials by their sum can change the prediction when the same constant is added to every score. We obtain a shift-invariant probability map by maximizing the score over the simplex with a convex regularizer. For optimization, the same regularizer $F_{a,b,\lambda}(\bw)=\sum_i\phi_{a,b,\lambda}(w_i)$, with $\phi'_{a,b,\lambda}=\log_{a,b,\lambda}$, defines the Bregman cost of an update.} Six
benefits are commonly attributed to such generalized logarithms: control of
(C1) tail behavior, (C2) penalty bounds, (C3) robustness to noisy labels and
outliers, (C4) exploration in reinforcement learning, (C5) heavy-tailed
generative and probabilistic modeling, and (C6) probability sparsity.
Section~\ref{sec:flexibility} examines these claims one by one: it identifies
the parameter responsible for each, proves what can be proved (including two
new results on label-noise tolerance and on Euler--Gaussian densities with a
tunable tail index), and formulates the rest as explicit research questions.
The remaining sections then provide the losses, computational tools and
algorithms that make this design usable.

\paragraph{Contributions.} The paper moves from a function to algorithms in the following steps.
\begin{enumerate}
\item \textbf{An analysis of six flexibility claims.} Section~\ref{sec:flexibility} maps tail control, penalty bounds, label-noise robustness, exploration, heavy tails and sparsity to the parameters $a$, $b$ and $\lambda$, separates proved statements from hypotheses (Table~\ref{tab:flex-claims}), and makes the trade-off between robustness and rare-class retention explicit.
\item \textbf{What the exponents and free scale add.} Section~\ref{subsec:euler-vs-ln-tsallis} isolates the property that distinguishes the Euler family from the natural and Tsallis logarithms. For $a<0<b$ the Euler exponential is defined on all of $\mathbb R$ and has {polynomial decay at the negative end and polynomial growth at the positive end, with independently selectable exponents $1/|a|$ and $1/b$.} We trace this property to its consequences: dual updates that never leave their domain, polynomial instead of exponential weight floors, { a softmax whose suppressed probabilities and gradients decay polynomially in a growing logit gap}, and curvature that protects small coordinates while staying moderate for large ones.
\item \textbf{Loss functions with an interpretable gradient shape and a statistical meaning.} Sections~\ref{sec:euler-CE} and~\ref{sec:canonical-softmax} show how the exponents attenuate or emphasize low-confidence examples, {characterize the proper composite loss and its excess-risk bound after posterior recovery}, {and distinguish the raw probability vector, the recovered posterior and the scalar potential that defines each logit loss.}
\item \textbf{Precise computation of the Euler exponential.} Section~\ref{sec:inverse-computation} gives an explicit bracket of bounded width for the solution, a global lower bound on the slope of the defining equation, and a safeguarded Newton method in logarithmic coordinates (Algorithm~\ref{alg:euler-exp}) { with an explicit iteration bound and log-error certificate in exact arithmetic, and a stated enclosure requirement for rounded computation}. Exact derivatives with respect to the input, both exponents and log-scale make the function usable inside gradient-based training.
\item \textbf{Mirror geometry with explicit guarantees.} Section~\ref{sec:Euler-MD} establishes well-posedness and convergence of exact Euler mirror descent, an explicit strong-convexity modulus and radius, and a polynomial lower bound on all weights. { Lemma~\ref{lem:audit-exact-ftrl} {identifies the common trajectories of exact simplex mirror descent and FTRL at fixed geometry and constant step size, with a matching regularizer anchor.}}
\item \textbf{FTRL with a free scale.} Section~\ref{sec:FTRL-3p} relates nine Euler FTRL constructions to mirror descent, proves which are exactly equivalent, {gives implementable update rules, and exhibits a scale that minimizes $R/c$ for fixed exponents. With $|a|=b=1/\ln N$, the associated leading regret constant approaches $\sqrt{2\sinh(1/2)}$ times the entropic bound while the inverse retains polynomial tails.}  Section~\ref{sec:ftrl-bandit} extends the analysis to bandit feedback, where the variance term is the sum of reciprocal gradient weights, and Section~\ref{sec:ftrl-choosing} gives metric, benchmark and scenario tables for choosing the geometry.
\item \textbf{Sixteen algorithms built on two primitives.} { The scalar inverse (Algorithm~\ref{alg:euler-exp}) supports the simplex normalizer (Algorithm~\ref{alg:euler-norm}); signed and shifted constructions require their own constraints and optimality conditions.} On them we build and analyze algorithms for online learning, sparse and schedule-free optimization, policy and value optimization, monotone games and self-play, implicit regularization, inexact computation and mixture-of-experts routing. Section~\ref{subsec:alg-guide} is a reader's guide to all of them.
\item \textbf{A clear statement of limits.} The costs of the family are stated as explicitly as its benefits, and Section~\ref{sec:new-experiments} separates what is proved, what is conditional, and what is only proposed, and lists experiments that can decide where the flexibility pays off.
\end{enumerate}

\paragraph{How to read this paper.}
{The main technical sections open with a short box} that states its question, its key idea and what to remember; long proofs and secondary results are collected in the appendices, so the main text can be read as a continuous story. Three reading paths are possible. \emph{Practitioners} can read Sections~\ref{sec:logab-properties}, \ref{sec:flexibility}, \ref{sec:euler-CE} and~\ref{sec:Euler-MD}, then jump to the application they need. \emph{Readers interested in online learning} should read all of Section~\ref{sec:Euler-MD}. \emph{Readers interested in theory} will find the proofs in Appendix~\ref{app:proofs} and the supplementary results in Appendices~\ref{app:supp-losses}--\ref{app:supp-apps}. Appendix~\ref{sec:notation-guide} is a notation guide that can be consulted at any time.

\paragraph{How the paper is organized.}
Section~\ref{sec:logab-properties} introduces the Euler logarithm, its special cases, free scale and two-level curvature, and Section~\ref{sec:flexibility} examines six flexibility claims. Sections~\ref{sec:euler-CE} and~\ref{sec:canonical-softmax} treat loss functions and the Euler softmax, and Section~\ref{sec:inverse-computation} explains how to compute and differentiate the Euler exponential and how to learn all three parameters. Section~\ref{sec:Euler-MD} reviews the established mirror-descent and FTRL algorithms and develops the Euler versions that make a clear contribution: exact mirror descent, Euler-Hedge FTRL with an optimal scale, adaptive FTRL, FTRL-Proximal and RDA, and optimistic FTRL, followed by bandit feedback and a guide to choosing the geometry. Sections~\ref{sec:extensions} and~\ref{sec:new-policy-scale} treat reinforcement learning and relative smoothness, Section~\ref{sec:phase2-games} games, Section~\ref{sec:phase2-barrier} implicit regularization and damping, and Section~\ref{sec:phase2-moe} mixture-of-experts routing. Section~\ref{sec:design-geometry} measures Euler potentials against design criteria for Bregman geometries and compares them with Adam, Shampoo and Muon, and Section~\ref{sec:new-experiments} summarizes the status of all results and proposes experiments. The appendices contain the notation guide, supplementary results (including further FTRL variants), the longer proofs, the special cases and related deformed logarithms, and the analysis of radially normalized updates.

\paragraph{Notation and scope.} Throughout, $\log_{a,b,\lambda}$ is the three-parameter link and $\exp_{a,b,\lambda}$ its inverse; $\nu$ denotes a normalization multiplier. We write $\log_{a,b}=\log_{a,b,1}$ for the unscaled Euler logarithm and
$\exp_{a,b}$ for its inverse; we always order the exponents as $a\le0<b$ (hence $a<b$). The cases $a<0=b$ are explicitly marked boundary extensions, and $(a,b)=(0,0)$ denotes the ordinary-logarithm limit. The magnitude of the negative exponent is simply $|a|$.
Vectors are bold lowercase and matrices bold uppercase. Appendix~\ref{sec:notation-guide}
collects the notation, parameter conventions, and guidance for reading the equations
and guarantees. Local definitions are retained beside the results that use them.

\paragraph{Relationship to the later algorithms.}
All algorithmic sections use the three-parameter link with a fixed scale
$\lambda>0$ unless stated otherwise; their constants depend on $\lambda$
through the mixture weights $\alpha_\lambda,\beta_\lambda$, the radius and the
strong-convexity modulus of Section~\ref{sec:Euler-MD}. Learned time-varying
scales additionally require the geometry-drift analysis of
Theorem~\ref{thm:new-drift}. Normalization multipliers are denoted $\nu$.

\section{The three-parameter scale-free Euler logarithm and its geometry}
\label{sec:logab-properties}
\begin{glance}
\textbf{Question.} What is the scaled Euler logarithm, and what do its three parameters do?\\
\textbf{Key idea.} {The link is a weighted average of the normalized power logarithms $(x^a-1)/a$ and $(x^b-1)/b$, with continuous zero-exponent limits.} The exponents fix the two tails; the scale $\lambda$ slides the point where one regime hands over to the other (Figure~\ref{fig:euler-link}).\\
\textbf{Remember.} For $a<0<b$ the inverse is defined on the whole real line and has polynomial tails with exponents $1/|a|$ and $1/b$. {At $a=0<b$, the inverse has a finite lower endpoint. This change in domain allows a simplex optimizer to assign a coordinate exactly zero.}
\end{glance}

The normalized, freely scaled Euler logarithm is
\begin{equation}
 \log_{a,b,\lambda}(x)=\frac{\lambda^k(x^a-1)-\lambda^{-k}(x^b-1)}
 {a\lambda^k-b\lambda^{-k}},\qquad k=\frac{a-b}{2},
 \label{logab}
\end{equation}
\label{deflogEL}
{where $x>0$ and $\lambda>0$. In the ordered domain $a\le0\le b$, $a<b$, the denominator is strictly negative. The coincident-exponent limit is addressed in Lemma~\ref{lem:audit-domains}.}
The name \emph{scale-free Euler} refers here to the freedom to choose the
argument scale $\lambda$. It does not imply that a learning algorithm is
invariant to rescaling losses or that its learning rate is parameter-free.
\rev{Throughout the paper} we reserve $\lambda$ for this scale; $\nu$ denotes a normalization or KKT multiplier.
Unless stated otherwise, $\lambda$ is fixed during the algorithm or theorem under discussion.

Let $\log_{a,b}(x)=(x^a-x^b)/(a-b)$ denote the original two-parameter Euler logarithm and
$\omega_0(x)=x\log_{a,b}'(x)$. Then
\begin{equation}
 \log_{a,b,\lambda}(x)=\frac{\log_{a,b}(\lambda x)-\log_{a,b}(\lambda)}
 {\omega_0(\lambda)},\qquad \log_{a,b,\lambda}(1)=0,\qquad \log_{a,b,\lambda}'(1)=1.
 \label{eq:scaled-Euler-identity}
\end{equation}
Thus $\lambda=1$ recovers exactly the original Euler logarithm.
To verify Eq.~\eqref{eq:scaled-Euler-identity}, expand its right-hand side and multiply numerator and denominator by $\lambda^{-(a+b)/2}>0$:
\[
 \frac{\log_{a,b}(\lambda x)-\log_{a,b}(\lambda)}{\lambda\log_{a,b}'(\lambda)}
 =\frac{\lambda^a(x^a-1)-\lambda^b(x^b-1)}{a\lambda^a-b\lambda^b}
 =\frac{\lambda^{(a-b)/2}(x^a-1)-\lambda^{-(a-b)/2}(x^b-1)}{a\lambda^{(a-b)/2}-b\lambda^{-(a-b)/2}},
\]
which is Eq.~\eqref{logab}.
The unscaled family has the Euler, Borges--Roditi and
Sharma--Taneja--Mittal history described in
\cite{Euler1779,Borges1998,sharma1975,mittal1975}.
The scaled algebraic form also agrees with the three-parameter logarithm
of Kaniadakis~\cite{kaniadakis2009}, with
$\kappa=(b-a)/2$, $r=(a+b)/2$, and scale $\lambda$.
Our claims concern its use as a machine-learning link and mirror map,
not the invention of this function.

\paragraph{A normalized mixture of power links.}
Its two single-power boundary members are $\log_{a,0}(x)=(x^a-1)/a$ and $\log_{0,b}(x)=(x^b-1)/b$, on which the scale cancels; in general $\log_{s,0}(x)=(x^s-1)/s$ for $s\ne0$ and $\log_{0,0}=\ln$.
In the ordered admissible regime $a\le0\le b$, $a<b$, define
\begin{equation}
 \alpha_\lambda=\frac{a\lambda^a}{a\lambda^a-b\lambda^b},\qquad
 \beta_\lambda=\frac{-b\lambda^b}{a\lambda^a-b\lambda^b},\qquad
 \log_{a,b,\lambda}=\alpha_\lambda\log_{a,0}+\beta_\lambda\log_{0,b},
 \quad \alpha_\lambda+\beta_\lambda=1.
 \label{eq:scaled-power-mixture}
\end{equation}
The weights are nonnegative and both are positive when $a<0<b$.
On either zero-exponent boundary the scale cancels exactly.
In the interior their odds satisfy
\[
 \frac{\beta_\lambda}{\alpha_\lambda}
 =\frac{b}{|a|}\lambda^{b-a}.
\]
Increasing $\lambda$ shifts weight toward the positive-power component.
Equivalently, in the interior,
\[
 \alpha_\lambda=\frac{|a|}{|a|+b\lambda^{b-a}}=\Bigl[1+\exp\bigl(\ln(b/|a|)+(b-a)\ln\lambda\bigr)\Bigr]^{-1},\qquad
 \beta_\lambda=1-\alpha_\lambda=\frac{b\lambda^{b-a}}{|a|+b\lambda^{b-a}};
\]
the logistic form avoids overflow and underflow for extreme $\lambda$ and is the one used in the algorithms.
For fixed $a<0<b$, $\log_{a,b,\lambda}\to\log_{a,0}$ as $\lambda\downarrow0$ and
$\log_{a,b,\lambda}\to\log_{0,b}$ as $\lambda\to\infty$, uniformly on compact
positive intervals. Nevertheless, every finite positive $\lambda$ retains
both asymptotic tails and a full-range inverse.

\begin{lemma}[Monotonicity, concavity, and inverse domains]\label{lem:audit-domains}
For $a<b$ and a defined normalization, $\log_{a,b,\lambda}'(x)>0$ for all $x>0$
if and only if $a\le0\le b$. In this regime $\log_{a,b,\lambda}''\le0$ everywhere
if and only if $b\le1$, with strict concavity except at $(a,b)=(0,1)$.
For every $\lambda>0$,
\begin{equation}
 \operatorname{range}(\log_{a,b,\lambda})=
 \begin{cases}
 \mathbb R,&a<0<b,\\
 (-1/b,\infty),&a=0<b,\\
 (-\infty,-1/a),&a<0=b.
 \end{cases}\label{eq:audit-ranges}
\end{equation}
For fixed $\lambda$, the admissible limit $(a,b)\to(0,0)$ is $\ln x$.
For coincident exponents $a=b=c$, the continuous extension, when
$1+c\ln\lambda\ne0$, is
\[
 \frac{x^c\ln x+(x^c-1)\ln\lambda}{1+c\ln\lambda}.
\]
It is globally increasing only for $c=0$.
\end{lemma}
\noindent\textit{The proof is given in Appendix~\ref{app:proofs}, p.~\pageref{prf:lem:audit-domains}.}
The main statistical regime is $a\le0\le b<1$; the main finite-radius,
full-range mirror regime is $-1<a<0<b\le1$.
The trace entropy $\sum_i p_i \log_{a,b,\lambda}(1/p_i)$ is distinct from the
mirror potential $\sum_i\int_1^{p_i}\log_{a,b,\lambda}(x)\,\mathrm dx$.
The former has vanishing zero-mass terms when $b<1$; the latter has
finite boundary values when $a>-1$.

\paragraph{Integral and series descriptions.}
For $t=\ln x$,
\begin{equation}
 \log_{a,b,\lambda}(e^t)=\int_0^t
  \left(\alpha_\lambda e^{au}+\beta_\lambda e^{bu}\right)\,\mathrm du
 =t\left(\alpha_\lambda e^{a\theta t}+\beta_\lambda e^{b\theta t}\right)
 \quad\text{for some }\theta\in(0,1).
 \label{eq:MVT-Euler-log}
\end{equation}
At $\lambda=1$ this also equals the divided difference
$\log_{a,b}(x)=x^c\ln x$ for some $c\in(a,b)$.
For general scale one must use the normalized rescaling, rather than
carry over an unscaled divided-difference formula unchanged.
With $\kappa=(b-a)/2$, $r=(a+b)/2$,
\begin{equation}
 \log_{a,b}(e^t)=e^{rt}\frac{\sinh(\kappa t)}{\kappa},\qquad
 \log_{a,b,\lambda}(e^t)=
 \frac{\log_{a,b}(e^{t+\ln\lambda})-\log_{a,b}(\lambda)}{\omega_0(\lambda)}.
 \label{AbeG}
\end{equation}
A cancellation-resistant evaluation uses the power mixture with
\texttt{expm1}, rather than subtracting two nearly equal unscaled links.
The exact, locally uniformly convergent log-coordinate series is
\begin{equation}
 \log_{a,b,\lambda}(x)=
 \sum_{n\ge1}\frac{\alpha_\lambda a^{n-1}+\beta_\lambda b^{n-1}}{n!}
 (\ln x)^n.
 \label{eq:logab-local-series}
\end{equation}
{For $n=1$, the coefficient of $\ln x$ is one, including on zero-exponent boundaries. For $n\ge2$, a component with zero exponent contributes zero.} At $\lambda=1$ its numerator
reduces to the complete homogeneous polynomial
$(b^n-a^n)/(b-a)$, recovering the unscaled series.

\begin{theorem}[Ordered exponents, scale, and local curvature]
\label{thm:ordered-parameters}
The link is symmetric under $a\leftrightarrow b$, including the
change $k\mapsto-k$. In the interior $a<0<b$,
\begin{align}
 \log_{a,b,\lambda}''(1)&=\alpha_\lambda(a-1)+\beta_\lambda(b-1),
 \label{eq:local-curvature-ordered}\\
 \log_{a,b,\lambda}(x)&\sim-\frac{\alpha_\lambda}{|a|}x^a
 \quad(x\downarrow0),\label{eq:left-tail-ordered}\\
 \log_{a,b,\lambda}(x)&\sim\frac{\beta_\lambda}{b}x^b
 \quad(x\to\infty).\label{eq:right-tail-ordered}
\end{align}
The local curvature equals $a+b-1$ only in the unscaled case $\lambda=1$.
\end{theorem}
\begin{proof}
Exchange of exponents changes the signs of both numerator and denominator
in~\eqref{logab}. Differentiation of~\eqref{eq:scaled-power-mixture}
gives the curvature; the dominant powers give the tails.
\end{proof}
\begin{remark}[Three parameter roles]\label{rem:interpret-ordered-parameters}
The exponent $a$ controls the small-coordinate tail and $b$ the
large-coordinate tail. The scale $\lambda$ changes their relative
contribution at a specified coordinate size, including near $x=1$,
without changing those limiting exponents.
These are three degrees of freedom in the interior, but not three
independent coefficients at every point. On the Tsallis boundaries
$\lambda$ is redundant.
\end{remark}

\subsection{Special cases and what the scale adds}
\label{subsec:euler-vs-ln-tsallis}
\begin{table}[htbp]
\centering\small
\caption{Named logarithms inside the scaled family. The usual interior
named forms occur at $\lambda=1$; arbitrary $\lambda$ gives their
normalized scaled extensions, with the curvature $h_{a,b,\lambda}=\log_{a,b,\lambda}'$.}\label{tab:special-cases}
\begin{tabular}{@{}L{.14\textwidth}L{.17\textwidth}L{.24\textwidth}L{.35\textwidth}@{}}
\hline
Family & Parameters & Curvature $h_{a,b,\lambda}(x)$ & Link and inverse/domain\\\hline
Natural & $(a,b)\to(0,0)$, fixed $\lambda$ & $x^{-1}$ &
$\ln x$, inverse $e^y$ on $\mathbb R$.\\
Tsallis $q<1$ & $(0,1-q)$, any $\lambda$ & $x^{-q}$ &
$(x^{1-q}-1)/(1-q)$; inverse $[1+(1-q)y]^{1/(1-q)}$ on its positive-base domain.\\
Amari / Tsallis $q>1$ & $(1-q,0)$, any $\lambda$ & $x^{-q}$ &
Same power formula; finite upper inverse endpoint.\\
Burg/log-barrier & $(-1,0)$, any $\lambda$ & $x^{-2}$ &
$1-x^{-1}$; inverse $(1-y)^{-1}$ for $y<1$.\\
Euclidean & $(0,1)$, any $\lambda$ & $1$ &
$x-1$; inverse $1+y$ for $y>-1$.\\
Kaniadakis & $(-\kappa,\kappa)$, $\lambda=1$ & $\frac12(x^{-\kappa-1}+x^{\kappa-1})$ &
$(x^\kappa-x^{-\kappa})/(2\kappa)$; full-range inverse.\\
$\gamma$-logarithm & $(-\gamma,2\gamma)$, $\lambda=1$ & $\frac13(x^{-\gamma-1}+2x^{2\gamma-1})$ &
$(x^{2\gamma}-x^{-\gamma})/(3\gamma)$; cubic inverse, Eq.~\eqref{eq:gamma-exp}.\\
Abe, $1<\sigma<2$ & $(1-\sigma,1-\sigma^{-1})$, $\lambda=1$ & $\frac{(\sigma-1)x^{-\sigma}+(1-\sigma^{-1})x^{-1/\sigma}}{\sigma-\sigma^{-1}}$ &
The unscaled Euler form at these exponents; inverse numerical in general.\\
KLS & $(r-\kappa,r+\kappa)$, $|r|<\kappa$, $\lambda=1$ & $\frac{(\kappa-r)x^{r-\kappa-1}+(\kappa+r)x^{r+\kappa-1}}{2\kappa}$ &
$x^r(x^\kappa-x^{-\kappa})/(2\kappa)$; full-range inverse.\\
Scaled Euler interior & $a<0<b$, $\lambda>0$ & $\alpha_\lambda x^{a-1}+\beta_\lambda x^{b-1}$ &
Eq.~\eqref{logab}; full-range certified inverse, Algorithm~\ref{alg:euler-exp}.\\\hline
\end{tabular}
\end{table}
The Tsallis logarithm is therefore a boundary special case, not a separate
competing construction. Its conventional parameter $q$ is unrelated to a
predicted probability $q_i$. Positive-part continuations used for sparse
optimization are KKT constructions, not inverses outside the link range.
Other related logarithms listed in Appendix~\ref{app:special-cases}
need not belong to this family.

\begin{figure}[ht]
\centering
\begin{tikzpicture}
\begin{axis}[width=0.47\textwidth,height=0.37\textwidth,xmin=0.02,xmax=4,ymin=-5,ymax=3,
 xlabel={$x$},ylabel={$\log_{a,b,\lambda}(x)$},legend pos=south east,legend style={font=\scriptsize},
 tick label style={font=\scriptsize},label style={font=\small},samples=160,domain=0.02:4]
\addplot[thick,black,dashed]{ln(x)}; \addlegendentry{$\ln x$}
\addplot[thick,blue]{2*(-(1/(1+0.25))*(x^(-0.5)-1)+(0.25/(1+0.25))*(x^(0.5)-1))}; \addlegendentry{$\lambda=0.25$}
\addplot[thick,red]{2*(-(1/2)*(x^(-0.5)-1)+(1/2)*(x^(0.5)-1))}; \addlegendentry{$\lambda=1$}
\addplot[thick,teal]{2*(-(1/(1+4))*(x^(-0.5)-1)+(4/(1+4))*(x^(0.5)-1))}; \addlegendentry{$\lambda=4$}
\end{axis}
\end{tikzpicture}\hfill
\begin{tikzpicture}
\begin{semilogxaxis}[width=0.47\textwidth,height=0.37\textwidth,xmin=0.001,xmax=1000,ymin=0.4,ymax=1.6,
 xlabel={$x$},ylabel={elasticity $\varepsilon_{a,b,\lambda}(x)$},legend style={font=\scriptsize},
 tick label style={font=\scriptsize},label style={font=\small},samples=200,domain=0.001:1000]
\addplot[thick,blue]{1-0.5*tanh(0.5*ln(0.25*x))}; \addlegendentry{$\lambda=0.25$}
\addplot[thick,red]{1-0.5*tanh(0.5*ln(x))}; \addlegendentry{$\lambda=1$}
\addplot[thick,teal]{1-0.5*tanh(0.5*ln(4*x))}; \addlegendentry{$\lambda=4$}
\addplot[black,dashed,domain=0.001:1000]{1};
\end{semilogxaxis}
\end{tikzpicture}
\caption{The scaled Euler logarithm for $(a,b)=(-\frac12,\frac12)$. Left: all curves pass through $(1,0)$ with unit slope, like $\ln x$ (dashed), but fall to $-\infty$ polynomially near zero and grow like $\sqrt x$ at infinity. Right: the elasticity moves from $1-a=1.5$ (small coordinates, steeper than entropy) to $1-b=0.5$ (large coordinates, flatter than entropy); the dashed line is entropy. Changing $\lambda$ only \emph{slides} this transition along the axis, here $\varepsilon=1$ at $x=1/\lambda$, without changing the two limiting levels.}
\label{fig:euler-link}
\end{figure}

\paragraph{Curvature, mobility, gradient weight, and elasticity.}
Define
\begin{align}
 h_{a,b,\lambda}(x)&=\log_{a,b,\lambda}'(x)
 =\alpha_\lambda x^{a-1}+\beta_\lambda x^{b-1},&
 m_{a,b,\lambda}(x)&=\frac1{h_{a,b,\lambda}(x)},\nonumber\\
 \omega_{a,b,\lambda}(x)&=xh_{a,b,\lambda}(x)
 =\alpha_\lambda x^a+\beta_\lambda x^b,&
 \varepsilon_{a,b,\lambda}(x)&=-\frac{x \log_{a,b,\lambda}''(x)}{\log_{a,b,\lambda}'(x)}.
 \label{eq:elasticity-def}
\end{align}

\begingroup
Here $h$ is the curvature of the scalar mirror potential, $m$ converts a small dual displacement into a local primal displacement, and $\omega$ weights a one-hot loss gradient under ordinary softmax. Elasticity $\varepsilon=-\mathrm d\ln h/\mathrm d\ln x$ measures the percentage decrease in curvature per percentage increase in $x$.\par
\endgroup
\begin{proposition}[Two elasticity levels and a movable transition]
\label{prop:elasticity}
For $a<0<b$,
\begin{equation}
 \varepsilon_{a,b,\lambda}(x)
 =(1-b)(1-\chi_\lambda(x))+(1-a)\chi_\lambda(x),\quad
 \chi_\lambda(x)=\frac1{1+\frac b{|a|}(\lambda x)^{b-a}}.
 \label{eq:elasticity-Euler}
\end{equation}
Elasticity decreases strictly from $1-a$ to $1-b$. In $\ln x$,
the logistic transition has center
$x_0=\lambda^{-1}(|a|/b)^{1/(b-a)}$ and width $1/(b-a)$.
The crossing $\varepsilon=1$ is at
$x_m=\lambda^{-1}(|a|/b)^{2/(b-a)}$; it generally differs from $x_0$.
At $x=1$,
$\varepsilon_1=1-\alpha_\lambda a-\beta_\lambda b$.
\end{proposition}
\begin{proof}
Divide $-x\log_{a,b,\lambda}''$ by $\log_{a,b,\lambda}'$.
The relative weight of its negative-power term is
$\alpha_\lambda x^a/(\alpha_\lambda x^a+\beta_\lambda x^b)$.
Its odds give the displayed logistic expression. Setting its value to
$1/2$, or setting $\varepsilon=1$, gives the two distinct locations.
\end{proof}
\begin{table}[htbp]
\centering\small
\caption{Limiting versus local elasticity.}\label{tab:elasticity-families}
\begin{tabular}{llccc}
\hline Family & Curvature $h_{a,b,\lambda}(x)$ & $\varepsilon(0^+)$ & $\varepsilon(1)$ & $\varepsilon(\infty)$\\\hline
Natural & $x^{-1}$ & $1$ & $1$ & $1$\\
Tsallis $(0,b)$ & $x^{b-1}$ & $1-b$ & $1-b$ & $1-b$\\
Amari $(a,0)$ & $x^{a-1}$ & $1-a$ & $1-a$ & $1-a$\\
Unscaled Kaniadakis & $\frac12(x^{-\kappa-1}+x^{\kappa-1})$ & $1+\kappa$ & $1$ & $1-\kappa$\\
Scaled Kaniadakis & $\alpha_\lambda x^{-\kappa-1}+\beta_\lambda x^{\kappa-1}$ & $1+\kappa$ & $1-\kappa\tanh(\kappa\ln\lambda)$ & $1-\kappa$\\
Scaled Euler & $\alpha_\lambda x^{a-1}+\beta_\lambda x^{b-1}$ & $1-a$ & $1-\alpha_\lambda a-\beta_\lambda b$ & $1-b$\\\hline
\end{tabular}
\end{table}
Moreover $\mathrm d\ln m/\mathrm d\ln x=\varepsilon$ and
$\mathrm d\ln\omega/\mathrm d\ln x=1-\varepsilon$. Hence the elasticity crossing $x_m=r/\lambda$ is also the unique minimizer of the gradient weight, so $x_m=q_m$, whereas at the logistic centre $x_0$ the two curvature terms are equal, $\alpha_\lambda x_0^{a-1}=\beta_\lambda x_0^{b-1}$. The two scales coincide if and only if $|a|=b$, the symmetric Kaniadakis subfamily.
Thus the same profile governs optimization mobility and ordinary-softmax
loss weighting. Its entire graph shifts in log-coordinate as $\lambda$
changes; the two limiting levels are controlled by $a,b$.

\begin{proposition}[Tail calculus of the scaled Euler inverse]\label{prop:euler-tails}
For $a<0<b$, let $\exp_{a,b,\lambda}$ denote the inverse of $\log_{a,b,\lambda}$.
It is a smooth increasing bijection $\mathbb R\to(0,\infty)$,
with $\exp_{a,b,\lambda}(0)=1$, and is convex if $b\le1$. In addition,
\begin{align}
 \exp_{a,b,\lambda}(y)&=\lambda^{-1}\exp_{a,b}
 \bigl(\omega_0(\lambda)y+\log_{a,b}(\lambda)\bigr),
 \label{defexpEuler}\\
 \exp_{a,b,\lambda}(y)&\sim(by/\beta_\lambda)^{1/b}\ (y\to+\infty),\qquad
 \exp_{a,b,\lambda}(y)\sim(|a||y|/\alpha_\lambda)^{-1/|a|}
 \ (y\to-\infty),\label{eq:euler-exp-tails}\\
 \frac{\mathrm d}{\mathrm dy}\ln \exp_{a,b,\lambda}(y)
 &=\frac1{\omega_{a,b,\lambda}(\exp_{a,b,\lambda}(y))}.
 \label{eq:euler-log-sensitivity}
\end{align}
\end{proposition}
\begin{proof}
Invert the rescaling identity and the two link tails.
The inverse function theorem gives $\exp_{a,b,\lambda}'=m\circ \exp_{a,b,\lambda}$,
and division by $\exp_{a,b,\lambda}$ gives logarithmic sensitivity.
\end{proof}
The tail exponents $1/|a|$ and $1/b$ are independent of $\lambda$;
their prefactors and finite-score behavior are not.
Logarithmic sensitivity tends to zero at both ends, with asymptotic
forms $x^{|a|}/\alpha_\lambda$ and $x^{-b}/\beta_\lambda$.
It equals one for the ordinary exponential.

\begin{table}[htbp]
\centering\small
\caption{Structural comparison; no empirical superiority is implied.}
\label{tab:euler-vs-ln-tsallis}
\begin{tabular}{L{.25\textwidth}L{.31\textwidth}L{.34\textwidth}}
\hline Pair & Inverse behavior & Available controls\\\hline
Natural & Full range; exponential tails & No deformation; temperature sets score scale.\\
Tsallis/GCE & Finite lower endpoint; sparse KKT extension & One power exponent; $\lambda$ cancels.\\
Amari & Finite upper endpoint; polynomial left tail & One power exponent; $\lambda$ cancels.\\
Scaled Kaniadakis & Full range; equal polynomial tail exponents & One tail exponent and a movable transition.\\
Scaled Euler interior & Full range; independently selectable polynomial tails $(|a||y|/\alpha_\lambda)^{-1/|a|}$ and $(by/\beta_\lambda)^{1/b}$ & Two exponents plus transition scale.\\\hline
\end{tabular}
\end{table}
Finite dual scores yield positive weights in exact arithmetic. Polynomial
retention is quantified in Theorem~\ref{thm:no-extinction}; it does not
guarantee fast recovery, adequate sampling, or freedom from numerical underflow.
{For a two-class score gap $D=s_{\rm high}-s_{\rm low}$ at unit temperature, the smaller probability $q$ satisfies $\log_{a,b,\lambda}(q)=\log_{a,b,\lambda}(1-q)-D$. As $D\to\infty$, $q\sim(|a|D/\alpha_\lambda)^{-1/|a|}$. Differentiating the gap equation gives $\mathrm dq/\mathrm dD=-[h_{a,b,\lambda}(q)+h_{a,b,\lambda}(1-q)]^{-1}\sim-q^{1+|a|}/\alpha_\lambda$, so the suppressed probability has a derivative of order $D^{-1-1/|a|}$.}
Performance comparisons require experiments.
The geometry cost is the ratio of potential range to a sufficient
strong-convexity modulus, not either quantity alone.

\begin{table}[htbp]
\centering\small
\caption{Reference geometry costs $R/\mu$ for the unscaled specialization
$\lambda=1$, using Theorem~\ref{thm:sharp-simplex-geometry}.
For free $\lambda$ recompute both constants. The second row gives the curvature; the smallest cost in each row is in bold.}\label{tab:geometry-cost}
\begin{tabular}{lccccc}
\hline $N$ & Natural & $(0,0.5)$ & $(-0.05,0.7)$ & $(-0.2,0.5)$ & $(-0.5,0.5)$\\
$h(x)$ & $x^{-1}$ & $x^{-1/2}$ & $\frac1{15}x^{-1.05}+\frac{14}{15}x^{-0.3}$ & $\frac27x^{-1.2}+\frac57x^{-0.5}$ & $\frac12(x^{-1.5}+x^{-0.5})$\\\hline
$10$ & $\mathbf{2.30}$ & $2.88$ & $3.13$ & $2.51$ & $4.78$\\
$100$ & $\mathbf{4.61}$ & $12.00$ & $10.52$ & $5.27$ & $18.60$\\
$1000$ & $\mathbf{6.91}$ & $40.83$ & $13.36$ & $9.25$ & $61.89$\\\hline
\end{tabular}
\end{table}
The remaining sections develop the full three-parameter family; the effect of
a free scale on the geometry cost is quantified in
Table~\ref{tab:near-optimal-lambda}.
\section{What does the Euler logarithm add? Six flexibility claims}
\label{sec:flexibility}
\begin{glance}
\textbf{Question.} The purpose of this paper is not to re-derive known algorithms but to ask what is gained when they are redesigned around $\log_{a,b,\lambda}$ rather than the natural logarithm. Which of the commonly claimed benefits are actually delivered, by which parameter, and under which conditions?\\
\textbf{Key idea.} Each claim corresponds to one ``knob'': the left exponent $a$ governs small arguments, the right exponent $b$ large ones, and the scale $\lambda$ places the transition. A claim is credible only if it can be traced to an explicit property of $\log_{a,b,\lambda}$ or $\exp_{a,b,\lambda}$.\\
\textbf{Remember.} Tail control, bounded penalties, sparsity and exploration floors are \emph{proved}; {symmetric-label-noise tolerance is quantified on the bounded boundary $a=0$} {where attenuation applies to all low-confidence examples, including correctly labeled rare-class examples}; benefits for heavy-tailed generative models and for exploration efficiency are \emph{hypotheses} that the paper sharpens but does not settle.
\end{glance}

Incorporating $\log_{a,b,\lambda}$ and $\exp_{a,b,\lambda}$ into a loss, a
softmax, a probability model or a mirror-descent update, each redesigned so
that its defining property survives the deformation, gives the model three
tunable parameters. It is often
claimed that such generalized logarithms let one control
\begin{enumerate}[label=\textbf{C\arabic*.},leftmargin=2.6em,itemsep=1pt]
\item \emph{tail behavior} of probabilities and weights;
\item \emph{penalty bounds}, i.e.\ how severely confident mistakes are punished;
\item \emph{robustness} to noisy labels and outliers;
\item \emph{exploration} in reinforcement learning;
\item \emph{heavy-tailed} generative and probabilistic modeling;
\item \emph{sparsity} of probability vectors.
\end{enumerate}
This section examines each claim, states precisely what the Euler family
delivers, and turns the remainder into research questions. Throughout,
$-1<a\le0<b\le1$ unless stated otherwise; recall from
Eq.~\eqref{eq:scaled-power-mixture} that
$\log_{a,b,\lambda}=\alpha_\lambda\log_{a,0}+\beta_\lambda\log_{0,b}$ with
weights $\alpha_\lambda+\beta_\lambda=1$ set by $\lambda$.

\subsection{C1: Tail behavior is controlled exactly, and separately on both sides}
For $a<0<b$, Proposition~\ref{prop:euler-tails} gives
\begin{equation}
 \exp_{a,b,\lambda}(y)\sim\Bigl(\frac{|a|\,|y|}{\alpha_\lambda}\Bigr)^{-1/|a|}\ (y\to-\infty),\qquad
 \exp_{a,b,\lambda}(y)\sim\Bigl(\frac{b\,y}{\beta_\lambda}\Bigr)^{1/b}\ (y\to+\infty),
 \label{eq:flex-tails}
\end{equation}
whereas $\exp(y)$ decays and grows exponentially. The left exponent $1/|a|$
{sets how quickly a large negative dual score becomes a small weight. The right exponent $1/b$ governs growth of the unnormalized inverse on large positive inputs. For simplex probabilities, the common multiplier constrains the inverse arguments to be at most zero. The approach of a dominant probability to one follows from the decay of the remaining probabilities. The scale $\lambda$ changes the tail prefactors and the transition location (Figure~\ref{fig:euler-link}).} In probability space the same
statement reads: the curvature behaves like $x^{-(1+|a|)}$ near zero and like
$x^{-(1-b)}$ at large $x$ (Theorem~\ref{thm:design-criteria}). \textbf{Status:}
proved. Its consequences are polynomial weight floors in mirror descent
(Section~\ref{sec:Euler-MD}), in adaptive FTRL
(Corollary~\ref{cor:adaptive-FTRL-floor}) and in policy updates
(Corollary~\ref{cor:Euler-softmax-update}).

\subsection{C2 and C3: Penalty growth and robustness to noisy labels}
Euler cross-entropy charges $\mathcal L_{a,b,\lambda}(q)=-\log_{a,b,\lambda}(q)$
for a true-class probability $q$ and multiplies the ordinary softmax gradient
by the weight $\omega_{a,b,\lambda}(q)=\alpha_\lambda q^a+\beta_\lambda q^b$
(Eq.~\eqref{eq:ECE-shape-design}). How these behave as $q\to0$, the regime of
confident mistakes and of mislabeled examples, decides both claims.

\begin{proposition}[Three regimes of penalty growth and label-noise tolerance]
\label{prop:flex-penalty}
Let $N\ge2$ classes and $0<q\le1$.
\begin{enumerate}[label=(\roman*),leftmargin=2.2em]
\item \emph{Bounded penalty.} If $a=0<b$, the scale cancels,
$\mathcal L_{0,b}(q)=(1-q^b)/b\in[0,1/b)$ and $\omega_{0,b}(q)=q^b\to0$: confident
mistakes cost at most $1/b$ and contribute vanishing gradient.
\item \emph{Logarithmic penalty.} In the limit $(a,b)\to(0,0)$,
$\mathcal L=-\ln q$ and $\omega\equiv1$ (ordinary cross-entropy).
\item \emph{Polynomial penalty.} If $a<0$, then
$\mathcal L_{a,b,\lambda}(q)\sim\alpha_\lambda q^{-|a|}/|a|$ and
$\omega_{a,b,\lambda}(q)\sim\alpha_\lambda q^{-|a|}$ as $q\to0$: confident mistakes
are punished, and emphasized, \emph{more} than by cross-entropy. For fixed
$(a,b)$ the prefactor $\alpha_\lambda$ decreases with $\lambda$, and the gradient
{weight has its minimum over $q>0$ at $q_m=r/\lambda$, with value $K/D_\lambda\le1$ (strictly below one whenever $\lambda\ne r$, since $\omega_{a,b,\lambda}(1)=\alpha_\lambda+\beta_\lambda=1$). Over probabilities, the minimum occurs at $\min\{q_m,1\}$ (Proposition~\ref{prop:omega-profile}).}
\item \emph{Symmetric label noise.} At $(a,b)=(0,1)$,
$\sum_{j=1}^N\mathcal L_{0,1}(q_j)=N-1$ for every probability vector, so Euler
cross-entropy is a symmetric loss and its risk minimizer is unchanged by
{symmetric label noise~\cite{Ghosh2017,Ma2020}. Each label is flipped with probability $\eta<(N-1)/N$, and a flip chooses uniformly among the other $N-1$ labels.} For $a=0<b<1$,
\begin{equation}
 \frac{N-N^{1-b}}{b}\le\sum_{j=1}^N\mathcal L_{0,b}(q_j)\le\frac{N-1}{b},
 \label{eq:flex-gce-sum}
\end{equation}
so the loss is the generalized cross-entropy $L_q$ of Zhang and
Sabuncu~\cite{ZhangSabuncu2018} with $q=b$, and their approximate
noise-tolerance bound applies with this interval.
\end{enumerate}
\end{proposition}

\begin{proof}
(i)--(iii) follow from $\log_{0,b}(q)=(q^b-1)/b$, the limit to $\ln$, and the
dominant term $\alpha_\lambda(q^a-1)/a$ as $q\to0$; monotonicity of
$\alpha_\lambda=|a|\lambda^{-|a|}/D_\lambda$ in $\lambda$ is elementary.
(iv) $\sum_j(1-q_j)=N-1$. For $0<b<1$,
$\sum_j(1-q_j^b)/b=(N-\sum_jq_j^b)/b$, and the concave sum $\sum_jq_j^b$ ranges
over $[1,N^{1-b}]$ on the simplex, with extremes at vertices and at the
uniform vector.
\end{proof}

\begingroup
In this noise model let $R$ be the clean risk, $R_\eta$ the noisy risk, and $B(\bq)=\sum_j\mathcal L_{0,b}(q_j)$. Conditional averaging over the corrupted label gives
\[
 \begin{aligned}
 R_\eta(f)&=(1-\eta)R(f)+\frac{\eta}{N-1}\mathbb E_{X,Y}\Bigl[\sum_{j\ne Y}\mathcal L_{0,b}(f_j(X))\Bigr]\\
 &=c_\eta R(f)+\frac{\eta}{N-1}\,\mathbb E_X B(f(X)),
 \qquad c_\eta=1-\frac{N\eta}{N-1}>0.
 \end{aligned}
\]
Here $f(X)$ is the probability vector at input $X$. If $f_\eta^*$ and $f^*$ minimize the noisy and clean risks over the same predictor class, comparison under $R_\eta$ and Eq.~\eqref{eq:flex-gce-sum} give
\[
 0\le c_\eta\bigl[R(f_\eta^*)-R(f^*)\bigr]\le\frac{\eta}{N-1}\,\mathbb E_X\bigl[B(f^*(X))-B(f_\eta^*(X))\bigr]\le\frac{\eta}{N-1}\Bigl(\frac{N-1}{b}-\frac{N-N^{1-b}}{b}\Bigr),
\]
and dividing by $c_\eta=(N-1-N\eta)/(N-1)$ yields
\[
 0\le R(f_\eta^*)-R(f^*)\le
 \frac{\eta\bigl(N^{1-b}-1\bigr)}{b(N-1-N\eta)}.
\]
At $b=1$ the upper bound is zero because the sum over labels is constant.\par
\endgroup

The proposition makes the trade-off explicit. Robustness to mislabeled
examples asks that confident mistakes be \emph{down}-weighted, which the
bounded boundary $a=0$ delivers; retaining rare but correct classes asks that
small true-class probabilities be \emph{up}-weighted, which $a<0$ delivers. No
single setting does both. The interior offers a compromise: with $a<0$ small,
the dip of $\omega_{a,b,\lambda}$ at $q_m=r/\lambda$ attenuates
moderately wrong examples, and $\lambda$ places this dip where suspected noisy
labels sit, while extreme mistakes are still emphasized polynomially.
{On the bounded boundary, the loss and its ordinary-softmax logit gradient are bounded. A parameter gradient also contains the network's logit Jacobian, so a uniform bound on parameter gradients for input outliers needs a bound on that Jacobian. \textbf{Status.} C2 follows from the scalar link. The C3 guarantee above applies to symmetric label noise on $a=0$. Robustness in the interior remains to be established.}

\begin{itemize}[leftmargin=1.5em,itemsep=1pt]
\item[\textbf{Q1.}] Does placing the attenuation dip by $\lambda$ at an estimated
noise level improve accuracy under asymmetric or instance-dependent label noise,
compared with GCE ($a=0$) and with bi-tempered losses~\cite{Amid2019}?
\item[\textbf{Q2.}] Can $(a,\lambda)$ be scheduled so that early training is
robust ($a\approx0$) and late training recovers rare classes ($a<0$)?
\end{itemize}

\subsection{C4: Tunable exploration in reinforcement learning}
In the Euler softmax policy update
(Corollary~\ref{cor:Euler-softmax-update}) an action whose accumulated
advantage trails the best one by $\Delta$ {has a polynomial lower bound when $\Delta$ is the gap between its dual score and the maximum dual score, including the initial dual scores. With \rev{$n$} actions, the best action has probability at least \rev{$1/n$}, giving $\pi_i\ge\exp_{a,b,\lambda}(\log_{a,b,\lambda}(\rev{1/n})-\Delta)$. This lower bound is asymptotic to $(|a|\Delta/\alpha_\lambda)^{-1/|a|}$ as $\Delta\to\infty$. Ordinary softmax gives the bound \rev{$e^{-\Delta}/n$}.} Thus $|a|$ sets the polynomial rate at which
neglected actions are abandoned and $\lambda$ the prefactor; the boundary
$a=0$ instead removes clearly inferior actions exactly, as in sparse
Tsallis-regularized MDPs~\cite{Lee2018sparseMDP}. {Exact statewise Euler policy steps using the current policy's exact advantages satisfy the conditions of the monotone value-improvement result in Proposition~\ref{prop:Euler-policy-improvement}.}
\textbf{Status:} the floors and improvement are proved; that they translate into
better exploration (fewer samples to find a good action) is \emph{not}.
Retained probability is not directed exploration, and the vanilla policy
gradient of an Euler softmax policy is \emph{damped} for rare actions by
$m_{a,b,\lambda}(\pi)\sim\pi^{1+|a|}/\alpha_\lambda$
(Proposition~\ref{prop:Euler-softmax-PG}).
\begin{itemize}[leftmargin=1.5em,itemsep=1pt]
\item[\textbf{Q3.}] In bandits and tabular MDPs with deceptive rewards, does
the polynomial floor shorten the time to recover a temporarily bad but
eventually optimal action, and how does the answer depend on $|a|$ and
$\lambda$?
\item[\textbf{Q4.}] Should exploration be tuned through the Euler kernel of a
mirror update, through an Euler entropy bonus, or both?
\end{itemize}

\subsection{C5: Heavy-tailed generative and probabilistic models}
Building an exponential family on $\exp_{a,b,\lambda}$ gives an
\emph{Euler exponential family}
\begin{equation}
 p_\theta(x)=\exp_{a,b,\lambda}\bigl(\langle\theta,T(x)\rangle-\psi(\theta)\bigr),
 \label{eq:flex-family}
\end{equation}
{a deformed exponential family~\cite{Naudts2011}. Here $T(x)$ is the sufficient-statistic vector and $\psi(\theta)$ is an additive normalizer, when it exists for the chosen reference measure. On a finite set with counting measure and $T(x)=\bolde_x$, this is the unit-temperature Euler softmax of Definition~\ref{def:Euler-softmax} in the full-range regime. The continuous example below uses multiplicative normalization and illustrates the same inverse-tail mechanism.}

\begin{proposition}[Euler--Gaussian densities with tunable tails]
\label{prop:flex-euler-gauss}
Let $-1<a<0<b\le1$, $\lambda,\sigma>0$, and
$p(x)=Z^{-1}\exp_{a,b,\lambda}\bigl(-x^2/(2\sigma^2)\bigr)$ on $\mathbb R$. Then
$Z<\infty$, the density is unimodal and symmetric, and
\begin{equation}
 p(x)\sim\frac1Z\Bigl(\frac{|a|\,x^2}{2\alpha_\lambda\sigma^2}\Bigr)^{-1/|a|}
 \qquad(|x|\to\infty).
 \label{eq:flex-gauss-tail}
\end{equation}
{For a nonnegative moment order $k$, $\mathbb E|X|^k<\infty$ exactly when $k<2/|a|-1$. In particular, the variance is finite if and only if $|a|<2/3$. The density has the same tail power as a Student-$t$ with $2/|a|-1$ degrees of freedom. At fixed $\sigma$, as $(a,b)\to(0,0)$ the normalized density converges in $L^1(\mathbb R)$ to that of $N(0,\sigma^2)$.}
\end{proposition}
\begin{proof}
$\exp_{a,b,\lambda}$ is positive and increasing on $\mathbb R$, so $p$ is positive,
symmetric and decreasing in $|x|$. The left tail in Eq.~\eqref{eq:flex-tails}
with $|y|=x^2/(2\sigma^2)$ gives Eq.~\eqref{eq:flex-gauss-tail}; the density is
bounded near zero and decays like $|x|^{-2/|a|}$ with $2/|a|>2$, so $Z<\infty$,
and $|x|^kp(x)$ is integrable iff $2/|a|-k>1$. Student-$t$ densities decay like
$|x|^{-(\nu+1)}$. {To justify the Gaussian limit, choose $p_0\in(0,1)$ and take $|a|\le p_0$. For $t\ge0$, the integral representation in Eq.~\eqref{eq:MVT-Euler-log} gives $-\log_{a,b,\lambda}(e^{-t})\le(e^{p_0t}-1)/p_0$. With $v=x^2/(2\sigma^2)$, it follows that $\exp_{a,b,\lambda}(-v)\le(1+p_0v)^{-1/p_0}$, an integrable envelope in $x$. The inverse converges pointwise to $e^{-v}$. Dominated convergence gives convergence of the normalizing constants and $L^1$ convergence of the normalized densities.}
\end{proof}
{The exponent $|a|$ sets the tail power. The normalized tail coefficient depends on $\alpha_\lambda$ and $Z$, while $b$ and $\lambda$ also shape the profile away from the mode. Since $\exp_{a,b,\lambda}(0)=\exp_{a,b,\lambda}'(0)=1$, the relative density satisfies $p(x)/p(0)=1-x^2/(2\sigma^2)+O(x^4)$ as $x\to0$. Thus $\sigma$ sets the Gaussian quadratic coefficient at the mode independently of the tail power. This local statement concerns the height-normalized profile.} \textbf{Status:} the tail calculus is proved;
benefits for learning are hypotheses. Natural tests are heavy-tailed priors or
likelihoods in variational autoencoders, heavy-tailed noise in diffusion
models, robust regression with an Euler--Gaussian likelihood, and attention or
output layers built from Eq.~\eqref{eq:flex-family}.
\begin{itemize}[leftmargin=1.5em,itemsep=1pt]
\item[\textbf{Q5.}] Can the normalizer $\psi(\theta)$ of
Eq.~\eqref{eq:flex-family} be computed and differentiated efficiently in
continuous spaces, e.g.\ by the one-dimensional root-finding used for the Euler
softmax?
\item[\textbf{Q6.}] Does an Euler--Gaussian likelihood with learned $(a,\lambda)$
match Student-$t$ robustness to outliers while keeping Gaussian-like
efficiency for clean data?
\end{itemize}

\subsection{C6: Probability sparsity}
{For a separable simplex potential and finite scores, a finite slope at zero allows a coordinate to satisfy the boundary KKT inequality and receive exactly zero mass.} On the boundary
$a=0<b\le1$ the Euler softmax is the $(1+b)$-entmax and the Euler FTRL decision
has an explicit support threshold (Proposition~\ref{prop:sparse-FTRL}); for
$a<0$ a shift by $\delta>0$ restores finite slope and yields gates with
computable thresholds for certainly active and inactive coordinates
(Propositions~\ref{prop:p2-shifted-gate} and~\ref{prop:gate-support}). Within
one family one therefore moves continuously from dense polynomial tails
($a<0$), through exact sparsity ($a=0$), to adjustable sparsity of shifted gates.
\textbf{Status:} proved. Sparsity of the probability vector is not the same as
saved computation, which depends on the dispatch rule
(Section~\ref{sec:phase2-moe}).

\subsection{Summary}
Table~\ref{tab:flex-claims} collects the six claims. The remainder of the paper
supplies the tools that make them usable: losses (Sections~\ref{sec:euler-CE}
and~\ref{sec:canonical-softmax}), certified computation
(Section~\ref{sec:inverse-computation}), and optimization algorithms whose Bregman
divergence is generated by the integral of $\log_{a,b,\lambda}$ (Sections~\ref{sec:Euler-MD}--\ref{sec:design-geometry}).

\begin{table}[ht]
\centering\small
\caption{Six flexibility claims for models built on $\log_{a,b,\lambda}$ and $\exp_{a,b,\lambda}$, with the formula that governs each.}
\label{tab:flex-claims}
\begin{tabular}{@{}L{.11\textwidth}L{.13\textwidth}L{.26\textwidth}L{.24\textwidth}L{.14\textwidth}@{}}
\hline
Claim & Knob & Governing formula & What is established & Status\\\hline
C1 Tails & $1/|a|$, $1/b$; $\lambda$ via prefactors & $\exp_{a,b,\lambda}(y)\sim(|a||y|/\alpha_\lambda)^{-1/|a|}$ $(y\to-\infty)$, $(by/\beta_\lambda)^{1/b}$ $(y\to\infty)$ & Polynomial tails, Eq.~\eqref{eq:flex-tails}; weight and policy floors & Proved\\
C2 Penalty bounds & $a=0$ vs.\ $a<0$; $b$ & $\mathcal L_{0,b}(q)=(1-q^b)/b<1/b$; $\mathcal L_{a,b,\lambda}(q)\sim\alpha_\lambda q^{-|a|}/|a|$ & Bounded, logarithmic or polynomial growth (Proposition~\ref{prop:flex-penalty}) & Proved\\
C3 Noisy labels & $a=0$, $b$; dip at $r/\lambda$ & $R(f_\eta^*)-R(f^*)\le\frac{\eta(N^{1-b}-1)}{b(N-1-N\eta)}$ & Symmetric at $(0,1)$; GCE bound for $0<b<1$ & Proved at $a=0$; Q1--Q2 open\\
C4 Exploration & $|a|$, $\alpha_\lambda$ & $\pi(\alpha|s)\ge\exp_{a,b,\lambda}\bigl(\log_{a,b,\lambda}(1/n)-2Y_K(s)\bigr)$ & Polynomial action floors with monotone improvement & Floors proved; Q3--Q4 open\\
C5 Heavy tails & $|a|$ (index), $\lambda$ (amplitude), $b$ (body) & $p(x)\sim Z^{-1}\bigl(|a|x^2/(2\alpha_\lambda\sigma^2)\bigr)^{-1/|a|}$ & Euler--Gaussian tail index $2/|a|$ (Proposition~\ref{prop:flex-euler-gauss}) & Tail proved; Q5--Q6 open\\
C6 Sparsity & $a=0$; shift $\delta$ & $[1+b(\theta_i/\tau-\nu)]_+^{1/b}$; shifted gate & Entmax boundary, shifted gates with thresholds & Proved\\\hline
\end{tabular}
\end{table}

\section{Euler cross-entropy as a loss function}
\label{sec:euler-CE}
\begin{glance}
\textbf{Question.} What changes when cross-entropy uses the Euler logarithm instead of $\ln$?\\
\textbf{Key idea.} The softmax gradient is the ordinary one multiplied by a \emph{gradient weight} $\omega_{a,b,\lambda}(q_y)$ of the true-class probability. The exponents decide whether hard examples are attenuated or emphasized, and $\lambda$ decides \emph{where} the attenuation is strongest.\\
{\textbf{Remember.} Under the statistical assumptions below, the population-optimal prediction is an invertible, ranking-preserving transformation of the true posterior. Composing the loss with that transformation gives a strictly proper loss in posterior coordinates.}
\end{glance}

A network produces logits $\bz$, and ordinary softmax gives
$q_i=e^{z_i}/\sum_j e^{z_j}$. Building the loss on
$\log_{a,b,\lambda}$ changes the weighting of examples.
This use of a scalar logarithm in a loss is distinct from using its
integral as a mirror potential, and from using its inverse in a
variational softmax; the same family connects all three constructions.
{For a target distribution $\bp\in\Delta_N$ and a positive prediction $\bq\in\Delta_N^\circ$,}
\begin{equation}
 \mathcal L_{a,b,\lambda}(\bp,\bq)
 =-\sum_i p_i \log_{a,b,\lambda}(q_i)
 =-\sum_i p_i\left[
 \alpha_\lambda\frac{q_i^a-1}{a}
 +\beta_\lambda\frac{q_i^b-1}{b}\right],
 \label{eq:EulerCE-general}
\end{equation}
with continuous power-link limits at zero exponents.
For a label $y$,
\begin{equation}
 \mathcal L_{a,b,\lambda}(q_y)=-\log_{a,b,\lambda}(q_y).
 \label{eq:EulerCE-onehot}
\end{equation}
The ordinary CE limit is $(a,b)\to(0,0)$ at fixed $\lambda$.
On the Tsallis/GCE boundary $a=0<b\le1$, the scale is redundant and
\begin{equation}
 \mathcal L_{0,b,\lambda}(q_y)=\frac{1-q_y^b}{b}\le\frac1b.
 \label{eq:bounded-GCE}
\end{equation}
This recovers the GCE family~\cite{ZhangSabuncu2018}.
Interior choices $a<0<b$ remain unbounded at zero for every finite
$\lambda>0$; moving the transition does not make them bounded robust losses.

\subsection{Backpropagation through ordinary softmax}
\label{subsec:EulerCE-backprop}
With the curvature $h_{a,b,\lambda}=\log_{a,b,\lambda}'$ of Eq.~\eqref{eq:elasticity-def},
\[
 \frac{\partial\mathcal L_{a,b,\lambda}}{\partial q_i}
 =-p_i h_{a,b,\lambda}(q_i),\qquad
 h_{a,b,\lambda}(x)=\alpha_\lambda x^{a-1}+\beta_\lambda x^{b-1}.
\]
Define the per-class gradient weights
\begin{equation}
 \omega_i=q_i h_{a,b,\lambda}(q_i)
 =\omega_{a,b,\lambda}(q_i)
 =\alpha_\lambda q_i^a+\beta_\lambda q_i^b.
 \label{eq:ri-Euler}
\end{equation}
The ordinary-softmax Jacobian gives
\begin{equation}
 \frac{\partial\mathcal L_{a,b,\lambda}}{\partial z_j}
 =q_j\sum_i p_i \omega_i-p_j\omega_j.
 \label{eq:general-logit-gradient}
\end{equation}
For a one-hot target this becomes
\begin{equation}
 \frac{\partial\mathcal L_{a,b,\lambda}}{\partial z_j}
 =\omega_{a,b,\lambda}(q_y)(q_j-\delta_{jy}).
 \label{eq:onehot-logit-gradient}
\end{equation}

\begingroup
The indices $i,j$ range over the $N$ classes, $y$ is the observed class and $\delta_{jy}$ equals one when $j=y$ and zero otherwise. For soft targets, put $C=\sum_i p_i \omega_i>0$ and $\widetilde p_i=p_i \omega_i/C$. Equation~\eqref{eq:general-logit-gradient} then reads $\nabla_{\bz}\mathcal L=C(\bq-\widetilde\bp)$, so the deformation changes both the effective target and the gradient scale. The gradient components sum to zero because a common logit shift leaves $\bq$ unchanged.\par
\endgroup
{For a one-hot target the effective target remains $\bolde_y$, and $C=\omega_{a,b,\lambda}(q_y)$ changes only the gradient magnitude.} No inverse root solve is required.
For a learned parameter $\xi\in\{a,b,\ell\}$, $\ell=\ln\lambda$,
\[
 \partial_\xi\mathcal L_{a,b,\lambda}
 =-\sum_i p_i\,\partial_\xi \log_{a,b,\lambda}(q_i)
\]
at fixed predictions. The exact partials in~\eqref{eq:dlog-dab}
include the derivatives of the mixing weights.
If $\bq$ depends on these parameters, the chain-rule contribution through
$\bq$ must also be included. Hard clipping requires its own derivative.

\subsection{Tail exponents and a freely placed attenuation region}
\label{subsec:two-parameter-gain}
The loss and its logit-gradient weight are
\begin{equation}
 \mathcal L_{a,b,\lambda}(q)
 =-\alpha_\lambda\log_{a,0}(q)-\beta_\lambda\log_{0,b}(q),\qquad
 \omega_{a,b,\lambda}(q)=\alpha_\lambda q^a+\beta_\lambda q^b.
 \label{eq:ECE-shape-design}
\end{equation}
The negative exponent restores increasing weight to sufficiently
small probabilities. The positive exponent permits an attenuation
region. The scale moves that region relative to an operating probability
such as $1/N$, without changing the two tail exponents.

\begin{proposition}[Gradient-weight minimum and parameter controls]
\label{prop:omega-profile}
Let $p=-a>0$, $d=p+b$, $r=(p/b)^{2/d}$,
$D_\lambda=p\lambda^{-p}+b\lambda^b$, and $K=p r^{-p}+b r^b$.
Then $\mathrm d\ln\omega/\mathrm d\ln q=1-\varepsilon(q)$ and
\begin{equation}
 q_m=\frac r\lambda,\qquad
 \omega_{\min}:=\min_{q>0}\omega_{a,b,\lambda}(q)=\frac K{D_\lambda}.
 \label{eq:omega-min}
\end{equation}
There is a strict interior attenuation minimum on $(0,1)$ precisely when
\begin{equation}
 \lambda>r.
 \label{eq:shape-feasibility}
\end{equation}
If $\lambda\le r$, the minimum on $(0,1]$ is $1$ at $q=1$.
For any desired positive location $q_m$ and fixed $a<0<b$, choose
\begin{equation}
 \lambda=\frac{(|a|/b)^{2/(b-a)}}{q_m}.
 \label{eq:shape-to-parameters}
\end{equation}
This fixes the location; its depth is then determined by
$K/D_\lambda$, and cannot be chosen independently using $\lambda$ alone.
\end{proposition}
\begin{proof}
The sign of $\omega'(q)$ is the sign of
$b^2(\lambda q)^b-p^2(\lambda q)^{-p}$.
It changes once, at $\lambda q=r$.
Substitution gives the minimum, and $\log_{a,b,\lambda}'(1)=1$ gives
$\omega(1)=1$. The location statement follows directly.
\end{proof}
At $\lambda=1$, the earlier two-parameter design is recovered:
with $\rho=p/b\in(0,1)$,
$m=\rho^{(1-\rho)/(1+\rho)}$, $q_m=\rho^{2/(p+b)}$.
{For a prescribed depth $m\in(0,1)$ and location $q_m\in(0,1)$, the function $\rho\mapsto\rho^{(1-\rho)/(1+\rho)}$ increases strictly from zero to one on $(0,1)$. Solve its scalar equation for $\rho$, then set}
\begin{equation}
d=\frac{2\ln\rho}{\ln q_m},\qquad
p=\frac{\rho d}{1+\rho},\qquad b=\frac{d}{1+\rho},
\qquad(\lambda=1).
\label{eq:unscaled-shape-design}
\end{equation}
Require $b<1$ and, when needed, $p<1$.
In the full family one can instead keep desired tail exponents and
place the transition directly with $\lambda$.

\begin{table}[htbp]
\centering\small
\caption{Effect of scale at fixed exponents $a=-0.2$, $b=0.5$.
These are analytic function values, not training results. The mixing weights are listed, and the deepest attenuation is in bold.}
\label{tab:independent-shape}
\begin{tabular}{ccccccc}
\hline $\lambda$ & $\alpha_\lambda$ & $\beta_\lambda$ & $q_m$ & $\omega_{\min}$ & $\varepsilon(1)$ & $\omega(0.01)$\\\hline
0.1 & 0.6672 & 0.3328 & 0.72951 & 0.99489 & 0.96704 & 1.70919\\
1 & 0.2857 & 0.7143 & 0.072951 & 0.67523 & 0.70000 & 0.78911\\
10 & 0.0739 & 0.9261 & 0.0072951 & 0.27684 & 0.55174 & 0.27827\\
100 & 0.0157 & 0.9843 & 0.00072951 & $\mathbf{0.09305}$ & 0.51097 & $\mathbf{0.13781}$\\
\hline
\end{tabular}
\end{table}
The same named interior subfamilies acquire a movable transition:
scaled Kaniadakis is no longer pinned at $q_m=1$, and scaled Abe or
$\gamma$ links need not retain their unscaled probability-domain profile.
Tsallis and Amari boundary links remain independent of $\lambda$.
As $\lambda\to\infty$, the link on a fixed positive operating interval
approaches the positive-power link, while the negative tail survives
ever closer to zero. Such limits are not uniform down to $q=0$.

Attenuation does not distinguish noisy labels from difficult clean
examples. Robustness, class-imbalance benefits and optimization speed
must be evaluated under specified data models.
{The positive floor bounds the multiplier $\omega$, so the one-hot Euler gradient norm is at least this floor times the ordinary cross-entropy gradient norm at the same logits. The probability-domain floor is $c_{a,b,\lambda}$ in Eq.~\eqref{eq:entropy-comparison-constant}. It supports the excess-risk bound and can approach zero as $\lambda$ varies.}

\begingroup
\paragraph{Statistical meaning.}
Assume $a\le0\le b<1$ and an interior true posterior $\bp$, with the ordinary logarithm understood at $(0,0)$. At the unique population optimum $\bq^*$, stationarity requires $p_i h_{a,b,\lambda}(q_i^*)=\nu_{\rm CE}$ for every class, where $\nu_{\rm CE}>0$ enforces $\sum_i q_i^*=1$. Therefore
\[
 p_i=\frac{m_{a,b,\lambda}(q_i^*)}{\sum_j m_{a,b,\lambda}(q_j^*)}.
\]
The mobility $m_{a,b,\lambda}=1/h_{a,b,\lambda}$ is strictly increasing in this regime, so class order is preserved. For an arbitrary interior prediction $\bq$, the same normalization defines the recovered posterior $\br=\mathcal T_{a,b,\lambda}^{-1}(\bq)$. Euler CE is strictly proper when expressed in these posterior coordinates. Its excess conditional risk is $D_{-H_{a,b,\lambda}}(\bp\Vert\br)$, where $H_{a,b,\lambda}$ is the minimum conditional risk. {For $a<0<b<1$, this excess is at least} $(c_{a,b,\lambda}/[2(1-a)])\|\bp-\br\|_1^2$ under Theorem~\ref{thm:EulerCE-regret-bound}. The corrected Euler score in Proposition~\ref{prop:proper-Euler-score} targets the posterior directly. Appendix~\ref{subsec:EulerCE-statistics} gives the derivations and the separate boundary results.\par
\endgroup

\subsection{Practical applications and parameter selection}
\label{subsec:Euler-applications}

\paragraph{Many-class learning and difficult examples.}
When initial true-class probabilities are near $1/N$, a bounded GCE loss can assign very small gradient weights. Equations~\eqref{eq:omega-min}--\eqref{eq:shape-to-parameters} allow the practitioner to choose tail exponents and use $\lambda$ to place $q_m$ relative to this operating range; the resulting minimum weight is given by Eq.~\eqref{eq:omega-min}. The result is a tunable compromise between attenuating moderately difficult examples and retaining their learning signal. To place the minimum at $1/N$, use $\lambda=N(|a|/b)^{2/(b-a)}$; this placement need not optimize a statistical or optimization bound.

\paragraph{Class imbalance and soft targets.}
Correctly labelled minority examples can initially have small true-class probabilities. An Euler loss with a prescribed gradient floor is a candidate for maintaining their contribution during training. This confidence-based mechanism complements class weighting; confidence alone does not identify a minority class or distinguish a difficult clean example from a mislabeled one. For soft targets, including teacher predictions, the general gradient in Eq.~\eqref{eq:general-logit-gradient} applies, and the population characterization identifies the resulting order-preserving target transformation.

\paragraph{Probability estimation.}
When downstream decisions use numerical probabilities, one can either apply the inverse link~\eqref{eq:posterior-recovery} after Euler-CE training or train with the strictly proper score~\eqref{eq:proper-Euler-score}. This connects the deformation to probability estimation through an explicit statistical target. The two options can be compared using held-out log loss, Brier score, and calibration curves, alongside classification accuracy.

\paragraph{Evaluation and tuning.}
A focused empirical study can compare CE, a tuned one-parameter GCE loss, a tuned Kaniadakis loss, Euler CE, and the proper Euler score using the same architecture, optimizer budget, data splits, and number of tuning trials. Ablations at fixed $(a,b)$ with varying $\lambda$, followed by exponents varied at fixed transition location, distinguish scale effects from tail effects. Useful outcomes include macro-averaged recall for class imbalance, accuracy under specified label-noise rates, convergence versus wall-clock time, and dispersion over repeated runs. Standard CE remains an explicit candidate through $(a,b)=(0,0)$. Validation selects whether and how much deformation benefits the task: the analytical controls above describe mechanisms, and do not by themselves establish empirical superiority over CE or other robust losses.

\begin{remark}[Probability and logit curvature]
The statistical convexity statement concerns probabilities. {For a binary logit contrast $z=z_y-z_{\rm other}$, $q=(1+e^{-z})^{-1}$} and the GCE loss~\eqref{eq:bounded-GCE} with $0<b\le1$ give, using $\mathrm dq/\mathrm dz=q(1-q)$ and $\frac{\mathrm d}{\mathrm dz}\frac{1-q^b}{b}=-q^b(1-q)$,
\[
\frac{\mathrm d^2}{\mathrm dz^2}\frac{1-q^b}{b}
=q^b(1-q)\bigl[(1+b)q-b\bigr].
\]
It is negative below $q=b/(1+b)$. Thus probability-space convexity and the population results do not impose convexity on a neural-network training problem. At $b=1$, the population GCE optimum is supported on the most probable classes.
\end{remark}

\paragraph{Natural gradients.} The Euclidean backpropagation signal above extends to natural-gradient backpropagation with the Fisher metric of the softmax output; exact, damped and diagonal versions are derived in Appendix~\ref{subsec:EulerCE-NG}.

\section{Euler softmax: normalization, posterior recovery, and convex logit losses}
\label{sec:canonical-softmax}
\begin{glance}
\textbf{Question.} How should probabilities be produced when the logarithm is not $\ln$?\\
\textbf{Key idea.} Define softmax variationally, with the Euler potential as regularizer in the role played by Shannon entropy. Normalization then becomes a one-dimensional root-finding problem for a single multiplier, rather than division by a sum.\\
\textbf{Remember.} {For the ordinary logarithm, the raw prediction equals the recovered posterior, and the log-normalizer equals the regularized maximum up to the additive convention for the potential. For Euler links these are two separate comparisons, one between vectors and one between scalars.}
\end{glance}

Naively substituting deformed exponentials for exponentials and dividing by their sum is one possible probability map, but it generally loses invariance to a common logit shift; this is a concrete instance of why the Euler pair cannot simply be swapped in for $(\ln,\exp)$. A variational definition preserves this basic property and connects the prediction map to Euler mirror descent. This section distinguishes the normalization multiplier, the convex conjugate, and the posterior recovered from Euler cross-entropy.

\subsection{Variational normalization and its Jacobian}

The scalar primitive and vector potential are
\[
 \phi_{a,b,\lambda}(x)=\alpha_\lambda\Phi_a(x)+\beta_\lambda\Phi_b(x),\qquad
 F_{a,b,\lambda}(\bp)=\sum_i\phi_{a,b,\lambda}(p_i),
\]
where $\Phi_s(x)=\int_1^x\log_{s,0}(u)\,\mathrm du$ is given explicitly in
Section~\ref{sec:Euler-MD}. Thus $\phi(1)=0$ and $\phi'=\log_{a,b,\lambda}$.
We use its closed convex extension on the simplex and $+\infty$ outside it
when forming the conjugate. The scale $\lambda$ and score temperature
$\tau$ are distinct: $\lambda$ changes coordinate-dependent curvature
through the mixture weights, whereas $\tau$ multiplies the whole potential.

\begin{definition}[Euler softmax]
\label{def:Euler-softmax}
Let $n\ge2$ and use $F_{a,b,\lambda}$ with $\lambda>0$ and $-1<a<0<b<1$, or with $a=0<b\le1$ for the sparse boundary case. For logits $\boldsymbol\theta\in\mathbb R^n$ and temperature $\tau>0$, let
\begin{equation}
\sigma_{a,b,\lambda,\tau}(\boldsymbol\theta)=\operatorname*{arg\,max}_{\boldsymbol\rho\in\Delta_n}\bigl\{\langle\boldsymbol\theta,\boldsymbol\rho\rangle-\tau F_{a,b,\lambda}(\boldsymbol\rho)\bigr\},
\qquad
{\sigma_{a,b,\lambda,\tau}(\boldsymbol\theta)_\alpha=\exp_{a,b,\lambda}\bigl(\theta_\alpha/\tau-\nu(\boldsymbol\theta)\bigr)\quad(a<0),}
\label{eq:Euler-softmax}
\end{equation}
{where the component formula is the interior KKT representation. The multiplier $\nu$ is in units of $\boldsymbol\theta/\tau$ and is determined by the unit-mass equation. On $a=0$ use the positive-part KKT formula in Proposition~\ref{prop:Euler-softmax}(iv). {At fixed $\lambda,\tau$, the map converges to ordinary softmax of $\boldsymbol\theta/\tau$ as $(a,b)\to(0,0)$. The equality-constrained score representation then reduces to subtracting the usual log-sum-exp.}}
\end{definition}

\begin{proposition}[Properties of the Euler softmax]
\label{prop:Euler-softmax}
For parts~(i)--(iii), let $\lambda>0$ and $-1<a<0<b<1$, $\boldsymbol\rho=\sigma_{a,b,\lambda,\tau}(\boldsymbol\theta)$, and $m_\alpha=m_{a,b,\lambda}(\rho_\alpha)=1/(\log_{a,b,\lambda})'(\rho_\alpha)$.
\begin{enumerate}[label=(\roman*)]
\item $\boldsymbol\rho$ is unique and strictly positive. It is invariant under $\boldsymbol\theta\mapsto\boldsymbol\theta+c\1$ and order-preserving: $\theta_\alpha>\theta_\beta$ if and only if $\rho_\alpha>\rho_\beta$.
\item Its Jacobian is
\begin{equation}
\frac{\partial\boldsymbol\rho}{\partial\boldsymbol\theta}=\frac1\tau\Bigl(\diag(\mathbf{m})-\frac{\mathbf{m}\mathbf{m}^T}{\1^T\mathbf{m}}\Bigr),
\label{eq:Euler-softmax-Jacobian}
\end{equation}
which is symmetric and positive semidefinite with null vector $\1$. {In the logarithmic limit, $\mathbf{m}\to\boldsymbol\rho$ evaluated at the limiting softmax probabilities, and Eq.~\eqref{eq:Euler-softmax-Jacobian} tends to the softmax Jacobian} $(\diag\boldsymbol\rho-\boldsymbol\rho\boldsymbol\rho^T)/\tau$.
\item As $\tau\to\infty$, $\boldsymbol\rho\to\1/n$. As $\tau\downarrow0$, it tends to the uniform distribution on $\arg\max_\alpha\theta_\alpha$, including the point mass when the maximizer is unique.
\item On $a=0<b\le1$, $\rho_\alpha=[1+b(\theta_\alpha/\tau-\nu)]_+^{1/b}$ equals $(1+b)$-entmax of $\boldsymbol\theta/\tau$ (sparsemax for $b=1$) and can have exact zeros. For $0<b<1$, Eq.~\eqref{eq:Euler-softmax-Jacobian} holds with $m_\alpha=\rho_\alpha^{1-b}$, interpreted as zero off the support. For $b=1$, use $m_\alpha=\mathbf1_{\{\rho_\alpha>0\}}$ on regions with fixed support; at support transitions, sparsemax need not be differentiable.
\end{enumerate}
\end{proposition}
\noindent\textit{The proof is given in Appendix~\ref{app:proofs}, p.~\pageref{prf:prop:Euler-softmax}.}

{ Until the sparse-boundary subsection, assume $-1<a<0<b<1$. Let $\bs=\bz/\tau$} denote dimensionless logits, $\bq=\sigma_{a,b,\lambda,1}(\bs)$, and $\nu=\nu(\bs)$, so that $\log_{a,b,\lambda}(q_i)=s_i-\nu$. Define
\begin{equation}
m_i=m_{a,b,\lambda}(q_i)=1/\log_{a,b,\lambda}'(q_i),\qquad Z_m=\sum_i m_i,\qquad r_i=m_i/Z_m.
\label{eq:softmax-recovered-posterior}
\end{equation}
Thus $\br=\mathcal T_{a,b,\lambda}^{-1}(\bq)$ is the recovered posterior of the statistical section. {The scalar $Z_m$ normalizes the inverse-derivative weights. The multiplier $\nu$ instead normalizes the Euler inverse outputs $q_i$.} Direct differentiation gives
\begin{equation}
\nabla_{\bs}\nu=\br,\qquad
\bJ:=\frac{\partial\bq}{\partial\bs}=\diag(\mathbf{m})-\frac{\mathbf{m}\mathbf{m}^T}{Z_m}.
\label{eq:normalizer-gradient}
\end{equation}
The probability Jacobian is positive definite on the zero-sum subspace and has kernel $\operatorname{span}\{\1\}$. In particular, all pairwise logit contrasts are identifiable, although a common shift is not.

\subsection{Euler cross-entropy under the matching softmax link}
\begin{proposition}[Convex Euler cross-entropy in Euler logits]
\label{prop:matched-Euler-loss}
In the full-range regime above, the single-label Euler cross-entropy composed with the Euler softmax satisfies
\begin{equation}
\mathcal L^{\rm logit}_{a,b,\lambda}(\bs;y)=-\log_{a,b,\lambda}(q_y)=\nu(\bs)-s_y,\qquad
\nabla_{\bs}\mathcal L^{\rm logit}_{a,b,\lambda}=\br-\bolde_y.
\label{eq:matched-loss-gradient}
\end{equation}
It is convex in $\bs$, strictly convex modulo common shifts, and its conditional expected gradient under posterior $\bp$ is $\br-\bp$. Its minimizers satisfy $\bq=\mathcal T_{a,b,\lambda}(\bp)$ and $\br=\bp$ for every interior posterior.
\end{proposition}
\noindent\textit{The proof is given in Appendix~\ref{app:proofs}, p.~\pageref{prf:prop:matched-Euler-loss}.}

\begingroup
The two Jacobians describe different objects. Equation~\eqref{eq:normalizer-gradient} gives $\partial\bq/\partial\bs$, whereas the matched loss has Hessian $\partial\br/\partial\bs=\nabla^2\nu$. In particular, for a logit perturbation $\bv$,
\[
 \bv^T\nabla^2\nu(\bs)\bv
 =\frac{1}{Z_m}\sum_i m_i m_i'\bigl(v_i-\br^T\bv\bigr)^2,
 \qquad m_i'=m_{a,b,\lambda}'(q_i)>0.
\]
Each summand measures the squared change of a logit relative to the mobility-weighted mean change. This quadratic form vanishes exactly for a common shift. For a posterior with a zero coordinate, $\br=\bp$ cannot hold at finite logits in the full-range regime. The Bayes risk is approached by taking positive posteriors converging to $\bp$ and using their optimal scores.\par
\endgroup
For physical logits $\bz$ at fixed $\tau$, the gradient in Eq.~\eqref{eq:matched-loss-gradient} is divided by $\tau$ and its Hessian by $\tau^2$. A general target distribution replaces $\bolde_y$ by $\bp$. This convexity is a property of the loss as a function of free logits; a neural network producing those logits remains a generally nonconvex parameterization. The recovered-posterior residual also differs from the weighted ordinary-softmax residual in Section~\ref{subsec:EulerCE-backprop}. Choosing the matching Euler link does not make raw $\bq$ a proper probability estimate.

\paragraph{A calibrated alternative.} The convex conjugate of the Euler potential yields a Fenchel--Young loss that is proper for the raw prediction and convex in the logits; it is developed in Appendix~\ref{subsec:FY-conjugate}.

\subsection{Numerical normalization and learned parameters}
For full-range parameters the strictly decreasing scalar function $G(\nu)=\sum_i\exp_{a,b,\lambda}(s_i-\nu)-1$ has the finite bracket
\begin{equation}
\min_i s_i-\log_{a,b,\lambda}(1/N)\le\nu\le\max_i s_i-\log_{a,b,\lambda}(1/N).
\label{eq:Euler-softmax-bracket}
\end{equation}
At the lower endpoint every term is at least $1/N$; at the upper endpoint every term is at most $1/N$. Since the largest term equals one at $\nu=\max_is_i$, the lower endpoint can be raised to $\max\{\max_is_i,\ \min_is_i-\log_{a,b,\lambda}(1/N)\}$; this shrinks the bracket whenever the score range exceeds $-\log_{a,b,\lambda}(1/N)$. Safeguarded Newton uses $G'(\nu)=-\sum_im_{a,b,\lambda}(\exp_{a,b,\lambda}(s_i-\nu))<0$, with bisection when a Newton proposal leaves the bracket. Inner inverse evaluations use the safeguarded routines developed below. A practical implementation checks both the inverse residual and total-mass residual; exact-arithmetic bracketing alone is not a floating-point error certificate. Log-domain evaluation is useful for very small probabilities. Jacobian--vector products require only diagonal multiplication, one inner product, and a rank-one correction after the probabilities have been computed.

For a deformation parameter $\xi\in\{a,b,\ell\}$, $\ell=\ln\lambda$, away from singular parameterizations and at fixed $\bs$, implicit differentiation gives
\begin{equation}
\frac{\partial q_i}{\partial\xi}
=-m_i\left[\partial_\xi \log_{a,b,\lambda}(q_i)-\sum_j r_j\partial_\xi \log_{a,b,\lambda}(q_j)\right],\qquad
\frac{\partial\nu}{\partial\xi}=-\sum_jr_j\partial_\xi \log_{a,b,\lambda}(q_j).
\label{eq:Euler-softmax-parameter}
\end{equation}
Here $\partial_\xi \log_{a,b,\lambda}(x)$ holds its scalar argument fixed; the normalization correction ensures $\sum_i\partial_\xi q_i=0$. At fixed logits, exponents and scale, $\partial_\tau\bq=-\bJ\bz/\tau^2$. These formulas allow differentiation through the normalized layer without treating the multiplier as constant.

\subsection{The sparse boundary}
{The radial map divides each $\exp_{a,b,\lambda}(s_i)$ by the sum of all such terms. Its Jacobian is generally nonsymmetric;} it is discussed in Appendix~\ref{app:radial-softmax}. On $a=0<b\le1$, the variational map becomes $(1+b)$-entmax~\cite{Peters2019}; it includes sparsemax at $b=1$. The Fenchel--Young construction remains valid on this sparse boundary. However, Eq.~\eqref{eq:matched-loss-gradient} is stated only for the full-range interior: if a label has zero probability, $\log_{a,b,\lambda}(q_y)=s_y-\nu$ need not hold because its KKT inequality can be strict. {For the Fenchel--Young loss, define $\gamma_i=\log_{0,b}(q_i)-s_i+\nu$ at the optimum, using $\log_{0,b}(0)=-1/b$. The KKT conditions give $\gamma_i\ge0$ and $\gamma_iq_i=0$. The conjugate expression in Eq.~\eqref{eq:Euler-FY} then equals $D_{F_{0,b}}(\bolde_y\Vert\bq)+\gamma_y$. More generally the conditional excess Fenchel--Young risk is $D_{F_{0,b}}(\bp\Vert\bq)+\langle\boldsymbol\gamma,\bp\rangle$. The additional term is zero for labels on the support and is positive when the corresponding KKT inequality is strict.} This distinction prevents an interior identity from being used incorrectly at a sparse prediction.

\section{Computing and differentiating the scaled Euler exponential}
\label{sec:inverse-computation}
\begin{glance}
\textbf{Question.} How do we evaluate $\exp_{a,b,\lambda}$ reliably and differentiate through it?\\
\textbf{Key idea.} Work in logarithmic coordinates $z=\ln x$. There the defining equation is monotone, has an explicit bracket of bounded width, and a slope bounded below, so a safeguarded Newton method converges with a certified error.\\
\textbf{Remember.} Algorithm~\ref{alg:euler-exp} is the scalar workhorse of the whole paper; all derivatives with respect to input, exponents and scale follow from its output alone.
\end{glance}

For $a<0<b$ and fixed $\lambda>0$, every $y\in\mathbb R$ has a unique
positive inverse $x=\exp_{a,b,\lambda}(y)$.
All solver constants below include $\lambda$; calling a two-parameter
inverse with the unchanged argument would compute a different function.
The Tsallis boundaries have the explicit inverses in
Table~\ref{tab:special-cases}.

Closed forms for special exponent pairs, the local series of the inverse around zero, and two integral identities are collected in Appendix~\ref{subsec:exp-closed}; the general case needs the numerical method below.

\subsection{Logarithmic coordinates and an explicit bracket}
\label{subsec:bisection-expab}
Let $x=e^z$. Instead of allowing a primal Newton step to become negative,
solve the monotone scalar equation
\begin{equation}
 G_{a,b,\lambda}(z)
 =\alpha_\lambda\frac{\operatorname{expm1}(az)}a
 +\beta_\lambda\frac{\operatorname{expm1}(bz)}b-y=0.
 \label{eq:Gz}
\end{equation}
Here $y$ is a fixed input to the solver, and
$G'_{a,b,\lambda}(z)=\alpha_\lambda e^{az}+\beta_\lambda e^{bz}$.
To form the mixing weights stably use their log-odds
$\ln(b/|a|)+(b-a)\ln\lambda$ rather than extreme powers of $\lambda$.

\begin{proposition}[Shape and bracketing of the log-coordinate equation]
\label{prop:exp-solver-shape}
Let $a<0<b$, $\lambda>0$, $p=|a|$, $d=p+b$,
$r=(p/b)^{2/d}$, and $c=\max\{p,b\}$. Then:
\begin{enumerate}[label=(\roman*),leftmargin=2.2em]
\item $G'(z)=\omega_{a,b,\lambda}(e^z)\ge
\omega_{\min}=(p r^{-p}+b r^b)/(p\lambda^{-p}+b\lambda^b)>0$, the minimum of Eq.~\eqref{eq:omega-min}.
If $z_*$ is the root, $|z-z_*|\le |G(z)|/\omega_{\min}$.
\item $|G''(z)|\le cG'(z)$. The sign of $G''$ changes from negative
to positive at $z_q=\ln(r/\lambda)$.
\item If $y=0$, $z_*=0$. Otherwise,
\begin{equation}
 [\ell,u]=
 \begin{cases}
 [b^{-1}\ln(1+by),\ b^{-1}\ln(1+by/\beta_\lambda)],&y>0,\\
 [-p^{-1}\ln(1+p|y|/\alpha_\lambda),\
  -p^{-1}\ln(1+p|y|)],&y<0
 \end{cases}
 \label{eq:exp-bracket}
\end{equation}
contains $z_*$. Its width is at most
$-\ln\beta_\lambda/b$ for $y>0$ and $-\ln\alpha_\lambda/p$ for $y<0$.
\end{enumerate}
\end{proposition}
\noindent\textit{The proof is given in Appendix~\ref{app:proofs}, p.~\pageref{prf:prop:exp-solver-shape}.}
The bracket width is bounded over $y$ for fixed parameters, but not
uniformly over all scales or exponents. This dependence matters in
solver-cost comparisons.

\subsection{Safeguarded Newton: the practical method}
\label{subsec:exp-newton}

\begingroup
\paragraph{Conventions used in Algorithms~\ref{alg:euler-exp}--\ref{alg:moe-step}.}
Within a listing, $\ell_E(x)=\log_{a,b,\lambda}(x)$, $E_E(y)=\exp_{a,b,\lambda}(y)$, $h_E=\ell_E'$ and $m_E=1/h_E$ use the triple declared in its inputs. The subscript $j$ on a function or normalizer selects block $j$'s triple; the subscript $v$ selects the expert prior's triple. A scalar function of a vector is applied coordinatewise. The notation $[r]_+=\max\{r,0\}$ and $\1/N$ respectively denote the positive part and the uniform vector. Use $\operatorname{hr}(\bg)=(\max_i g_i-\min_i g_i)/2$ and $\operatorname{ctr}(\bg)=\bg-(\max_i g_i+\min_i g_i)\1/2$ to measure or remove a common score offset.

$\mathsf{Norm}(\by;\epsilon)$ calls Algorithm~\ref{alg:euler-norm} with unit weights. It accepts finite scores and returns $(\bp,\nu,\hat\epsilon)$ on success, or a precision-limit status without a certified decision. A dot in an output tuple discards that field. For weighted calls write $\mathsf{Norm}(\by,\bD;\epsilon)$, where $\bD$ lists the positive coordinate weights. Use any fixed safeguard fraction in $(0,1/2)$, for example one quarter. A certified decision is feasible, is strictly positive on the dense branch, and has a verified stationarity residual at most the requested tolerance. The exact map $\mathcal N_{a,b,\lambda}$ returns only the optimizer. Numerical use of that map in an exact-arithmetic listing calls for the separate residual bounds cited there. A returned multiplier is the one used for the candidate and need not equal the exact root.
\par\endgroup

\paragraph{Which error should the solver control?}
An approximation $\hat x=e^{\hat z}$ of $x^*=\exp_{a,b,\lambda}(y)$ can be judged in two ways. The \emph{residual} (backward error)
\[
 g(\hat z)=\log_{a,b,\lambda}(\hat x)-y=G_{a,b,\lambda}(\hat z)
\]
measures how well $\hat x$ satisfies the defining equation; it is what a solver can evaluate. The \emph{solution error} (forward error) $|\hat z-z^*|=|\ln\hat x-\ln x^*|$ measures how far $\hat x$ is from the answer. Since $\hat x/x^*=e^{\hat z-z^*}$, a solution error $\epsilon$ means a relative error of at most $e^\epsilon-1\approx\epsilon$ in $x$. The next lemma connects the two measures exactly.

\begin{lemma}[Residual and solution error]
\label{lem:exp-residual}
Let $a<0<b$, let $J=[\ell,u]$ be an interval that contains both $\hat z$ and $z^*$, and let $s_J=\min_{t\in J}\omega_{a,b,\lambda}(e^t)$ and $S_J=\max_{t\in J}\omega_{a,b,\lambda}(e^t)$. Then
\begin{equation}
 s_J\,|\hat z-z^*|\ \le\ |g(\hat z)|\ \le\ S_J\,|\hat z-z^*|.
 \label{eq:exp-residual}
\end{equation}
Moreover $s_J$ is computable: with {$z_q=\ln q_m=2\ln(|a|/b)/(b-a)-\ln\lambda$}, it equals $\omega_{a,b,\lambda}(e^u)$ if $u\le z_q$, $\omega_{a,b,\lambda}(e^\ell)$ if $\ell\ge z_q$, and $\omega_{\min}$ otherwise.
\end{lemma}
\noindent\textit{The proof is given in Appendix~\ref{app:proofs}, p.~\pageref{prf:lem:exp-residual}.}

The lemma explains why Algorithm~\ref{alg:euler-exp} \emph{tests} the residual but \emph{guarantees} the solution error. Three reasons make the solution error the right target.
\begin{itemize}[leftmargin=1.5em,itemsep=1pt]
\item \emph{It is what the user needs.} The Euler exponential produces positive numbers that range over many orders of magnitude, from tiny tail probabilities to large weights. Relative accuracy is meaningful at every scale, and it is exactly what a bound on $|\ln\hat x-\ln x^*|$ provides.
\item \emph{The residual is not scale-free.} The same residual corresponds to very different errors, because the slope $\omega_{a,b,\lambda}$ varies strongly. For $(a,b,\lambda)=(-\frac12,\frac12,1)$ and $y=10^6$, we have $x^*\approx10^{12}$ and $\omega_{a,b,\lambda}(x^*)\approx5\times10^5$, so a residual of $10^{-10}$ means a relative error of about $2\times10^{-16}$. At $y\approx0$, where $\omega_{a,b,\lambda}\approx1$, the same residual means a relative error of $10^{-10}$.
\item \emph{A fixed residual tolerance can be unattainable.} In floating point, evaluating $\log_{a,b,\lambda}(\hat x)-y$ loses about $\mathrm u\,|y|$ to rounding, where $\mathrm u\approx1.1\times10^{-16}$ is the unit roundoff. {At $y=10^6$ the uncertainty of a direct double-precision evaluation can be of order $10^{-10}$. A test requiring a certified residual below $10^{-14}$ can therefore stall at that working precision.}
\end{itemize}
When the residual itself is what matters, it is available too: inside the Euler softmax the optimality conditions are stated in dual coordinates, $\log_{a,b,\lambda}(w_i)=y_i-\nu$, so the residual is the dual error. Algorithm~\ref{alg:euler-exp} therefore returns both $\hat x$ and $g(\hat z)$, and by Lemma~\ref{lem:exp-residual} a solution error $\epsilon$ implies $|g|\le S_J\,\epsilon$.

\paragraph{The algorithm.}
Algorithm~\ref{alg:euler-exp} keeps an interval $[\ell,u]$ that always contains $z^*=\ln x^*$, starting from the bracket of Proposition~\ref{prop:exp-solver-shape}. At each iteration it evaluates the residual at the current guess $z$, uses its sign to shrink the interval, and stops as soon as Lemma~\ref{lem:exp-residual} certifies $|z-z^*|\le\epsilon$. Otherwise it takes a Newton step, $z\leftarrow z-g/\omega_{a,b,\lambda}(e^z)$, and falls back to bisection when Newton would leave the interval or makes too little progress.

\begin{algorithm}[ht]
\caption{{\textsc{EulerExp}$(y;a,b,\lambda,\epsilon,\theta)$}: safeguarded Newton evaluation of $\exp_{a,b,\lambda}(y)$ in log coordinates}
\label{alg:euler-exp}
\begingroup\small
\begin{algorithmic}[1]
\Require Inverse input $y\in\mathbb R$; $a<0<b$, $\lambda>0$; log error $\epsilon>0$; safeguard $0<\theta<1/2$
\Ensure $(\hat z,e^{\hat z},G(\hat z))$, with $|\hat z-\ln \exp_{a,b,\lambda}(y)|\le\epsilon$ in exact arithmetic
\State Compute $\alpha_\lambda\gets[1+\exp(\ln(b/|a|)+(b-a)\ln\lambda)]^{-1}$ and $\beta_\lambda\gets1-\alpha_\lambda$.
\State Put $G(z)=\log_{a,b,\lambda} (e^z)-y$ $=\alpha_\lambda\operatorname{expm1}(az)/a+\beta_\lambda\operatorname{expm1}(bz)/b-y$ and $W(z)=G'(z)=\alpha_\lambda e^{az}+\beta_\lambda e^{bz}$.
\If{$y=0$}
 \State \Return $(0,1,0)$
\EndIf
\State Obtain $[\ell,u]$ from Eq.~\eqref{eq:exp-bracket}; set $z\gets(\ell+u)/2$.
\State $z_q\gets 2\ln(|a|/b)/(b-a)-\ln\lambda$ \Comment{minimum of $W$}
\Loop
 \If{$u-\ell\le2\epsilon$}
  \State $\hat z\gets(\ell+u)/2$; \Return $(\hat z,e^{\hat z},G(\hat z))$
 \EndIf
 \State $g\gets G(z)$.
 \If{$g=0$}
  \State \Return $(z,e^z,0)$
 \EndIf
 \State If $g>0$ set $u\gets z$; otherwise set $\ell\gets z$.
 \State $z_c\gets\min\{u,\max\{\ell,z_q\}\}$; $s\gets W(z_c)$.
 \If{$|g|\le\epsilon s$}
  \State \Return $(z,e^z,g)$
 \EndIf
 \State $z_N\gets z-g/W(z)$; $d\gets\theta(u-\ell)$.
 \State Use $z\gets z_N$ if $z_N\in[\ell+d,u-d]$; otherwise use $z\gets(\ell+u)/2$.
\EndLoop
\Guarantee By Proposition~\ref{prop:exp-newton}, the log-error bound holds after at most $\lceil\ln(w_0/(2\epsilon))/(-\ln(1-\theta))\rceil$ sign updates in exact arithmetic.
\end{algorithmic}
\endgroup
\end{algorithm}

The tolerance controls the logarithm of the answer, not the raw residual:
$|\hat x/\exp_{a,b,\lambda}(y)-1|\le\operatorname{expm1}(\epsilon)$.
Thus a requested relative error $\delta>0$ is obtained by choosing
$\epsilon=\ln(1+\delta)$. The log output $\hat z$ remains useful when $e^{\hat z}$
cannot be represented in the working floating-point format. All certificates
in the pseudocode are exact-arithmetic statements; the implementation contract
below specifies what is needed to retain them under rounding.

\begin{proposition}[Correctness and convergence of Algorithm~\ref{alg:euler-exp}]
\label{prop:exp-newton}
Let $a<0<b$, $\lambda>0$, $\epsilon>0$, $0<\theta<1/2$, $c=\max\{|a|,b\}$,
and let $w_0$ be the initial bracket width. In exact arithmetic:
\begin{enumerate}[label=(\roman*),leftmargin=2.2em]
\item Every successful return satisfies the stated log-error bound.
\item Each nonterminal sign update reduces the width by at least the factor
$1-\theta$. If $w_0>2\epsilon$, at most
\[
K=\left\lceil\frac{\ln(w_0/(2\epsilon))}{-\ln(1-\theta)}\right\rceil
\]
sign updates are needed; if $w_0\le2\epsilon$, none is needed. An additional
residual evaluation may be used to report the residual at the final midpoint.
\item Every accepted Newton step satisfies
$|e_{k+1}|\le(c/2)e^{c|e_k|}e_k^2$, where $e_k=z_k-z^*$.
This is a bound on Newton steps, not a claim of eventual quadratic convergence
of the safeguarded sequence: the central-interval test may continue to reject them.
\end{enumerate}
\end{proposition}
\noindent\textit{The proof is given in Appendix~\ref{app:proofs}, p.~\pageref{prf:prop:exp-newton}.}

\paragraph{Floating-point implementation contract.}
Stable evaluation and a rigorous certificate are different requirements.
Use \texttt{log1p} for $\ln(1+u)$ in the initial bracket and
\texttt{expm1} in the weighted formula near zero.
Neither an unscaled hyperbolic-function implementation nor separate exponentials are
overflow-safe for all inputs; scale the evaluation, or increase precision.
For a rigorous certificate, round bracket endpoints outwards, enclose the
residual by $[g_-,g_+]$, and use a verified positive lower bound $s_-$ on
the slope. Update the upper endpoint only if $g_->0$, and the lower endpoint
only if $g_+<0$. The residual test becomes
$\max\{|g_-|,|g_+|\}\le\epsilon s_-$.
If the sign is unresolved and the test fails, increase evaluation precision;
never choose a side from an uncertified sign. Account also for rounding in
the midpoint and in the conversion $\hat x=e^{\hat z}$, or return the log
enclosure itself. If no representable interior trial exists, increase
precision or return an explicit \textsc{precision-limit} status with the
current enclosure. A fixed precision implementation cannot promise successful
termination for every positive tolerance.

\paragraph{Empirical performance.}
A reproducible benchmark over exponents, scale, inputs and tolerances is needed before assigning an average iteration count. The convergence guarantee is an exact-arithmetic bound, not a measured runtime claim.

\paragraph{Practical advice.}
\begin{itemize}[leftmargin=1.5em,itemsep=1pt]
\item \emph{Work with $z=\ln x$ whenever possible.} For extreme arguments, $x$ itself can overflow or underflow in double precision while $z$ is perfectly representable. { Store $z$ where possible, but normalization, mobility evaluation, and downstream algorithms may also need primal weights; their dynamic range must be handled explicitly.}
\item \emph{Warm starts in vector operations.} Inside the Euler softmax (Algorithm~\ref{alg:euler-norm}) every coordinate is evaluated at $y_i-\nu$ for a sequence of multipliers $\nu$ that change little from one iteration to the next. { A previous $z_i$ may be used as a warm start after clipping it to the central fraction of the new bracket. The number of inner and outer iterations should be measured rather than assumed.}
\item \emph{Monotone Newton.} By part~(ii) of Proposition~\ref{prop:exp-solver-shape}, {undamped Newton converges monotonically when $z_0\ge z^*\ge z_q$ on the convex branch, or $z_0\le z^*\le z_q$ on the concave branch. These inequalities ensure that the entire interval between the start and the root lies on the same branch. For a bracket crossing $z_q$, the safeguarded iteration provides the stated general guarantee.}
\item \emph{Boundary cases.} For $a=0<b$ or $a<0=b$ the inverse has the closed form of Table~\ref{tab:special-cases}. Only arguments inside its domain are admissible: a positive-part extension used outside the domain defines a different function.
\end{itemize}

\subsection{Differentiating through the inverse and the scale}
\label{subsec:exp-derivatives}
For $x=\exp_{a,b,\lambda}(y)$,
\begin{equation}
 \frac{\mathrm dx}{\mathrm dy}
 =m_{a,b,\lambda}(x)
 =\frac{x}{\alpha_\lambda x^a+\beta_\lambda x^b},\qquad
 \frac{\mathrm d\ln x}{\mathrm dy}=\frac1{\omega_{a,b,\lambda}(x)}.
 \label{eq:expab-derivative}
\end{equation}
For parameter derivatives it is convenient to use $\ell=\ln\lambda$.
Define
\[
 J_s(x)=\partial_s\log_{s,0}(x)
 =\frac{s x^s\ln x-(x^s-1)}{s^2},\qquad
 J_0(x)=\tfrac12(\ln x)^2.
\]
Holding the other parameters and $x$ fixed, the exact interior partials are
\begin{equation}
 \begin{aligned}
 \partial_a \log_{a,b,\lambda}(x)
 &=\alpha_\lambda J_a(x)
   +\alpha_\lambda\beta_\lambda(1/a+\ell)
       [\log_{a,0}(x)-\log_{0,b}(x)],\\
 \partial_b \log_{a,b,\lambda}(x)
 &=\beta_\lambda J_b(x)
   -\alpha_\lambda\beta_\lambda(1/b+\ell)
       [\log_{a,0}(x)-\log_{0,b}(x)],\\
 \partial_\ell \log_{a,b,\lambda}(x)
 &=(b-a)\alpha_\lambda\beta_\lambda
       [\log_{0,b}(x)-\log_{a,0}(x)].
 \end{aligned}
 \label{eq:dlog-dab}
\end{equation}
These follow by differentiating the log-odds
$\ln(\beta_\lambda/\alpha_\lambda)=\ln(b/|a|)+(b-a)\ell$.
If derivatives in $\lambda$ rather than $\ell$ are required,
$\partial_\lambda f=\partial_\ell f/\lambda$.
Implicit differentiation gives, for each $\xi\in\{a,b,\ell\}$,
\begin{equation}
 \partial_\xi \exp_{a,b,\lambda}(y)
 =-m_{a,b,\lambda}(x)\,\partial_\xi \log_{a,b,\lambda}(x).
 \label{eq:du_da}
\end{equation}

\begingroup
In this equation $y$ is fixed and $x$ is the converged inverse. The link partial on the right holds $x$ fixed. The total derivative when $y$ also depends on $\xi$ is $\mathrm dx/\mathrm d\xi=m_{a,b,\lambda}(x)\{\mathrm dy/\mathrm d\xi-\partial_\xi\log_{a,b,\lambda}(x)\}$.\par
\endgroup
These are derivatives of the converged root. Differentiating a fixed
number of iterations need not produce the same result.
The formulas with $1/a$ and $1/b$ are stated for the interior;
{use the boundary formula for derivatives tangent to a single-power branch. A one-sided derivative into the two-power family differentiates the full extension, including the exponent dependence of its weights.}

\begin{corollary}[Monotonicity in scale, and the limits of exponent monotonicity]
\label{cor:exp-monotone-ab}
For fixed $a<0<b$ and $y\ne0$,
$\partial_\ell \exp_{a,b,\lambda}(y)<0$.
{At $\lambda=1$ and $y\ne0$, either exponent partial is strictly negative. At other fixed scales the weight derivatives can reverse the sign. At $y=0$ all these parameter derivatives vanish because the inverse equals one for every triple.}
\end{corollary}
\noindent\textit{The proof is given in Appendix~\ref{app:proofs}, p.~\pageref{prf:cor:exp-monotone-ab}.}
An inverse log-error $|\ln\hat x-\ln x|\le\epsilon$ bounds the mobility error:
\[
 \left|\ln\frac{m_{a,b,\lambda}(\hat x)}
 {m_{a,b,\lambda}(x)}\right|
 \le\max\{|1-a|,|1-b|\}\epsilon.
\]
Parameter derivatives are locally well behaved on compact interior
parameter/input domains, but no uniform absolute error bound follows
in extreme tails or degenerate limits.

\subsection{Learning all three parameters}
\label{sec:learn-parameters}
For full-range concave links and finite boundary potentials use
\begin{equation}
 -1<a<0,\qquad 0<b<1,\qquad \lambda>0.
 \label{eq:ab-domain}
\end{equation}
A convenient parameterization is
\begin{equation}
 a=-\operatorname{sig}(\alpha),\qquad
 b=\operatorname{sig}(\beta),\qquad
 \lambda=e^\ell,\qquad
 \operatorname{sig}(u)=\frac1{1+e^{-u}}.
 \label{eq:reparam}
\end{equation}
Here $\alpha,\beta$ are unconstrained training parameters, distinct
from the mixture weights $\alpha_\lambda,\beta_\lambda$.
For a differentiable outer objective,
\begin{equation}
 \partial_\alpha\mathcal L=a(1+a)\partial_a\mathcal L,\qquad
 \partial_\beta\mathcal L=b(1-b)\partial_b\mathcal L,\qquad
 \partial_\ell\mathcal L=\lambda\partial_\lambda\mathcal L.
 \label{eq:chain-reparam}
\end{equation}
Bounds on $\ell$ and margins from the exponent boundaries can help keep
conditioning and solver costs controlled. Such bounds are modeling choices.
Changing any geometry parameter during online learning requires
accounting for geometry drift; the fixed-geometry guarantees of
Section~\ref{sec:Euler-MD} do not cover arbitrary learned schedules.

\paragraph{A caution.} {At fixed predictions with positive target mass on a class of probability below one, the raw Euler CE strictly decreases with $\lambda$ in the two-power interior, and with $b$ on the bounded branch. For predictions produced by a learned Euler link, the total derivative also contains the prediction dependence.} Parameters should therefore be selected on validation data or with explicit regularization (Appendix~\ref{subsec:ab-evolution}).

\section{Euler mirror descent and FTRL: established algorithms in Euler geometry}
\label{sec:mirror-EG}\label{sec:Euler-MD}
\begin{glance}
\textbf{Question.} Which of the established mirror-descent (MD) and follow-the-regularized-leader (FTRL) algorithms gain something concrete when their Bregman divergence is generated by the Euler potential, the integral of $\log_{a,b,\lambda}$, rather than by negative entropy, the integral of $\ln$?\\
\textbf{Key idea.} Keep each algorithm exactly as practitioners know it---its loss approximation, learning-rate schedule and sparsity mechanism---and change only the mirror map. Every update then becomes one evaluation of $\exp_{a,b,\lambda}$ per coordinate, plus one scalar normalization on the simplex.\\
\textbf{Remember.} Five families are kept: exact mirror descent and exponentiated gradient, Euler-Hedge FTRL with an optimally chosen scale, AdaGrad-type adaptive FTRL, FTRL-Proximal and RDA with exact sparsity, and optimistic FTRL. {The full-range simplex constructions retain small positive weights on a polynomial scale. A boundary link or a shifted penalty can instead set coordinates exactly to zero. Additional} variants are collected in Appendix~\ref{app:supp-online}.
\end{glance}

\paragraph{Which algorithms, and why.}
FTRL and MD are the backbone of online learning, of large-scale sparse
prediction, and of adaptive optimization. FTRL chooses each action by
minimizing the sum of all past (linearized) losses plus a regularizer; MD
updates the previous action with a local Bregman step~\cite{Nemirowsky,BeckTeboulle2003}. They coincide for a
constant step size without active constraints, and on the simplex with
matched initialization, but differ for adaptive steps, composite penalties and
general constraints. Table~\ref{tab:sota-ftrl} lists the families that matter
most in practice or theory and records, for each, what the Euler link changes
and where it is treated. We keep a family in this section only if the Euler
link makes a clear, provable contribution to it; this criterion selects the
five families of the glance box. Three further families are important but are
not changed in a useful way by a separable Euler potential: parameter-free
(coin-betting) methods adapt to unknown comparator norms through a radial
potential, and full-matrix or Kronecker-factored preconditioners (Shampoo,
Muon) act on correlated matrix curvature; they are discussed in
Appendix~\ref{subsec:new-parameterfree} and Section~\ref{sec:design-geometry}.
KL trust-region methods for policies are the subject of
Section~\ref{sec:extensions}.

\begin{table}[ht]
\centering\footnotesize\setlength{\tabcolsep}{3pt}\renewcommand{\arraystretch}{1.15}
\caption{Established MD and FTRL families, the role of the Euler link, and where each is treated. The order roughly follows combined practical and theoretical importance.}
\label{tab:sota-ftrl}
\begin{tabular}{@{}L{.17\textwidth}L{.11\textwidth}L{.17\textwidth}L{.33\textwidth}L{.14\textwidth}@{}}
\hline
Family & Baseline link & Main use & What the Euler link adds & Treated in\\\hline
AdaGrad-type adaptive FTRL~\cite{DuchiAdaGrad2011,McMahan2017} & $w-1$ or $\ln w$ & Sparse and deep learning; basis of RMSProp/Adam & Scale-free regret with a polynomial, scale-independent coordinate floor; signed Euler-AdaGrad & Section~\ref{subsec:FTRL-adaptive}, Section~\ref{subsec:euler-adagrad}\\
FTRL-Proximal~\cite{McMahan2011,McMahan2013} & shifted quadratic & Billion-feature CTR and recommender models & Exact $\ell_1$ zeros with a tunable growth of active weights, $[br/\beta_\lambda]^{1/b}$ & Section~\ref{sec:new-da}\\
Optimistic FTRL/MD~\cite{Chiang2012,Rakhlin2013} & $\ln w$ or $w-1$ & Games, min-max training, predictable losses & Regret in forecast errors with polynomial floors; Euler Mirror-Prox and self-play & Section~\ref{sec:new-optimism}, Section~\ref{sec:phase2-games}\\
KL mirror descent / trust regions (MDPO, TRPO, PPO)~\cite{TomarMDPO2022,Schulman2015TRPO,Lan2022PMD} & $\ln\pi(\alpha|s)$ & RL and LLM post-training & Euler trust region retaining rare actions; certified improvement & Section~\ref{sec:extensions}\\
Entropic MD, EG and Hedge~\cite{EG,FreundSchapire1997} & $\ln w$ & Experts, boosting, routing & The logarithmic limit; Euler version with polynomial floors and optimal scale & Sections~\ref{NABEG1}, \ref{sec:FTRL-3p}\\
Parameter-free FTRL (coin betting)~\cite{OrabonaPal2018} & radial coin betting & Tuning-free online learning & No direct gain from a separable kernel & Appendix~\ref{subsec:new-parameterfree}\\
Full-matrix and Kronecker MD (Shampoo)~\cite{Gupta2018} & matrix or spectral & Foundation-model pretraining & Complementary; position- vs.\ history-dependent curvature & Section~\ref{sec:design-geometry}\\
RDA and FOBOS / composite MD~\cite{Xiao2010,DuchiSinger2009} & $w-1$ & Structured sparsity & Signed Euler-RDA with dead zone and bounded-domain regret & Section~\ref{sec:new-da}\\\hline
\end{tabular}
\end{table}
Two common statements about these families need qualification. First, the
``$O(1)$ regret'' of optimistic methods in games refers to the \emph{sum} of the
players' regrets in two-player zero-sum games~\cite{Rakhlin2013,Syrgkanis2015};
individual regrets in general games are polylogarithmic~\cite{Daskalakis2021},
and adversarial forecasts can make optimism worse than no forecast
(Remark~\ref{rem:aoftrl}). Second, PPO enforces a trust region by ratio
clipping~\cite{Schulman2017PPO} rather than by an exact KL mirror step, so an
Euler trust region modifies MDPO-type updates directly and PPO-type updates
only through a surrogate.

\subsection{Euler mirror descent}
\paragraph{Intuition: mirror descent works in two spaces.}
For readers new to mirror descent it helps to keep two spaces in view. The \emph{primal} space contains the objects being learned: probability vectors, mixture weights or policies, here $\bw\in\Delta_N$. The \emph{dual} space contains the coordinates $\bz=\nabla F_{a,b,\lambda}(\bw)=\log_{a,b,\lambda}(\bw)\in\mathbb R^N$, in which gradients are added linearly. {One mirror step maps to dual coordinates and subtracts $\eta\bg$. On the simplex it also subtracts a common scalar from the dual scores before applying $\exp_{a,b,\lambda}$. That scalar is chosen so the resulting primal weights sum to one.} The potential therefore decides how an additive change of a dual coordinate becomes a change of probability: to first order $\mathrm dw_i=m_{a,b,\lambda}(w_i)\,\mathrm dz_i$, with the mobility $m_{a,b,\lambda}=1/h_{a,b,\lambda}$. Three choices show the range.
\begin{itemize}[leftmargin=1.5em]
\item \emph{Entropy}, $(a,b)\to(0,0)$: $m(w)=w$. Exponentiated gradient multiplies weights by $e^{-\eta g_i}$, so a coordinate with persistently large losses decays exponentially, $w_i\propto\exp(-\eta\sum_sg_{i,s})$.
\item \emph{Euclidean}, $(a,b)=(0,1)$: $\log_{0,1}(w)=w-1$ and $m\equiv1$; the step is a Euclidean projection, which sets small coordinates exactly to zero.
\item \emph{Euler interior}, $a<0<b$: $m_{a,b,\lambda}(w)\sim w^{1+|a|}/\alpha_\lambda$ as $w\downarrow0$, so small coordinates move cautiously and decay only polynomially (Section~\ref{subsec:euler-vs-ln-tsallis}). The ratio of the head to the tail part of the curvature, $(b/|a|)(\lambda w)^{b-a}$, increases with $w$, so the exponent $b$ increasingly shapes the geometry of large coordinates.
\end{itemize}
Thus $|a|$ controls how strongly small coordinates resist extinction, which protects rarely good actions against transient bad gradients, while $b$ shapes concentration among large coordinates. The exponents set limiting powers, while $\lambda$ changes their mixture weights and shifts the transition. Their effects on curvature at a finite coordinate remain coupled. Theorem~\ref{thm:pnorm-mixture} states this mixture at the level of Hessians.

Vectors are bold lowercase, matrices bold uppercase, and scalar nonlinearities act componentwise. Let $L$ be a differentiable objective, $\bg_t=\nabla L(\bw_t)$, and $\eta_t>0$. The Euler geometry is specified directly by
\[
 \phi_{a,b,\lambda}(x)=\int_1^x\log_{a,b,\lambda}(r)\,\mathrm dr,\qquad
 F_{a,b,\lambda}(\bw)=\sum_i\phi_{a,b,\lambda}(w_i),\qquad
 \nabla F_{a,b,\lambda}(\bw)=\log_{a,b,\lambda}(\bw).
\]
No additional link function is needed. We use $a<0<b$ for an inverse defined on all real dual coordinates; finite-range boundary cases require the KKT conditions described below.

For a convex feasible set $K$, Euler mirror descent minimizes the linearized objective plus a geometry-dependent displacement penalty:
\begin{equation}
 \bw_{t+1}=\arg\min_{\bw\in K}\left\{
 \eta_t\langle\bg_t,\bw\rangle+D_{F_{a,b,\lambda}}(\bw\Vert\bw_t)\right\}.
 \label{Eq-1a}
\end{equation}

\begingroup
Here $K$ is the feasible set, $\eta_t$ is the step size, and the second argument $\bw_t$ of the Bregman divergence is the current iterate. The equivalent dual-coordinate optimality condition is
\[
 0\in\eta_t\bg_t+\nabla F_{a,b,\lambda}(\bw_{t+1})
 -\nabla F_{a,b,\lambda}(\bw_t)+N_K(\bw_{t+1}),
\]
provided the potential is differentiable at both iterates and the normal-cone representation applies. Here $N_K(\bw)=\{\bn:\langle\bn,\bv-\bw\rangle\le0\text{ for all }\bv\in K\}$. For a positive simplex solution the normal cone contributes precisely the common multiplier.\par
\endgroup
The Bregman divergence is
\begin{equation}
 D_{F_{a,b,\lambda}}(\bw\Vert\bv)=F_{a,b,\lambda}(\bw)-F_{a,b,\lambda}(\bv)
 -\langle\log_{a,b,\lambda}(\bv),\bw-\bv\rangle.
 \label{Bregman1}
\end{equation}
It is the gap between the potential and its tangent plane at the anchor $\bv$~\cite{Bregman1967,MD1}. Since $\log_{a,b,\lambda}'>0$, the potential is strictly convex, so the gap is nonnegative and vanishes only when its arguments agree. Its derivatives in the first argument expose the Euler link explicitly:
\begin{align}
 \nabla_{\bw}D_{F_{a,b,\lambda}}(\bw\Vert\bv)&=\log_{a,b,\lambda}(\bw)-\log_{a,b,\lambda}(\bv),\nonumber\\
 \nabla^2_{\bw}D_{F_{a,b,\lambda}}(\bw\Vert\bv)&=\bH_{a,b,\lambda}(\bw)
 =\diag\{\log_{a,b,\lambda}'(w_i)\}.
 \label{Bregman2}
\end{align}
This Hessian defines a chosen optimization metric; it is not automatically the Fisher information matrix of a statistical model.

On the positive orthant, when the minimizer is interior, stationarity gives the exact discrete update
\begin{empheq}[box=\fbox]{equation}
 \bw_{t+1}=\exp_{a,b,\lambda}\!\left(\log_{a,b,\lambda}(\bw_t)-\eta_t\bg_t\right).
 \label{f-1fDT}
\end{empheq}
Thus a gradient step is taken in Euler dual coordinates and mapped back by the inverse Euler logarithm. On a constrained set, normal-cone terms must be included; in particular, simplex normalization introduces a common scalar multiplier. Equation~\eqref{f-1fDT} linearizes $L$ and is not an implicit proximal step on the original objective~\cite{MD1,shalev2011}.

The corresponding interior mirror flow, with time scale $\eta_{\rm c}>0$, is
\begin{equation}
 \frac{\mathrm d}{\mathrm dt}\log_{a,b,\lambda}(\bw(t))
 =-\eta_{\rm c}\nabla L(\bw(t)),\qquad
 \dot{\bw}=-\eta_{\rm c}\bH_{a,b,\lambda}(\bw)^{-1}\nabla L(\bw).
 \label{f-1fCT}
\end{equation}
A forward-Euler step in the \emph{primal} flow produces mirrorless mirror descent (MMD)~\cite{MMD2021}:
\begin{equation}
 \bw_{t+1}=\left[\bw_t-\eta_t\diag\{m_{a,b,\lambda}(w_{i,t})\}\bg_t\right]_+,
 \qquad m_{a,b,\lambda}(x)=\frac1{\log_{a,b,\lambda}'(x)}.
 \label{diagMD}
\end{equation}
Here $[x]_+=\max(0,x)$ componentwise. The matrix is a diagonal inverse Hessian, not a reciprocal of a vector derivative. Exact discrete MD and this primal discretization agree to first order for small steps but are generally different algorithms. Clipping is an additional feasibility operation; if it becomes active, interior-flow theory alone does not establish convergence of the resulting iteration. Appendix~\ref{sec:notation-guide} lists these objects together, and the final appendix analyzes the MD--MMD comparison.

\subsection{Ordinary exponentiated gradient as the logarithmic limit}
For $(a,b)=(0,0)$, $\phi_{0,0}(x)=x\ln x-x+1$ and $\log_{0,0}(x)=\ln x$. The flow becomes
\begin{equation}
 \frac{\mathrm d}{\mathrm dt}\ln\bw(t)=-\eta_{\rm c}\nabla L(\bw(t)).
 \label{MDEG1}
\end{equation}
The discrete dual step is ordinary unnormalized exponentiated gradient~\cite{EG}:
\begin{equation}
 \bw_{t+1}=\exp(\ln\bw_t-\eta_t\bg_t)
 =\bw_t\odot\exp(-\eta_t\bg_t).
 \label{EGU}
\end{equation}
Here $\odot$ denotes componentwise multiplication. On the simplex, EG solves, with the Kullback--Leibler (I-)divergence~\cite{Csiszar1975},
\begin{equation}
 \bw_{t+1}=\arg\min_{\bw\in\Delta_N}
 \{\eta_t\langle\bg_t,\bw\rangle+D_{\rm KL}(\bw\Vert\bw_t)\}.
 \label{eq:exact-EG-simplex-problem}
\end{equation}
Its scalar solution and equivalent implementation are
\begin{align}
 w_{i,t+1}&=\frac{w_{i,t}e^{-\eta_tg_{i,t}}}{\sum_jw_{j,t}e^{-\eta_tg_{j,t}}},
 \label{eq:exact-EG-scalar}\\
 \widetilde{\bw}_{t+1}&=\bw_t\odot e^{-\eta_t\bg_t},\qquad
 \bw_{t+1}=\frac{\widetilde{\bw}_{t+1}}{\1^T\widetilde{\bw}_{t+1}}.
 \label{EG}
\end{align}
This radial normalization is exact for ordinary EG because translation of an ordinary exponential's argument becomes a common multiplicative factor~\cite{KW1995,Helmbold98}. Euler exponentials generally lack that property. Consequently, their exact simplex update must solve for the multiplier inside $\exp_{a,b,\lambda}$, as shown next. Choosing $(a,b,\lambda)$ changes the geometry; comparative speed, robustness, and predictive performance still require task-specific evidence.

\subsection{Exact simplex-constrained Euler mirror descent}
\label{NABEG1}

Let
\begin{equation}
F_{a,b,\lambda}(\bw)=\sum_{i=1}^{N}\phi_{a,b,\lambda}(w_i),\qquad
\phi_{a,b,\lambda}'(x)=\log_{a,b,\lambda}(x),
\label{eq:Euler-potential-vector}
\end{equation}

where
\begin{equation}
 \begin{aligned}
 \phi_{a,b,\lambda}(x)&=\alpha_\lambda\Phi_a(x)+\beta_\lambda\Phi_b(x),\\
 \Phi_s(x)&=
 \begin{cases}
 \dfrac{x^{s+1}-1}{s(s+1)}-\dfrac{x-1}{s},&s\notin\{0,-1\},\\
 x\ln x-x+1,&s=0,\\
 x-1-\ln x,&s=-1.
 \end{cases}
 \end{aligned}\label{eq:scaled-potential}
\end{equation}
{In Eq.~\eqref{eq:scaled-potential}, $s$ stands for $a$ or $b$, and $\Phi_s(0)=1/(1+s)$ when $s>-1$. Hence $\phi(0)=\alpha_\lambda/(1+a)+\beta_\lambda/(1+b)$ for $a>-1$. At $a\le-1$ with $a<0<b$, the extended value at zero is $+\infty$. Additive constants preserve the derivative, and affine terms cancel in the Bregman divergence.}

Given $\bg_t=\nabla L(\bw_t)$, the exact constrained mirror step is
\begin{equation}
\bw_{t+1}=\operatorname*{arg\,min}_{\bw\in\Delta_N^{\circ}}
\left\{\eta_t\langle\bg_t,\bw\rangle+D_{F_{a,b,\lambda}}(\bw\Vert\bw_t)\right\}.
\label{eq:exact-Euler-simplex-MD}
\end{equation}
The stationarity conditions give, componentwise,
\begin{align}
\log_{a,b,\lambda}(w_{i,t+1})&=\log_{a,b,\lambda}(w_{i,t})-\eta_t g_{i,t}-\nu_t,\label{eq:Euler-simplex-KKT}\\
w_{i,t+1}&=\exp_{a,b,\lambda}\!\left(\log_{a,b,\lambda}(w_{i,t})-\eta_t g_{i,t}-\nu_t\right),\label{eq:Euler-simplex-exact-update}
\end{align}
where the single scalar $\nu_t$ is chosen from
\begin{equation}
\Phi_t(\nu):=\sum_{i=1}^{N}\exp_{a,b,\lambda}\!\left(s_{i,t}-\nu\right)-1=0,
\qquad s_{i,t}:=\log_{a,b,\lambda}(w_{i,t})-\eta_t g_{i,t}.
\label{eq:lambda-root}
\end{equation}

\begin{theorem}[Well-posedness of the exact simplex step]
\label{thm:Euler-simplex-wellposed}
{Assume $a<0<b$ and fixed $\lambda>0$, so that $\exp_{a,b,\lambda}:\mathbb{R}\to(0,\infty)$ is continuous and strictly increasing. The subproblem in Eq.~\eqref{eq:exact-Euler-simplex-MD} is posed on the positive simplex, so it is well defined even when the extended potential is infinite at zero.} For every $\bw_t\in\Delta_N^{\circ}$, every finite gradient $\bg_t$, and every $\eta_t>0$, Eq.~\eqref{eq:lambda-root} has exactly one solution $\nu_t\in\mathbb{R}$. Consequently Eq.~\eqref{eq:Euler-simplex-exact-update} defines a unique point $\bw_{t+1}\in\Delta_N^{\circ}$.
\end{theorem}
\noindent\textit{The proof is given in Appendix~\ref{app:proofs}, p.~\pageref{prf:thm:Euler-simplex-wellposed}.}

The scalar multiplier can itself be found safely by bisection or safeguarded Newton; unlike post-normalization, this root solve gives the exact constrained Euler mirror step. The bracket in \eqref{eq:audit-multiplier-bound} supplies a finite interval immediately from the current gradient.
The normalized map is also differentiable. For fixed exponents, let $w_i=\exp_{a,b,\lambda}(s_i-\nu)$ with $\sum_iw_i=1$, and put $m_i=1/\log_{a,b,\lambda}'(w_i)>0$. Implicit differentiation gives
\begin{equation}
\frac{\partial\nu}{\partial s_j}=\frac{m_j}{\sum_i m_i},\qquad
\frac{\partial w_i}{\partial s_j}=m_i\delta_{ij}-\frac{m_i m_j}{\sum_k m_k}.
\label{eq:normalized-Euler-Jacobian}
\end{equation}
Thus gradients through the exact normalization can be computed from the converged weights in $\mathcal O(N)$ operations per Jacobian--vector product. The common-shift direction is annihilated, as required by simplex normalization. When the exponents are differentiated as well, their direct inverse-function contributions must also be included in the normalization equation.

\begin{theorem}[Convergence of exact Euler mirror descent]
\label{thm:Euler-MD-convergence}
Assume $-1<a<0<\rev{b\le 1}$, fixed $\lambda>0$, that $L$ is convex on the closed simplex
$\Delta_N$ and differentiable on $\Delta_N^{\circ}$ (which, by
Theorem~\ref{thm:Euler-simplex-wellposed}, contains every iterate
$\bw_t$ generated by the exact update), and that
$\|\nabla L(\bw_t)\|_2\le G$ for all $t$. Let $\bw_1\in\Delta_N^{\circ}$
and let $\bw^*\in\Delta_N$ be any comparator with
$D:=D_{F_{a,b,\lambda}}(\bw^*\Vert\bw_1)<\infty$. For the exact update
\eqref{eq:exact-Euler-simplex-MD} with a constant step size $\eta>0$
and $\bar{\bw}_T=T^{-1}\sum_{t=1}^{T}\bw_t$,
\begin{equation}
L(\bar{\bw}_T)-L(\bw^*)\le \frac{D}{\eta T}+\frac{\eta G^2}{2}.
\label{eq:Euler-MD-bound}
\end{equation}
For $D>0$ and $G>0$, choosing $\eta=\sqrt{2D/(G^2T)}$ gives
\begin{equation}
L(\bar{\bw}_T)-L(\bw^*)\le G\sqrt{\frac{2D}{T}}=\mathcal{O}(T^{-1/2}).
\label{eq:Euler-MD-rate}
\end{equation}

\begingroup
Here $D$ is the initial divergence to the comparator and $G$ bounds the Euclidean gradient norm. The first term of Eq.~\eqref{eq:Euler-MD-bound} comes from telescoping the divergences. The second uses $h_{a,b,\lambda}(w_i)\ge1$ for $0<w_i\le1$ and $b\le1$, which makes the potential $1$-strongly convex in the Euclidean norm on the simplex.\par
\endgroup
\end{theorem}
\noindent\textit{The proof is given in Appendix~\ref{app:proofs}, p.~\pageref{prf:thm:Euler-MD-convergence}.}

\paragraph{Three-parameter simplex curvature and an explicit radius.}
\label{subsec:sharp-simplex-geometry}
The link is normalized by $h_{a,b,\lambda}(1)=1$ for every scale.
Thus comparisons at different $\lambda$ alter curvature across coordinates,
rather than merely multiplying the entire potential by a constant.

\begin{theorem}[Scaled modulus and exact uniform-center radius]
\label{thm:sharp-simplex-geometry}
Let $-1<a<0<b\le1$, $\lambda>0$, and $N\ge2$. Put $p=|a|$,
$r=(p/b)^{2/(p+b)}$, $D_\lambda=p\lambda^{-p}+b\lambda^b$,
and $K=p r^{-p}+b r^b$. Define
\begin{align}
 c_{a,b,\lambda}
 &:=\min_{0<x\le1}\omega_{a,b,\lambda}(x)
 =\begin{cases}1,&\lambda\le r,\\K/D_\lambda,&\lambda>r,\end{cases}
 \label{eq:entropy-comparison-constant}\\
 \mu_{a,b,\lambda}(N)
 &:=\max\{c_{a,b,\lambda},\
 \alpha_\lambda2^{|a|}+\beta_\lambda N^{-b}\}.
 \label{eq:l1-modulus}
\end{align}
Then $F_{a,b,\lambda}$ is $\mu_{a,b,\lambda}(N)$-strongly convex in
$\|\cdot\|_1$ on $\Delta_N$. For $\bx\in\Delta_N$ and
$\by\in\Delta_N^\circ$,
\begin{equation}
 \begin{aligned}
 D_{F_{a,b,\lambda}}(\bx\Vert\by)&\ge
 c_{a,b,\lambda}D_{\rm KL}(\bx\Vert\by),\\
 D_{F_{a,b,\lambda}}(\bx\Vert\by)&\ge
 \frac{\mu_{a,b,\lambda}(N)}2\|\bx-\by\|_1^2.
 \end{aligned}
 \label{eq:Euler-KL-comparison}
\end{equation}
This sufficient modulus exceeds $N^{-b}$ for each fixed interior triple.
The exact uniform-center radius is
\begin{equation}
 \begin{aligned}
 R_{a,b,\lambda}(N)
 &:=\max_{\bx\in\Delta_N}
 D_{F_{a,b,\lambda}}(\bx\Vert\1/N)\\
 &=\alpha_\lambda\frac{N^{|a|}-1}{|a|(1-|a|)}
 +\beta_\lambda\frac{1-N^{-b}}{b(1+b)}.
 \end{aligned}
 \label{eq:bregman-radius}
\end{equation}
It equals $\max F-\min F$ and is attained at every vertex.
For fixed $\lambda$ and $N$, as $(a,b)\to(0,0)$ within the
admissible cone, $R\to\ln N$ and $\mu\to1$.
For fixed $b,\lambda$, as $a\downarrow-1$,
$R\sim (N-1)/[(1+a)(1+b\lambda^{b+1})]$.
\end{theorem}
\noindent\textit{The proof is given in Appendix~\ref{app:proofs}, p.~\pageref{prf:thm:sharp-simplex-geometry}.}
\begin{remark}[Loss weighting and optimization stability]
\label{rem:interpret-sharp-simplex-geometry}
When $\lambda>r$, $c$ is exactly the global minimum gradient weight
$\omega_{\min}$ of Eq.~\eqref{eq:omega-min}; otherwise the minimum over
probabilities is $1$. The same weight enters loss attenuation,
entropy comparison, and clipping-free primal step estimates.
The mixture term in $\mu$ additionally uses the simplex tangent
constraint. These are sufficient moduli, not asserted optimal moduli.
\end{remark}

\begin{corollary}[Dimension-explicit convergence]\label{cor:sharp-rate}
Under Theorem~\ref{thm:Euler-MD-convergence}, if
$\|\nabla L(\bw_t)\|_\infty\le G_\infty$ and $\bw_1=\1/N$,
write $\mu=\mu_{a,b,\lambda}(N)$ and
$D=D_{F_{a,b,\lambda}}(\bw^*\Vert\bw_1)$. Then
\begin{equation}
 L(\bar{\bw}_T)-L(\bw^*)\le\frac{D}{\eta T}
 +\frac{\eta G_\infty^2}{2\mu}.
 \label{eq:sharp-rate}
\end{equation}
For $G_\infty>0$, choosing
$\eta=\sqrt{2\mu R_{a,b,\lambda}(N)/(G_\infty^2T)}$ gives
$G_\infty\sqrt{2R_{a,b,\lambda}(N)/(\mu T)}$.
\end{corollary}
\begin{proof}
Apply the three-point identity and Young's inequality in the
$(\ell_1,\ell_\infty)$ norm pair, telescope the Bregman terms,
and use $D\le R$ and convexity of $L$.
\end{proof}
\begin{remark}[Comparing geometries]\label{rem:rate-comparison}
The diagnostic is $R/\mu$, not $\mu$ alone.
For $N=100$, $a=-1/2$, $b=1/2$, $\lambda=1$, the values are
$R=18.6$, $\mu=1$, compared with $\ln100$ for ordinary entropy.
For fixed interior $a,b,\lambda$, the radius still grows as
$N^{|a|}$. Neither polynomial retention nor a tunable scale
establishes a uniformly better worst-case regret constant.
\end{remark}

\paragraph{An explicit scale for a sufficient regret bound.}
Let
$A=(N^p-1)/(p(1-p))$ and $B=(1-N^{-b})/(b(1+b))$,
where $p=|a|\in(0,1)$. Then $A>B$ and the unique minimizer of
$R_{a,b,\lambda}(N)/c_{a,b,\lambda}$ over $\lambda>0$ is
\begin{equation}
 \lambda_*=
 \left(\frac{p^2 A}{b^2 B}\right)^{1/(p+b)}
 =\left(\frac pb\frac{1+b}{1-p}
 \frac{N^p-1}{1-N^{-b}}\right)^{1/(p+b)}>r.
 \label{eq:scaled-optimal-lambda}
\end{equation}
Indeed on $\lambda\le r$, $c=1$ and $R=\alpha_\lambda A+
\beta_\lambda B$ decreases. On $\lambda>r$,
$R/c=(pA\lambda^{-p}+bB\lambda^b)/K$,
whose derivative changes from negative to positive at $\lambda_*$.
This optimizes a sufficient bound using $c$, not necessarily the
stronger bound using $\mu$ or the exact optimal modulus.
Since the leading optimized regret is proportional to $\sqrt{R/c}$,
percentage reductions in $R/c$ must not be reported as equal percentage
reductions in regret.
For $p=b=s\in(0,1)$,
\[
 \lambda_*=\sqrt N\left(\frac{1+s}{1-s}\right)^{1/(2s)},\qquad
 \frac{R_{\lambda_*}}{c_{\lambda_*}}
 =\frac{N^{s/2}-N^{-s/2}}{s\sqrt{1-s^2}}.
\]
Taking $s=1/\ln N<1$ yields a geometry cost of order $\ln N$;
this is a dimension-dependent choice, not a statement for arbitrary
fixed exponents.

\medskip\noindent\textbf{Polynomial weight floor and tunable concentration.}\
The Euler geometry offers a concrete advantage over ordinary EG when retaining small positive weights is desirable. This is relevant to online allocation and expert weighting, where currently unfavorable coordinates may remain useful later. The following bound quantifies how the deformation controls concentration.

\begin{theorem}[Weight floor of exact Euler simplex mirror descent]
\label{prop:weight-floor}\label{thm:no-extinction}
Assume $a<0<b$, fixed $\lambda>0$, $\bw_1\in\Delta_N^\circ$, $\eta_t>0$, and $\|\bg_t\|_\infty\le G$. Set $A_T=\sum_{t=1}^T\eta_t$ and
$C_1=\max_j\log_{a,b,\lambda}(w_{j,1})-\min_i\log_{a,b,\lambda}(w_{i,1})$.
Then every exact simplex iterate satisfies
\begin{equation}
w_{i,T+1}\ge B_T:=\exp_{a,b,\lambda}\!\left(\log_{a,b,\lambda}(1/N)-C_1-2GA_T\right).
\label{eq:weight-floor}
\end{equation}
If $G>0$ and $A_T\to\infty$, the lower bound has the asymptotic form
\begin{equation}
B_T\sim\left[\frac{2G|a|}{\alpha_\lambda}A_T\right]^{-1/|a|}.
\label{eq:weight-floor-asymptotic}
\end{equation}
For constant step size, $A_T=\eta T$. Ordinary entropic EG gives the corresponding exponential lower bound $N^{-1}\exp(-C_1-2GA_T)$, and exponential decay is attainable under bounded constant gradients.
\end{theorem}
\noindent\textit{The proof is given in Appendix~\ref{app:proofs}, p.~\pageref{prf:prop:weight-floor}.}
\begin{remark}[Interpretation and potential use]
The floor quantifies retention under bounded updates in exact arithmetic. It is potentially useful when a currently weak action or expert may become valuable later. It does not guarantee a large enough sampling probability, fast recovery, or protection from floating-point underflow; those are separate design and evaluation questions.
\end{remark}

For a reproducible illustration, take $N=4$, $(a,b,\lambda)=(-0.5,0.5,1)$, $\eta=0.5$, and $\bg_t=\bolde_1$. At $T=1000$, the exact Euler iterate has $w_1\approx3.98\times10^{-6}$ and the lower bound is $9.97\times10^{-7}$. Ordinary EG gives $w_1\approx2.37\times10^{-218}$. The Euler floor therefore preserves a much larger representable weight in this example. Log-domain implementations also allow EG to retain very small weights; the distinction established here is their polynomial versus exponential scale. The floor supplies a mechanism for controlling concentration, while recovery after a change of environment also depends on subsequent gradients and step sizes.

The one-parameter Amari family also has a polynomial floor, but no free
transition scale. With the full family one may first choose $a$ for
the floor exponent and $b$ for the upper-tail elasticity, and then tune
curvature at a reference coordinate using $\lambda$.
For a target $\varepsilon_1\in(1-b,1-a)$, let
$\chi_*=(\varepsilon_1-(1-b))/(b-a)\in(0,1)$. Then
\begin{equation}
 \text{floor exponent}=\frac1{|a|},\qquad
 \lambda=\left(\frac{|a|}{b}\frac{1-\chi_*}{\chi_*}\right)^{1/(b-a)}.
 \label{eq:floor-curvature-design}
\end{equation}
At $\lambda=1$ this reduces to $\varepsilon_1=1-a-b$.
For curvature prescribed at $x_*>0$, divide the displayed scale by $x_*$.
The exponent of the polynomial floor remains unchanged, but its
prefactor involves $\alpha_\lambda$ and must be assessed as well.

For implementation, the exact-step multiplier obeys
\begin{equation}\label{eq:audit-multiplier-bound}
-\eta_t\max_i g_{i,t}\le\nu_t\le-\eta_t\min_i g_{i,t}.
\end{equation}
The normalization sum is at least one at the left endpoint and at most one at the right endpoint, giving an immediate bracket. The preceding floor also sharpens by replacing $2GA_T$ with $\sum_{t=1}^T\eta_t(\max_i g_{i,t}-\min_i g_{i,t})$, as is clear from its pairwise proof. For unnormalized MD, direct telescoping gives $w_{i,T+1}\ge \exp_{a,b,\lambda}(\log_{a,b,\lambda}(w_{i,1})-\sum_t\eta_t|g_{i,t}|)$.

\medskip\noindent\textbf{Boundary regime.}\ If $a=0<b$, the mirror map has range $(-1/b,\infty)$ rather than all of $\mathbb{R}$. A dual proposal can therefore leave the range and an interior inverse-map formula need not describe a boundary optimum. In that regime one must specify an active-set/KKT treatment, a generalized inverse, or an explicit clipping rule. The full-range theorem above is deliberately not claimed for such finite-range clipped variants.

\medskip\noindent\textbf{Radial normalization.}\ A cheaper alternative to the exact step divides an unconstrained mirror step by its sum, as ordinary exponentiated gradient does. For Euler potentials this is a different algorithm: it follows a different continuous-time flow and generally converges only to stationary points under additional assumptions. Its analysis, together with the related clipped variable-metric update, is collected in Appendix~\ref{app:MMD-convergence}. { In the logarithmic limit, exact dual mirror descent and its radial normalization reduce to exponentiated gradient~\eqref{EGU}. The clipped primal variable-metric Euler step instead uses $[w_i(1-\eta g_i)]_+$ before normalization and agrees only to first order in the step size.}

\subsection{Certified normalization: the shared primitive}
\label{sec:FTRL}
Every simplex algorithm in this section, and in the rest of the paper, needs the
multiplier $\nu$ that restores unit mass. The cheap shortcut of ordinary
exponentiated gradient, dividing by the sum, is not available, because
$\exp_{a,b,\lambda}$ does not turn an additive shift into a common factor.
\paragraph{The normalization oracle and its certificate.}
\label{subsec:FTRL-oracle}
Every exact simplex update below---FTRL, mirror descent, optimistic, policy and
value updates, Mirror-Prox---solves, for scores $\by\in\mathbb R^N$ and
positive weights $D_i$ (default $D_i=1$), the scalar equation
\begin{equation}
 S(\nu):=\sum_{i=1}^N\exp_{a,b,\lambda}\bigl((y_i-\nu)/D_i\bigr)=1,\qquad
 w_i=\exp_{a,b,\lambda}\bigl((y_i-\nu)/D_i\bigr).
 \label{eq:oracle-equation}
\end{equation}
$S$ is continuous and strictly decreasing with
$S'(\nu)=-\sum_im_{a,b,\lambda}(w_i)/D_i$. For $D_i\equiv1$,
Eq.~\eqref{eq:weighted-normalization} and Eq.~\eqref{eq:Euler-softmax-bracket}
give the certified bracket
\begin{equation}
 \nu^*\in[\nu_{\rm lo},\nu_{\rm hi}],\qquad
 \nu_{\rm lo}=\max\Bigl\{\max_iy_i,\ \min_iy_i-\log_{a,b,\lambda}(\tfrac1N)\Bigr\},\quad
 \nu_{\rm hi}=\max_iy_i-\log_{a,b,\lambda}(\tfrac1N),
 \label{eq:oracle-bracket}
\end{equation}
of width $\min\{\operatorname{range}(\by),\,-\log_{a,b,\lambda}(1/N)\}$, invariant under
adding a common constant to the scores. In plain words, Algorithm~\ref{alg:euler-norm}
guesses the single unknown $\nu$, adds up the weights, moves $\nu$ up if the
total exceeds one and down otherwise, and accepts a Newton step only when it
lands safely inside the shrinking bracket. {Before stopping, the mass defect is assigned to a largest weight, which has the smallest scalar link curvature among the uncorrected weights for $b\le1$. The weighted objective has coordinate curvature $D_i h_{a,b,\lambda}(w_i)$. The solver reports a dual residual for the resulting feasible vector.}

\begin{algorithm}[ht]
\caption{{\textsc{EulerNorm} with scores, weights and tolerance}: weighted simplex normalization with a dual-residual certificate}
\label{alg:euler-norm}
\begingroup\small
\begin{algorithmic}[1]
\Require Scores $\by\in\mathbb R^N$, $N\ge1$; weights $D_i>0$ (default one); $a\le0<b\le1$, $\lambda>0$; tolerance $\epsilon>0$; safeguard $0<\theta<1/2$
\Ensure $(\bp,\nu,\hat\epsilon)$, with $\bp\in\Delta_N$ and centered dual residual at most $\epsilon$, or \/\textsc{precision-limit} in rounded computation
\State Use $E(u)=\exp_{a,b,\lambda}(u)$ for $a<0$ and $E(u)=[1+bu]_+^{1/b}$ for $a=0$.
\If{$N=1$}
 \State \Return $((1),y_1,0)$
\EndIf
\State $\ell\gets\max\{\max_i y_i,\ \min_i(y_i-D_i \log_{a,b,\lambda}(1/N))\}$; $u\gets\max_i(y_i-D_i \log_{a,b,\lambda}(1/N))$; $\nu\gets(\ell+u)/2$.
\Loop
 \State $w_i\gets E((y_i-\nu)/D_i)$ for all $i$; $S\gets\sum_i w_i$.
 \State Choose the first largest weight $j$; set $\bp\gets\bw+(1-S)\bolde_j$.
 \If{$p_j>0$}
  \State $\hat\epsilon\gets D_j| \log_{a,b,\lambda}(p_j)-\log_{a,b,\lambda}(w_j)|/2$.
  \State If $\hat\epsilon\le\epsilon$, \Return $(\bp,\nu,\hat\epsilon)$.
 \EndIf
 \State If $S>1$ set $\ell\gets\nu$; otherwise set $u\gets\nu$.
 \State $\nu_N\gets\nu+\frac{S-1}{\sum_{i:w_i>0}\rev{m_{a,b,\lambda}(w_i)}/D_i}$; $d\gets\theta(u-\ell)$.
 \State Use $\nu\gets\nu_N$ if $\nu_N\in[\ell+d,u-d]$; otherwise use $\nu\gets(\ell+u)/2$.
\EndLoop
\Guarantee The loop above uses exact arithmetic. With rounded inverses enclose $S$, check feasibility, and use the full residual in Eq.~\eqref{eq:normalization-full-residual}. Refine if the sign or certificate is unresolved; return \textsc{precision-limit} if refinement cannot certify the decision.
\end{algorithmic}
\endgroup
\end{algorithm}

{At $a=0$, the solver evaluates the positive-part KKT extension. When $b=1$, the Newton proposal uses the slope from the positive coordinates at a threshold. Bracket updates still use the sign of the exact mass defect.} Each sign update contracts the bracket by at least $1-\theta$. {In the logarithmic limit, equal weights $D_i=D$ give ordinary softmax of $\by/D$, with $\nu=D\ln\sum_i\exp(y_i/D)$. For unequal weights the common multiplier still solves $\sum_i\exp((y_i-\nu)/D_i)=1$ and generally requires a scalar solve.}

\begin{lemma}[Certified normalization]
\label{lem:certified-norm}
\begingroup
Let $N\ge2$, $a\le0<b\le1$, $D_i>0$, $L_D=\sum_iD_i^{-1}$ and $D_{\max}=\max_iD_i$. Write $\nu^*$ for the root of Eq.~\eqref{eq:oracle-equation}, using the KKT extension for $a=0$. On the bracket of Algorithm~\ref{alg:euler-norm}, let $w_i$ be the evaluated weights, $S=\sum_iw_i$, $j$ any largest-weight index and $\bp=\bw+(1-S)\bolde_j$. Then
\begin{enumerate}[label=(\roman*),leftmargin=2.2em]
\item $|1-S|\le L_D|\nu-\nu^*|$. If $L_D|\nu-\nu^*|\le1/(2N)$, then $\bp\in\Delta_N$ with $p_j\ge1/(2N)$.
\item Whenever $p_j>0$, the feasible candidate has a stationarity residual of at most $\hat\epsilon=\tfrac12 D_j|\log_{a,b,\lambda}(p_j)-\log_{a,b,\lambda}(w_j)|$ for the objective $\sum_iD_i\phi_{a,b,\lambda}(p_i)-\langle\by,\bp\rangle$. At $a=0$ each zero coordinate satisfies $y_i-\nu\le-D_i/b$.
\item Under the distance condition in (i),
$\hat\epsilon\le\tfrac12D_{\max}(2N)^{1-a}L_D|\nu-\nu^*|$.
\end{enumerate}
In exact arithmetic, for a fixed safeguard fraction, the algorithm terminates after $O(1+\log(1+W/\rho))$ iterations, where $W$ is the initial bracket width and
\[
 \rho=\min\left\{\frac1{2NL_D},\,
 \frac{2\epsilon}{D_{\max}(2N)^{1-a}L_D}\right\}.
\]
Once the bracket width is at most $\rho$, every trial in the bracket is feasible after the repair and passes the residual test.\par
\endgroup
\end{lemma}
\noindent\textit{The proof is given in Appendix~\ref{app:proofs}, p.~\pageref{prf:lem:certified-norm}.}
{If the weighted potential is $\mu_D$-strongly convex in $\ell_1$ on the simplex, the returned distribution satisfies $\|\bp-\bp^*\|_1\le\hat\epsilon/\mu_D$. One may use $\mu_D=(\min_i D_i)\mu$ whenever the unweighted potential has modulus $\mu$. Strong monotonicity gives this conversion from the centered dual residual, as in Lemma~\ref{lem:stage2-oracle-error}.}
Dividing an approximate output by its sum is not a valid repair: the dual
perturbation $\log_{a,b,\lambda}(w_i/S)-\log_{a,b,\lambda}(w_i)$ is unbounded as $w_i\downarrow0$
when $a<0$.

For inexact inner solves, a full dual-residual certificate
(Eq.~\eqref{eq:normalization-full-residual}) and a worked example are given in
Appendix~\ref{subsec:supp-oracle}.

\subsection{Euler-Hedge FTRL and the optimal scale}
\label{sec:FTRL-3p}
In online convex optimization the learner plays $\bw_t\in\Delta_N$, then
observes a loss gradient $\bg_t$, and is judged by its regret
$\mathrm{Reg}_T(\bx)=\sum_{t\le T}\langle\bg_t,\bw_t-\bx\rangle$ against a fixed
comparator $\bx$. On the simplex only coordinate differences matter, so we use
the half-range and the centered gradient
\begin{equation}
 \sigma_t=\tfrac12\bigl(\max_ig_{i,t}-\min_ig_{i,t}\bigr),\qquad
 \bar\bg_t=\bg_t-\tfrac12\bigl(\max_ig_{i,t}+\min_ig_{i,t}\bigr)\1 ,
 \label{eq:centered-online-gradients}
\end{equation}
and write $\operatorname{hr}$ and $\operatorname{ctr}$ for these two maps. For
losses $(100,101,100)$, $\|\bg_t\|_\infty=101$ but $\sigma_t=1/2$. Throughout
$\mu=\mu_{a,b,\lambda}(N)$ and $R=R_{a,b,\lambda}(N)$ are the constants of
Theorem~\ref{thm:sharp-simplex-geometry}.

\label{subsec:FTRL-exact}
\begin{definition}[Euler FTRL]
\label{def:Euler-FTRL}
For a nonincreasing positive sequence $\eta_t$, let $\bw_1=\1/N$ and
\begin{equation}
 \bw_{t+1}=\operatorname*{arg\,min}_{\bw\in\Delta_N}
 \Bigl\{\Bigl\langle\sum_{s\le t}\bar\bg_s,\bw\Bigr\rangle
 +\frac{F_{a,b,\lambda}(\bw)}{\eta_t}\Bigr\},\qquad
 w_{i,t+1}=\exp_{a,b,\lambda}(-\eta_th_{i,t}-\nu_t),
 \label{eq:Euler-FTRL}
\end{equation}
{where $\bh_t=\sum_{s\le t}\bar\bg_s$ accumulates the centered gradients, and $\nu_t$ enforces unit mass. The inverse formula uses the full-range regime or the positive-part KKT extension at $a=0$.}
\end{definition}
\begin{equation}
 \sum_i\exp_{a,b,\lambda}(-\eta_th_{i,t}-\nu_t)=1
 \label{eq:FTRL-scalar-solve}
\end{equation}
is one call of Algorithm~\ref{alg:euler-norm}; we write
$\mathcal N_{a,b,\lambda}(\by)=\argmin_{\bp\in\Delta_N}\{F_{a,b,\lambda}(\bp)-\langle\by,\bp\rangle\}$
for the exact decision, so that $\bw_{t+1}=\mathcal N_{a,b,\lambda}(-\eta_t\bh_t)$.

\begin{lemma}[Fixed-step equivalence and adaptive dual stabilization]
\label{lem:audit-exact-ftrl}
With uniform initialization and constant $\eta$, exact Euler simplex MD and
Eq.~\eqref{eq:Euler-FTRL} generate identical iterates; for an arbitrary interior
$\bw_1$ the same holds with regularizer $D_{F_{a,b,\lambda}}(\cdot\Vert\bw_1)$.
For that anchored regularizer and varying steps, with $r_t=\eta_t/\eta_{t-1}\in(0,1]$,
\begin{equation}
 \nabla F_{a,b,\lambda}(\bw_{t+1})=r_t\nabla F_{a,b,\lambda}(\bw_t)
 +(1-r_t)\nabla F_{a,b,\lambda}(\bw_1)-\eta_t\bar\bg_t-c_t\1,\qquad t\ge2,
 \label{eq:audit-variable-ftrl}
\end{equation}
where $c_t$ enforces unit mass and the first step is the anchored mirror step
with rate $\eta_1$.
\end{lemma}
\begin{proof}
Summing the exact MD KKT equations at constant step gives the anchored FTRL
optimality equations; for varying steps, multiply the FTRL equations at $t-1$
by $r_t$ and subtract. Uniqueness identifies the iterates.
\end{proof}
With a decreasing rate, the factor $r_t\le1$ shrinks the dual displacement from
the anchor before the new gradient step, so the state is exactly what FTRL with
the new rate produces. This is an Euler instance of dual
stabilization~\cite{Fang2022}; Section~\ref{subsec:p2-dual-damping} gives its
continuous-time form.

\begin{theorem}[Fixed-step Euler FTRL regret]
\label{thm:Euler-FTRL-regret}
For any $\eta>0$, finite gradient stream and $\bx\in\Delta_N$,
$\mathrm{Reg}_T(\bx)\le R/\eta+(\eta/2\mu)\sum_{t\le T}\sigma_t^2$. If
$\sigma_t\le G$, $G>0$, the rate $\eta=\sqrt{2\mu R/(G^2T)}$ gives
$\mathrm{Reg}_T(\bx)\le G\sqrt{2RT/\mu}$; the logarithmic limit recovers
$G\sqrt{2T\ln N}$.
\end{theorem}
\noindent\textit{The proof is given in Appendix~\ref{app:proofs}, p.~\pageref{prf:thm:Euler-FTRL-regret}.}
{The first term is the comparator's regularization cost and the second bounds the cost of changing the decision. After optimizing the step size, the geometry enters through $R/\mu$. Corollary~\ref{cor:ftrl3-regret} optimizes the explicit alternative $R/c$, where $c\le\mu$.}

\paragraph{What the free scale changes in a regret bound.}
\label{subsec:lambda-improvements}
In this subsection $0<p<1$, $0<q\le1$ and $N\ge2$; put
$r=(p/q)^{2/(p+q)}$. Since
$\omega_{a,b,\lambda}'=h_{a,b,\lambda}(1-\varepsilon_{a,b,\lambda})$, the
crossing $\varepsilon_{a,b,\lambda}=1$ and the minimum of the gradient weight
are the \emph{same point}, $w_{\min}(\lambda)=r/\lambda$
(Proposition~\ref{prop:omega-profile}). It equals $1/\lambda$ only when $p=q$,
and placing it at $1/N$ requires $\lambda=Nr$, which is a different objective
from optimizing regret.

\begin{corollary}[Regret of fixed-step Euler FTRL and the optimal scale]
\label{cor:ftrl3-regret}
Let $c_{a,b,\lambda}$ and $R_{a,b,\lambda}(N)$ be the sufficient modulus and
uniform-center radius of Theorem~\ref{thm:sharp-simplex-geometry}. For
fixed-step FTRL initialized at $\1/N$, or the equivalent fixed-step simplex MD
(Section~\ref{subsubsec:ftrl3-hedge}), and convex losses with
$\|\bg_t\|_\infty\le G$, $G>0$,
\begin{equation}
 \operatorname{Regret}_T(\bu)\le\frac{R_{a,b,\lambda}(N)}{\eta}
 +\frac{\eta}{2c_{a,b,\lambda}}\sum_{t=1}^T\|\bg_t\|_\infty^2
 \le\frac{R_{a,b,\lambda}(N)}{\eta}+\frac{\eta G^2T}{2c_{a,b,\lambda}}.
 \label{eq:verified-regret}
\end{equation}
The choice $\eta=\sqrt{2c_{a,b,\lambda}R_{a,b,\lambda}(N)/(G^2T)}$ gives
$G\sqrt{2TR_{a,b,\lambda}(N)/c_{a,b,\lambda}}$, and the ratio $R/c$ is
minimized over $\lambda$ uniquely at $\lambda_*$ of
Eq.~\eqref{eq:scaled-optimal-lambda}. The modulus $c_{a,b,\lambda}$ may be
replaced by the larger $\mu_{a,b,\lambda}(N)$.
\end{corollary}
\begin{proof}
This is the standard strongly convex FTRL/MD bound~\cite{McMahan2017} in the
$(\ell_1,\ell_\infty)$ pair, with
$D_{F_{a,b,\lambda}}(\bu\Vert\1/N)=F_{a,b,\lambda}(\bu)-F_{a,b,\lambda}(\1/N)
\le R_{a,b,\lambda}(N)$ and the modulus of
Theorem~\ref{thm:sharp-simplex-geometry}.
\end{proof}
Improvements in $R/c$ enter the regret constant only through a square root,
and $c_{a,b,\lambda}$ is nonincreasing in $\lambda$: scale tuning balances
range against curvature rather than improving strong convexity. For
$p=q=s\in(0,1)$, Eq.~\eqref{eq:scaled-optimal-lambda} gives
$\lambda_*=\sqrt N[(1+s)/(1-s)]^{1/(2s)}$ and
$R_{\lambda_*}/c_{\lambda_*}=(N^{s/2}-N^{-s/2})/(s\sqrt{1-s^2})$. For the
dimension-dependent choice $s=1/\ln N<1$, $\lambda_*\sim e\sqrt N$ and
\[
 \frac{R_{\lambda_*}}{c_{\lambda_*}}
 =\frac{2\sinh(1/2)\,\ln N}{\sqrt{1-(\ln N)^{-2}}},
\]
which proves an $\mathcal O(\sqrt{T\ln N})$ bound within the factor
$\sqrt{2\sinh(1/2)}\approx1.02$ of entropy asymptotically; arbitrary fixed
exponents need not give this rate. Table~\ref{tab:near-optimal-lambda}
compares this choice with the unscaled one.

\begin{table}[ht]
\centering\small
\caption{Dimension-dependent exponents $|a|=b=1/\ln N$: geometry constant
$R_{a,b,\lambda}(N)/\mu_{a,b,\lambda}(N)$ relative to entropy, and retained
weight at dual value $-50$, for $\lambda=1$ and for the scale $\lambda_*$ of
Eq.~\eqref{eq:scaled-optimal-lambda}. For comparison $e^{-50}\approx1.9\times10^{-22}$. The smaller geometry constant in each row is in bold.}
\label{tab:near-optimal-lambda}
\begin{tabular}{rcccccc}
\hline
 & & \multicolumn{2}{c}{$\lambda=1$} & \multicolumn{3}{c}{$\lambda=\lambda_*$}\\
$N$ & $\ln N$ & $R/(\mu\ln N)$ & $\exp_{a,b,1}(-50)$ & $\lambda_*$ & $R/(\mu\ln N)$ & $\exp_{a,b,\lambda_*}(-50)$\\\hline
$10$ & $2.30$ & $1.74$ & $1.7\times10^{-4}$ & $9.23$ & $\mathbf{1.16}$ & $7.8\times10^{-6}$\\
$10^2$ & $4.61$ & $1.36$ & $6.9\times10^{-7}$ & $27.6$ & $\mathbf{1.07}$ & $1.1\times10^{-8}$\\
$10^3$ & $6.91$ & $1.28$ & $9.3\times10^{-9}$ & $86.6$ & $\mathbf{1.05}$ & $4.9\times10^{-11}$\\
$10^4$ & $9.21$ & $1.25$ & $2.7\times10^{-10}$ & $273$ & $\mathbf{1.05}$ & $5.2\times10^{-13}$\\
$10^6$ & $13.8$ & $1.22$ & $1.0\times10^{-12}$ & $2723$ & $\mathbf{1.05}$ & $3.7\times10^{-16}$\\
\hline
\end{tabular}
\end{table}
The scale therefore trades part of the polynomial retention at a fixed dual
value for a geometry constant within $5$--$16\%$ of entropy; the retained
weights remain many orders of magnitude above the exponential value.

\subsection{Adaptive (AdaGrad-type) Euler FTRL}
\label{subsec:FTRL-adaptive}
The natural adaptive rate $\eta_t=\eta_0/\sqrt{\delta+S_t}$ faces one
difficulty: FTRL chooses $\bw_t$ before seeing $\bg_t$, so the curvature bound
$\eta_{t-1}\sigma_t^2/(2\mu)$ can be arbitrarily large after a jump. The
stability cost of a round, however, never exceeds $2\sigma_t$; taking the
smaller bound in every round (range capping) removes all assumptions on
gradient size.

\begin{theorem}[Adaptive Euler FTRL with range-capped stability]
\label{thm:Euler-adaptive}
Let $S_t=\sum_{s\le t}\sigma_s^2$, $\eta_0>0$, $\delta\ge0$,
$\eta_t=\eta_0/\sqrt{\delta+S_t}$, use Eq.~\eqref{eq:Euler-FTRL} with exact
decisions, and keep the uniform iterate while $\delta+S_t=0$. For every finite
gradient stream,
\begin{align}
 \mathrm{Reg}_T(\bx)&\le\frac{R}{\eta_0}\sqrt{\delta+S_T}
 +\sum_{t=1}^T\min\Bigl\{\frac{\eta_0\sigma_t^2}{2\mu\sqrt{\delta+S_{t-1}}},\,2\sigma_t\Bigr\}
 \label{eq:audit-adaptive-predictable}\\
 &\le\frac{R}{\eta_0}\sqrt{\delta+S_T}+C_{\eta_0}\bigl(\sqrt{\delta+S_T}-\sqrt\delta\bigr),
 \qquad C_{\eta_0}=\frac{\eta_0}{2\mu}+\sqrt{\frac{\eta_0^2}{4\mu^2}+4}.
 \label{eq:Euler-adaptive-bound}
\end{align}

\begingroup
Here $S_0=0$ and $\bw_t$ uses $S_{t-1}$ before the current gradient is revealed. If $\delta+S_{t-1}=0$, interpret the minimum as $2\sigma_t$, including zero when $\sigma_t=0$.\par
\endgroup
With $\eta_0=\sqrt{2\mu R}$,
\begin{equation}
 \mathrm{Reg}_T(\bx)\le\Bigl(\sqrt{2R/\mu}+\sqrt{R/(2\mu)+4}\Bigr)\sqrt{\delta+S_T}.
 \label{eq:scale-free-Euler-bound}
\end{equation}
For $\delta=0$ the iterates are invariant under positive rescaling of the whole
gradient stream and under common-coordinate offsets in every round.
\end{theorem}
\noindent\textit{The proof is given in Appendix~\ref{app:proofs}, p.~\pageref{prf:thm:Euler-adaptive}.}
The regret grows with the accumulated squared half-ranges rather than with
$\sqrt T$ times a worst-case bound: it is $o(\sqrt T)$ when $S_T=o(T)$, and
individual gradients may be arbitrarily large. {With $\delta=0$ the iterates are invariant to the units of the losses while the geometry still has its own scale $\lambda$. When losses are multiplied by $c>0$ and $\delta>0$, replacing $\delta$ by $c^2\delta$ preserves the same decisions.}

\begin{corollary}[A scale-independent coordinate floor]
\label{cor:adaptive-FTRL-floor}
Under Theorem~\ref{thm:Euler-adaptive}, every coordinate satisfies
\begin{equation}
 w_{i,t+1}\ge \exp_{a,b,\lambda}\bigl(\log_{a,b,\lambda}(1/N)-2\eta_0\sqrt t\bigr)
 \sim\Bigl[\frac{2\eta_0|a|\sqrt t}{\alpha_\lambda}\Bigr]^{-1/|a|}>0 .
 \label{eq:adaptive-FTRL-floor}
\end{equation}
\end{corollary}
\begin{proof}
Coordinate differences of the KKT equations are at most
\[
 \eta_t\operatorname{range}(\bh_t)\le2\eta_0\sum_{s\le t}\sigma_s/\sqrt{\delta+S_t}\le2\eta_0\sqrt t
\]
by Cauchy--Schwarz, and some coordinate is at least $1/N$.
The asymptotic is the left tail of Proposition~\ref{prop:euler-tails}.
\end{proof}
Entropic FTRL gives only $N^{-1}\exp(-2\eta_0\sqrt t)$. The floor is a lower
bound, not a trajectory order, and does not by itself guarantee fast recovery.

{When the played decision $\bw_t$ is computed with certified tolerance $\epsilon_t$ in the score-normalized objective of Eq.~\eqref{eq:oracle-equation}, round $t$ adds at most} $\operatorname{hr}(\bg_t)\min\{2,\epsilon_t/\mu\}$ to the regret
(Lemma~\ref{lem:stage2-oracle-error} in Appendix~\ref{subsec:supp-oracle}).

\begin{algorithm}[ht]
\caption{\textsc{AdaEulerFTRL}: adaptive Euler FTRL with a rate set by past loss ranges}
\label{alg:adaptive-ftrl}
\begingroup\small
\begin{algorithmic}[1]
\Require $N\ge2$; $-1<a<0<\rev{b\le 1}$, $\lambda>0$; $\eta_0>0$; offset $\delta\ge0$; online loss vectors $\bg_t$; optional played-round tolerances $\epsilon_t>0$
\Ensure Played decisions $\bw_t\in\Delta_N$ with optional solver certificates
\State $\bh\gets\0$; $S\gets0$.
\For{$t=1,2,\ldots$}
 \If{$\delta+S=0$}
  \State $\bw_t\gets\1/N$.
 \Else
  \State $\eta\gets\eta_0/\sqrt{\delta+S}$; $\by\gets-\eta\bh$.
  \State Set $\bw_t\gets\mathcal N_{a,b,\lambda}(\by)$, or use $\mathsf{Norm}(\by;\epsilon_t)$.
  \State If the numerical call fails, return the played prefix with \textsc{precision-limit}.
 \EndIf
 \State Play $\bw_t$, then observe $\bg_t$.
 \State $\bh\gets\bh+\operatorname{ctr}(\bg_t)$; $S\gets S+\operatorname{hr}(\bg_t)^2$.
\EndFor
\Guarantee Theorem~\ref{thm:Euler-adaptive}; a numerical decision adds the round-$t$ cost of Lemma~\ref{lem:stage2-oracle-error}. The optional default $\eta_0=\sqrt{2\mu R}$ uses the simplex constants in Theorem~\ref{thm:sharp-simplex-geometry}.
\end{algorithmic}
\endgroup
\end{algorithm}
Hedge is the limit $(a,b)\to(0,0)$, where Algorithm~\ref{alg:euler-norm} becomes
ordinary softmax. The exponents and scale change the normalization map, the
geometry constants and hence the default $\eta_0$; for $N=1$ play $(1)$.

\paragraph{A guaranteed minimum probability.} If an application prescribes a fixed minimum probability $\epsilon$, adaptive mirror descent on the floored simplex $K_\epsilon$ achieves regret $2\sqrt{B_\epsilon(\delta+S_T)/\mu}$, where $B_\epsilon$ is the Bregman diameter of $K_\epsilon$ (Theorem~\ref{thm:adaptive-Euler-OMD}, Appendix~\ref{subsec:FTRL-floor}). Full-simplex FTRL needs no floor and pays only the radius $R$; OMD on $K_\epsilon$ pays for the floor twice, through $B_\epsilon$ and a comparator bias.

\subsection{FTRL-Proximal, RDA and exact sparsity}
\label{sec:new-da}\label{sec:recent-euler}
Sparse online learning is where FTRL has had its largest industrial impact.
Exact zeros require a potential with \emph{finite} slope at zero, which the
full-range Euler link does not have. Three constructions restore it: shifting
the argument (FTRL-Proximal on the nonnegative orthant), symmetrizing a shifted
potential (signed RDA), and moving to the boundary $a=0$ (sparse simplex
FTRL). In each, the Euler exponents still shape the \emph{active} weights.
\subsubsection{Euler FTRL-Proximal and composite mirror descent}
\label{subsubsec:ftrl3-shifted}
FTRL-Proximal with per-coordinate adaptive rates~\cite{McMahan2011,McMahan2013}
is the version of FTRL with the largest practical impact, notably in
large-scale click-through-rate prediction, where it combines AdaGrad-style
learning rates with exact sparsity and small memory over billions of features.
Its Euler analogue follows.
To obtain exact coordinate sparsity together with Bregman stabilization on
$\mathbb R^N_{\ge0}$, use the shifted link $w\mapsto\log_{a,b,\lambda}(1+w)$,
which vanishes at $w=0$ with unit slope $h_{a,b,\lambda}(1)=1$, and its potential
$\tilde F_{a,b,\lambda}(\bw)=\sum_i\int_0^{w_i}\log_{a,b,\lambda}(1+r)\,\mathrm dr$. For signed
variables, $F_{\rm sym}(\bw)=\sum_i\phi_{a,b,\lambda}(1+|w_i|)$ is convex and
differentiable with derivative $\operatorname{sign}(w_i)\log_{a,b,\lambda}(1+|w_i|)$;
Section~\ref{subsec:signed-rda} uses a general shift $\varrho$. These are Euler
analogues of composite FTRL~\cite{McMahan2011}. With $\ell_1$ and potential
weights $\rho_1,\rho_2\ge0$, FTRL-Proximal solves
\begin{equation}
 \bw_{t+1}^{\text{FTRL-Prox}}=\argmin_{\bw\in\mathbb R^N_{\ge0}}
 \Bigl\{\langle\bg_{1:t},\bw\rangle+t\rho_1\|\bw\|_1+\rho_2\tilde F_{a,b,\lambda}(\bw)
 +\sum_{s=1}^t\sigma_sD_{\tilde F_{a,b,\lambda}}(\bw\Vert\bw_s)\Bigr\}.
 \label{eq:ftrl-prox}
\end{equation}

\begingroup
Here $\bg_{1:t}=\sum_{s=1}^t\bg_s$ and the nonnegative $\sigma_s$ are proximal regularizer increments, distinct from the half-ranges in Eq.~\eqref{eq:centered-online-gradients}. For nonincreasing positive $\eta_t$, set $\sigma_1=\eta_1^{-1}$ and $\sigma_s=\eta_s^{-1}-\eta_{s-1}^{-1}$ for $s\ge2$. Their cumulative sum is $\eta_t^{-1}$, so $\rho_2+\eta_t^{-1}$ is the coefficient multiplying the coordinate link in the optimality condition.\par
\endgroup
With the cumulative proximal state
$z_{t,i}=g_{1:t,i}-\sum_{s\le t}\sigma_s\log_{a,b,\lambda}(1+w_{s,i})$, the exact solution is
the inverse-link thresholding update
\begin{equation}
{
 w_{t+1,i}^{\text{FTRL-Prox}}=
 \exp_{a,b,\lambda}\Bigl(\frac{[-z_{t,i}-t\rho_1]_+}{\rho_2+\eta_t^{-1}}\Bigr)-1,
 \qquad w_{t+1,i}^{\text{FTRL-Prox}}=0\iff -z_{t,i}\le t\rho_1 .}
 \label{eq:euler-ftrl-prox}
\end{equation}
Composite MD (FOBOS) applies the penalty only to the single-step update,
\begin{equation}
 \bw_{t+1}^{\text{FOBOS}}=\argmin_{\bw\in\mathbb R^N_{\ge0}}
 \Bigl\{\langle\bg_t,\bw\rangle+\rho_1\|\bw\|_1+\rho_2\tilde F_{a,b,\lambda}(\bw)
 +\tfrac1{\eta_t}D_{\tilde F_{a,b,\lambda}}(\bw\Vert\bw_t)\Bigr\},
 \label{eq:fobos}
\end{equation}
{with solution
$w_{t+1,i}=\exp_{a,b,\lambda}\bigl([\eta_t^{-1}\log_{a,b,\lambda}(1+w_{t,i})-g_{t,i}-\rho_1]_+
/(\rho_2+\eta_t^{-1})\bigr)-1$. In both updates the positive part is applied to the evidence before inversion. This tests the KKT inequality at zero and keeps the inverse input in range even when $a=0$.} Because FTRL-Proximal tests the
\emph{cumulative} evidence $-z_{t,i}$ against $t\rho_1$, its sparsity pattern
can differ from FOBOS; no universal ordering of the number of nonzeros
follows. On $\Delta_N$ the penalty $\|\bw\|_1=1$ is constant and cannot induce
sparsity.

\label{sec:pseudocode}
\paragraph{Per-coordinate adaptive Euler FTRL-Proximal.}
Choose $\varphi_+(w)=\phi_{a,b,\lambda}(1+w)$ on $w\ge0$ or
$\varphi_\pm(w)=\phi_{a,b,\lambda}(1+|w|)$ on $\mathbb R$, write
$\ell=\varphi'$ and $\mathcal F(\bw)=\sum_i\varphi(w_i)$. With
$\eta_0,\delta_0>0$, Algorithm~\ref{alg:ao-ftrl-prox} exactly minimizes
\[
 \langle\bG_t+\vm_{t+1},\bw\rangle+t\rho_1\|\bw\|_1
 +(\rho_2+\delta_0/\eta_0)\mathcal F(\bw)
 +\sum_{s=1}^t\sum_i\sigma_{s,i}D_\varphi(w_i\Vert w_{s,i}).
\]
{In this objective $\bG_t=\sum_{s=1}^t\bg_s$, $\vm_{t+1}$ is the next gradient forecast, and $\sigma_{s,i}$ is the increment of coordinate $i$'s regularizer weight. The initial origin weight $\delta_0/\eta_0$ is essential}: the increments
telescope to $(\delta_0+\sqrt{v_{t,i}})/\eta_0-\delta_0/\eta_0$, not to the full
coefficient. The shifted geometry is used whether or not $\rho_1>0$.

\begin{algorithm}[ht]
\caption{\textsc{AO-EulerProx}: adaptive optimistic shifted Euler FTRL-Proximal}
\label{alg:ao-ftrl-prox}
\begingroup\small
\begin{algorithmic}[1]
\Require Dimension $N$ and horizon $T$; $-1<a\rev{\le}0<b\le1$, $\lambda>0$; $\eta_0,\delta_0>0$; penalties $\rho_1,\rho_2\ge0$; signed or nonnegative domain; gradients $\bg_t$ and hints $\widehat\bg_{t+1}$
\Ensure Decisions $\bw_1,\ldots,\bw_{T+1}$; the decision after round $t$ uses the supplied forecast for round $t+1$
\State Set $k(w)=\operatorname{sign}(w)\log_{a,b,\lambda}(1+|w|)$ with $k(0)=0$.
\State $\bw_1\gets\0$; $z_i\gets0$, $v_i\gets0$, $A_i\gets\delta_0/\eta_0$ for every $i$.
\For{$t=1,\ldots,T$}
 \State Play $\bw_t$; observe $\bg_t$ and obtain $\widehat\bg_{t+1}$ before the next loss.
 \For{$i=1,\ldots,N$}
  \State $v_i\gets v_i+g_{t,i}^2$; $A_i^+\gets(\delta_0+\sqrt{v_i})/\eta_0$.
  \State $z_i\gets z_i+g_{t,i}-(A_i^+-A_i)k(w_{t,i})$; $A_i\gets A_i^+$.
  \State $q_i\gets z_i+\widehat g_{t+1,i}$; $D_i\gets\rho_2+A_i$.
  \If{nonnegative domain}
   \State $d_i\gets1$; $r_i\gets[-q_i-t\rho_1]_+$.
  \Else
   \State $d_i\gets-\operatorname{sign}(q_i)$; $r_i\gets[|q_i|-t\rho_1]_+$.
  \EndIf
  \State $w_{t+1,i}\gets d_i[\exp_{a,b,\lambda}(r_i/D_i)-1]$ \Comment{zero if $r_i=0$}
 \EndFor
\EndFor
\State \Return $(\bw_1,\ldots,\bw_{T+1})$
\Guarantee With exact scalar inverses, each successor minimizes the cumulative objective in Section~\ref{sec:pseudocode}. The initial decision is the origin.
\end{algorithmic}
\endgroup
\end{algorithm}

\subsubsection{Signed sparse weights: Euler regularized dual averaging}
\label{subsec:signed-rda}
Euler logarithms act on positive arguments, so signed coefficients need a
different construction. For a shift $\varrho>0$, which generalizes the shift
$1$ of Section~\ref{subsubsec:ftrl3-shifted},
\begin{equation}
 \psi_{a,b,\lambda,\varrho}(w)=\int_0^{|w|}\bigl[\log_{a,b,\lambda}(\varrho+s)-\log_{a,b,\lambda}(\varrho)\bigr]\,\mathrm ds
 \label{eq:new-signed-potential}
\end{equation}
{is strictly convex and $C^2$ on $\mathbb R$, with second derivative $h_{a,b,\lambda}(\varrho+|w|)$ including at the origin. For $a\le0<b$ this potential grows superlinearly as $|w|\to\infty$. With $\beta_t>0$, $A_t\ge0$ and $\rho_1\ge0$, the composite unconstrained problem}
\begin{equation}
 \min_{\bw\in\mathbb R^N}\Bigl\{\langle\bs_t,\bw\rangle
 +\beta_t\sum_i\psi_{a,b,\lambda,\varrho}(w_i)+A_t\rho_1\|\bw\|_1\Bigr\}
 \label{eq:new-signed-ftrl}
\end{equation}
has the unique coordinate solution
\begin{empheq}[box=\fbox]{equation}
 w_{i,t+1}=-\operatorname{sign}(s_{i,t})
 \Bigl[\exp_{a,b,\lambda}\Bigl(\log_{a,b,\lambda}(\varrho)+\frac{[|s_{i,t}|-A_t\rho_1]_+}{\beta_t}\Bigr)-\varrho\Bigr].
 \label{eq:new-soft-threshold}
\end{empheq}
Thus $|s_{i,t}|\le A_t\rho_1$ gives an exact zero; the magnitude map
$r\mapsto \exp_{a,b,\lambda}(\log_{a,b,\lambda}(\varrho)+r/\beta_t)-\varrho$ has slope
$m_{a,b,\lambda}(\varrho)/\beta_t$ at $r=0$ and grows like
$[br/(\beta_\lambda\beta_t)]^{1/b}$. Hence $\rho_1$ sets the dead zone,
$(a,b,\lambda,\varrho)$ the initial slope and $b$ the growth of strongly
supported features. This is the Euler analogue of $\ell_1$-RDA~\cite{Xiao2010}, built on Nesterov's dual averaging~\cite{Nesterov2009DA};
its logarithmic limit $\operatorname{sign}(w)\ln(1+|w|/\varrho)$ is not the
hypentropy map $\operatorname{arsinh}(w/\varrho)$~\cite{Ghai2020}. A regret
bound requires a bounded domain, so Algorithm~\ref{alg:euler-rda} has an
optional box radius $w_{\max}$; separability makes the capped magnitude the
exact constrained minimizer.

\begin{algorithm}[ht]
\caption{\textsc{EulerRDA}: signed sparse dual averaging with an optional box constraint}
\label{alg:euler-rda}
\begingroup\small
\begin{algorithmic}[1]
\Require Dimension $N$; $-1<a\le0<b\le1$, $\lambda>0$; shift $\varrho>0$; penalty $\rho_1\ge0$; $w_{\max}\in(0,\infty]$; predictable weights $\omega_t>0$; nondecreasing $\beta_t>0$; online convex losses $\ell_t$
\Ensure Signed decisions $\bw_t$ in the box of radius $w_{\max}$
\State $\bs\gets\0$; $A\gets0$; $\bw_1\gets\0$.
\State Set $\kappa_{\max}\gets \log_{a,b,\lambda}(\varrho+w_{\max})-\log_{a,b,\lambda}(\varrho)$ for a finite box and $+\infty$ otherwise.
\For{$t=1,2,\ldots$}
 \State Play $\bw_t$; observe $\bg_t\in\partial\ell_t(\bw_t)$.
 \State $\bs\gets\bs+\omega_t\bg_t$; $A\gets A+\omega_t$.
 \For{$i=1,\ldots,N$}
  \State $r_i\gets (|s_i|-A\rho_1)/\beta_t$.
  \If{$r_i\le0$}
   \State $w_{t+1,i}\gets0$.
  \ElsIf{$r_i\ge\kappa_{\max}$}
   \State $w_{t+1,i}\gets-\operatorname{sign}(s_i)w_{\max}$.
  \Else
   \State Solve $\alpha_\lambda\varrho^a\operatorname{expm1}(ad)/a+\beta_\lambda\varrho^b\operatorname{expm1}(bd)/b=r_i$ for $d\in(0,\ln(1+w_{\max}/\varrho))$ and set $v\gets\varrho\operatorname{expm1}(d)$, which equals $\exp_{a,b,\lambda}(\log_{a,b,\lambda}(\varrho)+r_i)-\varrho$ without cancellation.
   \State $w_{t+1,i}\gets-\operatorname{sign}(s_i)v$.
  \EndIf
 \EndFor
\EndFor
\Guarantee Exact composite FTRL decisions. For a finite box the regret bound is Proposition~\ref{prop:stage2-rda-regret}.
\end{algorithmic}
\endgroup
\end{algorithm}

\begin{proposition}[A bounded-domain guarantee for Euler-RDA]
\label{prop:stage2-rda-regret}
{Let $w_{\max}<\infty$, choose $0<\beta_0\le\beta_1$, and put $H_{a,b,\lambda,\varrho}(\bw)=\sum_i\psi_{a,b,\lambda,\varrho}(w_i)$,}
$h_{\min}=h_{a,b,\lambda}(\varrho+w_{\max})>0$, and use exact decisions and
convex losses. For every $\bu\in K_{\max}=[-w_{\max},w_{\max}]^N$, with
$J_t=\omega_t(\|\bg_t\|_2+\rho_1\sqrt N)$,
\begin{equation}
 \sum_{t=1}^T\omega_t\{\ell_t(\bw_t)+\rho_1\|\bw_t\|_1-\ell_t(\bu)-\rho_1\|\bu\|_1\}
 \le\beta_TH_{a,b,\lambda,\varrho}(\bu)+\frac1{2h_{\min}}\sum_{t=1}^T\frac{J_t^2}{\beta_{t-1}} .
 \label{eq:stage2-rda-regret}
\end{equation}
If $J_t\le J$ and $\beta_t=\gamma\sqrt{t+1}$, this is at most
$\gamma\sqrt{T+1}H_{a,b,\lambda,\varrho}(\bu)+J^2\sqrt T/(h_{\min}\gamma)$.
\end{proposition}
\noindent\textit{The proof is given in Appendix~\ref{app:proofs}, p.~\pageref{prf:prop:stage2-rda-regret}.}
For numerical work, threshold equality belongs to the zero branch, and a small
positive magnitude should be computed without subtracting $\varrho$ from an
approximate inverse: solve for $d>0$ in
$\alpha_\lambda\varrho^a\operatorname{expm1}(ad)/a+\beta_\lambda\varrho^b\operatorname{expm1}(bd)/b=r/\beta_t$
\rev{(with $r=[|s_i|-A\rho_1]_+$, i.e.\ $r/\beta_t=r_i$ in Algorithm~\ref{alg:euler-rda})}
and set $v=\varrho\operatorname{expm1}(d)$; the left side is increasing, so
$[0,\ln(1+w_{\max}/\varrho)]$ brackets the uncapped root. If the implemented
decision has a composite stationarity residual $\zeta_t$, add
$\sum_tJ_t\zeta_t/(\beta_{t-1}h_{\min})$ to Eq.~\eqref{eq:stage2-rda-regret}.

\subsubsection{Sparse Euler FTRL on the simplex boundary}
\label{subsec:FTRL-sparse}
Exact zeros require the boundary $a=0<b\le1$, where
$\phi_{0,b}'(0)=\log_{0,b}(0)=-1/b$ is finite and the scale $\lambda$ cancels.

\begin{proposition}[Sparse Euler FTRL]
\label{prop:sparse-FTRL}
Let $a=0<b\le1$, $\eta>0$ and $\bh\in\mathbb R^N$. The minimizer of
$\langle\bh,\bw\rangle+F_{0,b}(\bw)/\eta$ over $\Delta_N$ is unique and equals
\begin{equation}
 w_i=\bigl[1+b(-\eta h_i-\nu)\bigr]_+^{1/b},\qquad\sum_iw_i=1 .
 \label{eq:sparse-FTRL}
\end{equation}
(i) With $h_*=\min_jh_j$ and $w_*=\max_jw_j$,
$w_i=[w_*^b-b\eta(h_i-h_*)]_+^{1/b}$; hence $w_i=0$ iff
$h_i-h_*\ge w_*^b/(b\eta)$, and $h_i-h_*\ge1/(b\eta)$ is sufficient.
(ii) For $b=1$ the map is sparsemax~\cite{Martins2016}; for $0<b<1$ it is
$(1+b)$-entmax~\cite{Peters2019}. (iii) $F_{0,b}$ is $N^{-b}$-strongly convex
in $\ell_1$ with radius $R_{0,b}(N)=(1-N^{-b})/[b(1+b)]$, and
Theorems~\ref{thm:Euler-FTRL-regret} and~\ref{thm:Euler-adaptive} hold with
$(\mu,R)=(N^{-b},R_{0,b}(N))$.
\end{proposition}
\noindent\textit{The proof is given in Appendix~\ref{app:proofs}, p.~\pageref{prf:prop:sparse-FTRL}.}
For $a<0$ the corresponding coordinate is
$w_i=\exp_{a,b,\lambda}(\log_{a,b,\lambda}(w_*)-\eta\Delta h_i)\sim[|a|\eta\Delta h_i/\alpha_\lambda]^{-1/|a|}$,
so the negative exponent is a sparsity--retention dial: $a=0$ gives exact zeros
with re-entry whenever $h_i-h_*<w_*^b/(b\eta_t)$, while $a<0$ gives polynomial
retention. At the boundary, recursive simplex MD generally differs from FTRL,
as Euclidean greedy and lazy projections do.

\subsection{Optimistic Euler FTRL and mirror descent}
\label{sec:new-optimism}
Optimistic methods act on a forecast $\widehat\bg_t$ of the next gradient,
available before feedback, and then correct~\cite{Rakhlin2013}; a common choice
is $\widehat\bg_t=\bg_{t-1}$. Predictability is a property of the loss sequence
and the forecast, not of the link.
\subsubsection{Optimistic mirror descent: prediction-error regret}
Take a compact convex $K$ on which
$F_{a,b,\lambda}$ is differentiable and $\mu$-strongly convex in $\|\cdot\|_1$
(e.g.\ $K_\epsilon$), an anchor $\bz_1\in K$ and constant $\eta>0$, and use two
prox steps with the \emph{same anchor}:
\begin{equation}
 \bp_t=\operatorname*{arg\,min}_{\bp\in K}\{\eta\langle\widehat\bg_t,\bp\rangle+D_{F_{a,b,\lambda}}(\bp\Vert\bz_t)\},\qquad
 \bz_{t+1}=\operatorname*{arg\,min}_{\bp\in K}\{\eta\langle\bg_t,\bp\rangle+D_{F_{a,b,\lambda}}(\bp\Vert\bz_t)\}.
 \label{eq:new-oomd-play}
\end{equation}
This is Eq.~\eqref{eq:euler-omd} written with the played point first.

\begin{theorem}[Euler optimistic OMD: prediction-error regret]
\label{thm:new-oomd}
For every $\bp^*\in K$,
\begin{equation}
 \sum_{t=1}^T\langle\bg_t,\bp_t-\bp^*\rangle\le
 \frac{D_{F_{a,b,\lambda}}(\bp^*\Vert\bz_1)}{\eta}
 +\frac{\eta}{2\mu}\sum_{t=1}^T\|\bg_t-\widehat\bg_t\|_\infty^2 .
 \label{eq:new-oomd-regret}
\end{equation}
On simplices the norm may be replaced by the half-range of the prediction
error; on the full simplex with interior initialization, boundary comparators
are admitted by continuity.
\end{theorem}
\noindent\textit{The proof is given in Appendix~\ref{app:proofs}, p.~\pageref{prf:thm:new-oomd}.}
{With divergence $D=D_{F_{a,b,\lambda}}(\bp^*\Vert\bz_1)>0$ and an error budget $V_T\ge\sum_t\|\bg_t-\widehat\bg_t\|_\infty^2$ with $V_T>0$, the rate $\eta=\sqrt{2\mu D/V_T}$ gives $\sqrt{2DV_T/\mu}$.} The theorem does not prove
last-iterate convergence in games or gradient- and comparator-variation
guarantees~\cite{Xie2024,Zhao2026Variation}. In nonstationary environments a positive
floor cannot cancel old accumulated losses; history pruning, discounting,
fixed-share mixing or restarts remain necessary~\cite{herbsterwarmuth,Mhaisen2025}.

\subsubsection{Adaptive optimistic Euler FTRL}
\label{subsec:new-adaptive-oftrl}
Range capping removes the need to know the size of the prediction errors. Let
\begin{equation}
 \widetilde\sigma_t=\operatorname{hr}(\bg_t-\widehat\bg_t),\qquad
 \widetilde E_t=\sum_{s\le t}\widetilde\sigma_s^2,\qquad
 \eta_t=\frac{\eta_0}{\sqrt{\delta+\widetilde E_{t-1}}},\qquad
 \bp_t=\mathcal N_{a,b,\lambda}\bigl(-\eta_t(\bL_{t-1}+\widehat\bg_t)\bigr),
 \label{eq:aoftrl}
\end{equation}
with $\bL_{t-1}=\sum_{s<t}\bg_s$; the rate uses only past errors.

\begin{theorem}[Adaptive optimistic Euler FTRL]
\label{thm:new-adaptive-oftrl}
Let $N\ge2$, fixed exponents with either $-1<a<0<\rev{b\le 1}$ and $(\mu,R)$ from
Theorem~\ref{thm:sharp-simplex-geometry}, or $a=0<b\le1$ and $(\mu,R)$ from
Proposition~\ref{prop:sparse-FTRL}; let $\eta_0,\delta>0$. For exact decisions,
every finite gradient and forecast sequence and every $\bu\in\Delta_N$,
\begin{align}
 \sum_{t=1}^T\langle\bg_t,\bp_t-\bu\rangle
 &\le\frac{R}{\eta_0}\sqrt{\delta+\widetilde E_{T-1}}
 +\sum_{t=1}^T\min\Bigl\{\frac{\eta_0\widetilde\sigma_t^2}{2\mu\sqrt{\delta+\widetilde E_{t-1}}},\,2\widetilde\sigma_t\Bigr\}
 \label{eq:aoftrl-capped}\\
 &\le\Bigl(\frac{R}{\eta_0}+C_{\eta_0}\Bigr)\sqrt{\delta+\widetilde E_T},
 \label{eq:aoftrl-bound}
\end{align}
with $C_{\eta_0}$ of Theorem~\ref{thm:Euler-adaptive}. Iterates are unchanged by
common offsets of $\bg_t$ or $\widehat\bg_t$.
\end{theorem}
\noindent\textit{The proof is given in Appendix~\ref{app:proofs}, p.~\pageref{prf:thm:new-adaptive-oftrl}.}

\begin{remark}[What the theorem buys]
\label{rem:aoftrl}
{With $\widehat\bg_t=0$ and $\delta>0$, the decisions coincide with adaptive Euler FTRL after aligning the round indices, and Eq.~\eqref{eq:aoftrl-capped} gives the same stability sum with regularization cost $R\sqrt{\delta+S_{T-1}}/\eta_0$;} with
perfect forecasts regret is at most $(R/\eta_0)\sqrt\delta$; with
$\widehat\bg_t=\bg_{t-1}$ it depends on $\sum_t\operatorname{hr}(\bg_t-\bg_{t-1})^2$,
small for slowly drifting losses. The cap bounds a round's stability term, not
its instantaneous regret. In the full-range case
$p_{i,t}\ge \exp_{a,b,\lambda}(\log_{a,b,\lambda}(1/N)-\eta_t\operatorname{range}(\bL_{t-1}+\widehat\bg_t))>0$.
Arbitrarily bad forecasts can be much worse than none: for $N=2$,
$\bg_t=(0,1)$ and $\widehat\bg_t=(0,-2t)$, symmetry gives $p_{2,t}\ge1/2$ and
regret at least $T/2$. Competing with both a forecast and a no-forecast
baseline needs an additional aggregation mechanism.
\end{remark}

\begin{algorithm}[ht]
\caption{\textsc{AO-EulerFTRL}: adaptive optimistic Euler FTRL with the rate fixed before the current loss}
\label{alg:aoftrl}
\begingroup\small
\begin{algorithmic}[1]
\Require $N\ge2$; $-1<a<0<\rev{b\le 1}$ or $a=0<b\le1$, with $\lambda>0$; $\eta_0,\delta>0$; online losses $\bg_t$ and predictable hints $\widehat\bg_t$; optional solver tolerances $\epsilon_t>0$
\Ensure One decision $\bp_t\in\Delta_N$ before each loss is revealed
\State $\bh\gets\0$; $E\gets0$.
\For{$t=1,2,\ldots$}
 \State Obtain $\widehat\bg_t$ from pre-round data, for example the previous loss vector (zero at $t=1$).
 \State $\eta\gets\eta_0/\sqrt{\delta+E}$; $\by\gets-\eta[\bh+\operatorname{ctr}(\widehat\bg_t)]$.
 \State Set $\bp_t\gets\mathcal N_{a,b,\lambda}(\by)$, or use $\mathsf{Norm}(\by;\epsilon_t)$.
 \State If the numerical call fails, return the played prefix with \textsc{precision-limit}.
 \State Play $\bp_t$, then observe $\bg_t$.
 \State $E\gets E+\operatorname{hr}(\bg_t-\widehat\bg_t)^2$.
 \State $\bh\gets\bh+\operatorname{ctr}(\bg_t)$.
\EndFor
\Guarantee The forecast-error regret of Theorem~\ref{thm:new-adaptive-oftrl}, with the solver cost in Lemma~\ref{lem:stage2-oracle-error}.
\end{algorithmic}
\endgroup
\end{algorithm}
Decisions are invariant under rescaling gradients and forecasts by $c>0$ when
$\delta$ is replaced by $c^2\delta$; $\delta=0$ is not covered.

\subsection{What the Euler geometry adds, and what it does not}
\label{sec:takeaways}
Across the five families the pattern is the same. The algorithm keeps its
established structure and guarantee type; the Euler link changes (i) the scale
at which unfavorable coordinates are abandoned, from exponential to polynomial
with exponent $1/|a|$; (ii) the geometry constant $R/\mu$, which a well-chosen
{$\lambda$ brings to within about $4.5$--$16\%$ of the entropy geometry constant for the dimension-dependent parameters in Table~\ref{tab:near-optimal-lambda};} and (iii) the shape of active weights in
sparse methods through $b$ and $\lambda$. The cost is one scalar root-solve per
simplex step and three parameters to select. FTRL and MD coincide only under
specified geometry, constraint, initialization and schedule conditions;
retained probability is not by itself faster recovery or better exploration;
and hints, adaptivity and discounting need their own analyses. Further variants
(lazy and proximal equivalences, time-varying FTRL, schedule-free averaging,
weighted dual averaging, and learning the geometry online) are collected in
Appendix~\ref{app:supp-online}.  Sections~\ref{sec:ftrl-bandit} and~\ref{sec:ftrl-choosing} add bandit feedback and a practical guide to choosing the geometry.

\subsection{Euler FTRL with bandit feedback}
\label{sec:ftrl-bandit}
The full-information guarantees above use the global $\ell_1$ modulus $\mu_{a,b,\lambda}(N)$. Importance-weighted bandit feedback produces loss estimates with second moments of order $1/w_{t,i}$, which a global modulus cannot absorb; a \emph{local} bound is needed. For Euler geometry it takes a particularly transparent form.

\begin{lemma}[Local-norm stability for nonnegative losses]
\label{lem:bandit-local-norm}
Let $a<0\le b\le1$ or $(a,b)\to(0,0)$, $\lambda>0$, $\eta>0$, $\bw_t\in\Delta_N^\circ$ and $\bg\in\mathbb R^N_{\ge0}$, and let $\bw_{t+1}$ be the exact simplex Euler mirror step from $\bw_t$. Then
\begin{equation}
 \langle\bg,\bw_t-\bw_{t+1}\rangle-\frac1\eta D_{F_{a,b,\lambda}}(\bw_{t+1}\Vert\bw_t)
 \le\frac\eta2\sum_{i=1}^Ng_i^2\,m_{a,b,\lambda}(w_{t,i}).
 \label{eq:bandit-local-norm}
\end{equation}
\end{lemma}
\noindent\textit{The proof is given in Appendix~\ref{app:proofs}, p.~\pageref{prf:lem:bandit-local-norm}.}

In the $N$-armed adversarial bandit with oblivious losses $\boldsymbol\ell_t\in[0,1]^N$, the learner plays $I_t\sim\bw_t$ and forms $\hat g_{t,i}=\ell_{t,i}\mathbb 1\{I_t=i\}/w_{t,i}\ge0$, so that $\mathbb E_t[\hat g_{t,i}^2]=\ell_{t,i}^2/w_{t,i}$. Euler FTRL is run on the raw estimates, $\bw_{t+1}=\mathcal N_{a,b,\lambda}(-\eta\sum_{s\le t}\hat\bg_s)$; centering would destroy their nonnegativity. Because
\begin{equation}
 \frac{m_{a,b,\lambda}(x)}{x}=\frac1{\omega_{a,b,\lambda}(x)},
 \label{eq:bandit-m-over-x}
\end{equation}
the variance factor is the reciprocal of the \emph{same} gradient weight that shapes Euler cross-entropy (Section~\ref{subsec:two-parameter-gain}). Define the bandit variance constant
\begin{equation}
 V_{a,b,\lambda}(N)=\sup_{\bw\in\Delta_N^\circ}\sum_{i=1}^N\frac1{\omega_{a,b,\lambda}(w_i)}
 \le\frac{N^{1-|a|}}{\alpha_\lambda}\qquad(-1\le a<0\le b),
 \label{eq:bandit-variance}
\end{equation}
where the bound follows from $\omega_{a,b,\lambda}(x)\ge\alpha_\lambda x^a$ and $\sum_iw_i^{|a|}\le N^{1-|a|}$, with equality on the boundary $b=0$; in the logarithmic limit $V=N$.

\begin{proposition}[Euler FTRL with bandit feedback]
\label{prop:euler-bandit}
Let $-1<a<0\le b\le1$, $\lambda>0$, a constant rate $\eta>0$ and uniform initialization. For every $\bu\in\Delta_N$,
\begin{equation}
 \mathbb E\sum_{t=1}^T\langle\boldsymbol\ell_t,\bw_t-\bu\rangle
 \le\frac{R_{a,b,\lambda}(N)}{\eta}+\frac\eta2\sum_{t=1}^T\mathbb E\sum_{i=1}^N\frac{\ell_{t,i}^2}{\omega_{a,b,\lambda}(w_{t,i})}
 \le\frac{R_{a,b,\lambda}(N)}{\eta}+\frac{\eta T}{2}V_{a,b,\lambda}(N).
 \label{eq:euler-bandit}
\end{equation}
The rate $\eta=\sqrt{2R/(VT)}$ gives expected regret at most $C^{\rm band}_{a,b,\lambda}(N)\sqrt T$ with
\[
 C^{\rm band}_{a,b,\lambda}(N)=\sqrt{2R_{a,b,\lambda}(N)\,V_{a,b,\lambda}(N)}.
\]
For $a=-1$ the radius is infinite; comparing with $\tilde\bu_\gamma=(1-\gamma)\bu+\gamma\1/N$ adds $\gamma T$ and replaces $R$ by $B_\gamma=D_{F_{-1,b,\lambda}}(\tilde\bu_\gamma\Vert\1/N)$.
\end{proposition}
\noindent\textit{The proof is given in Appendix~\ref{app:proofs}, p.~\pageref{prf:prop:euler-bandit}.}

\begin{corollary}[What the three parameters do in the bandit bound]
\label{cor:bandit-shape}
For $p=|a|\in(0,1)$ and $0<b\le1$,
\begin{equation}
 R_{a,b,\lambda}(N)\,V_{a,b,\lambda}(N)\le
 \underbrace{\frac{N-N^{1-p}}{p(1-p)}}_{\text{independent of }b,\lambda}
 +\underbrace{\frac{b}{p}\lambda^{b-a}\,\frac{1-N^{-b}}{b(1+b)}\,N^{1-p}}_{O(N^{1-p})}.
 \label{eq:bandit-RV}
\end{equation}
The leading term is minimized at $p=\frac12$, where it equals $4(N-\sqrt N)$, the Tsallis-INF exponent~\cite{Audibert2009,Zimmert2021}. A positive bulk exponent $b$ and the scale enter only through $\beta_\lambda/\alpha_\lambda=(b/p)\lambda^{b-a}$, multiplying a lower-order term; a small $\lambda$ makes this cost vanish. For $a=-1$, $B_\gamma V\le(N-1)\ln(1/\gamma)+(\beta_\lambda/\alpha_\lambda)/[b(1+b)]$, which is minimized by the pure barrier $\alpha_\lambda=1$.
\end{corollary}
\begin{proof}
Multiply Eq.~\eqref{eq:bregman-radius} by the bound in Eq.~\eqref{eq:bandit-variance}; for $a=-1$ use $V\le1/\alpha_\lambda$ and the explicit form of $B_\gamma$ given before Table~\ref{tab:ftrl-bandit}.
\end{proof}

Proposition~\ref{prop:euler-bandit} is a pseudo-regret bound for oblivious adversaries with a tuned constant rate; anytime rates $\eta_t\propto t^{-1/2}$ follow by the usual FTRL argument. The best-of-both-worlds guarantees of Tsallis-INF~\cite{Zimmert2021} are not claimed for $b>0$.

\subsection{Choosing the geometry for FTRL: metrics, benchmark tables and scenarios}
\label{sec:ftrl-choosing}
This subsection collects, in one place, the quantities that decide convergence and stability of Euler FTRL on $\Delta_N$, evaluates them for representative geometries, and states where the Euler geometry should and should not be used. Two locations organize the curvature profile: the centre of the logistic transition between the barrier and bulk terms, $x_0=\lambda^{-1}(|a|/b)^{1/(b-a)}$, where $\alpha_\lambda x_0^a=\beta_\lambda x_0^b$ (Proposition~\ref{prop:elasticity}), and the minimum $q_m=r/\lambda$ of the gradient weight (Proposition~\ref{prop:omega-profile}). Centering the transition at $1/N$ requires $\lambda=N(|a|/b)^{1/(b-a)}$; placing $q_m$ there requires $\lambda=Nr$. Neither placement optimizes a regret bound.

\begin{table}[htbp]

\centering\footnotesize\setlength{\tabcolsep}{3pt}\renewcommand{\arraystretch}{1.2}
\caption{Metrics of the Euler geometry and their role in FTRL on $\Delta_N$.}
\label{tab:ftrl-metrics}
\begin{tabular}{@{}L{.15\textwidth}L{.28\textwidth}L{.1\textwidth}L{.3\textwidth}L{.11\textwidth}@{}}
\hline
Metric & Formula & Set by & Role in FTRL & Where\\\hline
Curvature & $h_{a,b,\lambda}=\alpha_\lambda x^{a-1}+\beta_\lambda x^{b-1}$ & $a,b,\lambda$ & Local metric of the Bregman step; nonincreasing for $b\le1$ & Eq.~\eqref{eq:elasticity-def}\\
Mobility & $m_{a,b,\lambda}=1/h_{a,b,\lambda}\sim x^{1-a}/\alpha_\lambda$ ($x\downarrow0$) & $a$ (small), $b$ (large) & Primal response to a dual step; local stability $\tfrac\eta2\sum_ig_i^2m(w_i)$ & Lemma~\ref{lem:bandit-local-norm}\\
Gradient weight & $\omega_{a,b,\lambda}=\alpha_\lambda x^a+\beta_\lambda x^b$ & $a,b,\lambda$ & Bandit variance $\sum_i1/\omega(w_i)$; Euler-CE weight & Eq.~\eqref{eq:bandit-m-over-x}\\
Elasticity & $\varepsilon=-\mathrm d\ln h/\mathrm d\ln x$, from $1-a$ to $1-b$ & $a,b$ (levels), $\lambda$ (place) & How fast curvature relaxes from boundary to bulk & Prop.~\ref{prop:elasticity}\\
Transition & $x_0=\lambda^{-1}(|a|/b)^{1/(b-a)}$; $q_m=r/\lambda$ & $\lambda$ & Where barrier and bulk terms balance; where $\omega$ is smallest & Props.~\ref{prop:elasticity}, \ref{prop:omega-profile}\\
Floors & $c=\min_{(0,1]}\omega$; $\omega_{\min}=K/D_\lambda$ & $a,b,\lambda$ & KL comparison $D_F\ge cD_{\rm KL}$; clipping-free steps & Eq.~\eqref{eq:entropy-comparison-constant}\\
$\ell_1$ modulus & $\mu=\max\{c,\alpha_\lambda2^{|a|}+\beta_\lambda N^{-b}\}$ & $a,b,\lambda,N$ & Full-information stability $\eta\operatorname{hr}(\bg_t)^2/(2\mu)$ & Thm.~\ref{thm:sharp-simplex-geometry}\\
Radius & $R_{a,b,\lambda}(N)$ & $a,b,\lambda,N$ & Diameter term $R/\eta$; infinite at $a=-1$ & Thm.~\ref{thm:sharp-simplex-geometry}\\
Full-information cost & $\sqrt{2R/\mu}$ & all & Regret $\le\sqrt{2R/\mu}\,G\sqrt T$; minimized over $\lambda$ at $\lambda_*$ & Cor.~\ref{cor:ftrl3-regret}\\
Bandit cost & $\sqrt{2RV}$, $V\le N^{1-|a|}/\alpha_\lambda$ & $a$ (leading), $\lambda$ (lower order) & Expected regret $\le\sqrt{2RVT}$ & Prop.~\ref{prop:euler-bandit}\\
Centre curvature & $h(1/N)=\alpha_\lambda N^{1-a}+\beta_\lambda N^{1-b}$ & $\lambda$ & Stiffness at the start; Newton conditioning & ---\\
Retention floor & $w_i\gtrsim(|a|\Delta/\alpha_\lambda)^{-1/|a|}$ for score gap $\Delta$ & $a$ (power), $\lambda$ (prefactor) & Polynomial rather than exponential suppression & Thm.~\ref{thm:no-extinction}\\
Dual-step stability & $|\ln w'-\ln w|\le|z'-z|/\omega_{\min}$ & $a,b,\lambda$ & No coordinate changes by more than a fixed factor per bounded dual step & Thm.~\ref{thm:design-criteria}(v)\\
Self-concordance & $M_{\rm sc}=\sqrt{1+b\lambda^{b+1}}$ at $a=-1$; $\infty$ for $-1<a\le0$ & $a=-1$, $\lambda$ & Dikin-ellipsoid control of curvature & Prop.~\ref{prop:barrier-sc}\\
Link range & $\mathbb R$ for $a<0<b$; $(-1/b,\infty)$ at $a=0$ & $a$ & Interior iterates versus exact zeros (entmax) & Lemma~\ref{lem:audit-domains}, Prop.~\ref{prop:sparse-FTRL}\\
Cost per update & one scalar root for $\nu$ & --- & $O(N)$ per sweep, certified & Alg.~\ref{alg:euler-norm}, Lemma~\ref{lem:certified-norm}\\\hline
\end{tabular}
\end{table}

\paragraph{Benchmark constants.}
Tables~\ref{tab:ftrl-fullinfo}--\ref{tab:ftrl-structure} evaluate these metrics; they are analytic upper-bound constants, not measured regrets. The geometries are Shannon entropy $(0,0)$; the Tsallis-INF/Amari boundary $(-\frac12,0)$; the unscaled interior $(-\frac12,\frac12,1)$; Hybrid~B $(-\frac12,1,2^{-2/3})$, for which $\alpha_\lambda=\frac12$; the dimension-dependent choice $|a|=b=1/\ln N$ with $\lambda_*$ of Eq.~\eqref{eq:scaled-optimal-lambda}; the Burg barrier $(-1,0)$; and Hybrid~A $(-1,1,\lambda)$ with $\alpha_\lambda=1/(N\ln(1/\gamma))$, the scale that caps its comparator divergence near $1.5$. For $a=-1$ the comparator is shrunk with $\gamma=(NT)^{-1/2}$, $T=10^9$, and
\[
 B_\gamma=\alpha_\lambda\bigl[(N-1)\ln(1/\gamma)-\ln(N-(N-1)\gamma)\bigr]+\beta_\lambda\frac{\sum_i\tilde u_i^{\,b+1}-N^{-b}}{b(b+1)};
\]
the bias $\gamma T=\sqrt{T/N}$ is not included.

\begin{table}[htbp]

\centering\small
\caption{Full-information Euler FTRL: leading regret constant $\sqrt{2R_{a,b,\lambda}(N)/\mu_{a,b,\lambda}(N)}$ (regret $\le$ constant $\times\,G\sqrt T$ with $G\ge\operatorname{hr}(\bg_t)$). Bold marks the smallest value in each column.}
\label{tab:ftrl-fullinfo}
\begin{tabular}{llrrrrr}
\hline
Geometry $(a,b,\lambda)$ & Curvature $h(x)$ & $N=10$ & $10^2$ & $10^3$ & $10^4$ & $10^6$\\\hline
Shannon $(0,0)$ & $x^{-1}$ & $\mathbf{2.15}$ & $\mathbf{3.04}$ & $\mathbf{3.72}$ & $\mathbf{4.29}$ & $\mathbf{5.26}$\\
Euler, $|a|=b=1/\ln N$, $\lambda_*$ & $\alpha_{\lambda_*}x^{a-1}+\beta_{\lambda_*}x^{b-1}$ & $2.31$ & $3.14$ & $3.82$ & $4.40$ & $5.37$\\
Euler $(-\frac12,\frac12,1)$ & $\frac12(x^{-3/2}+x^{-1/2})$ & $3.09$ & $6.10$ & $11.13$ & $19.93$ & $63.22$\\
Hybrid~B $(-\frac12,1,2^{-2/3})$ & $\frac12x^{-3/2}+\frac12$ & $3.10$ & $6.22$ & $11.41$ & $20.48$ & $65.03$\\
Tsallis-INF $(-\frac12,0)$ & $x^{-3/2}$ & $3.50$ & $7.14$ & $13.16$ & $23.66$ & $75.17$\\
Burg, Hybrid~A ($a=-1$) & $\alpha_\lambda x^{-2}+\beta_\lambda$ & \multicolumn{5}{c}{infinite radius; requires a floored domain}\\\hline
\end{tabular}
\end{table}

\begin{table}[htbp]

\centering\small
\caption{Bandit Euler FTRL (Proposition~\ref{prop:euler-bandit}): constant $C^{\rm band}=\sqrt{2RV}$, with $B_\gamma$ in place of $R$ for $a=-1$. Bold marks the smallest value in each column.}
\label{tab:ftrl-bandit}
\begin{tabular}{llrrrrr}
\hline
Geometry $(a,b,\lambda)$ & Curvature $h(x)$ & $N=10$ & $10^2$ & $10^3$ & $10^4$ & $10^6$\\\hline
Shannon (Exp3) & $x^{-1}$ & $\mathbf{6.79}$ & $30.35$ & $117.54$ & $429.19$ & $5256.5$\\
Tsallis-INF $(-\frac12,0)$ & $x^{-3/2}$ & $7.40$ & $\mathbf{26.83}$ & $\mathbf{88.02}$ & $\mathbf{281.42}$ & $\mathbf{2827.0}$\\
Hybrid~B $(-\frac12,1,2^{-2/3})$ & $\frac12x^{-3/2}+\frac12$ & $7.59$ & $27.02$ & $88.20$ & $281.60$ & $2827.2$\\
Euler $(-\frac12,\frac12,1)$ & $\frac12(x^{-3/2}+x^{-1/2})$ & $7.78$ & $27.28$ & $88.48$ & $281.89$ & $2827.5$\\
Euler, $|a|=b=1/\ln N$, $\lambda_*$ & $\alpha_{\lambda_*}x^{a-1}+\beta_{\lambda_*}x^{b-1}$ & $10.15$ & $38.57$ & $142.91$ & $511.12$ & $6136.6$\\
Burg $(-1,0)$, $\gamma$-shrunk & $x^{-2}$ & $14.24$ & $49.98$ & $166.10$ & $547.07$ & $5877.0$\\
Hybrid~A $(-1,1)$ & $\alpha_\lambda x^{-2}+\beta_\lambda$ & $17.47$ & $61.25$ & $203.45$ & $670.03$ & $7197.8$\\\hline
\end{tabular}
\end{table}

\begin{table}[htbp]

\centering\footnotesize\setlength{\tabcolsep}{3.5pt}
\caption{Structural metrics at $N=10^3$ ($T=10^9$ for the $a=-1$ rows; $B_\gamma$ in parentheses). ``Floor'' is the exponent $1/|a|$ of polynomial retention.}
\label{tab:ftrl-structure}
\begin{tabular}{lcccccccc}
\hline
Geometry & $\alpha_\lambda$ & Floor & $h(1/N)$ & $R$ & $\mu$ & $V$ & $M_{\rm sc}$ & Zeros\\\hline
Shannon & --- & exponential & $10^3$ & $6.91$ & $1$ & $10^3$ & $\infty$ & no\\
Euler, $1/\ln N$, $\lambda_*=86.6$ & $0.216$ & $6.91$ & $875$ & $5.98$ & $0.82$ & $1706$ & $\infty$ & no\\
Euler $(-\frac12,\frac12,1)$ & $0.5$ & $2$ & $1.58\times10^4$ & $61.9$ & $1$ & $63.2$ & $\infty$ & no\\
Hybrid~B & $0.5$ & $2$ & $1.58\times10^4$ & $61.5$ & $0.94$ & $63.2$ & $\infty$ & no\\
Tsallis-INF & $1$ & $2$ & $3.16\times10^4$ & $122.5$ & $1.41$ & $31.6$ & $\infty$ & no\\
Burg & $1$ & $1$ & $10^6$ & $(1.38\times10^4)$ & $2$ & $1$ & $1$ & no\\
Hybrid~A, $\lambda=117.5$ & $7.2\times10^{-5}$ & $1$ & $73.4$ & $(1.50)$ & $0.017$ & $1.38\times10^4$ & $117.5$ & no\\
Sparse $(0,\frac12)$ & $0$ & exact zeros & $31.6$ & $1.29$ & $0.032$ & $\infty$ & $\infty$ & yes\\\hline
\end{tabular}
\end{table}

\paragraph{What the tables show.}
(i) \emph{Full information.} Shannon has the smallest worst-case constant. The dimension-dependent Euler geometry stays within $2$--$8\%$ of it for $N\le10^6$ while retaining polynomial floors with exponent $\ln N$; fixed interior exponents pay a radius that grows like $N^{|a|}$. (ii) \emph{Bandits.} The ordering reverses, as Corollary~\ref{cor:bandit-shape} predicts: Tsallis-INF is best for $N\ge10^2$, and a bulk exponent costs only a lower-order term. Hybrid~B is within $2.6\%$ of Tsallis-INF at $N=10$ and $0.007\%$ at $N=10^6$, while its radius and uniform-centre curvature are halved and large coordinates acquire Euclidean curvature $\beta_\lambda=\frac12$; no bandit regret improvement is claimed. (iii) \emph{Barrier hybrids.} Raising $\lambda$ in Hybrid~A shrinks the comparator divergence from $1.38\times10^4$ to $1.50$ at $N=10^3$, but the variance constant grows from $1$ to $1/\alpha_\lambda=1.38\times10^4$; the bandit constant becomes worse than the pure barrier ($203$ versus $166$). The scale is useful for curvature conditioning and stiffness at the start, not for this bound. (iv) \emph{Sparse boundary.} At $a=0$ exact zeros are available, but $\mu=N^{-b}$ and $V=\infty$: the boundary serves sparse decisions, not importance-weighted bandits.

\begin{table}[htbp]

\centering\footnotesize\setlength{\tabcolsep}{3pt}\renewcommand{\arraystretch}{1.22}
\caption{Scenario guide for Euler FTRL. The last column records what is \emph{not} guaranteed.}
\label{tab:ftrl-scenarios}
\begin{tabular}{@{}L{.17\textwidth}L{.2\textwidth}L{.19\textwidth}L{.18\textwidth}L{.19\textwidth}@{}}
\hline
Scenario & Recommended geometry & FTRL variant & Supporting result & Caveat\\\hline
Full-information experts, allocation, routing weights & $|a|=b=1/\ln N$, $\lambda=\lambda_*$; or Shannon & Fixed-step or adaptive Euler FTRL & Thms.~\ref{thm:Euler-FTRL-regret}, \ref{thm:Euler-adaptive}; Table~\ref{tab:near-optimal-lambda} & Retention is not recovery; Hedge is simpler when retention is irrelevant\\
Unknown or heavy-tailed loss scale & Any interior triple; $\eta_0=\sqrt{2\mu R}$ & Adaptive, range-capped (Alg.~\ref{alg:adaptive-ftrl}) & Thm.~\ref{thm:Euler-adaptive}, Cor.~\ref{cor:adaptive-FTRL-floor} & Constants still scale with $R/\mu$\\
Slowly varying losses, games & Interior triple & Optimistic adaptive (Alg.~\ref{alg:aoftrl}) & Thm.~\ref{thm:new-adaptive-oftrl} & Bad forecasts can hurt; no last-iterate claim\\
Adversarial bandits & $a=-\frac12$, $b\in[0,1]$, small $\lambda$ & Raw importance-weighted estimates & Prop.~\ref{prop:euler-bandit}, Cor.~\ref{cor:bandit-shape} & No best-of-both-worlds claim for $b>0$\\
Bandits needing local Newton control & $a=-1$, $b\in(0,1]$, small $\lambda$ for $M_{\rm sc}\approx1$ & Shrunk comparator or floored domain & Prop.~\ref{prop:barrier-sc} & Larger $\lambda$ trades variance for diameter\\
Sparse decisions (MoE gates, attention, feature subsets) & Boundary $a=0<b\le1$; $\lambda$ irrelevant & Sparse FTRL & Prop.~\ref{prop:sparse-FTRL} & $\mu=N^{-b}$; unsuitable for bandit estimates\\
Signed sparse weights (CTR, recommenders) & Shifted potential, $-1<a\le0<b\le1$ & FTRL-Proximal, Euler-RDA (Algs.~\ref{alg:ao-ftrl-prox}, \ref{alg:euler-rda}) & Prop.~\ref{prop:stage2-rda-regret} & Needs a bounded box for regret\\
Nonstationary experts needing retention & $a<0$ chosen by the floor power $1/|a|$ & Any of the above with restarts or discounting & Thm.~\ref{thm:no-extinction} & A floor cannot cancel old evidence\\
\hline
\end{tabular}
\end{table}

\paragraph{A selection recipe.}
(1) Choose $a$ from the feedback model: $|a|=1/\ln N$ for full information, $a=-\frac12$ for importance-weighted bandits, $a=-1$ when a self-concordant barrier is needed, and $a=0$ for exact sparsity. (2) Choose $b\in(0,1]$ for the bulk: $b=1$ gives the curvature floor $\beta_\lambda$ for large coordinates (Theorem~\ref{thm:design-criteria}(iii)); a small $b$ keeps the potential close to entropy there. (3) Choose $\lambda$ by the quantity that matters: $\lambda_*$ for the full-information constant, a small $\lambda$ for the bandit constant and for $M_{\rm sc}\approx1$, $\lambda=N(|a|/b)^{1/(b-a)}$ to centre the curvature transition at $1/N$, or $\lambda=Nr$ to place the gradient-weight minimum there. (4) Set the rate from the constants, $\eta=\sqrt{2\mu R/(G^2T)}$, $\eta_0=\sqrt{2\mu R}$ for the adaptive rate or $\eta=\sqrt{2R/(VT)}$ for bandits, and certify each normalization with Algorithm~\ref{alg:euler-norm}.

\section{Policy mirror descent and relative smoothness}
\label{sec:extensions}
\begin{glance}
\textbf{Question.} What does the Euler metric add to policy optimization and to smooth optimization?\\
\textbf{Key idea.} Used as a trust region, the Euler divergence limits how far a policy moves in one step and keeps rare actions alive; used as a mirror potential, its singular curvature can \emph{match} objectives that are too steep for Euclidean methods.\\
\textbf{Remember.} {Exact tabular Euler policy steps with exact advantages improve value monotonically. With bounded advantage errors, an explicit test can certify improvement. For convex objectives the relative-smoothness argument gives an $O(1/T)$ last-iterate objective bound.}
\end{glance}

Three established lines of work replace Euclidean geometry by one adapted to
the variables: $p$-norm algorithms, which interpolate between gradient descent
and multiplicative updates~\cite{Grove2001,Gentile2003}; policy mirror descent,
a statewise mirror step with the advantage as linear objective and a Bregman
trust region~\cite{TomarMDPO2022}; and relative or generalized
smoothness~\cite{Bauschke2017,Lu2018,Zhang2020clipping}. This section shows what
the three-parameter Euler metric adds to each.

The Euler Hessian is a weighted average of two separable power Hessians, a retaining one of order $1+a$ and a gentler one of order $1+b$, with weights $\alpha_\lambda,\beta_\lambda$ (Theorem~\ref{thm:pnorm-mixture}, Appendix~\ref{subsec:pnorm}). This two-component structure is what the next subsections exploit.

\subsection{Policy mirror descent with an Euler trust region}
\label{subsec:MDPO}
{Take a finite MDP with bounded rewards, a fixed action set $\mathcal A$ of size $n\ge2$, and discount $0\le\gamma<1$. Use $-1<a<0<b\le1$, $\lambda>0$, a strictly positive tabular policy $\pi_k$, and step $t_k>0$. Write $A^{\pi_k}(s,\alpha)=Q^{\pi_k}(s,\alpha)-V^{\pi_k}(s)$ for the exact advantage of action $\alpha$ at state $s$. Its mean under $\pi_k(\cdot|s)$ is zero. The statewise Euler step and its solution are}
\begin{empheq}[box=\fbox]{equation}
 \pi_{k+1}(\cdot|s)=\operatorname*{arg\,max}_{\pi\in\Delta_{|\mathcal A|}}
 \Bigl\{\langle A^{\pi_k}(s,\cdot),\pi\rangle-\frac{D_{F_{a,b,\lambda}}(\pi\Vert\pi_k(\cdot|s))}{t_k}\Bigr\},
 \label{eq:euler-mdpo}
\end{empheq}
\begin{empheq}[box=\fbox]{equation}
 \pi_{k+1}(\alpha|s)=\exp_{a,b,\lambda}\bigl(\log_{a,b,\lambda}(\pi_k(\alpha|s))+t_kA^{\pi_k}(s,\alpha)-\nu_{k,s}\bigr),
 \label{eq:euler-mdpo-closed}
\end{empheq}

\begingroup
Here $\nu_{k,s}$ is the multiplier in the scaled dual scores, chosen to make the action probabilities at state $s$ sum to one. Full range keeps the solution positive. The objective in Eq.~\eqref{eq:euler-mdpo} uses a divergence penalty, with $t_k$ setting its strength.
\par\endgroup
{The Euler softmax policy update accumulates step-weighted advantages in dual coordinates. With the ordinary logarithm the update multiplies each action probability by $\exp(t_k A^{\pi_k}(s,\alpha))$ before renormalizing.}

\begin{proposition}[Policy improvement and a quantitative step bound]
\label{prop:Euler-policy-improvement}
Let $\mu=\mu_{a,b,\lambda}(|\mathcal A|)$ and let $\sigma_s$ be the half-range of
$A^{\pi_k}(s,\cdot)$. With $\pi'=\pi_{k+1}(\cdot|s)$ and $\pi=\pi_k(\cdot|s)$,
\begin{equation}
 \langle A^{\pi_k}(s,\cdot),\pi'\rangle
 =\frac{D_{F_{a,b,\lambda}}(\pi'\Vert\pi)+D_{F_{a,b,\lambda}}(\pi\Vert\pi')}{t_k}\ge\frac{\mu}{t_k}\|\pi'-\pi\|_1^2\ge0,
 \qquad\|\pi'-\pi\|_1\le\min\{2,t_k\sigma_s/\mu\}.
 \label{eq:audit-policy-improvement}
\end{equation}
Hence $V^{\pi_{k+1}}\ge V^{\pi_k}$ componentwise, and
$D_{F_{a,b,\lambda}}\ge c_{a,b,\lambda}D_{\rm KL}$.
\end{proposition}
\begin{proof}
Pair the statewise KKT equation with $\pi'-\pi$; the multiplier cancels and
$\langle A^{\pi_k},\pi\rangle=0$. The divergence sum is at least
$\mu\|\pi'-\pi\|_1^2$, while
$\langle A^{\pi_k},\pi'-\pi\rangle\le\sigma_s\|\pi'-\pi\|_1$. Then
$T_{\pi_{k+1}}V^{\pi_k}\ge V^{\pi_k}$ and monotonicity of the Bellman operator
give improvement; the KL comparison is Theorem~\ref{thm:sharp-simplex-geometry}.
\end{proof}

\begin{proposition}[Approximate-advantage error certificate]
\label{prop:approximate-Euler-policy}
Suppose $\|\widehat A(s,\cdot)-A^{\pi_k}(s,\cdot)\|_\infty\le e_s$ and a feasible
$\widetilde\pi(\cdot|s)$ {has objective value within $\zeta_s\ge0$ of the maximum in Eq.~\eqref{eq:euler-mdpo} after replacing $A^{\pi_k}$ by $\widehat A$. Thus $\zeta_s$ is measured in advantage units. If the numerical solver instead bounds the gap in the objective multiplied by $t_k$, divide that bound by $t_k$.} Then
$\langle A^{\pi_k}(s,\cdot),\widetilde\pi(\cdot|s)\rangle\ge-t_ke_s^2/(2\mu)-\zeta_s$ and
\begin{equation}
 V^{\widetilde\pi}-V^{\pi_k}\ge-\frac{\max_s\{t_ke_s^2/(2\mu)+\zeta_s\}}{1-\gamma}\1 .
 \label{eq:approximate-policy-value}
\end{equation}
\end{proposition}
\begin{proof}
Compare the approximate prox objective with its value at $\pi_k$, bound
$|\langle\widehat A-A^{\pi_k},\widetilde\pi-\pi_k\rangle|$ by $e_sd_s$ and use
$D\ge\mu d_s^2/2$; maximize over $d_s$. {Let $\bP_{\widetilde\pi}$ be the state-transition matrix under the candidate policy. The performance difference is obtained by multiplying the vector of statewise advantage gains by $(\bI-\gamma\bP_{\widetilde\pi})^{-1}=\sum_{j\ge0}\gamma^j\rev{\bP_{\widetilde\pi}^{j}}$. This nonnegative matrix has row sums $1/(1-\gamma)$, which gives the stated componentwise bound.}
\end{proof}
The ratio $t_k/\mu$ thus controls both the maximum policy change and the
sensitivity to advantage error. For any feasible candidate $\bp$, the direct
test
\begin{equation}
 L_s(\bp)=\langle\widehat A(s,\cdot),\bp-\pi_k(\cdot|s)\rangle-e_s\|\bp-\pi_k(\cdot|s)\|_1\ge0
 \label{eq:stage3-direct-policy-test}
\end{equation}
{certifies $\langle A^{\pi_k}(s,\cdot),\bp\rangle\ge0$ directly for the feasible candidate. In the test, $e_s$ bounds the advantage error and the norm measures the candidate's distance from the old policy. Algorithm~\ref{alg:euler-pmd} tries finitely many step sizes and retains the old action distribution whenever no candidate passes. The fallback has exactly zero advantage gain at that state.}
Its guarantee requires valid advantage-error bounds; an estimated uncertainty
without coverage does not establish them.

\begin{algorithm}[ht]
\caption{\textsc{EulerPMD}: certified policy mirror descent that accepts a verified gain or keeps the old policy}
\label{alg:euler-pmd}
\begingroup\small
\begin{algorithmic}[1]
\Require Finite discounted MDP with at least two actions, $0\le\gamma<1$; $-1<a<0<b\le1$, $\lambda>0$; positive policy $\pi_0$; trial steps $t_k>0$; solver tolerance $\epsilon>0$; integer $J\ge0$
\Ensure Each state uses a verified candidate or keeps its action distribution
\For{$k=0,1,\ldots$}
 \State For the frozen $\pi_k$, obtain estimates $\widehat A_k(s,\cdot)$ and bounds $e_{k,s}$ on their $\ell_\infty$ errors relative to the true advantages.
 \State Begin with $\pi_{k+1}\gets\pi_k$.
 \For{each state $s$}
  \State $\bp_0\gets\pi_k(\cdot|s)$; $\bu\gets\log_{a,b,\lambda}(\bp_0)$; $\bar A\gets\operatorname{ctr}(\widehat A_k(s,\cdot))$.
  \For{$j=0,\ldots,J$}
   \State $\eta\gets t_k2^{-j}$; call $\mathsf{Norm}(\bu+\eta\bar A;\epsilon)$ (centering changes only the multiplier and improves conditioning).
   \State If the call fails, continue to the next trial.
   \State Let $\bp$ be the returned decision; $\bd\gets\bp-\bp_0$.
   \State Test a verified lower bound on $\langle\widehat A_k(s,\cdot),\bd\rangle-e_{k,s}\|\bd\|_1$.
   \State If the bound is nonnegative, set $\pi_{k+1}(\cdot|s)\gets\bp$ and exit the trial loop.
  \EndFor
 \EndFor
\EndFor
\Guarantee If the stated advantage bounds hold, $V^{\pi_{k+1}}\ge V^{\pi_k}$ componentwise. The fallback has zero displacement and zero true gain.
\end{algorithmic}
\endgroup
\end{algorithm}

\subsection{Euler softmax policies}
\label{subsec:Euler-softmax}
\begin{corollary}[Euler softmax policy update]
\label{cor:Euler-softmax-update}
Under the assumptions of Section~\ref{subsec:MDPO}, with $n=|\mathcal A|$ and
logits satisfying $\pi_k(\cdot|s)=\sigma_{a,b,\lambda,1}(\boldsymbol\theta_k(s,\cdot))$,
the exact step~\eqref{eq:euler-mdpo-closed} is
\begin{equation}
 \boldsymbol\theta_{k+1}(s,\cdot)=\boldsymbol\theta_k(s,\cdot)+t_kA^{\pi_k}(s,\cdot),\qquad
 \pi_{k+1}(\cdot|s)=\sigma_{a,b,\lambda,1}(\boldsymbol\theta_{k+1}(s,\cdot)).
 \label{eq:Euler-softmax-update}
\end{equation}
{From a uniform $\pi_0$, let $Y_K(s)=\sum_{k<K}t_k\operatorname{hr}(A^{\pi_k}(s,\cdot))$. Each action probability at state $s$ is bounded below by $\exp_{a,b,\lambda}(\log_{a,b,\lambda}(1/n)-2Y_K(s))$. As $Y_K(s)\to\infty$, this lower bound is asymptotic to $[2|a|Y_K(s)/\alpha_\lambda]^{-1/|a|}$. In the logarithmic limit the corresponding lower bound is $n^{-1}e^{-2Y_K(s)}$.}
\end{corollary}
\begin{proof}
The multiplier absorbs the constant in $\log_{a,b,\lambda}(\pi_k)=\boldsymbol\theta_k-\nu_k\1$.
Logit differences change by at most $2t_k\sigma_{s,k}$ per step, and some action
has probability at least $1/n$; use the left tail of
Proposition~\ref{prop:euler-tails}.
\end{proof}

\begin{proposition}[Euler softmax policy gradient]
\label{prop:Euler-softmax-PG}
For the tabular parameterization
$\pi_\theta(\cdot|s)=\sigma_{a,b,\lambda,1}(\boldsymbol\theta(s,\cdot))$, initial law
$\rho$ and normalized discounted occupancy $d^{\pi_\theta}_\rho$,
\begin{equation}
 \frac{\partial V^{\pi_\theta}(\rho)}{\partial\theta(s,\alpha)}
 =\frac{d^{\pi_\theta}_\rho(s)}{1-\gamma}m_{s,\alpha}\bigl[Q^{\pi_\theta}(s,\alpha)-\bar Q(s)\bigr],\qquad
 \bar Q(s)=\frac{\sum_\beta m_{s,\beta}Q^{\pi_\theta}(s,\beta)}{\sum_\beta m_{s,\beta}},
 \label{eq:Euler-softmax-PG}
\end{equation}
{Here $m_{s,\alpha}=m_{a,b,\lambda}(\pi_\theta(\alpha|s))$, $V^{\pi_\theta}(\rho)=\sum_s\rho(s)V^{\pi_\theta}(s)$, and $d^{\pi_\theta}_\rho(s)=(1-\gamma)\sum_{k\ge0}\gamma^k\Pr_{\rho,\pi_\theta}(S_k=s)$. The value baseline $\bar Q(s)$ is a mobility-weighted mean. In the logarithmic limit the mobilities equal the action probabilities and $\bar Q(s)=V^{\pi_\theta}(s)$.}
\end{proposition}
\begin{proof}
Combine the policy-gradient theorem with the Jacobian~\eqref{eq:Euler-softmax-Jacobian} at unit temperature.
\end{proof}
The vanilla gradient carries the factor
$m_{s,\alpha}\sim\pi_\theta(\alpha|s)^{1+|a|}/\alpha_\lambda$, which suppresses
rare-action logits more strongly than softmax; it suggests, but does not prove,
a rare-action slowdown~\cite{Agarwal2021,Mei2020}. The mirror update~\eqref{eq:Euler-softmax-update}
removes this factor from the dual step and is the form covered by the
improvement certificate. {For a state with positive occupancy, the categorical Fisher natural-gradient direction in logits has representative $d^{\rm Fisher}_\alpha=(\pi_\alpha/m_\alpha)A_\alpha$, up to a common shift and the overall factor $1/(1-\gamma)$ when the Fisher metric uses normalized occupancy. Indeed the unit-temperature Jacobian in Eq.~\eqref{eq:Euler-softmax-Jacobian} sends this representative to the probability-space vector $(\pi_\alpha A_\alpha)_\alpha$ since $\sum_\alpha\pi_\alpha A_\alpha=0$. The direction coincides with the Euler mirror direction up to this common scaling and shift in the logarithmic limit.} With the
entropy bonus $-\tau F_{a,b,\lambda}$, the regularized greedy policy is
$\sigma_{a,b,\lambda,\tau}(Q(s,\cdot))$; at $a=0$ it is sparse, as in
Tsallis-regularized MDPs~\cite{Lee2018sparseMDP}.

\subsection{Relative and generalized smoothness}
\label{subsec:gensmooth}
\begin{theorem}[Euclidean smoothness implies relative smoothness]
\label{thm:relative-smoothness-Euler}
Let $-1<a<0<\rev{b\le 1}$, $\lambda>0$, $\Lambda>0$, and let $L$ be convex and $C^2$ near
$\Delta_N$ with $\nabla^2L\preceq\Lambda\bI$. Then $\Lambda F_{a,b,\lambda}-L$
is convex on $\Delta_N$, and exact Euler MD~\eqref{eq:exact-Euler-simplex-MD}
with $\eta=1/\Lambda$ from $\bw_1=\1/N$ satisfies
\begin{equation}
 L(\bw_{T+1})-L(\bw^*)\le\frac{\Lambda D_{F_{a,b,\lambda}}(\bw^*\Vert\bw_1)}{T}
 \le\frac{\Lambda R_{a,b,\lambda}(N)}{T}.
 \label{eq:relsmooth-rate}
\end{equation}

\begingroup
Here $\bw^*$ is a minimizer on the closed simplex and $T$ counts updates from $\bw_1$. Relative smoothness means $L(\bu)\le L(\bw)+\langle\nabla L(\bw),\bu-\bw\rangle+\Lambda D_{F_{a,b,\lambda}}(\bu\Vert\bw)$ for an interior anchor $\bw$ and a feasible $\bu$. The bound applies to $\bw_{T+1}$ itself because objective descent lets its error be bounded by the mean of the errors after the $T$ updates.
\par\endgroup
\end{theorem}
\begin{proof}
On $(0,1]$, $h_{a,b,\lambda}\ge1$ (proof of Theorem~\ref{thm:Euler-MD-convergence}),
so $\nabla^2(\Lambda F_{a,b,\lambda}-L)\succeq\0$. Convexity, relative
smoothness and the three-point identity give
$\eta[L(\bw_{t+1})-L(\bw^*)]\le D(\bw^*\Vert\bw_t)-D(\bw^*\Vert\bw_{t+1})$, and
the step decreases $L$; sum~\cite{Bauschke2017,Lu2018}.
\end{proof}

\begin{corollary}[Relative smoothness for a boundary-singular objective]
\label{cor:Euler-relative-power}
For $\tau>0$, $\bc\in\mathbb R^N$ and
$L(\bw)=\langle\bc,\bw\rangle-\frac{\tau}{|a|(1+a)}\sum_iw_i^{1+a}$, the convex
function $L$ is $L_F$-smooth relative to $F_{a,b,\lambda}$ on the simplex with
$L_F=\tau/\alpha_\lambda$, and exact Euler MD with $\eta=1/L_F$ from the uniform
point satisfies $L(\bw_{T+1})-\min_{\Delta_N}L\le L_FR_{a,b,\lambda}(N)/T$.
\end{corollary}
\begin{proof}
$\nabla^2L=\tau\diag(w_i^{a-1})\preceq(\tau/\alpha_\lambda)\nabla^2F_{a,b,\lambda}$,
since $h_{a,b,\lambda}(w)\ge\alpha_\lambda w^{a-1}$; extend by continuity and
repeat the previous argument.
\end{proof}
The diversity-promoting power term has Euclidean curvature $\tau w^{a-1}$,
unbounded at the boundary, so no global Euclidean smoothness constant exists;
the retaining Euler component has the same singularity and absorbs it.
Entropic MD has curvature $1/w$ and an unbounded ratio $\tau w^a$. At
$\lambda=1$, $L_F=\tau(b-a)/|a|$; increasing the weight $\alpha_\lambda$ by
decreasing $\lambda$ reduces $L_F$ at the price of a larger radius. Extensions
to generalized smoothness~\cite{Yu2025MD,Zhang2020clipping} require checking
the norm, domain and step policy for the objective at hand.

\subsection{Practical selection and implementation}
\label{subsec:practical-online}
The strongest immediate applications have small or moderate simplex dimension
and a reason to control small weights: expert aggregation, ensemble and
mixture weights, constrained allocation, and tabular action distributions
(Table~\ref{tab:practical-constructions}). Store cumulative centered gradients
for FTRL, monitor the spread of the KKT residual $\log_{a,b,\lambda}(w_i)+\eta h_i$ (or
$\log_{a,b,\lambda}(w_i')-\log_{a,b,\lambda}(w_i)+\eta g_i$ for MD), use log-domain evaluation for
very small probabilities, and report solver evaluations and wall-clock time.
Choose a retention exponent $1/|a|$, an upper elasticity through $b$, and
then the scale by Eq.~\eqref{eq:scaled-optimal-lambda} or
Eq.~\eqref{eq:floor-curvature-design}; inspect the resulting $R/\mu$ and tune
against held-out metrics with equal budgets for all baselines.

\begin{table}[ht]
\centering\small
\caption{Selecting an Euler construction by its intended behavior.}
\label{tab:practical-constructions}
\begin{tabular}{@{}L{.16\textwidth}L{.21\textwidth}L{.26\textwidth}L{.26\textwidth}@{}}
\hline
Task & Construction & Core formula & Quantities to select or monitor\\\hline
Online expert or mixture weights & Adaptive full-simplex FTRL (Algorithm~\ref{alg:adaptive-ftrl}) & $\bw_{t+1}=\mathcal N_{a,b,\lambda}\bigl(-\eta_t\sum_{s\le t}\operatorname{ctr}(\bg_s)\bigr)$ & $R/\mu$ and $\lambda_*$, cumulative squared half-ranges, smallest weight, solver residual\\
Minimum-probability allocation & Adaptive OMD on $K_\epsilon$ & $w_{i,t+1}=\max\{\epsilon,\exp_{a,b,\lambda}(\cdot)\}$ & Floor $\epsilon$, Bregman diameter, comparator bias\\
Many-class prediction & Euler CE, posterior recovery or proper Euler score & $\mathcal L_{a,b,\lambda}=-\log_{a,b,\lambda}(q_y)$; $r_i=m_i/\sum_jm_j$ & Attenuation depth and location, accuracy, NLL, Brier score\\
Tabular policy learning & Euler softmax policy update~\eqref{eq:euler-mdpo-closed} & $\pi_{k+1}=\sigma_{a,b,\lambda,1}(\boldsymbol\theta_k+t_kA^{\pi_k})$ & $t_k/\mu$, advantage uncertainty, minimum action probability\\
Sparse experts or actions & Sparse FTRL or boundary softmax, $a=0$ & $w_i=[1+b(-\eta h_i-\nu)]_+^{1/b}$ & Threshold $1/(b\eta_t)$, support size, re-entry\\\hline
\end{tabular}
\end{table}

\section{Policy mirror descent at scale and value mirror descent}
\label{sec:new-policy-scale}
\begin{glance}
\textbf{Question.} What survives when policies are neural networks, and how does Euler geometry enter value-based planning?\\
\textbf{Key idea.} Exact statewise Euler steps become \emph{targets} that a network is trained to fit; value mirror descent alternates an Euler policy step with a single Bellman backup.\\
\textbf{Remember.} {For a neural policy, the value bound depends on the largest statewise fitting and advantage errors. For tabular planning with exact Bellman backups, a computable prox radius sets the value mirror-descent step size.} LLM-scale surrogates are collected in Appendix~\ref{app:supp-apps}.
\end{glance}

\subsection{From exact distribution updates to neural policies}
{Let $\bp_s^*$ be the exact maximizer of the statewise penalized objective in Eq.~\eqref{eq:euler-mdpo} using $\widehat A_k$. One can fit the explicit targets with $\min_\theta\mathbb E_{s\sim\chi_k}D_{F_{a,b,\lambda}}(\pi_\theta(\cdot|s)\Vert\bp_s^*)$, where $\chi_k$ is the chosen state or prefix sampling law. A shared neural policy may fit the targets unevenly~\cite{Sherman2025}. Write $\eta_k=t_k$ for the prox step size. At a positive exact target, its KKT identity shows that the unscaled prox objective gap at state $s$ is exactly $\zeta_s=D_{F_{a,b,\lambda}}(\pi_\theta(\cdot|s)\Vert\bp_s^*)/\eta_k$. If the statewise advantage error is also at most $e_s$, Proposition~\ref{prop:approximate-Euler-policy} yields}
\begin{equation}
 V^{\pi_\theta}-V^{\pi_k}\ge-\frac{\max_s\{\eta_ke_s^2/(2\mu)+\zeta_s\}}{1-\gamma}\1 ;
 \label{eq:new-neural-certificate}
\end{equation}
the maximum over states is demanding and is not supplied by a
trajectory-averaged training loss. Two regularization roles must be kept
separate: a prox penalty $D_{F_{a,b,\lambda}}(\pi\Vert\pi_k)/\eta_k$ controls one
step, whereas a persistent term $\tau F_{a,b,\lambda}(\pi)$ changes the target
MDP.

\paragraph{Large-language-model policies.} Three proposed constructions for LLM policy optimization are developed in Appendix~\ref{subsec:new-opmd}--\ref{subsec:new-gbmpo}: an Euler version of online policy mirror descent, whose regression target is a \emph{difference} of deformed logarithms; a ratio-based penalty that keeps the categorical metric at the reference (Proposition~\ref{prop:new-two-metrics}); and an analytic Euler instance of group-based policy optimization with unbiased sampled penalties. They are proposals, not established improvements.

\subsection{Value mirror descent with an Euler policy kernel}
\label{sec:new-vmd}
VMD~\cite{JiaLan2026VMD} performs a policy prox step with a current value
iterate followed by one Bellman backup. In reward notation, with
$0\le r\le R_{\max}$, $0<\gamma<1$ and
$Q_{V_t}(s,\alpha)=r(s,\alpha)+\gamma\sum_{s'}P(s'|s,\alpha)V_t(s')$,

\begin{empheq}[box=\fbox]{equation}
 \pi_{t+1}(\alpha|s)=\exp_{a,b,\lambda}\bigl(\log_{a,b,\lambda}(\pi_t(\alpha|s))+\eta_tQ_{V_t}(s,\alpha)-\nu_{t,s}\bigr),\qquad
 V_{t+1}(s)=\sum_\alpha\pi_{t+1}(\alpha|s)Q_{V_t}(s,\alpha).
 \label{eq:new-vmd-euler}
\end{empheq}

\begingroup
Here $P(s'|s,\alpha)$ is the transition probability, $\pi_{t+1}$ is computed using the frozen value vector $V_t$, and $\nu_{t,s}$ enforces unit mass. The second update applies one Bellman backup under $\pi_{t+1}$. For a value vector $V$, write $(TV)(s)=\max_\alpha Q_V(s,\alpha)$ and $(T_\pi V)(s)=\sum_\alpha\pi(\alpha|s)Q_V(s,\alpha)$.
\par\endgroup
Since $V_t$ need not equal $V^{\pi_t}$, the PMD improvement result does not
transfer by relabeling.

\begin{theorem}[Bellman consistency of basic Euler VMD]
\label{thm:new-vmd}
Assume exact transition expectations and exact prox solutions,
{$-1<a<0<b\le1$, $\lambda>0$, at least two actions, $\eta_t>0$, $V_0=0$ and $\pi_0>0$. With}
\begin{equation}
 B_t=\max_{s,\alpha}D_{F_{a,b,\lambda}}(\bolde_\alpha\Vert\pi_t(\cdot|s)),\qquad
 e_t=\|V^*-V_t\|_\infty,
 \label{eq:new-vmd-radius}
\end{equation}
every $B_t$ is finite, $0\le V_t\le V^*$, and
\begin{equation}
 e_{t+1}\le\gamma e_t+B_t/\eta_t .
 \label{eq:new-vmd-contraction}
\end{equation}
If $B_t/\eta_t\to0$ then $V_t\to V^*$; for $0<q<1$, $C>0$ and
$\eta_t\ge B_t/(Cq^t)$,
\begin{equation}
 e_t\le\gamma^te_0+C\sum_{j<t}\gamma^{t-1-j}q^j,
 \label{eq:new-vmd-rate}
\end{equation}
{For $t\ge1$, the bound equals $\gamma^te_0+C(\gamma^t-q^t)/(\gamma-q)$ when $q\ne\gamma$, and $\gamma^te_0+Ct\gamma^{t-1}$ when $q=\gamma$. The constant $q$ is the targeted decay factor of the greedy-backup error and $e_0=\|V^*\|_\infty$.}
\end{theorem}
\begin{proof}
Positivity follows from the infinite boundary slope, and finiteness of
$\phi_{a,b,\lambda}(0)$ gives $B_t<\infty$. Comparing the prox objective with
$\bolde_{\alpha^*}$ for a maximizing action gives
$0\le(TV_t)(s)-V_{t+1}(s)\le B_t/\eta_t$. Nonnegative rewards and induction give
$0\le V_{t+1}=T_{\pi_{t+1}}V_t\le TV_t\le TV^*=V^*$; the contraction of $T$ gives
Eq.~\eqref{eq:new-vmd-contraction}, and iteration the rate.
\end{proof}
The radius is attained at the least likely action:
\begin{equation}
 B_t=\max_s\Bigl\{(n-1)\phi_{a,b,\lambda}(0)-F_{a,b,\lambda}(\pi_t(\cdot|s))
 +\langle\pi_t(\cdot|s),\log_{a,b,\lambda}(\pi_t(\cdot|s))\rangle-\min_\alpha \log_{a,b,\lambda}(\pi_t(\alpha|s))\Bigr\},
 \label{eq:vmd-radius}
\end{equation}
{with $n=|\mathcal A|$. The term $-\min_\alpha\log_{a,b,\lambda}(\pi_t(\alpha|s))$ comes from evaluating the affine part of the Bregman divergence at a vertex of the simplex. Since $T_{\pi_{t+1}}V^*\ge T_{\pi_{t+1}}V_t=V_{t+1}$ and $T_{\pi_{t+1}}V^*\le V^*$, the Bellman residual at $V^*$ is at most $e_{t+1}$. Hence $\|V^*-V^{\pi_{t+1}}\|_\infty\le e_{t+1}/(1-\gamma)$.}
$B_t$ need not stay bounded as policies approach the boundary, so the schedule
may demand large steps; increasing $\eta$ helps only if the radius does not grow
faster. {With a persistent per-state reward penalty $-\tau\mathcal R_{a,b,\lambda}$ and $\tau\ge0$, $0\le V^*-V^*_\tau\le[\tau R/(1-\gamma)]\1$ componentwise. The bound follows because $0\le\mathcal R_{a,b,\lambda}(\pi(\cdot|s))\le R$ at every state and under every policy.}

\begin{proposition}[Certified inexact VMD and a current-policy stopping test]
\label{prop:stage3-vmd-certificate}
Under Theorem~\ref{thm:new-vmd} with feasible positive approximate prox solutions
whose normalization scores have full residuals $\epsilon_{t,s}$, put
$d_t=2\max_s\epsilon_{t,s}$. Then $0\le V_t\le V^*$ and
$e_{t+1}\le\gamma e_t+(B_t+d_t)/\eta_t$, so
$\eta_t=(\widehat B_t+2\epsilon_N)/(Cq^t)$ with verified $\widehat B_t\ge B_t$ and
$\epsilon_{t,s}\le\epsilon_N$ enforces the same forcing term. Independently of
the update, for any $V$ and $\pi$,
\begin{equation}
 \|V^*-V^\pi\|_\infty\le\frac{\|TV-V\|_\infty+\|T_\pi V-V\|_\infty}{1-\gamma}.
 \label{eq:stage3-current-policy-stop}
\end{equation}
\end{proposition}
\begin{proof}
The scaled prox objective has suboptimality at most $2\epsilon_{t,s}$, which
bounds the greedy error by $(B_t+d_t)/\eta_t$; repeat the contraction argument.
The stopping bound follows from the two contractions and the triangle inequality.
\end{proof}

\begin{algorithm}[ht]
\caption{\textsc{EulerVMD}: value mirror descent with certified prox errors and a Bellman stopping test}
\label{alg:euler-vmd}
\begingroup\small
\begin{algorithmic}[1]
\Require Finite known MDP $(P,r,\gamma)$ with $0<\gamma<1$, $0\le r\le R_{\max}$, $n\ge2$ actions; $-1<a<0<b\le1$, $\lambda>0$; $C>0$, $0<q<1$; solver tolerance $\epsilon_N>0$ and policy-error target $\varepsilon>0$
\Ensure Current $(\pi_t,V_t)$ and status \textsc{success} or \textsc{precision-limit}
\State $V_0\gets\0$; $\pi_0(\cdot|s)\gets\1/n$ for all states.
\For{$t=0,1,\ldots$}
 \State Freeze $Q_t(s,\alpha)\gets r(s,\alpha)+\gamma\sum_{s'}P(s'|s,\alpha)V_t(s')$.
 \State $[\bv_o]_s\gets\max_\alpha Q_t(s,\alpha)$; $[\bv_p]_s\gets\sum_\alpha\pi_t(\alpha|s)Q_t(s,\alpha)$.
 \State $B_{\rm stop}\gets(\|\bv_o-V_t\|_\infty+\|\bv_p-V_t\|_\infty)/(1-\gamma)$.
 \State If a verified upper bound on $B_{\rm stop}$ is at most $\varepsilon$, \Return $(\pi_t,V_t,\textsc{success})$.
 \State Bound $B_t=\max_{s,\alpha}\rev{D_{F_{a,b,\lambda}}}(\bolde_\alpha\Vert\pi_t(\cdot|s))$ from above by $\widehat B_t$.
 \State $\eta\gets(\widehat B_t+2\epsilon_N)/(Cq^t)$.
 \For{each state $s$}
  \State $\by_s\gets\log_{a,b,\lambda}(\pi_t(\cdot|s))+\eta[Q_t(s,\cdot)-[\bv_o]_s\1]$.
  \State Use $\mathsf{Norm}(\by_s;\epsilon_N)$ to get $\pi_{t+1}(\cdot|s)$.
  \State If certification fails, \Return $(\pi_t,V_t,\textsc{precision-limit})$.
 \EndFor
 \State $V_{t+1}(s)\gets\sum_\alpha\pi_{t+1}(\alpha|s)Q_t(s,\alpha)$ for all $s$.
\EndFor
\Guarantee On success, $\|V^*-V^{\pi_t}\|_\infty\le\varepsilon$. The forcing bound in Proposition~\ref{prop:stage3-vmd-certificate} assumes exact Bellman backups.
\end{algorithmic}
\endgroup
\end{algorithm}
{Centering $Q$ before multiplying by $\eta_t$ removes a common score offset, which the multiplier would otherwise absorb. The stopping test needs certified bounds on both Bellman residuals, including any rounding in the value backup.}
{In exact arithmetic with the stated forcing schedule, $V_t\to V^*$ and the on-policy residual of the returned candidate also vanishes: for $t\ge1$, $\|T_{\pi_t}V_t-V_t\|_\infty\le\gamma\|V_t-V_{t-1}\|_\infty$. Sampled transitions, stabilized and variance-reduced VMD}~\cite{JiaLan2026VMD},
and noisy Bellman backups need their own error terms; the theorem covers only
the algorithm it states.


\section{Learning in games: Euler self-play, Mirror-Prox, and equilibrium certificates}
\label{sec:phase2-games}\label{sec:games-extension}
\begin{glance}
\textbf{Question.} Does Euler geometry help learning in games?\\
\textbf{Key idea.} Geometry alone cannot stop cycling: a conserved Bregman energy shows that. Convergence must come from a mechanism, such as an extra gradient evaluation (Mirror-Prox) or a persistent anchor; Euler geometry then shapes retention and constants.\\
\textbf{Remember.} {For a monotone Lipschitz game operator and the stated step size, the averaged Mirror-Prox prediction has an $O(1/T)$ gap. With a fixed positive anchor weight, the last-iterate divergence contracts toward the regularized solution. The anchor introduces a separately bounded bias for the original game.}
\end{glance}

\subsection{Setting and the appropriate question}
Recent work studies regularization in self-play alignment~\cite{Tang2026RSPO},
and Anagnostides et al.~\cite{Anagnostides2026} show that FTRL with
permutation-invariant regularizers, a class containing the separable Euler
potential, can take exponential time in potential games. This rules out a
universal fast-convergence claim based only on choosing $(a,b,\lambda)$. The
precise question is whether an Euler kernel improves a specified stability or
adaptation property while the convergence mechanism is preserved.

Players $j=1,\ldots,M$ choose $\bp^j\in\Delta_{n_j}$ with their own triples
$(a_j,b_j,\lambda_j)$; {write $\bz=(\bp^j)_{j=1}^M\in K=\prod_j\Delta_{n_j}$ for the joint profile and $F_{\ba,\bb,\blambda}(\bz)=\sum_jF_{a_j,b_j,\lambda_j}(\bp^j)$ for the block potential. The game operator is $\mathcal G(\bz)=(\nabla_{\bp^j}L_j(\bz))_j$, where $L_j$ is player $j$'s expected cost and $\bg^j$ is its gradient in that player's block.} The interior mirror flow with
$\bM_j=\diag(m_{a_j,b_j,\lambda_j}(p_i^j))$ is
\begin{equation}
 \dot\bp^j=-\bM_j(\bg^j-c_j\1),\qquad c_j=\frac{\1^T\bM_j\bg^j}{\1^T\bM_j\1},
 \label{eq:p2-game-flow}
\end{equation}
a replicator-type flow in the entropic limit; for $|a_j|>0$ small
probabilities respond with mobility $\sim p^{1+|a_j|}/\alpha_{\lambda_j}$,
which preserves availability but can delay recovery.

\begingroup
The correction $c_j$ subtracts the mobility-weighted mean cost in block $j$. In particular, $\1^T\dot\bp^j=0$, so the flow preserves the total mass of each player\textquotesingle s strategy.
\par\endgroup

\begin{proposition}[Changing Euler geometry alone does not remove zero-sum cycling]
\label{prop:p2-conservation}
In a finite zero-sum game $\min_\bx\max_\by\bx^T\bA\by$ with a fully mixed
equilibrium $(\bx^*,\by^*)$, exact interior mirror flow with fixed Euler
potentials conserves
\begin{equation}
 \mathcal E(t)=D_{F_x}(\bx^*\Vert\bx(t))+D_{F_y}(\by^*\Vert\by(t)),
 \label{eq:p2-conserved-energy}
\end{equation}
so a trajectory with $\mathcal E(0)>0$ cannot converge to the equilibrium.
\end{proposition}
\begin{proof}
$\frac{\mathrm d}{\mathrm dt}D(\bu\Vert\bw)=\langle\nabla^2F(\bw)\dot\bw,\bw-\bu\rangle$;
the two players contribute $-\langle\bA\by,\bx-\bx^*\rangle$ and
$\langle\bA^T\bx,\by-\by^*\rangle$, whose sum
{$(\bx^*)^T\bA\by-\bx^T\bA\by^*$ vanishes because fully mixed equilibrium strategies make every opposing pure action yield the same game value. This argument conserves the energy along any interior trajectory and rules out its convergence to that equilibrium from positive initial energy.}
\end{proof}

\subsection{Euler Mirror-Prox and its anchored last-iterate version}
\label{subsec:anchored-mp}
{Assume $n_j\ge2$, $-1<a_j\le0<b_j\le1$ and $\lambda_j>0$ in every block. Let $\mathcal G$ be monotone and $\Lambda_G$-Lipschitz in the Euclidean norm, with $\Lambda_G>0$ a valid upper bound. Each diagonal curvature of the Euler Hessian is at least one on the positive simplex, so the block potential is $1$-strongly convex in the Euclidean norm on $K$. Initialize dense blocks positively. On sparse blocks $a_j=0$, use the finite right derivative $\log_{0,b_j}(0)=-1/b_j$ at zero.} Euler
Mirror-Prox~\cite{Nemirovski2004MP} uses a common anchor for two operator
evaluations:
\begin{equation}
 \widetilde\bz_t=\operatorname*{arg\,min}_{\bz\in K}\{\eta\langle\mathcal G(\bz_t),\bz\rangle+D(\bz\Vert\bz_t)\},\qquad
 \bz_{t+1}=\operatorname*{arg\,min}_{\bz\in K}\{\eta\langle\mathcal G(\widetilde\bz_t),\bz\rangle+D(\bz\Vert\bz_t)\},
 \label{eq:p2-mirrorprox-first}
\end{equation}
with $D=D_{F_{\ba,\bb,\blambda}}$.

\begin{theorem}[A monotone-game guarantee for Euler Mirror-Prox]
\label{thm:p2-mirrorprox}
With exact computations, {$0<\eta\le1/\Lambda_G$, $T\ge1$, the average $\overline\bz_T=T^{-1}\sum_{t=1}^T\widetilde\bz_t$, and initial radius $R_0=\sup_{\bw\in K}D(\bw\Vert\bz_1)<\infty$,}
\begin{equation}
 \operatorname{Gap}_D(\overline\bz_T):=\sup_{\bw\in K}\langle\mathcal G(\bw),\overline\bz_T-\bw\rangle\le\frac{R_0}{\eta T}.
 \label{eq:p2-mirrorprox-gap}
\end{equation}
For bilinear zero-sum games this bounds the saddle gap of the averaged
strategies; it is not a last-iterate or general-sum Nash guarantee.
\end{theorem}
\begin{proof}
Apply the three-point inequality to the second step with comparator $\bw$ and
to the first with comparator $\bz_{t+1}$. The cross term
$\eta\langle\mathcal G(\widetilde\bz_t)-\mathcal G(\bz_t),\widetilde\bz_t-\bz_{t+1}\rangle\le\eta\Lambda_G\rho_1\rho_2$
{is bounded by the sum of the two subtracted divergences, which is at least $\tfrac12(\rho_1^2+\rho_2^2)$ for $\rho_1=\|\widetilde\bz_t-\bz_t\|_2$ and $\rho_2=\|\widetilde\bz_t-\bz_{t+1}\|_2$.} Use monotonicity, telescope, and take the supremum.
\end{proof}

\begin{lemma}[Certified inexact Mirror-Prox]
\label{lem:stage3-mp-error}
{Under the hypotheses of Theorem~\ref{thm:p2-mirrorprox}, suppose the approximate block outputs are feasible, positive in dense blocks, and have full stationarity certificates $\epsilon^P_{t,j}$ and $\epsilon^C_{t,j}$. Evaluate $\mathcal G$ at those actual profiles. Then, with}
$\delta_t=2\sum_j(\epsilon^P_{t,j}+\epsilon^C_{t,j})$,
$\operatorname{Gap}_D(\overline\bz_T)\le(R_0+\sum_{t\le T}\delta_t)/(\eta T)$.
\end{lemma}
\begin{proof}
{In the first prox inequality the residual is paired with the displacement from $\widetilde\bz_t$ to $\bz_{t+1}$, and in the second with the displacement from $\bz_{t+1}$ to the comparator. After centering, its $\ell_\infty$ norm in each block is at most the certificate. The $\ell_1$ diameter of that block is two, so the two prox inequalities together incur at most $\delta_t$. The remaining proof telescopes as in the exact case.}
\end{proof}

\begin{algorithm}[ht]
\caption{\textsc{EulerMP}: Euler Mirror-Prox with two stages per round and a certified average}
\label{alg:euler-mp}
\begingroup\small
\begin{algorithmic}[1]
\Require $M$ simplex blocks $\Delta_{n_j}$, $n_j\ge2$; per-block triples $-1<a_j\le0<b_j\le1$, $\lambda_j>0$; monotone $\Lambda_G$-Lipschitz operator $\mathcal G$ in Euclidean norm; $0<\eta\le1/\Lambda_G$; horizon $T\ge1$; positive tolerances $\epsilon^P_{t,j},\epsilon^C_{t,j}$
\Ensure Average of prediction profiles and a dual-gap bound, or \textsc{precision-limit}
\State $\bp^j_1\gets\1/n_j$ for all $j$; $\bz_1\gets(\bp^j_1)_j$; $\overline\bz_0\gets\bz_1$.
\State $E_0\gets0$; $R_0\gets\sum_j R_{a_j,b_j,\lambda_j}(n_j)$.
\For{$t=1,\ldots,T$}
 \State $\bg\gets\mathcal G(\bz_t)$; $\bu^j\gets \rev{\log_{a_j,b_j,\lambda_j}(\bp^j_t)}$ for all $j$.
 \State On $a_j=0$, evaluate the link at zero by $\rev{\log_{0,b_j}(0)=-1/b_j}$.
 \State For every $j$, get $(\widetilde\bp^j_t,\cdot,\hat\epsilon^P_{t,j})$ from $\mathsf{Norm}_j(\bu^j-\eta\bg^j;\epsilon^P_{t,j})$.
 \State If any prediction call fails, \Return \textsc{precision-limit}.
 \State Assemble $\widetilde\bz_t\gets(\widetilde\bp^j_t)_j$; $\widetilde\bg\gets\mathcal G(\widetilde\bz_t)$.
 \State For every $j$, get $(\bp^j_{t+1},\cdot,\hat\epsilon^C_{t,j})$ from $\mathsf{Norm}_j(\bu^j-\eta\widetilde\bg^j;\epsilon^C_{t,j})$.
 \State If any correction call fails, \Return \textsc{precision-limit}.
 \State $\bz_{t+1}\gets(\bp^j_{t+1})_j$; $\overline\bz_t\gets\overline\bz_{t-1}+(\widetilde\bz_t-\overline\bz_{t-1})/t$.
 \State $E_t\gets E_{t-1}+2\sum_j(\hat\epsilon^P_{t,j}+\hat\epsilon^C_{t,j})$.
\EndFor
\State \Return $(\overline\bz_T,(R_0+E_T)/(\eta T),\textsc{success})$
\Guarantee The second output bounds the dual gap of the first (Lemma~\ref{lem:stage3-mp-error}). Both prox calls in a round use the same anchor coordinates $\bu^j$.
\end{algorithmic}
\endgroup
\end{algorithm}

The average guarantee does not imply last-iterate convergence. A persistent
anchor supplies the missing dissipation. Fix a positive reference
$\bq\in K^\circ$ and $\tau>0$, let
$\mathcal G_\tau(\bz)=\mathcal G(\bz)+\tau[\nabla F(\bz)-\nabla F(\bq)]$, and keep the
anchor inside both prox subproblems,
$\argmin_{\bz\in K}\{\eta\langle\mathcal G(\cdot),\bz\rangle+\eta\tau D(\bz\Vert\bq)+D(\bz\Vert\bz_t)\}$.
Each block is one normalization of a dual convex combination:
\begin{equation}
 \bp^j=\mathcal N_{a_j,b_j,\lambda_j}\Bigl(\frac{\log_{a_j,b_j,\lambda_j}(\bp^j_t)+\eta\tau \log_{a_j,b_j,\lambda_j}(\bq^j)-\eta\bg^j}{1+\eta\tau}\Bigr).
 \label{eq:amp-closed}
\end{equation}

\begin{theorem}[Linear last-iterate convergence of anchored Euler Mirror-Prox]
\label{thm:anchored-mp}
Let $-1<a_j<0<b_j\le1$, $\lambda_j>0$, $\mathcal G$ monotone and $\Lambda_G$-Lipschitz,
$0<\eta\le1/\Lambda_G$. The anchored variational inequality
{$\langle\mathcal G_\tau(\bp^*_\tau),\bw-\bp^*_\tau\rangle\ge0$ for every $\bw\in K$ has a unique, interior}
solution, and the iterates started at $\bz_1=\bq$ satisfy
\begin{equation}
 D(\bp^*_\tau\Vert\bz_{t+1})\le\frac{D(\bp^*_\tau\Vert\bz_t)}{1+\eta\tau},\qquad
 D(\bp^*_\tau\Vert\bz_T)\le\frac{R_\bq}{(1+\eta\tau)^{T-1}},
 \label{eq:amp-rate}
\end{equation}
with $R_\bq=\sup_{\bw\in K}D(\bw\Vert\bq)<\infty$. With
$G_{\max}=\sup_K\|\mathcal G\|_2$,
\begin{equation}
 \sup_{\bw\in K}\langle\mathcal G(\bw),\bp^*_\tau-\bw\rangle\le\tau R_\bq,\qquad
 \sup_{\bw\in K}\langle\mathcal G(\bw),\bz_T-\bw\rangle\le\tau R_\bq+G_{\max}\sqrt{2D(\bp^*_\tau\Vert\bz_T)}.
 \label{eq:amp-gap}
\end{equation}
\end{theorem}
\noindent\textit{The proof is given in Appendix~\ref{app:proofs}, p.~\pageref{prf:thm:anchored-mp}.}
The factor $(1+\eta\tau)^{-1}$ is the discrete counterpart of the continuous
rate of Theorem~\ref{thm:games-dissipation}; the price is the bias $\tau R_\bq$
{in the original game's gap. The reference may encode a pretrained policy. Increasing $\tau$ speeds the bound on the contraction of the anchored divergence and increases the upper bound on this bias.}

\begin{proposition}[Inexact anchored steps and a computable stopping bound]
\label{prop:stage3-anchored-error}
With $c=(1+\eta\tau)^{-1}$ and full certificates of the normalized scores in
Eq.~\eqref{eq:amp-closed}, $D(\bp^*_\tau\Vert\bz_{t+1})\le cD(\bp^*_\tau\Vert\bz_t)
+2\sum_j(\epsilon^P_{t,j}+\epsilon^C_{t,j})$. Hence
$U_1=\widehat R_\bq\ge R_\bq$, $U_{t+1}=cU_t+2\sum_j(\epsilon^P_{t,j}+\epsilon^C_{t,j})$
bound the equilibrium divergence and
$\operatorname{Gap}_D(\bz_t)\le\tau\widehat R_\bq+\overline G\sqrt{2U_t}$ for any
verified $\overline G\ge G_{\max}$, e.g.\
$\overline G=\|\mathcal G(\bq)\|_2+\Lambda_G\sqrt{2M}$, and
$R_\bq=\sum_j\max_iD_{F_{a_j,b_j,\lambda_j}}(\bolde_i\Vert\bq^j)$ by
Eq.~\eqref{eq:vmd-radius}.
\end{proposition}
\begin{proof}
Multiplying a normalized stationarity equation by $1+\eta\tau$ gives a residual
$(1+\eta\tau)\epsilon$ in the anchored subproblem; repeat the contraction proof
and divide. A convex function of its first argument is maximal at a vertex.
\end{proof}

\begin{algorithm}[ht]
\caption{\textsc{AnchoredEulerMP}: anchored Euler Mirror-Prox with a last-iterate certificate}
\label{alg:anchored-mp}
\begingroup\small
\begin{algorithmic}[1]
\Require Simplex blocks and operator as in Algorithm~\ref{alg:euler-mp}, with $a_j<0$ in every block; positive reference $\bq$; $0<\eta\le1/\Lambda_G$, anchor strength $\tau>0$; $\epsilon_0>0$
\Require Verified $\widehat R_\bq\ge\sup_{\bw\in K}D_F(\bw\Vert\bq)$ and $\overline G\ge\sup_{\bw\in K}\|\mathcal G(\bw)\|_2$; target $\varepsilon>\tau\widehat R_\bq$
\Ensure Current profile and a dual-gap bound on success, or \textsc{precision-limit}
\State $\bz_1\gets\bq$; $c\gets(1+\eta\tau)^{-1}$; $U_1\gets\widehat R_\bq$.
\For{$t=1,2,\ldots$}
 \State $B_t\gets\tau\widehat R_\bq+\overline G\sqrt{2U_t}$.
 \State If a verified upper bound $\widehat B_t$ on $B_t$ is at most $\varepsilon$, \Return $(\bz_t,\widehat B_t,\textsc{success})$.
 \State $\epsilon_t\gets\epsilon_0 c^{2(t-1)}/(4M)$; $\bg\gets\mathcal G(\bz_t)$.
 \State For each $j$, $\bs^j\gets\rev{\log_{a_j,b_j,\lambda_j}(\bp^j_t)+\eta\tau\log_{a_j,b_j,\lambda_j}(\bq^j)}$.
 \State For each $j$, get $(\widetilde\bp^j_t,\cdot,\hat\epsilon^P_{t,j})$ from $\mathsf{Norm}_j(c[\bs^j-\eta\bg^j];\epsilon_t)$.
 \State If any prediction call fails, \Return \textsc{precision-limit}.
 \State $\widetilde\bz_t\gets(\widetilde\bp^j_t)_j$; $\widetilde\bg\gets\mathcal G(\widetilde\bz_t)$.
 \State For each $j$, get $(\bp^j_{t+1},\cdot,\hat\epsilon^C_{t,j})$ from $\mathsf{Norm}_j(c[\bs^j-\eta\widetilde\bg^j];\epsilon_t)$.
 \State If any correction call fails, \Return \textsc{precision-limit}.
 \State $\bz_{t+1}\gets(\bp^j_{t+1})_j$; $U_{t+1}\gets cU_t+2\sum_j(\hat\epsilon^P_{t,j}+\hat\epsilon^C_{t,j})$.
\EndFor
\Guarantee $U_t$ bounds the divergence to the anchored solution: $D_F(\bp^*_\tau\Vert\bz_t)\le U_t\le c^{t-1}\bigl(\widehat R_\bq+\epsilon_0/(c(1-c))\bigr)\to0$. The returned gap bound is for the original game (Proposition~\ref{prop:stage3-anchored-error}).
\end{algorithmic}
\endgroup
\end{algorithm}
The tolerance schedule makes $U_t\to0$ geometrically, so the stopping test
eventually succeeds for any $\varepsilon>\tau\widehat R_\bq$; a warm restart from
$\bz\ne\bq$ must reset $U_1$ to $\sup_\bw D(\bw\Vert\bz)$. For a fixed Euler
softmax with temperature $\tau$, adding the two variational inequalities gives
the one-step score stability
{$\|\bp(\bz+\bolde)-\bp(\bz)\|\le\|\bolde\|_*/(\tau\mu)$, where $\mu$ is the strong-convexity modulus in the primal norm and $\|\cdot\|_*$ is its dual. For a full-range link, an infinitesimal score perturbation $\mathrm d\bz$ gives $\mathrm d\ln p_i=(m_i/p_i)(\mathrm dz_i-\sum_j m_j\mathrm dz_j/\sum_j m_j)/\tau$. Thus relative local sensitivity for a small coordinate scales as $p_i^{|a|}/(\tau\alpha_\lambda)$ times its centered score perturbation.}

\subsection{Equilibrium certificates and optimistic self-play}
{A joint distribution $\mathcal P$ on action profiles is an $\varepsilon$-coarse correlated equilibrium (CCE) if $\mathbb E_{\boldsymbol\alpha\sim\mathcal P}[c_j(\boldsymbol\alpha)-c_j(\alpha_j',\boldsymbol\alpha_{-j})]\le\varepsilon$ for every player $j$ and fixed alternative action $\alpha_j'$. Here $c_j$ is a pure-profile cost and $\boldsymbol\alpha_{-j}$ denotes the other players' actions. A correlated equilibrium also tests deviations that depend on the action recommended to the player.}

\begin{proposition}[Euler external regret yields an empirical CCE]
\label{prop:games-cce}
If each player's full-information linearized external regret is at most
$\mathcal B_j(T)$, then {$\overline{\mathcal P}_T=T^{-1}\sum_{t=1}^T\bigotimes_j\bp^j_t$} is an
$\varepsilon_T$-CCE with $\varepsilon_T=\max_j\max\{0,\mathcal B_j(T)\}/T$; for
fixed-step Euler FTRL,
$\mathcal B_j(T)=R_j/\eta_j+(\eta_j/2\mu_j)\sum_t(\sigma^j_t)^2$.
\end{proposition}
\begin{proof}
Multilinearity identifies $T$ times the CCE violation against $\alpha_j'$ with
$\sum_t\langle\bg^j_t,\bp^j_t-\bolde_{\alpha_j'}\rangle$;
use Theorem~\ref{thm:Euler-FTRL-regret}.
\end{proof}
In a zero-sum game the joint gap
$\operatorname{Gap}(\bx,\by)=\max_{\by'}\bx^T\bA\by'-\min_{\bx'}(\bx')^T\bA\by$ of the
averaged strategies is at most $[\mathcal B_x(T)+\mathcal B_y(T)]/T$; an
$O(T^{-1})$ gap must not be called $O(T^{-1})$ cumulative regret. Cautious
Optimism~\cite{Soleymani2025Cautious} adapts the pace non-monotonically; its
rates must be rechecked for an Euler implementation. Taking the regularization
weight from accumulated prediction errors as in
Theorem~\ref{thm:new-adaptive-oftrl} gives Algorithm~\ref{alg:selfplay}, {in which each update uses only the player's own forecast and full-information feedback. Aggregating the players' regret statistics gives the signed worst deviation gain}
\begin{equation}
 \hat\varepsilon_T=\max_j\frac1T\Bigl[\sum_{t\le T}\langle\bg^j_t,\bp^j_t\rangle-\min_\alpha\sum_{t\le T}g^j_{\alpha,t}\Bigr].
 \label{eq:selfplay-certificate}
\end{equation}

\begin{proposition}[Certificate and rate for optimistic Euler self-play]
\label{prop:selfplay}
{In Algorithm~\ref{alg:selfplay}, $\overline{\mathcal P}_T$ is} a $\max\{\hat\varepsilon_T,0\}$-CCE and
\begin{equation}
 \hat\varepsilon_T\le\max_j\frac1T\Bigl[\Bigl(\frac{R_{\bq^j}}{\eta_{0,j}}+C_{\eta_{0,j}}\Bigr)\sqrt{\delta+\widetilde E^j_T}+Q^j_T\Bigr],
 \qquad Q^j_T=\sum_t\operatorname{hr}(\bg^j_t)\min\{2,\hat\epsilon^j_t/\mu_j\},
 \label{eq:selfplay-rate}
\end{equation}
where $R_{\bq^j}=\max_{\bp}D_{F_{a_j,b_j,\lambda_j}}(\bp\Vert\bq^j)$ and $\mu_j=\mu_{a_j,b_j,\lambda_j}(n_j)$.
\end{proposition}
\begin{proof}
The anchored potential $D(\cdot\Vert\bq^j)$ is nonnegative, vanishes at its
minimum, has range $R_{\bq^j}$ and the same Hessian as $F_{a_j,b_j,\lambda_j}$,
which is all that the proof of Theorem~\ref{thm:new-adaptive-oftrl} uses; apply it
to the realized loss and forecast sequence and add
Lemma~\ref{lem:stage2-oracle-error}.
\end{proof}

\begin{algorithm}[ht]
\caption{\textsc{EulerSelfPlay}: certified optimistic Euler self-play with a finite budget}
\label{alg:selfplay}
\begingroup\small
\begin{algorithmic}[1]
\Require Finite normal-form cost game with $M$ players and each player's full cost vector against the opponents' current mixed actions; per-player $-1<a_j<0<b_j<1$, $\lambda_j>0$, positive reference $\bq^j\in\Delta_{n_j}$, rate $\eta_{0,j}>0$ and tolerances $\epsilon^j_t>0$; $\delta>0$, target $\varepsilon>0$, budget $T_{\max}\ge1$
\Ensure History defining the empirical joint play, a CCE certificate, and a status
\State $\bL^j\gets\0$, $E^j\gets0$, $c^j\gets0$ for all $j$; history $\mathcal H\gets\varnothing$; $e_0\gets+\infty$.
\For{$t=1,\ldots,T_{\max}$}
 \For{each player $j$}
  \State Choose pre-round hint $\widehat\bg^j_t$; $\eta\gets\eta_{0,j}/\sqrt{\delta+E^j}$.
  \State $\by^j\gets\rev{\log_{a_j,b_j,\lambda_j}(\bq^j)}-\eta(\bL^j+\widehat\bg^j_t)$.
  \State Obtain $\bp^j_t$ using $\mathsf{Norm}_j(\by^j;\epsilon^j_t)$.
  \State If the call fails, \Return $(\mathcal H,e_{t-1},\textsc{precision-limit})$.
 \EndFor
 \State Play $(\bp^j_t)_j$ simultaneously; append this profile to $\mathcal H$.
 \For{each player $j$}
  \State Observe the full cost vector $\bg^j_t$; $c^j\gets c^j+\langle\bg^j_t,\bp^j_t\rangle$.
  \State $\bL^j\gets\bL^j+\bg^j_t$; $E^j\gets E^j+\operatorname{hr}(\bg^j_t-\widehat\bg^j_t)^2$.
 \EndFor
 \State $e_t\gets\max\{0,\max_j(c^j-\min_i L_i^j)/t\}$.
 \State If $e_t\le\varepsilon$, \Return $(\mathcal H,e_t,\textsc{success})$.
\EndFor
\State \Return $(\mathcal H,e_{T_{\max}},\textsc{budget-exhausted})$
\Guarantee Average the product distributions of the stored profiles with equal weights. For a nonempty history this joint distribution is an $e_t$-coarse correlated equilibrium; an empty history carries no finite certificate.
\end{algorithmic}
\endgroup
\end{algorithm}
The certificate concerns time-averaged correlated play, not the last iterate;
{sampling from $\overline{\mathcal P}_T$ means drawing one round uniformly and then independently drawing each player's action using that round's marginals. If a solver fails before any completed round, the algorithm returns an empty history and no finite certificate.}

\paragraph{Regularized responses.} Prior-regularized Euler best responses, the dissipation they introduce in monotone flows (Theorem~\ref{thm:games-dissipation}), applications to preference games and multi-agent RL, and mixed or dilated Euler kernels are discussed in Appendix~\ref{subsec:supp-games}.

\section{Implicit bias and dual damping}
\label{sec:phase2-barrier}\label{sec:phase2-bridges}
\begin{glance}
\textbf{Question.} Which solution does an Euler method select when many fit the data, and how should regularization be applied?\\
\textbf{Key idea.} Exact Euler steps move only along directions allowed by the data, so the limit is a Bregman projection of the starting point; growing regularization acts as damping in dual coordinates.\\
{\textbf{Remember.} With nonnegative coordinate gradients, primal decay gives an exponential upper bound. With bounded gradients, damping dual displacement about a positive anchor gives a time-uniform positive lower bound.}
\end{glance}

{For $-1<a<0<b$ the potential is finite at zero and its right slope is $-\infty$. At $(a,b)=(-1,0)$ the scalar potential is $x-1-\ln x$, which generates the Burg divergence and is the log-barrier up to an affine term\rev{,}} whose stationarity equations are exactly the perturbed complementarity conditions of an interior-point method. This bridge, and the reason generic Euler potentials are not self-concordant, are discussed in Appendix~\ref{subsec:supp-barrier}.

\subsection{Implicit bias of Euler mirror flow and its discretization}
\label{subsec:p2-implicitbias}\label{subsec:implicit-md}
Mirror geometry controls which interpolating solution an optimization
trajectory selects~\cite{Sun2023Implicit,Jacobs2025Implicit}; a mixed-power
Euler potential is not homogeneous, so classification max-margin theorems cannot
be imported unchecked. For $L(\bw)=\frac12\|\bA\bw-\by\|_2^2$ on the positive
orthant, consider the flow
$\frac{\mathrm d}{\mathrm dt}\nabla F_{a,b,\lambda}(\bw)=-\bA^T(\bA\bw-\by)$,
$\bw(0)=\bw_0>\0$.

\begin{proposition}[Conditional Euler Bregman-projection bias]
\label{prop:p2-implicitprojection}
If the flow stays positive and converges to a positive $\bw_\infty$ with
$\bA\bw_\infty=\by$, then $\bw_\infty$ uniquely solves
\begin{equation}
 \min_{\bw>0:\,\bA\bw=\by}D_{F_{a,b,\lambda}}(\bw\Vert\bw_0).
 \label{eq:p2-projection-bias}
\end{equation}
\end{proposition}
\begin{proof}
For $\bd\in\ker\bA$, $\bd^T[\nabla F(\bw(t))-\nabla F(\bw_0)]$ is constant, hence
zero, so $\nabla F(\bw_\infty)-\nabla F(\bw_0)\in\operatorname{range}(\bA^T)$, the
optimality condition of Eq.~\eqref{eq:p2-projection-bias}.
\end{proof}
The selected solution depends on the triple and on the initialization through
$\log_{a,b,\lambda}(w_{0,i})$; convergence itself is not proved for every dataset.

\begin{proposition}[Euler mirror descent for positive interpolation]
\label{prop:implicit-md}
Let $-1<a<0<b\le1$, $\lambda>0$, $\bw_0\in(0,w_{\max}]^n$, $\|\bA\|_2>0$,
$0<\eta\le h_{a,b,\lambda}(w_{\max})/\|\bA\|_2^2$, and
\begin{equation}
 \bw_{t+1}=\min\bigl\{\exp_{a,b,\lambda}\bigl(\log_{a,b,\lambda}(\bw_t)-\eta\bA^T(\bA\bw_t-\by)\bigr),\,w_{\max}\bigr\}.
 \label{eq:implicit-md}
\end{equation}

\begingroup
The minimum is coordinatewise. Since $h_{a,b,\lambda}$ is nonincreasing, $h_{a,b,\lambda}(w_{\max})$ lower bounds the curvature on the box. The squared operator norm $\|\bA\|_2^2$ is the Euclidean smoothness constant of the least-squares objective. Their ratio bounds the relative-smoothness step.
\par\endgroup
(i) $L(\bw_t)$ is nonincreasing and
$L(\bw_T)-L(\bar\bw)\le D_{F_{a,b,\lambda}}(\bar\bw\Vert\bw_0)/(\eta T)$ for every
$\bar\bw\in(0,w_{\max}]^n$. {(ii) If the cap has not modified any proposal through the update producing $\bw_t$, then $\log_{a,b,\lambda}(\bw_t)-\log_{a,b,\lambda}(\bw_0)\in\operatorname{range}(\bA^T)$. This invariant holds over the entire uncapped prefix.} (iii) If
the cap is never active and $\bw_t\to\bw_\infty>0$ with $\bA\bw_\infty=\by$, then
$\bw_\infty$ is the Bregman projection~\eqref{eq:p2-projection-bias}.
\end{proposition}
\noindent\textit{The proof is given in Appendix~\ref{app:proofs}, p.~\pageref{prf:prop:implicit-md}.}
{As $q\downarrow0$, deleting a coordinate costs $D_{\phi_{a,b,\lambda}}(0\Vert q)=\int_0^q s h_{a,b,\lambda}(s)\,\mathrm ds\sim\alpha_\lambda q^{1+a}/(1+a)$. In the opposite direction, for a fixed $x>0$ the divergence $D_{\phi_{a,b,\lambda}}(x\Vert q)\sim x\alpha_\lambda q^{a}/|a|$ diverges, since its tangent term uses the slope at $q$. The order of the divergence arguments explains how a small initialization can bias the selected interpolator.}

\begin{algorithm}[ht]
\caption{\textsc{EulerMD-Fit}: exact Euler mirror descent for positive interpolation, with an explicit exit status}
\label{alg:implicit-md}
\begingroup\small
\begin{algorithmic}[1]
\Require Data $\bA\in\mathbb R^{m\times n}$, $\by\in\mathbb R^m$; $-1<a<0<b\le1$, $\lambda>0$; $\bw_0\in(0,w_{\max}]^n$ for finite $w_{\max}>0$; loss target $\varepsilon>0$; budget $T_{\max}\ge1$
\Ensure A tuple (iterate, loss, cap flag, status). The cap flag records whether a proposal was changed.
\State Set $L(\bw)=\frac12\|\bA\bw-\by\|_2^2$ and $\text{capped}\gets\text{false}$.
\State If $L(\bw_0)\le\varepsilon$, \Return $(\bw_0,L(\bw_0),\text{capped},\textsc{target-reached})$.
\State If $\bA=0$, \Return $(\bw_0,L(\bw_0),\text{capped},\textsc{zero-operator})$.
\State Choose $M\ge\|\bA\|_2^2$, for example $\|\bA\|_F^2$.
\State $\eta\gets h_{a,b,\lambda}(w_{\max})/M$; $u_{\max}\gets\log_{a,b,\lambda}(w_{\max})$.
\For{$t=0,\ldots,T_{\max}-1$}
 \State $\bg\gets\bA^T(\bA\bw_t-\by)$; $\bz\gets\log_{a,b,\lambda}(\bw_t)-\eta\bg$.
 \State If some $z_i>u_{\max}$, set $\text{capped}\gets\text{true}$.
 \State $w_{t+1,i}\gets \exp_{a,b,\lambda}(\min\{z_i,u_{\max}\})$ for each $i$.
 \State If $L(\bw_{t+1})\le\varepsilon$, \Return $(\bw_{t+1},L(\bw_{t+1}),\text{capped},\textsc{target-reached})$.
\EndFor
\State \Return $(\bw_{T_{\max}},L(\bw_{T_{\max}}),\text{capped},\textsc{budget-exhausted})$
\Guarantee In exact arithmetic, Proposition~\ref{prop:implicit-md} gives the capped fitting rate. The dual invariant holds throughout an uncapped prefix.
\end{algorithmic}
\endgroup
\end{algorithm}
For a rounded uncapped update with dual defect $\mathbf e_t$, the invariant
becomes $\log_{a,b,\lambda}(\bw_T)-\log_{a,b,\lambda}(\bw_0)=-\eta\sum_{t<T}\bA^T(\bA\bw_t-\by)+\sum_{t<T}\mathbf e_t$;
even summable defects can change the limiting projection.

The Euclidean gradient flow in $\theta_i$ with $w_i=\chi(\theta_i)$ matches the
Euler mirror flow exactly when $\theta(w)=\int^w\sqrt{h_{a,b,\lambda}(s)}\,\mathrm ds$.
Near zero, $\theta(w)-\theta(0^+)\sim\frac2{1+a}\sqrt{\alpha_\lambda}\,w^{(1+a)/2}$,
finite although $\log_{a,b,\lambda}(w)\to-\infty$, while at large $w$,
$\theta(w)\sim\frac2{1+b}\sqrt{\beta_\lambda}\,w^{(1+b)/2}$; at the Burg endpoint
both distances are infinite. The scale thus redistributes the two equivalent
power reparameterizations, which remain coupled through one potential.

\subsection{Dual averaging as damping, and where decay should act}
\label{subsec:p2-dual-damping}\label{subsec:p2-decay}
For continuous anchored FTRL on the positive orthant,
$\bw(t)=\argmin_{\bw>0}\{\langle\bs(t),\bw\rangle+\beta(t)D_{F_{a,b,\lambda}}(\bw\Vert\bw_0)\}$
with $\dot\bs=\bg$, the dual displacement
$\bz=\log_{a,b,\lambda}(\bw)-\log_{a,b,\lambda}(\bw_0)$ satisfies
\begin{equation}
 \beta\bz=-\bs,\qquad\dot\bz=-\frac{\bg}{\beta}-\frac{\dot\beta}{\beta}\bz .
 \label{eq:p2-dual-damping}
\end{equation}
The first term is a gradient step of size $1/\beta$; the second damps the
displacement at rate $\dot\beta/\beta$ ($\gamma/t$ for $\beta=\beta_0t^\gamma$).
{For differentiable $\beta(t)>0$ and $\bg(t)=\nabla L(\bw(t))$, the resulting mirror flow has speed $1/\beta(t)$ and instantaneous objective $L+\dot\beta(t)D_{F_{a,b,\lambda}}(\cdot\Vert\bw_0)$. This gives the continuous counterpart of the stabilization in Lemma~\ref{lem:audit-exact-ftrl}.} It matters whether decay acts on weights or
on dual coordinates:
\begin{equation}
 \dot\bw=-\bH_{a,b,\lambda}^{-1}\nabla L(\bw)-\kappa\bw
 \quad\text{versus}\quad
 \dot\bw=-\bH_{a,b,\lambda}^{-1}\{\nabla L(\bw)+\kappa[\nabla F(\bw)-\nabla F(\bw_0)]\},
 \label{eq:p2-two-decays}
\end{equation}
{with $\bH_{a,b,\lambda}=\nabla^2F_{a,b,\lambda}$ and decay coefficient $\kappa>0$. For the primal-decay flow, the equivalent objective on the orthant is $L(\bw)+\kappa\sum_i Q(w_i)$, where $Q'(w)=wh_{a,b,\lambda}(w)=\omega_{a,b,\lambda}(w)$. Its scalar primitive is}
\begin{equation}
 Q(w)=\frac{\alpha_\lambda}{1+a}w^{1+a}+\frac{\beta_\lambda}{1+b}w^{1+b},\qquad
 Q''(w)=a\alpha_\lambda w^{a-1}+b\beta_\lambda w^{b-1},
 \label{eq:p2-decay-penalty}
\end{equation}
concave below and convex above $q_m=r/\lambda$ of
Proposition~\ref{prop:omega-profile}; for entropic EG it is the $\ell_1$ penalty.

\begin{proposition}[Primal decay erases retention; dual damping preserves it]
\label{prop:decay-vs-damping}
Let $-1<a<0<b$, $\lambda,\kappa>0$, and let $\bw(t)$ solve one of the flows with
$\bw(0)=\bw_0>\0$. (i) If $\partial_iL\ge0$ along the first flow,
$w_i(t)\le w_{0,i}e^{-\kappa t}$. (ii) If $|\partial_iL|\le G$ along the second,
\begin{equation}
 \exp_{a,b,\lambda}(\log_{a,b,\lambda}(w_{0,i})-G/\kappa)\le w_i(t)\le \exp_{a,b,\lambda}(\log_{a,b,\lambda}(w_{0,i})+G/\kappa).
 \label{eq:damping-interval}
\end{equation}
(iii) If $L$ is convex, the second flow exists for all time and
$L+\kappa D(\cdot\Vert\bw_0)$ has a positive minimizer $\bw^*_\kappa$, then
$D(\bw^*_\kappa\Vert\bw(t))\le e^{-\kappa t}D(\bw^*_\kappa\Vert\bw_0)$.
\end{proposition}
\noindent\textit{The proof is given in Appendix~\ref{app:proofs}, p.~\pageref{prf:prop:decay-vs-damping}.}
The discrete exact prox step for the linearized loss plus the anchor penalty is
\begin{equation}
 \log_{a,b,\lambda}(\bw_{t+1})=\frac{\log_{a,b,\lambda}(\bw_t)+\eta\kappa \log_{a,b,\lambda}(\bw_0)-\eta\bg_t}{1+\eta\kappa},
 \label{eq:discrete-dual-damping}
\end{equation}
{on the positive orthant. On the simplex, subtract a common scalar from the right-hand dual coordinates before inversion and solve for unit mass, as in Eq.~\eqref{eq:amp-closed}.} The statement concerns Euler mirror methods; it does
not say that decoupled weight decay is wrong for Euclidean training.

\paragraph{Inexact steps.} {For simplex MD with a fixed step $\eta$, Appendix~\ref{subsec:p2-inexact} shows that centered KKT residual bounds $\epsilon_t$ add at most $(2/\eta)\sum_t\epsilon_t$ to regret. Choosing $\epsilon_t\le\epsilon_0\eta/t^2$ gives a total addition of at most $\pi^2\epsilon_0/3$} (Corollary~\ref{cor:certified-md}, Algorithm~\ref{alg:certified-md}); objective-gap errors require a different, transferred analysis.

\section{Dynamic sparsity and mixture-of-experts routing}
\label{sec:phase2-moe}
\begin{glance}
\textbf{Question.} Can Euler geometry produce sparse routing in mixture-of-experts layers?\\
\textbf{Key idea.} Shifting the potential by $\delta$ makes its slope at zero finite, so exact zeros appear, while the exponents and $\lambda$ still shape the active weights.\\
\textbf{Remember.} {Score gaps bound the active set before expert evaluation. Exact zeros save computation when the implementation dispatches tokens only to active experts.}
\end{glance}

In a mixture-of-experts (MoE) layer a router maps token scores $z_i(x)$ to
weights over $N$ experts. Three components must be kept apart: the
\emph{probability map}, the \emph{dispatch rule} deciding which experts run, and
the \emph{capacity mechanism} handling overload. Sparse probability maps
interact with learned score scales, so a sparsity-capable map does not by itself
reduce participating experts~\cite{Brogueira2026Routing}; population-level
balancing with convex potentials is studied in~\cite{Chen2026Balance}. The
unshifted Euler gate
$\bp(x)=\argmax_{\bp\in\Delta_N}\{\langle\bz(x),\bp\rangle-\tau F_{a,b,\lambda}(\bp)\}$
{can give entmax-type zeros at $a=0$. For $-1<a<0<b\le1$ and finite scores, the marginal cost of moving a coordinate away from zero is $-\infty$ in the potential, so a maximizing gate has positive mass in every coordinate.} Two coherent remedies are
a shifted gate and a two-level design.

\subsection{A shifted Euler gate with exact zeros}
\label{subsec:p2-shifted}
{Use $N\ge2$, $-1<a\le0<b\le1$, $\lambda>0$, and $\delta>0$. For a probability $p\in[0,1]$ define}
\begin{equation}
 \phi^\delta_{a,b,\lambda}(p)=\int_0^p\bigl[\log_{a,b,\lambda}(s+\delta)-\log_{a,b,\lambda}(\delta)\bigr]\,\mathrm ds,\qquad
 F^\delta_{a,b,\lambda}(\bp)=\sum_i\phi^\delta_{a,b,\lambda}(p_i)
 =D_{F_{a,b,\lambda}}(\bp+\delta\1\Vert\delta\1),
 \label{eq:p2-shifted-potential}
\end{equation}
{The shifted scalar potential has derivative zero at the origin and second derivative $h_{a,b,\lambda}(p+\delta)>0$. On $[0,1]$, the curvature is between $h_{a,b,\lambda}(1+\delta)$ and $h_{a,b,\lambda}(\delta)$. The last equality expresses the shifted potential as an ordinary Euler Bregman divergence in translated coordinates.}

\begin{proposition}[Sparse shifted-Euler normalization and singleton threshold]
\label{prop:p2-shifted-gate}
For $\tau>0$ the unique maximizer of $\langle\bz,\bp\rangle-\tau F^\delta_{a,b,\lambda}(\bp)$
over $\Delta_N$ is
\begin{equation}
 p_i=\begin{cases}\exp_{a,b,\lambda}\bigl(\log_{a,b,\lambda}(\delta)+(z_i-\nu)/\tau\bigr)-\delta,&z_i>\nu,\\0,&z_i\le\nu,\end{cases}
 \qquad\sum_ip_i=1,
 \label{eq:p2-shifted-gate}
\end{equation}

\begingroup
Here $z_i$ is the score of expert $i$, $\tau$ multiplies the regularizer and $\nu$ is the multiplier in score units. The mass sum is continuous in $\nu$ and strictly decreasing wherever it is positive. The equation therefore determines a unique $\nu$ and a unique routing vector.
\par\endgroup
{The support is a singleton exactly when}
$z_{(1)}-z_{(2)}\ge\tau[\log_{a,b,\lambda}(1+\delta)-\log_{a,b,\lambda}(\delta)]$.
\end{proposition}
\begin{proof}
KKT gives $z_i-\nu=\tau[\log_{a,b,\lambda}(p_i+\delta)-\log_{a,b,\lambda}(\delta)]$ on the support and
$z_i\le\nu$ off it. A singleton support puts unit mass on a top-score
coordinate, which fixes $\nu$.
\end{proof}
Let $\kappa_\delta(m)=\log_{a,b,\lambda}(m+\delta)-\log_{a,b,\lambda}(\delta)$, increasing in $m$, the
score gap per unit temperature at which an active expert carries mass $m$, and
$\kappa_\delta=\kappa_\delta(1)$.

\begin{proposition}[Guaranteed-active and guaranteed-inactive experts]
\label{prop:gate-support}
With $k$ active experts,
\begin{equation}
 z_{(1)}-\tau\kappa_\delta(1)\le\nu^*\le z_{(1)}-\tau\kappa_\delta(1/k)\le z_{(1)}-\tau\kappa_\delta(1/N).
 \label{eq:gate-twosided}
\end{equation}
Hence every expert with $z_i>z_{(1)}-\tau\kappa_\delta(1/N)$ is active and every
expert with $z_i\le z_{(1)}-\tau\kappa_\delta(1)$ is inactive. {For each fixed $m\in(0,1]$, as $\delta\downarrow0$, $\kappa_\delta(m)\to m^b/b$ if $a=0$, and $\kappa_\delta(m)\sim\alpha_\lambda\delta^{-|a|}/|a|$ if $a<0$. The thresholds decrease as $\delta$ increases, since $\partial_\delta\kappa_\delta(m)=h_{a,b,\lambda}(m+\delta)-h_{a,b,\lambda}(\delta)\le0$.}
\end{proposition}
\begin{proof}
The top expert has weight in $[1/k,1]$ and satisfies
$z_{(1)}-\nu^*=\tau\kappa_\delta(p_{(1)})$; $\kappa_\delta$ is increasing. The limit
uses $\log_{a,b,\lambda}(\delta)\sim-\alpha_\lambda\delta^{a}/|a|$.
\end{proof}
The shift therefore interpolates between ``always dense'' and ``readily sparse'',
with $|a|$ and $\alpha_\lambda$ controlling how fast; Table~\ref{tab:gate-thresholds}
gives typical values at $\lambda=1$. Because a trained router can rescale its
scores, these thresholds should be compared with observed score gaps.

\begin{table}[ht]
\centering\footnotesize\setlength{\tabcolsep}{3pt}
\caption{Score gaps, in units of $\tau$, that decide the support of the shifted gate with $N=8$ experts at $\lambda=1$: {a scaled gap strictly below $\kappa_\delta(1/8)$ certifies activity, and a scaled gap at least $\kappa_\delta(1)$ certifies inactivity.}}
\label{tab:gate-thresholds}
\begin{tabular}{llcccccc}
\hline
 & & \multicolumn{3}{c}{certainly active below $\kappa_\delta(1/8)$} & \multicolumn{3}{c}{certainly inactive above $\kappa_\delta(1)$}\\
$(a,b)$ & Threshold $\kappa_\delta(m)$ & $\delta=1$ & $\delta=0.1$ & $\delta=0.01$ & $\delta=1$ & $\delta=0.1$ & $\delta=0.01$\\\hline
$(0,0)$ & $\ln(1+m/\delta)$ & 0.118 & 0.811 & 2.603 & 0.693 & 2.398 & 4.615\\
$(0,0.5)$ & $2[(m+\delta)^{1/2}-\delta^{1/2}]$ & 0.121 & 0.316 & 0.535 & 0.828 & 1.465 & 1.810\\
$(-0.2,0.5)$ & $\frac{10}{7}[(m+\delta)^{0.5}-\delta^{0.5}-(m+\delta)^{-0.2}+\delta^{-0.2}]$ & 0.120 & 0.565 & 1.838 & 0.777 & 1.909 & 3.456\\
$(-0.5,0.5)$ & $(m+\delta)^{1/2}-\delta^{1/2}-(m+\delta)^{-1/2}+\delta^{-1/2}$ & 0.118 & 1.212 & 7.546 & 0.707 & 2.941 & 9.910\\\hline
\end{tabular}
\end{table}

On the active support $S$, differentiating the KKT equations gives, with
$\widetilde m_i=m_{a,b,\lambda}(p_i+\delta)$,
\begin{equation}
 \frac{\partial p_i}{\partial z_j}=\frac1\tau\Bigl(\widetilde m_i\delta_{ij}
 -\frac{\widetilde m_i\widetilde m_j}{\sum_{k\in S}\widetilde m_k}\Bigr),\qquad i,j\in S,
 \label{eq:p2-shifted-jacobian}
\end{equation}
{where $\delta_{ij}$ is the Kronecker delta. The derivative entries involving strictly inactive coordinates are zero. At a support boundary $z_i=\nu$, the classical Jacobian can fail to exist. Algorithm~\ref{alg:moe-step} uses the adjacent fixed-support derivative that treats threshold coordinates as inactive. Once the support is known, the score-side Jacobian--vector product costs $O(|S|)$ operations. The forward scan across all scores and expert execution are separate costs. In a fixed-support region the gate gradient to inactive scores is zero.}

\paragraph{Retention and load balancing.} A two-level router keeps a dense, never-zero expert-availability state learned by Euler FTRL and feeds it, together with capacity prices, into the sparse gate. {For frozen scores, the prices follow projected ascent on the dual of a fractional capacity problem, with step at most $\tau h_{a,b,\lambda}(1+\delta)/B$ for a batch of $B$ tokens (Proposition~\ref{prop:price-ascent}). The capacity constraints in that problem bound total routing mass.} The design, its bracket lemma and Algorithm~\ref{alg:moe-step} are in Appendix~\ref{subsec:p2-twolevel}.

\section{Euler potentials as designed Bregman geometries and comparison with Adam, Shampoo and Muon}
\label{sec:design-geometry}
\begin{glance}
\textbf{Question.} How good is an Euler potential as a designed geometry, and how does it relate to Adam, Shampoo and Muon?\\
\textbf{Key idea.} Euler curvature depends on \emph{where} the iterate is; Adam-type curvature depends on the \emph{history} of gradients. The two are complementary and combine in Euler-AdaGrad.\\
\textbf{Remember.} {On the simplex, the Euler kernel controls the response of a small probability to a gradient. For signed network weights, the shifted construction gives a coordinatewise update whose behavior can be compared with adaptive diagonal and matrix preconditioners.}
\end{glance}

A Bregman potential is a design object: its Hessian is the local metric, its
gradient difference from an anchor is the restoring force, and its inverse
Hessian is the preconditioner. Adam and Shampoo admit quadratic local step
models once their states are frozen, which does not identify their complete
recurrences with FTRL; Muon applies a nonlinear matrix orthogonalization and is
not a quadratic mirror map. Euler curvature depends on the \emph{position} of
the iterate, Adam-type curvature on the \emph{history} of gradients, and the two
can be combined.

\subsection{Design criteria and what the Euler family delivers}
\label{subsec:design-criteria}
{The six criteria are (C1) strong convexity, (C2) bounded curvature on regions separated from the boundary, (C3) steepness at zero, (C4) a gradient map onto all of dual space, (C5) curvature matching with the loss, and (C6) control of curvature change along a step. For an exact power curvature $h(x)=c x^{-\upsilon}$ on the positive axis with $c>0$ and $\upsilon>0$, the link has a divergent negative derivative at the boundary, namely $\phi'(0^+)=-\infty$ exactly when $\upsilon\ge1$. This follows by integrating $h$ near zero. The self-concordance ratio $|h'(x)|/h(x)^{3/2}$ is proportional to $x^{\upsilon/2-1}$. It is bounded near zero for $\upsilon\ge2$ and on the whole positive axis only for $\upsilon=2$. Entropy and the Burg barrier correspond to powers $1$ and $2$, respectively.}

\begin{theorem}[Euler potentials and the design criteria]
\label{thm:design-criteria}
Let $-1<a<0<b\le1$, $\lambda>0$ and $\omega_{\min}=K/D_\lambda$ of
Eq.~\eqref{eq:omega-min}.
\begin{enumerate}[label=(\roman*),leftmargin=2.2em]
\item (C3, C4) $\nabla F_{a,b,\lambda}=\log_{a,b,\lambda}$ is a smooth bijection
$(0,\infty)^N\to\mathbb R^N$; the Tsallis, Amari and log-barrier boundaries are
not.
\item (C2 vs C3) $h_{a,b,\lambda}(x)\sim\alpha_\lambda x^{-(1+|a|)}$ as $x\downarrow0$;
the boundary exponent can be placed anywhere in $(1,2)$.
\item (C1, C2) $h_{a,b,\lambda}$ is nonincreasing, so
$h_{a,b,\lambda}(w_{\max})\le h_{a,b,\lambda}(x)\le h_{a,b,\lambda}(x_0)$ on
$[x_0,w_{\max}]$; on the simplex $F_{a,b,\lambda}$ is
$\mu_{a,b,\lambda}(N)$-strongly convex in $\ell_1$; for $b=1$,
$h_{a,1,\lambda}=\alpha_\lambda x^{a-1}+\beta_\lambda\ge\beta_\lambda$ globally.
\item (C5) $t\mapsto\ln h_{a,b,\lambda}(e^t)$ is convex with slope
{$-\varepsilon_{a,b,\lambda}(e^t)$ increasing from $-(1-a)$ to $-(1-b)$. Changing $\lambda$ translates this slope profile in $t=\ln x$. For the log curvature itself there is also a vertical shift: $\ln h_{a,b,\lambda}(e^t)=\ln h_{a,b,1}(e^{t+\ln\lambda})-\ln h_{a,b,1}(\lambda)$.}
\item (C6) For $z=\log_{a,b,\lambda}(x)$, $z'=\log_{a,b,\lambda}(x')$,
\begin{equation}
 |\ln x'-\ln x|\le\frac{|z'-z|}{\omega_{\min}},\qquad
 |\ln h_{a,b,\lambda}(x')-\ln h_{a,b,\lambda}(x)|\le\frac{1-a}{\omega_{\min}}|z'-z|,
 \label{eq:dual-step-stability}
\end{equation}
uniformly on the positive axis.
\item (C6) On $[x_0,\infty)$,
$|h_{a,b,\lambda}'|\le2Mh_{a,b,\lambda}^{3/2}$ with
$M=(1-a)/(2\sqrt{x_0\omega_{\min}})$; at $(-1,0)$ this holds globally with $M=1$.
\end{enumerate}
\end{theorem}
\noindent\textit{The proof is given in Appendix~\ref{app:proofs}, p.~\pageref{prf:thm:design-criteria}.}
Part (v) is a first-order counterpart of self-concordance: if the dual step is
bounded, no coordinate and no curvature changes by more than a fixed factor,
wherever the iterate is; it is global exactly for full-range pairs and the
natural logarithm. On the Tsallis branch $\omega_{0,b}(x)=x^b\to0$ and
coordinates can reach zero in a finite dual step.  At the barrier endpoint, Proposition~\ref{prop:barrier-sc} strengthens part~(vi): for $a=-1$ and $0\le b\le1$, self-concordance holds on the whole positive axis with $M=\sqrt{1+b\lambda^{b+1}}$. Table~\ref{tab:design-scorecard}
summarizes.

\begin{table}[ht]
\centering\footnotesize\setlength{\tabcolsep}{3.5pt}\renewcommand{\arraystretch}{1.2}
\caption{Design criteria for separable potentials; ``box'' means on $[x_0,w_{\max}]$.}
\label{tab:design-scorecard}
\begin{tabular}{@{}L{.15\textwidth}L{.13\textwidth}L{.08\textwidth}L{.1\textwidth}L{.05\textwidth}L{.1\textwidth}L{.13\textwidth}L{.13\textwidth}@{}}
\hline
Potential & Curvature $h(x)$ & C1 & C2 & C3 & C4 & C5 profile & C6\\\hline
Quadratic (Adam, Shampoo) & $1$ & global & global & no & $\mathbb R$, no positivity & history-fitted & trivial\\
Entropy & $x^{-1}$ & box & box, $x^{-1}$ & yes & $\mathbb R$ & one slope & global, $e^\Delta$\\
Tsallis $(0,b)$ & $x^{b-1}$ & box & box & no & half-line & one slope & fails at $0$\\
Log barrier $(-1,0)$ & $x^{-2}$ & box & box, $x^{-2}$ & yes & half-line & one slope & self-concordant\\
Euler interior, $\lambda>0$ & $\alpha_\lambda x^{a-1}+\beta_\lambda x^{b-1}$ & box; simplex & box, $x^{-(1+|a|)}$ & yes & $\mathbb R$ & two slopes, movable & global, Eq.~\eqref{eq:dual-step-stability}\\
Euler with $b=1$ & $\alpha_\lambda x^{a-1}+\beta_\lambda$ & global, $\beta_\lambda$ & box & yes & $\mathbb R$ & Euclidean plus barrier & global\\\hline
\end{tabular}
\end{table}

\subsection{Choosing and adapting the parameters}
\label{subsec:adapting-exponents}
\emph{Matching.} If the diagonal loss curvature behaves like $x^{-\nu_0}$ for
{small and $x^{-\nu_\infty}$ for large coordinates, with $1<\nu_0<2$ and $0\le\nu_\infty<1$ for a finite boundary value, take}
\begin{equation}
 a=1-\nu_0,\qquad b=1-\nu_\infty,
 \label{eq:exponent-matching}
\end{equation}
{and use $\lambda$ to place the transition between the two regimes. The endpoint $\nu_0=2$ gives $a=-1$, for which the potential diverges at zero. Matching diagonal asymptotic powers motivates the choice; a matrix inequality on the full feasible region is needed for a relative-smoothness bound.}

\begin{proposition}[Relative smoothness for the Euclidean-plus-barrier kernel]
\label{prop:relsmooth-a1}
{Let $-1\le a<0$, $\lambda>0$ and $c_0,c_1\ge0$ with $c_0+c_1>0$. Assume $L$ is twice differentiable on the positive orthant and $\nabla^2L(\bw)\preceq\diag(c_0+c_1w_i^{-(1+|a|)})$ for all positive $\bw$. The matrix order compares curvature in every direction, including directions that combine several coordinates.} Then
$\nabla^2L\preceq L_{\rm rel}\nabla^2F_{a,1,\lambda}$ with
$L_{\rm rel}=\max\{c_0/\beta_\lambda,\,c_1/\alpha_\lambda\}$. If also
$\nabla^2L\succeq\diag(c_0'+c_1'w_i^{-(1+|a|)})$ with $c_0',c_1'>0$, then $L$ is
$\min\{c_0'/\beta_\lambda,c_1'/\alpha_\lambda\}$-strongly convex relative to
$F_{a,1,\lambda}$. For convex $L$, the exact mirror step $1/L_{\rm rel}$ gives the
last-iterate bound $L(\bw_T)-L(\bar\bw)\le L_{\rm rel}D_{F_{a,1,\lambda}}(\bar\bw\Vert\bw_0)/T$.
\end{proposition}
\begin{proof}
$\nabla^2F_{a,1,\lambda}=\diag(\beta_\lambda+\alpha_\lambda w_i^{-(1+|a|)})$ and
$c_0+c_1t\le\max\{c_0/\beta_\lambda,c_1/\alpha_\lambda\}(\beta_\lambda+\alpha_\lambda t)$;
the rate is the argument of Proposition~\ref{prop:implicit-md}~\cite{Bauschke2017,Lu2018}.
\end{proof}
{Least squares uses the constant curvature term. Burg-relative Poisson-type losses~\cite{Bauschke2017} motivate the singular term at $a=-1$. If $c_0,c_1>0$, the unique finite optimal scale for this bound satisfies $\alpha_\lambda=c_1/(c_0+c_1)$ and $\beta_\lambda=c_0/(c_0+c_1)$, giving $L_{\rm rel}=c_0+c_1$. When one coefficient is zero, the corresponding pure-power limit gives the infimum.}

\emph{Hessian matching for general exponents.} The same calculation extends to any pair $a<0<b$ and quantifies how much a mis-set scale costs.

\begin{proposition}[Exact Hessian matching and robustness to the scale]
\label{prop:hessian-match}
Let $a<0<b$ and let a separable objective have curvature $f''(x)=c_1x^{a-1}+c_2x^{b-1}$ on $(0,\infty)$ with $c_1,c_2>0$. With
\begin{equation}
 \lambda^\star=\Bigl(\frac{|a|\,c_2}{b\,c_1}\Bigr)^{1/(b-a)},
 \label{eq:lambda-match}
\end{equation}
$\nabla^2f=(c_1+c_2)\nabla^2F_{a,b,\lambda^\star}$ on the positive orthant, so the relative condition number is one. For $\lambda=\gamma\lambda^\star$, the relative condition number of $f$ with respect to $F_{a,b,\lambda}$ on $(0,\infty)$ equals $\max\{\gamma,\gamma^{-1}\}^{b-a}$, and on any subinterval it is at most this value.
\end{proposition}
\begin{proof}
$\alpha_\lambda/\beta_\lambda=(|a|/b)\lambda^{a-b}$ equals $c_1/c_2$ exactly at $\lambda^\star$. In general $f''/h_{a,b,\lambda}=(c_1+c_2u)/(\alpha_\lambda+\beta_\lambda u)$ with $u=x^{b-a}\in(0,\infty)$ is monotone between $c_1/\alpha_\lambda$ and $c_2/\beta_\lambda$, whose ratio is $c_1\beta_\lambda/(c_2\alpha_\lambda)=\gamma^{b-a}$.
\end{proof}
With exact matching, $f$ equals $(c_1+c_2)F_{a,b,\lambda^\star}$ plus an affine term, and one exact mirror step with rate $1/(c_1+c_2)$ reaches the minimizer; otherwise the relative-smoothness rates of~\cite{Bauschke2017,Lu2018} apply with $\kappa_{\rm rel}\le\max\{\gamma,\gamma^{-1}\}^{b-a}$. For $b=1$ this recovers the optimal scale of Proposition~\ref{prop:relsmooth-a1}. Table~\ref{tab:hessian-match} compares potentials on three composite curvature profiles.

\begin{table}[htbp]

\centering\footnotesize
\caption{Relative condition number $\kappa_{\rm rel}=\sup f''/h\,/\inf f''/h$ on the stated interval. The exact scales are $\lambda^\star=100$ $(a,b)=(-1,1)$ in~I, $\lambda^\star=12^{-1/4}\approx0.537$ $(-1,3)$ in~II and $\lambda^\star=100^{2/3}\approx21.5$ $(-\frac12,1)$ in~III; the mis-set scale is $2\lambda^\star$ in~I and~III and $1.5\lambda^\star$ in~II.}
\label{tab:hessian-match}
\begin{tabular}{@{}L{.14\textwidth}L{.16\textwidth}L{.19\textwidth}L{.22\textwidth}L{.19\textwidth}@{}}
\hline
Potential & Curvature $h(x)$ & I: $10^{-4}x^{-2}+1$, $x\in[10^{-6},1]$ & II: $0.2x^{-2}+0.05x^2$, $x\in[10^{-3},10^2]$ & III: $0.01x^{-3/2}+2$, $x\in[10^{-4},1]$\\\hline
Euclidean & $1$ & $1.00\times10^8$ & $1.00\times10^6$ & $4.98\times10^3$\\
Entropy & $x^{-1}$ & $5.00\times10^3$ & $2.01\times10^5$ & $18.2$\\
Burg barrier & $x^{-2}$ & $1.00\times10^4$ & $2.50\times10^7$ & $2.01\times10^4$\\
Euler, $\lambda=1$ & $\frac{|a|x^{a-1}+bx^{b-1}}{|a|+b}$ & $5.00\times10^3$ & $12.0$ & $67.0$\\
Euler, mis-set $\lambda$ & $\alpha_\lambda x^{a-1}+\beta_\lambda x^{b-1}$ & $4.00$ & $5.06$ & $2.82$\\
Euler, $\lambda=\lambda^\star$ & $\frac{c_1x^{a-1}+c_2x^{b-1}}{c_1+c_2}$ & $\mathbf{1}$ & $\mathbf{1}$ & $\mathbf{1}$\\\hline
\end{tabular}
\end{table}

\emph{Scheduling.} Exponents $|a|=b=1/\ln N$ with the scale $\lambda_*$ of
Eq.~\eqref{eq:scaled-optimal-lambda} {retain polynomial tails. Across the dimensions in Table~\ref{tab:near-optimal-lambda}, the sufficient geometry ratio $(R/\mu)/\ln N$ is about $1.045$--$1.157$, corresponding to a leading regret factor about $1.022$--$1.076$ after taking the square root. With $\lambda=1$, the geometry ratio is about $1.22$--$1.74$.}

\emph{Selecting online.} A portfolio of geometries with an exponential-weights
{master adds $O(\sqrt{T\ln M})$ regret for $M$ fixed candidates and master losses in $[0,1]$ (Eq.~\eqref{eq:new-meta-bound});} slowly varying parameters pay the
drift of Theorem~\ref{thm:new-drift}; gradient-based learning
(Section~\ref{sec:learn-parameters}) suits loss or link parameters selected on
validation data, but there is no theorem that it approaches the best optimizer
geometry.

\subsection{Combining both kinds of adaptivity: Euler-AdaGrad}
\label{subsec:euler-adagrad}
For signed coordinates, combine a history-dependent diagonal scale with the
shifted potential $\psi=\psi_{a,b,\lambda,\varrho}$ of
Eq.~\eqref{eq:new-signed-potential}, $k(w)=\psi'(w)=\operatorname{sign}(w)[\log_{a,b,\lambda}(\varrho+|w|)-\log_{a,b,\lambda}(\varrho)]$.
Fix $\varrho,w_{\max},\eta>0$ and declared gradient bounds $G_i>0$, and let
\begin{equation}
 d_{t,i}=\sqrt{G_i^2+\sum_{s<t}g_{s,i}^2},\qquad L_{t-1,i}=\sum_{s<t}g_{s,i},\qquad
 \psi_t(\bw)=\frac1\eta\sum_id_{t,i}\psi(w_i).
 \label{eq:euler-adagrad-potential}
\end{equation}
The ideal action minimizes $\langle\bL_{t-1},\bw\rangle+\psi_t(\bw)$ over
$[-w_{\max},w_{\max}]^N$:
\begin{empheq}[box=\fbox]{equation}
 w^*_{t,i}=-\operatorname{sign}(L_{t-1,i})\min\Bigl\{w_{\max},\,
 \exp_{a,b,\lambda}\Bigl(\log_{a,b,\lambda}(\varrho)+\frac{\eta|L_{t-1,i}|}{d_{t,i}}\Bigr)-\varrho\Bigr\}.
 \label{eq:euler-adagrad-update}
 \end{empheq}
This is adaptive dual averaging, not recursive AdaGrad, and uses no moving
averages or momentum.

\begin{theorem}[Exact and residual-controlled Euler-AdaGrad regret]
\label{thm:euler-adagrad}
{Under the preceding construction, take $-1<a\le0<b\le1$ and $\lambda>0$. Let $\ell_t$ be convex on the box, $\bg_t\in\partial\ell_t(\bw_t)$ and $|g_{t,i}|\le G_i$. Set $h_{\min}=h_{a,b,\lambda}(\varrho+w_{\max})>0$, and let the played coordinate have}
normalized residual $\sigma_{t,i}=|q_{t,i}|$ in the interior, $[q_{t,i}]_+$ at
$w_{\max}$ and $[-q_{t,i}]_+$ at $-w_{\max}$, where
$q_{t,i}=k(w_{t,i})+\eta L_{t-1,i}/d_{t,i}$. For every $\bu$ in the box,
\begin{equation}
 \sum_{t=1}^T[\ell_t(\bw_t)-\ell_t(\bu)]\le
 \sum_i\Bigl[\frac{\psi(u_i)}{\eta}+\frac{\eta}{h_{\min}}\Bigr]\sqrt{G_i^2+\sum_{t\le T}g_{t,i}^2}
 +\sum_{t,i}|g_{t,i}|\min\{2w_{\max},\sigma_{t,i}/h_{\min}\}.
 \label{eq:euler-adagrad-regret}
\end{equation}
\end{theorem}
\begin{proof}
$\psi\ge0$, $d_{t,i}$ is nondecreasing, and $\psi''\ge h_{\min}$ on the box, so
the nondecreasing-regularizer FTRL bound~\cite[Section~7]{Orabona2019} gives
$\frac1\eta\sum_i\psi(u_i)d_{T+1,i}+\frac\eta{2h_{\min}}\sum_{t,i}g_{t,i}^2/d_{t,i}$.
{For the predictable denominators, put $A_t=d_{t,i}^2$ and $x_t=g_{t,i}^2\le G_i^2$. The left rectangle $x_t/\sqrt{A_t}$ exceeds the integral of $A^{-1/2}$ on $[A_t,A_{t+1}]$ by at most $G_i^2(A_t^{-1/2}-A_{t+1}^{-1/2})$. Summing shows $\sum_t g_{t,i}^2/d_{t,i}\le2(d_{T+1,i}-G_i)+G_i^2(1/G_i-1/d_{T+1,i})\le2d_{T+1,i}$.} Strong monotonicity of $k$ plus the
box normal cone bounds $|w_{t,i}-w^*_{t,i}|$ by $\sigma_{t,i}/h_{\min}$; use
convexity.
\end{proof}
For $b<1$, $h_{\min}\to0$ as $w_{\max}\to\infty$, so the cap cannot simply be
dropped. As $\varrho\to\infty$ with $\eta=\eta'h_{a,b,\lambda}(\varrho)$ the update
tends to clipped quadratic dual averaging; at small evidence the magnitude is
$m_{a,b,\lambda}(\varrho)r+O(r^2)$ {where $r=\eta|L_{t-1,i}|/d_{t,i}$, with $m_{a,b,\lambda}(\varrho)\sim\varrho^{1+|a|}/\alpha_\lambda$ as $\varrho\downarrow0$ in the full-range interior. Scaling a coordinate's gradient history and its bound by the same positive factor preserves its exact update for that history;} a
pretrained network should be optimized in translated form
$\bw=\bw_0+\mathbf u$ rather than reset to zero.

\begin{algorithm}[ht]
\caption{\textsc{EulerAdaGrad}: residual-controlled signed Euler FTRL with AdaGrad scaling}
\label{alg:euler-adagrad}
\begingroup\small
\begin{algorithmic}[1]
\Require Dimension $N$ and horizon $T\ge1$; $-1<a\le0<b\le1$, $\lambda>0$; shift $\varrho>0$; box radius $w_{\max}>0$; rate $\eta>0$; bounds $G_i>0$ and tolerances $\epsilon_{t,i}>0$; online convex losses
\Ensure Played decisions and status \textsc{success}, \textsc{precision-limit} or \textsc{bound-violation}
\State Define $k(w)=\operatorname{sign}(w)[\log_{a,b,\lambda}(\varrho+|w|)-\log_{a,b,\lambda}(\varrho)]$ and $\kappa_{\max}=k(w_{\max})$.
\State Set $L_i\gets0$ and $v_i\gets G_i^2$ for all $i$.
\For{$t=1,\ldots,T$}
 \For{$i=1,\ldots,N$}
  \State $d_i\gets\sqrt{v_i}$; $r_i\gets\min\{\eta|L_i|/d_i,\kappa_{\max}\}$.
  \State Set $w_{t,i}\gets-\operatorname{sign}(L_i)u_i$, where $u_i\in[0,w_{\max}]$ solves $\log_{a,b,\lambda}(\varrho+u_i)-\log_{a,b,\lambda}(\varrho)=r_i$. Use $0$ or $w_{\max}$ at the endpoints.
  \State Form $q_i=k(w_{t,i})+\eta L_i/d_i$. Verify feasibility and the residual bound $\epsilon_{t,i}$ using $|q_i|$ inside the box, $[q_i]_+$ at its upper end, and $[-q_i]_+$ at its lower end.
  \State If verification fails after refinement, \Return the rounds through $t-1$ with \textsc{precision-limit}.
 \EndFor
 \State Play $\bw_t$; observe a subgradient $\bg_t$ of the round loss at $\bw_t$.
 \State If some $|g_{t,i}|>G_i$, \Return the played rounds through $t$ with \textsc{bound-violation}.
 \State $L_i\gets L_i+g_{t,i}$; $v_i\gets v_i+g_{t,i}^2$ for all $i$.
\EndFor
\State \Return the played rounds with \textsc{success}.
\Guarantee On success, Theorem~\ref{thm:euler-adagrad} applies with residual cost bounded by $\sum_{t,i}G_i\min\{2w_{\max},\epsilon_{t,i}/h_{a,b,\lambda}(\varrho+w_{\max})\}$.
\end{algorithmic}
\endgroup
\end{algorithm}

\subsection{Comparison with Adam, Shampoo and Muon}
\label{subsec:vs-adam}
AdaGrad~\cite{DuchiAdaGrad2011} accumulates squared gradients; Adam~\cite{Kingma2015}
uses bias-corrected moving averages and admits an FTRL reading~\cite{Zheng2017}; its original regret argument has known
counterexamples~\cite{Reddi2018Stage4}; AdamW decouples weight
decay~\cite{Loshchilov2019Stage4}. Shampoo~\cite{Gupta2018} and
K-FAC~\cite{MartensGrosse2015} use structured matrix preconditioners. Muon
{uses a Newton--Schulz-type iteration to approximate an orthogonalized momentum direction. If $\bM=\bU_r\boldsymbol\Sigma_r\bV_r^T$ is the compact singular value decomposition over the $r$ positive singular values, then $-\bU_r\bV_r^T$ is one minimizer of $\langle\bM,\bD\rangle_F$ over $\|\bD\|_{\rm op}\le1$. The minimum is $-\sum_{i=1}^r\sigma_i(\bM)$ by operator--nuclear norm duality. The practical iteration approximates this direction before layer-dependent scaling}~\cite{Jordan2024Muon,Liu2025MuonStage4}; such methods are steepest
descent under layerwise operator norms~\cite{Bernstein2024,Crawshaw2025}.

\begin{table}[ht]
\centering\footnotesize\setlength{\tabcolsep}{3.5pt}\renewcommand{\arraystretch}{1.2}
\caption{Adaptive preconditioners, norm-based directions and Euler geometries: distinct mechanisms.}
\label{tab:vs-adaptive}
\begin{tabular}{@{}L{.15\textwidth}L{.21\textwidth}L{.18\textwidth}L{.18\textwidth}L{.2\textwidth}@{}}
\hline
Method & Curvature depends on & Natural domain & Cost per step & Typical guarantee\\\hline
AdaGrad, Adam & gradient history & signed weights & $O(N)$ & data-dependent regret (AdaGrad)\\
Shampoo, K-FAC & gradient or activation covariances & weight matrices & matrix roots per layer & structured regret (Shampoo)\\
Muon & singular directions & weight matrices & a few matrix products & spectral steepest descent\\
Entropic MD & position, one slope & simplex & $O(N)$ & $\sqrt{T\ln N}$ regret\\
Euler MD/FTRL & position, two slopes, movable transition; $h_{a,b,\lambda}=\alpha_\lambda w^{a-1}+\beta_\lambda w^{b-1}$ & simplex, orthant, policies & $O(N)$ per solver sweep & regret, last-iterate, retention\\
Euler-AdaGrad & position and history: $h_{a,b,\lambda}(\varrho+|w|)$ and $d_{t,i}$ & signed box & $O(N)$ per solver sweep & Theorem~\ref{thm:euler-adagrad}\\\hline
\end{tabular}
\end{table}
An advantage of Euler geometry can be expected for variables that are
probabilities or positive quantities (policies, mixture, attention and routing
weights), where it adds tunable retention, exact boundary sparsity and the
bounded relative change~\eqref{eq:dual-step-stability}, and for problems with a
verified power-law curvature profile. None is claimed for generic hidden-layer
weights, whose difficulty is correlated anisotropic curvature addressed by
Shampoo and Muon; a fixed Euler potential also does not adapt to gradient
scale, which must be added by Algorithm~\ref{alg:adaptive-ftrl} or
Algorithm~\ref{alg:euler-adagrad}. A natural division of labour keeps a
structured optimizer for unconstrained matrices and uses Euler updates or links
for distributions. A spectral potential
$\operatorname{tr}\phi_{a,b,\lambda}(\mathbf X)$ on positive definite matrices
{would extend the scalar kernel to spectral coordinates~\cite{Lewis1996,Tsuda2005}. If a dual update is diagonal in a fixed eigenbasis, scalar inverse asymptotics apply to each eigenvalue. Retention and convergence for general noncommuting updates remain open.}

\section{A research and evaluation program}
\label{sec:new-experiments}
\begin{glance}
\textbf{Question.} What is proved, what is conditional, and what still needs experiments?\\
{\textbf{Remember.} The theory predicts specific differences in retention, curvature, sparsity and certificate accuracy. The proposed experiments measure when those differences improve outcomes under matched computational and tuning budgets.}
\end{glance}

\subsection{What is proved, what is conditional, and what is only proposed}
\label{subsec:status}
Table~\ref{tab:status} sorts the algorithmic results. {The theorems provide testable quantities: minimum weights, support thresholds, geometry-dependent rates and equilibrium or Bellman residuals. Experiments can measure those quantities alongside task performance.} The symmetry $a\leftrightarrow b$ means experiments should
use ordered exponents $a\le0<b$, and the scale should be searched in
log-coordinates, since $\lambda$ is unidentifiable on the single-power
boundaries.

\begin{table}[ht]
\centering\small
\caption{Status of the main algorithmic results of Sections~\ref{sec:Euler-MD}--\ref{sec:design-geometry}.}
\label{tab:status}
\begin{tabular}{L{.14\textwidth}L{.78\textwidth}}
\hline
Status & Results\\\hline
Proved &
FTRL--MD equivalences and the optimal sufficient scale (Section~\ref{sec:FTRL-3p}, Corollary~\ref{cor:ftrl3-regret}); certified normalization (Lemma~\ref{lem:certified-norm}); scale-free adaptive and optimistic regret (Theorems~\ref{thm:Euler-adaptive} and~\ref{thm:new-adaptive-oftrl}); schedule-free conversion (Theorem~\ref{thm:new-sf}); geometry drift (Theorem~\ref{thm:new-drift}); Euler VMD (Theorem~\ref{thm:new-vmd}); averaged and last-iterate Mirror-Prox (Theorems~\ref{thm:p2-mirrorprox} and~\ref{thm:anchored-mp}); self-play certificate (Proposition~\ref{prop:selfplay}); {capped interpolation rate and uncapped dual invariant (Proposition~\ref{prop:implicit-md});} decay versus damping (Proposition~\ref{prop:decay-vs-damping}); residual-based inexact MD (Corollary~\ref{cor:certified-md}); design criteria and Euler-AdaGrad (Theorems~\ref{thm:design-criteria} and~\ref{thm:euler-adagrad}); shifted-gate support (Propositions~\ref{prop:p2-shifted-gate}, \ref{prop:gate-support}); bandit Euler FTRL (Proposition~\ref{prop:euler-bandit}); exact self-concordance of barrier hybrids (Proposition~\ref{prop:barrier-sc}); Hessian matching with scale robustness (Proposition~\ref{prop:hessian-match}).\\
Conditional or transferred &
Monotone policy improvement under declared advantage-error bounds (Algorithm~\ref{alg:euler-pmd}); {projection selection conditional on convergence to a positive interpolator, and an uncapped trajectory in the discrete case (Propositions~\ref{prop:p2-implicitprojection} and~\ref{prop:implicit-md});} objective-error robustness transferred from~\cite{Schlisselberg2026}; capacity prices for frozen scores (Proposition~\ref{prop:price-ascent}).\\
Proposed, not analyzed &
neural OPMD and Euler-GBMPO surrogates; online learning of $(a,b,\lambda)$; staged reduction of the anchor weight; joint training of the two-level router; stochastic, delayed or neural feedback in games.\\\hline
\end{tabular}
\end{table}

\subsection{Experiments that would separate the mechanisms}
\label{subsec:experiments}
Each row of Table~\ref{tab:new-experiments} isolates one mechanism. For each proposed comparison,
{use matched tuning budgets, repeated runs and uncertainty estimates. For optimization, record both oracle calls and wall-clock time, and include validation-tuned steps or steps matched by a specified curvature constant. For links, report the chosen temperature and curvature at a reference distribution. For simplex regret, report $R/\mu$ separately from its square-root contribution to the regret bound.}

\begin{table}[ht]
\centering\small
\caption{Experiments that distinguish the proposed mechanisms.}
\label{tab:new-experiments}
\begin{tabular}{L{.2\textwidth}L{.35\textwidth}L{.35\textwidth}}
\hline
Study & Controls and baselines & Primary measurements\\\hline
Free scale & $\lambda=1$, $\lambda_*$, validation-tuned $\lambda$ at fixed $(a,b)$ & Regret, $R/\mu$, smallest weight, solver cost\\
Bandits & Exp3, Tsallis-INF, Hybrid~B, barrier hybrids & Regret, $\sum_i1/\omega(w_{t,i})$, smallest weight\\
Predictable experts & OMD, optimistic OMD, entropic and Euler FTRL & Regret, prediction-error sum, invariances\\
Sparse and changing experts & Entmax boundary, dense interior, restarts & Support size, re-entry delay, dynamic regret\\
Convex schedule-free learning & With and without interpolation & Objective versus gradient calls, dependence on $\vartheta$\\
Tabular RL and VMD & Exact and noisy Bellman quantities; PMD, VI & Value error, prox radius $B_t$, policy-value gap\\
LLM group policy learning & KL, probability-$\ell_2$, one-parameter kernels, Euler, ratio penalty & Held-out reward, seed variation, divergence cost\\
Games & Mirror-Prox, optimistic FTRL, anchored Mirror-Prox & {Current and averaged gaps; anchored divergence contraction $1/(1+\eta\tau)$ where the anchored solution is known}\\
Weight decay & Primal decay versus dual damping & Smallest coordinate, recovery after loss changes\\
Solver accuracy & Fixed tolerance versus $\epsilon_0\eta/t^2$ & Regret, total solver work, residual sum\\
Sparse MoE routing & Top-$k$ softmax, entmax, shifted gate, two-level router & Executed experts, score gaps versus Table~\ref{tab:gate-thresholds}, overload, starvation\\
Neural optimizers & AdaGrad, AdamW, Shampoo, Muon versus Euler-AdaGrad and Euler links & Matched-token loss, wall-clock, memory, failed certificates\\\hline
\end{tabular}
\end{table}

For language models, test three interventions separately: switching the
optimizer to Euler-AdaGrad, redesigning a probability link while keeping AdamW or a
matrix optimizer, and changing only a probability-space penalty. For large
vocabularies, separate full-sum divergences from frozen-proposal estimators and
report the KL to the reference as a common diagnostic. Begin with a fixed triple
and exact normalization on a small problem, verify the residuals of
Algorithm~\ref{alg:euler-norm}, and only then add optimism, interpolation or
learned geometry.

\subsection{Open questions}
\label{subsec:notation-open}
\begin{itemize}[leftmargin=1.5em,itemsep=1pt]
\item \emph{Learning the geometry.} An update rule for $(a,b,\lambda)$ whose regret
gain offsets its accumulated drift, compared with the best fixed triple in
hindsight. The scale is the most promising single parameter to adapt, because it
changes the constants without changing the tail exponents.
\item \emph{Retention versus recovery.} A dynamic-regret or recovery-time bound
in terms of $|a|$, $\alpha_\lambda$, the loss variation and the smallest relevant
probability, guiding restarts, forgetting or exploration.
\item \emph{Last iterates beyond exact monotone play.} Driving the anchor weight
$\tau\to0$ with a guaranteed complexity, and stochastic, delayed or neural
feedback.
\item \emph{Barrier bridges with complexity control.} Whether power geometries
near the Burg endpoint give useful accuracy--cost tradeoffs on problems with
boundary optima.
\item \emph{Implicit bias with a statistical question.} Whether the Bregman
projection selected by a given triple recovers structure under a specified data
model.
\item \emph{Sparse MoE.} A joint analysis of support adaptation, expert coverage
and hard capacity with trained scores, with measured wall-clock cost.

\item \emph{Multi-parameter Tempesta Group Logarithms\rev{.}} In this paper we deeply analyzed applications of \rev{the} free-scaled Euler logarithm. An open problem is to apply a wide class of flexible multi-parameter Tempesta group logarithms and their exponentials \cite{Tempesta2015,tempesta2016,Cichocki_Tempestalog,Cichocki-Tempesta}.

\end{itemize}

\section{Conclusions}
\label{sec:conclusions}

\paragraph{A common three-parameter geometry with interpretable controls.}
A probability close to zero carries different operational meanings across machine learning. It may belong to a mislabeled example whose influence should be limited, to a disfavored action worth retaining for future exploration, or to an expert that should receive no computation on the current token. The normalized, freely scaled Euler logarithm turns these competing goals into explicit choices that act consistently on loss functions, probability maps and optimization geometry. In the full-range interior $a<0<b$, its inverse maps every real score to a strictly positive weight with independently selectable tail powers $1/|a|$ and $1/b$, and the scale $\lambda$ moves the transition between boundary and bulk curvature while the value and slope of the link at one stay fixed. On the probability simplex all inverse arguments are nonpositive, so the negative exponent determines how fast suppressed coordinates decay; the finite-slope boundary $a=0$, or a shifted potential, instead lets a finite score gap produce an exact zero.

\paragraph{Classification, calibration and label noise.}
Under ordinary softmax, Euler cross-entropy multiplies the cross-entropy gradient by the explicit weight $\omega_{a,b,\lambda}$ of the true-class probability. On the bounded boundary this attenuates low-confidence predictions and yields a quantitative clean-risk guarantee under symmetric label noise; the same weighting also attenuates hard or rare clean examples, which makes the robustness--imbalance trade-off explicit. In the full-range statistical regime, composing the loss with the matching Euler softmax gives a loss that is convex in the logits, preserves class rankings, and recovers the class posterior at an interior population optimum by normalizing the coordinate mobilities. When the predicted probabilities themselves should target the posterior, including for sparse prediction maps, the Fenchel--Young loss provides that calibration directly.

\paragraph{Selecting the geometry across full-information, bandit, smooth and barrier regimes.}
On the interior simplex, fixed-step mirror descent and FTRL generate the same trajectory under matched anchors and keep a polynomial floor on every coordinate in terms of its accumulated score disadvantage. The metric, benchmark and scenario tables of Section~\ref{sec:ftrl-choosing} turn these properties into task-specific rules.
\begin{itemize}[leftmargin=1.6em,itemsep=2pt]
\item \emph{Full information.} Entropy has the smallest worst-case constant. The dimension-dependent choice $|a|=b=1/\ln N$ at the bound-minimizing scale stays within $2$--$8\%$ of it up to $N=10^6$, and within about $2\%$ asymptotically, while retaining polynomial floors.
\item \emph{Bandit feedback.} Because the mobility divided by the coordinate is the reciprocal gradient weight, the variance term of Euler FTRL is $\sum_i1/\omega_{a,b,\lambda}(w_i)$. Its leading order depends on $a$ alone and is smallest at the Tsallis-INF value $a=-\frac12$; the bulk exponent and the scale enter only through a lower-order term, so bulk curvature can be halved at a cost of $0.007\%$ in the leading constant at $N=10^6$.
\item \emph{Relative smoothness and barrier stability.} For separable two-power objectives a closed-form scale matches the Hessian exactly, and a scale mis-set by a factor $\gamma$ costs at most $\max\{\gamma,\gamma^{-1}\}^{b-a}$ in the relative condition number. At the barrier endpoint $a=-1$, $0\le b\le1$, the hybrids are globally self-concordant with constant $\sqrt{1+b\lambda^{b+1}}$, which exposes a trade-off between comparator divergence and Dikin-ellipsoid stability.
\item \emph{When simpler or different geometries suffice.} Hedge remains the simpler choice when retention is not needed; the sparse boundary does not suit importance-weighted bandits; and generic hidden-layer weight matrices call for covariance or spectral preconditioners such as Adam, Shampoo or Muon.
\end{itemize}

\paragraph{From certified computation to empirical validation.}
All sixteen algorithms rest on two certified scalar primitives, log-coordinate inversion (Algorithm~\ref{alg:euler-exp}) and weighted simplex normalization (Algorithm~\ref{alg:euler-norm}), and a residual certificate links the accuracy of each numerical update to the regret, Bellman or equilibrium-gap guarantee of the outer algorithm. The applications also supply runtime diagnostics, including a Bellman residual for a returned policy, an equilibrium gap for averaged or anchored play, and score thresholds for activating an expert. Under matched tuning and compute budgets, the natural measurements are recovery time after an expert's fortunes reverse, calibration on rare classes, bandit regret against Tsallis-INF, and the number of executed experts at a fixed prediction error, each reported together with normalization time. Two theoretical directions remain open: adapting the scale online while controlling the accumulated change of geometry, and extending the framework to the multi-parameter Tempesta group logarithms~\cite{Cichocki_Tempestalog,Cichocki-Tempesta}. Both continue the insight of Professor Amari, to whom this paper is dedicated: the geometry used to measure an update shapes what a learning system ultimately learns.

\appendix
\section{Notation and guide to mathematical statements}
\label{sec:notation-guide}
\subsection{One ordered parameter pair, one scale, and one link}
\label{subsec:notation-kernel}
{Use ordered exponents $a\le0<b$, positive scale $\lambda$, and positive link argument $x$. The notation $|a|=-a$ gives the magnitude of the left exponent. The explicitly marked extensions are $a<0=b$ and the logarithmic limit $(a,b)=(0,0)$. Individual results may narrow these ranges, for example to $-1<a<0<b\le1$ for finite-radius simplex optimization.} With
$k=(a-b)/2$,
\begin{equation}
 \log_{a,b,\lambda}(x)=\frac{\lambda^k(x^a-1)-\lambda^{-k}(x^b-1)}{a\lambda^k-b\lambda^{-k}}
 =\alpha_\lambda\log_{a,0}(x)+\beta_\lambda\log_{0,b}(x),
 \label{eq:notation-kernel}
\end{equation}
{The two normalized power links are $\log_{s,0}(x)=(x^s-1)/s$ for $s\ne0$ and $\log_{0,b}(x)=(x^b-1)/b$ for $b\ne0$. A zero power is interpreted as $\ln x$. For $a<b$, the weights are $\alpha_\lambda=|a|\lambda^a/(|a|\lambda^a+b\lambda^b)$ and $\beta_\lambda=b\lambda^b/(|a|\lambda^a+b\lambda^b)$. They sum to one. At $a=b=0$ use $\ln x$ directly.}
The unscaled link is $\log_{a,b}=\log_{a,b,1}$; the ordinary logarithm is
the limit $(a,b)\to(0,0)$, and $\lambda$ cancels on both single-power
boundaries. {The inverse $\exp_{a,b,\lambda}$ accepts all real inputs for $a<0<b$ and in the natural-logarithm limit. Its domain is $(-1/b,\infty)$ for $a=0<b$ and $(-\infty,-1/a)$ for $a<0=b$. A positive-part continuation on $a=0$ assigns zero by the simplex KKT rule when the score reaches or falls below $-1/b$.} The scale $\lambda$ is kept throughout, including Appendix~\ref{app:MMD-convergence}; normalization multipliers are always denoted by $\nu$.

\subsection{Functions derived from the link}
\label{subsec:notation-five}
\begin{align}
 \phi_{a,b,\lambda}(x)&=\int_1^x\log_{a,b,\lambda}(r)\,\mathrm dr,\qquad
 F_{a,b,\lambda}(\bp)=\sum_i\phi_{a,b,\lambda}(p_i),
 \label{eq:notation-potential}\\
 D_{F_{a,b,\lambda}}(\bp\Vert\bq)&=F_{a,b,\lambda}(\bp)-F_{a,b,\lambda}(\bq)-\langle \log_{a,b,\lambda}(\bq),\bp-\bq\rangle,
 \label{eq:notation-divergence}\\
 h_{a,b,\lambda}(x)&=\log_{a,b,\lambda}'(x)=\alpha_\lambda x^{a-1}+\beta_\lambda x^{b-1},\qquad
 m_{a,b,\lambda}=1/h_{a,b,\lambda}.
 \label{eq:notation-curvature}
\end{align}

\begingroup
Here $\bp,\bq$ are probability vectors and $\log_{a,b,\lambda}(\bq)$ means coordinatewise application. The formula for $D_F(\bp\Vert\bq)$ linearizes $F$ at the second argument. In the full-range regime that argument must be positive. When $a>-1$, the first argument may be on the boundary by continuous extension.
\par\endgroup
The exact interior simplex step is
\begin{equation}
 p_{i,t+1}=\exp_{a,b,\lambda}\bigl(\log_{a,b,\lambda}(p_{i,t})-\eta_tg_{i,t}-\nu_t\bigr),\qquad\sum_ip_{i,t+1}=1,
 \label{eq:notation-step}
\end{equation}
{Here $i$ indexes the simplex coordinates, $t$ indexes the update, $g_{i,t}$ is the loss-gradient component, and $\eta_t>0$ is the step size. The common multiplier $\nu_t$ is determined by the mass equation in dual coordinates. In the logarithmic limit this is equivalent to dividing an unnormalized exponential proposal by its sum.} Table~\ref{tab:notation-five} lists the derived functions.

\begin{table}[ht]
\centering\small
\caption{Functions derived from the Euler link, and their entropic limits.}
\label{tab:notation-five}
\begin{tabular}{L{.17\textwidth}L{.25\textwidth}L{.12\textwidth}L{.35\textwidth}}
\hline
Name & Definition & At $(0,0)$ & Question it answers\\\hline
Potential $\phi_{a,b,\lambda}$ & {$\int_1^x\log_{a,b,\lambda}(u)\,\mathrm du$} & $x\ln x-x+1$ & Which regularizer and divergence?\\
Curvature $h_{a,b,\lambda}$ & $\log_{a,b,\lambda}'$ & $1/x$ & Cost of moving a coordinate of size $x$.\\
Mobility $m_{a,b,\lambda}$ & $1/h_{a,b,\lambda}$ & $x$ & Response to a unit dual step; $\sim x^{1+|a|}/\alpha_\lambda$ at $0$.\\
Gradient weight $\omega_{a,b,\lambda}$ & $xh_{a,b,\lambda}(x)$ & $1$ & {Ordinary-softmax loss weight; global minimum $K/D_\lambda$ at $r/\lambda$ for $a<0<b$. On $(0,1]$ the minimizer is $\min\{r/\lambda,1\}$.}\\
Elasticity $\varepsilon_{a,b,\lambda}$ & $-xh'/h$ & $1$ & Local power; from $1-a$ at $0$ to $1-b$ at $\infty$.\\\hline
\end{tabular}
\end{table}

\subsection{Symbol table}
\label{subsec:notation-table}
\small
\begin{longtable}{L{.27\textwidth}L{.65\textwidth}}
\caption{Notation and scope.}\label{tab:complete-notation}\\
\hline Symbol & Meaning\\\hline\endfirsthead
\hline Symbol & Meaning (continued)\\\hline\endhead
\hline\endfoot
$a,b,\lambda$; $p=|a|$, $q=b$ & Ordered exponents and scale; $p,q$ are used in Sections~\ref{sec:inverse-computation} and~\ref{sec:FTRL-3p} and in Appendices~\ref{app:supp-online} and~\ref{app:MMD-convergence}.\\
$\alpha_\lambda,\beta_\lambda$ & {Mixture weights. In the interior $a<0<b$, their odds are $\beta_\lambda/\alpha_\lambda=(b/|a|)\lambda^{b-a}$.}\\
$r$, $D_\lambda$, $K$ & $(|a|/b)^{2/(b-a)}$, $|a|\lambda^{-|a|}+b\lambda^b$, $|a|r^{-|a|}+br^b$.\\
$\omega_{\min}$ & Global minimum $K/D_\lambda$ of $\omega_{a,b,\lambda}$, attained at $q_m=r/\lambda$ (Eq.~\eqref{eq:omega-min}).\\
$\omega_0$, $\ell$ & Unscaled gradient weight $\omega_0(x)=x\log_{a,b}'(x)$, so that $\omega_{a,b,\lambda}(x)=\omega_0(\lambda x)/\omega_0(\lambda)$ (Eq.~\eqref{eq:scaled-Euler-identity}); log-scale $\ell=\ln\lambda$.\\
$\Phi_s$ & Power potential of Eq.~\eqref{eq:scaled-potential}, $\Phi_s''(x)=x^{s-1}$, $s\in\{a,b\}$; $\phi_{a,b,\lambda}=\alpha_\lambda\Phi_a+\beta_\lambda\Phi_b$.\\
$d_{a,b,\lambda}$, $D_{-\log_{a,b,\lambda}}$ & Scalar Bregman gap of $\phi_{a,b,\lambda}$; scalar Bregman gap of the concave link (Appendix~\ref{subsec:EulerCE-statistics}).\\
$\omega_i$, $\mathcal T_{a,b,\lambda}$, $\br$ & Per-class gradient weight $\omega_i=\omega_{a,b,\lambda}(q_i)$ (Eq.~\eqref{eq:ri-Euler}); population link $\bp\mapsto\bq^*$; recovered posterior $\br=\mathcal T_{a,b,\lambda}^{-1}(\bq)$ with coordinates $r_i=m_{a,b,\lambda}(q_i)/\sum_jm_{a,b,\lambda}(q_j)$ (Eq.~\eqref{eq:posterior-recovery}).\\
$H_{a,b,\lambda}$, $\mu^{H}_{a,b,\lambda}$, $r_b$ & Bayes risk (generalized entropy) of Euler cross-entropy; its $\ell_1$ strong-concavity modulus $c_{a,b,\lambda}/(1-a)$; GCE exponent $1/(1-b)$ (Appendix~\ref{subsec:EulerCE-statistics}).\\
$A_{a,b,\lambda}$ & Simplex conjugate $\max_{\Delta_N}\{\langle\bs,\cdot\rangle-F_{a,b,\lambda}\}$, Eq.~\eqref{eq:Euler-conjugate}.\\
$\mathcal P_{a,b,\lambda}$ & Shared prox solver $\argmin_{\Delta_N}\{\eta\langle\bz,\cdot\rangle+D_{F_{a,b,\lambda}}(\cdot\Vert\bq)\}$, Eq.~\eqref{eq:new-prox}.\\
$\widehat L$, $\mathcal N^{\rm rad}$, $\bH_{a,b,\lambda}$ & Zero-homogeneous lift $L(\bu/\1^T\bu)$; radial normalization $\bu/\1^T\bu$; diagonal Hessian $\diag\{h_{a,b,\lambda}(w_i)\}$ (Appendix~\ref{app:MMD-convergence}).\\
$\Delta_N,\Delta_N^\circ$; $\1$, $\bolde_i$ & Closed and open simplex; all-ones and coordinate vectors.\\
$\operatorname{hr},\operatorname{ctr}$ & Half-range $\operatorname{hr}(\bg)=\frac12(\max_ig_i-\min_ig_i)$ and centered vector $\operatorname{ctr}(\bg)=\bg-\frac12(\max_ig_i+\min_ig_i)\1$, Eq.~\eqref{eq:centered-online-gradients}; both ignore common offsets.\\
$c_{a,b,\lambda}$, $\mu_{a,b,\lambda}(N)$, $R_{a,b,\lambda}(N)$ & {Entropy-comparison factor $c=\min_{(0,1]}\omega_{a,b,\lambda}$ (also the clipping-free threshold of Theorem~\ref{thm:clipping-inactive}), sufficient $\ell_1$ modulus $\mu$, and exact uniform-center radius $R$} (Theorem~\ref{thm:sharp-simplex-geometry}); $\lambda_*$ minimizes $R/c$. In the algorithms, $\mu=\mu_{a,b,\lambda}(N)$ and $R=R_{a,b,\lambda}(N)$.\\
$V_{a,b,\lambda}(N)$, $C^{\rm band}$ & Bandit variance constant $\sup_{\Delta_N}\sum_i1/\omega_{a,b,\lambda}(w_i)\le N^{1-|a|}/\alpha_\lambda$ and bandit regret constant $\sqrt{2RV}$ (Proposition~\ref{prop:euler-bandit}).\\
$x_0$, $M_{\rm sc}$ & Centre $\lambda^{-1}(|a|/b)^{1/(b-a)}$ of the curvature transition; self-concordance constant $\sqrt{1+b\lambda^{b+1}}$ at $a=-1$ (Proposition~\ref{prop:barrier-sc}).\\
$\mathcal R_{a,b,\lambda}$, $\mathcal N_{a,b,\lambda}$ & Centered regularizer $F-\min F$; exact simplex decision $\mathcal N_{a,b,\lambda}(\by)=\argmin_{\Delta_N}\{F_{a,b,\lambda}-\langle\by,\cdot\rangle\}$, {approximated with a verified residual by Algorithm~\ref{alg:euler-norm}.}\\
$\sigma_{a,b,\lambda,\tau}$ & Variational Euler softmax (Definition~\ref{def:Euler-softmax}).\\
$\nu,\nu_t$ & Normalization (KKT) multipliers.\\
$\sigma_t,S_t$ & Loss half-range $\operatorname{hr}(\bg_t)$ and $S_t=\sum_{s\le t}\sigma_s^2$.\\
$\eta_t,\eta_0,\delta$ & Step size, adaptive-rate constant, initial scale in $\eta_0/\sqrt{\delta+S_t}$.\\
$\rho_1,\rho_2$ & $\ell_1$ and potential weights of composite FTRL; other $\rho$ with subscripts ($\rho_{ab}$, $\rho_\ell$, $\rho_b$, $\rho_\theta$, $\rho_h$) are penalty weights.\\
$\beta_t,\omega_t,A_t$ & Regularizer weight, averaging weights and their sum in dual averaging.\\
$\varrho$, $\psi_{a,b,\lambda,\varrho}$, $H_{a,b,\lambda,\varrho}$ & Shift, signed shifted potential and its separable sum; $\varrho$ is also the boundary shift of the regularized metric~\eqref{eq:audit-regularized-metric}.\\
$\delta$ (MoE), $\kappa_\delta(m)$ & Gate shift; score gap $\log_{a,b,\lambda}(m+\delta)-\log_{a,b,\lambda}(\delta)$.\\
$\kappa$ (Section~\ref{sec:phase2-bridges}) & Decay coefficient in primal decay and dual damping.\\
$K_\epsilon$, $B_\epsilon$ & Floored simplex and its Bregman diameter bound.\\
$\pi,P,r,\gamma$; $V,Q,A$ & {Policy, transition probabilities, immediate reward and discount. $V^\pi$, $Q^\pi$ and $A^\pi=Q^\pi-V^\pi$ describe the evaluated policy. $V_t$ and $Q_{V_t}$ are the VMD iterates. Actions use index $\alpha$.}\\
$\mathcal G,\mathcal G_\tau,\Lambda_G,R_\bq$ & Game operator, anchored operator, Lipschitz constant, anchored radius.\\
$\hat\epsilon$, $\epsilon_t$, $D_K$ & Normalization certificate, residual tolerances, set diameter.\\
{$\log_{a,b,\lambda}$, $\exp_{a,b,\lambda}$, $h_{a,b,\lambda}$, \rev{$m_{a,b,\lambda}$}} & {\rev{Euler link, inverse, curvature and mobility (written in full throughout all sections, tables, and algorithms). Subscripts $j$ and $v$ select the block and prior triples.}}\\
{$\mathsf{Norm}$, $\mathcal N_{a,b,\lambda}$} & {Certified solver returning a decision, multiplier and residual, or an explicit failure status; exact simplex optimizer used in exact-arithmetic listings.}\\
{$\epsilon$, $\hat\epsilon$, $\varepsilon$} & {Requested solver accuracy, verified dual residual and application-level stopping target. For Algorithm~\ref{alg:euler-exp}, $\epsilon$ is a log-coordinate error.}\\
\multicolumn{2}{@{}l}{\emph{Variables inside Algorithms~\ref{alg:euler-exp}--\ref{alg:moe-step}: inputs and online data}}\\[1pt]
$\bg_t$, $g_{t,i}$ & Loss vector or (sub)gradient revealed after round $t$; coordinate $i$.\\
$\widehat\bg_t$, $\vm_{t+1}$ & Optimistic hint: a prediction of the next loss, formed before it is observed (Alg.~\ref{alg:ao-ftrl-prox}, Alg.~\ref{alg:aoftrl}, Alg.~\ref{alg:selfplay}).\\
$y$; $\by$ & Scalar argument of $\exp_{a,b,\lambda}$ (Alg.~\ref{alg:euler-exp}); score vector passed to $\mathcal N_{a,b,\lambda}$ (Alg.~\ref{alg:euler-norm}).\\
$\epsilon$, $\theta$; $T$, $T_{\max}$ & Solver tolerance and safeguard fraction $\theta\in(0,1/2)$; horizon and iteration budget.\\
\textsc{success}, \textsc{precision-limit}, \ldots & Exit status returned when a run stops; the certificate of the completed prefix is kept.\\
\multicolumn{2}{@{}l}{\emph{Scalar inverse and simplex normalization (Algorithms~\ref{alg:euler-exp}--\ref{alg:euler-norm})}}\\[1pt]
$z=\ln x$; $\hat z$, $\hat x$ & Log coordinate of $x=\exp_{a,b,\lambda}(y)$; returned log-estimate and value $\hat x=e^{\hat z}$.\\
$G_{a,b,\lambda}(z)$, $g$ & Inverse residual $\alpha_\lambda\operatorname{expm1}(az)/a+\beta_\lambda\operatorname{expm1}(bz)/b-y$, Eq.~\eqref{eq:Gz}; increasing in $z$ with root $\ln\exp_{a,b,\lambda}(y)$; $g$ is its current value.\\
$[\ell,u]$ & {Bracket containing the scalar root}: of $G_{a,b,\lambda}$ in Alg.~\ref{alg:euler-exp}, of {$S(\nu)-1=0$ in Alg.~\ref{alg:euler-norm}}.\\
$z_q$, $\omega_{\min}$, $s$ & Log-minimizer and minimum of $\omega_{a,b,\lambda}$; lower bound on the slope of $G_{a,b,\lambda}$ over the bracket, so that $|g|/s$ bounds the log error.\\
$z_N$, $\nu_N$ & {Newton candidates, accepted only in the central interval $[\ell+d,u-d]$ with $d=\theta(u-\ell)$.}\\
$w_i$, $S$, $j$ & Unnormalized weights $\exp_{a,b,\lambda}((y_i-\nu)/D_i)$; their sum $S(\nu)$, decreasing in $\nu$; index of the largest weight, used to restore $\sum_ip_i=1$.\\
$D_i$ & {Positive coordinate weights multiplying each scalar potential in weighted simplex normalization. The local symbol $D_i=\rho_2+A_{t,i}$ in Alg.~\ref{alg:ao-ftrl-prox} weights a separable orthant or signed-box subproblem.}\\
\multicolumn{2}{@{}l}{\emph{Accumulators of FTRL, dual averaging and AdaGrad}}\\[1pt]
$\bh_t$ & {Cumulative centered loss $\sum_{s\le t}\operatorname{ctr}(\bg_s)$. Adaptive FTRL uses $-\eta_t\bh_t$ for the next decision. Optimistic FTRL adds the current centered hint to the history available before play.}\\
$\widetilde E_t$ & Cumulative squared forecast error $\sum_{s\le t}\operatorname{hr}(\bg_s-\widehat\bg_s)^2$; sets the optimistic rate.\\
$\bz_t$, $z_{t,i}$ & {Cumulative proximal dual state, stored as $z_i$ in Alg.~\ref{alg:ao-ftrl-prox}. The scalar $z=\ln x$ is the inverse solver coordinate.}\\
$q_i$ & {Optimistic score $z_i+\widehat g_{t+1,i}$ (Alg.~\ref{alg:ao-ftrl-prox}).}\\
$v_{t,i}$, $A_{t,i}$, $\sigma_{t,i}$ & Sum of squared gradients; adaptive weight $(\delta_0+\sqrt{v_{t,i}})/\eta_0$; its increment $A_{t,i}-A_{t-1,i}$.\\
$\bs_t$ & Weighted dual sum $\sum_{s\le t}\omega_s\bg_s$ of RDA (Alg.~\ref{alg:euler-rda}).\\
$L_i$, $v_i$, $G_i$ & {Per-coordinate cumulative gradient, offset plus accumulated squares, and declared gradient bound. The predictable denominator is $d_i=\sqrt{v_i}$ before the current gradient (Alg.~\ref{alg:euler-adagrad}).}\\
$r$, $r_i$; $\kappa_{\max}$ & {Signed dual-averaging evidence divided by $\beta_t$, or the capped AdaGrad-scaled magnitude in the corresponding listing;} evidence cap $\log_{a,b,\lambda}(\varrho+w_{\max})-\log_{a,b,\lambda}(\varrho)$, reached when $|w|=w_{\max}$.\\
$\bx_t$, $\by_t$, $\bz_t$, $\vartheta$, $\xi_t$ & Schedule-free average, evaluation point, online FTRL point, interpolation weight and sample (Alg.~\ref{alg:sf-ftrl}).\\
\multicolumn{2}{@{}l}{\emph{Reinforcement learning (Algorithms~\ref{alg:euler-pmd}--\ref{alg:euler-vmd})}}\\[1pt]
$\widehat A_k$, $e_{k,s}$ & Estimated advantage and its declared error bound at state $s$.\\
$t_k$, $J$, $L_s(\bp)$ & Policy step, backtracking depth and verified improvement test, Eq.~\eqref{eq:stage3-direct-policy-test}.\\
$V_t$, $Q$ & {Value iterate and frozen action values $Q_{V_t}(s,\alpha)=r(s,\alpha)+\gamma\sum_{s'}P(s'|s,\alpha)V_t(s')$ in value mirror descent.}\\
$r_o$, $r_p$ & {Optimal and on-policy Bellman residual norms, computed from $\bv_o=TV_t$ and $\bv_p=T_{\pi_t}V_t$. Their sum divided by $1-\gamma$ bounds $\|V^*-V^{\pi_t}\|_\infty$.}\\
$\widehat B_t$, $C$, $q$, $\epsilon_N$ & {Verified upper bound on the current radius $B_t$ in Eq.~\eqref{eq:vmd-radius}; error target $Cq^t$ used to set the step size; normalization tolerance.}\\
\multicolumn{2}{@{}l}{\emph{Games and Mirror-Prox (Algorithms~\ref{alg:euler-mp}--\ref{alg:selfplay})}}\\[1pt]
$K=\prod_{j=1}^M\Delta_{n_j}$; $\bp^j$, $\bz$ & Product of $M$ simplices; block $j$ of a profile $\bz=(\bp^j)_j$.\\
$\bu^j$ & Dual coordinates $\log_{a_j,b_j,\lambda_j}(\bp^j_t)$ of the current block.\\
$\widetilde\bz_t$, $\overline\bz_t$ & Prediction (extrapolation) point and running average of predictions.\\
$E_t$, $R_0$, $B_T$ & {Accumulated certified solver error; initial radius $\sum_jR_{a_j,b_j,\lambda_j}(n_j)$; averaged gap bound $(R_0+E_T)/(\eta T)$. In the anchored listing $B_t$ bounds the current gap instead.}\\
$\operatorname{Gap}_D$, $G_{\max}$, $\overline G$ & Dual gap $\sup_{\bw\in K}\langle\mathcal G(\bw),\bz-\bw\rangle$; operator bound $\sup_K\|\mathcal G\|_2$ and a verified upper bound.\\
$\tau$, $\bq$, $c$, $U_t$ & Anchor strength and anchor; contraction $1/(1+\eta\tau)$; certified bound on the divergence to the anchored equilibrium.\\
$\bL^j$, $c^j$, $e_t$, $\overline{\mathcal P}_t$ & {Cumulative loss vector and realized loss of player $j$; nonnegative CCE certificate; empirical joint play} $\frac1t\sum_{s\le t}\bigotimes_j\bp^j_s$.\\
\multicolumn{2}{@{}l}{\emph{Positive interpolation and mixture-of-experts routing (Algorithms~\ref{alg:implicit-md} and~\ref{alg:moe-step})}}\\[1pt]
$\bA$, $\by$, $L(\bw)$, $M$ & Data, targets, loss $\frac12\|\bA\bw-\by\|_2^2$, curvature bound $M\ge\|\bA\|_2^2$ (Alg.~\ref{alg:implicit-md}).\\
$\widetilde\bz_j$, $\mathcal C_j$, $S_j$ & Adjusted token scores, candidate experts, support of $\bp_j$.\\
$\bu_j$, $\bar u_j$, $\widetilde m_i$ & Upstream gradient; its mobility-weighted mean; shifted mobility $m_{a,b,\lambda}(p_{j,i}+\delta)$.\\
$\boldsymbol\xi_t$, $C_i$, $\eta_\xi$ & Capacity prices, expert capacities, price step.\\
$\bv_t$, $\bH_t$, $c_v$, $\eta_v$ & Expert prior, its cumulative losses, prior coupling, prior rate.\\
\end{longtable}
\normalsize

\subsection{Symbols with more than one meaning}
\label{subsec:notation-overload}
Whenever a symbol is reused in different sections (for example $\delta$ for the adaptive offset and the MoE shift, $r$ for $(|a|/b)^{2/(b-a)}$ and a reward, and $q$ for the Tsallis index and the VMD decay factor), its meaning is declared next to the equation that uses it; Table~\ref{tab:notation-overload} lists all such cases.
{\small\setlength{\LTpre}{4pt}
\begin{longtable}{L{.07\textwidth}L{.85\textwidth}}
\caption{Overloaded symbols.}\label{tab:notation-overload}\\
\hline
Letter & Meanings\\\hline
$\lambda$ & Euler scale throughout (main text and appendices); $\lambda_F$ is a Fisher damping constant. A multiplier is never denoted by $\lambda$.\\
$\alpha$ & Mixture weight $\alpha_\lambda$; action index $\alpha$; Amari and Tempesta parameters in Appendix~\ref{app:special-cases}.\\
$\beta$ & $\beta_\lambda$ mixture weight; $\beta_t$ dual-averaging weight.\\
$\sigma$ & $\sigma_t$ half-range; $\sigma_{a,b,\lambda,\tau}$ Euler softmax; $\sigma_s$, $\sigma_{t,i}$ proximal increments (Section~\ref{sec:FTRL-3p}, Algorithm~\ref{alg:ao-ftrl-prox}); residual $\sigma_{t,i}$ (Algorithm~\ref{alg:euler-adagrad}).\\
$\kappa$ & Kaniadakis parameter; gate threshold $\kappa_\delta$; evidence cap $\kappa_{\max}$; decay coefficient; proper-score correction $\kappa_{a,b,\lambda}$.\\
$\delta$ & Initial adaptive scale; gate shift; Kronecker $\delta_{ij}$.\\
$\omega$ & Gradient weight $\omega_{a,b,\lambda}$, its values $\omega_i=\omega_{a,b,\lambda}(q_i)$ and its minimum $\omega_{\min}$; unscaled weight $\omega_0$; averaging weight $\omega_t$; mixture weight $\omega_k$.\\
$r$ & Transition constant $(|a|/b)^{2/(b-a)}$; recovered posterior $\br$ with coordinates $r_i$ (Section~\ref{sec:canonical-softmax}, Appendix~\ref{app:supp-losses}); GCE exponent $r_b$; reward $r$ and reward vector $\br$ (Appendix~\ref{app:supp-apps}); local residual or evidence $r$, $r_i$, $\br_t$; Bellman residuals $r_o$, $r_p$.\\
$q$ & $q=b$ in Section~\ref{sec:FTRL-3p}; anchor or reference $\bq$; optimistic score $q_i$; {the VMD forcing-term decay factor $q$; the Bellman contraction factor is $\gamma$.}\\
$R$ & Radius $R_{a,b,\lambda}(N)$; anchored radius $R_\bq$; regularizer $\mathcal R_{a,b,\lambda}$; initial Mirror-Prox radius $R_0$.\\
$\nu$ & Normalization (KKT) multiplier $\nu,\nu_t$ {for simplex normalization throughout the main text and appendices; state distributions are written $d^{\pi}$. Algorithm~\ref{alg:selfplay} denotes empirical play by $\overline{\mathcal P}_t$.}\\
$\xi$ & Random sample $\xi_t$ (Algorithm~\ref{alg:sf-ftrl}); capacity prices $\boldsymbol\xi_t$ (Algorithm~\ref{alg:moe-step}).\\
$A$ & Dual-averaging weight sum $A_t$; adaptive weight $A_{t,i}$; advantage $A$, $\widehat A_k$; data matrix $\bA$.\\
$c$, $C$ & Modulus $c_{a,b,\lambda}$; anchored contraction $c$; realized loss $c^j$; prior coupling $c_v$; VMD schedule constant $C$; capacity $C_i$; candidate set $\mathcal C_j$.\\
$D$ & Coordinate weight $D_i$; constant $D_\lambda$; divergences $D_F$, $D(\cdot\Vert\cdot)$; diameter $D_K$.\\
$E$, $e$ & Solver-error sum $E_t$; forecast-error sum $\widetilde E_t$; CCE violation $e_t$; advantage error bound $e_{k,s}$; coordinate vector $\bolde_i$.\\
$G$, $g$ & Inverse residual $G_{a,b,\lambda}(z)$ and its value $g$ (Algorithm~\ref{alg:euler-exp}); loss or gradient $\bg_t$; gradient bound $G_i$; game operator $\mathcal G$ and $G_{\max}$.\\
$h$ & Curvature $h_{a,b,\lambda}$; cumulative centered loss $\bh_t$; hidden features $\bh_n$ (Appendix~\ref{subsec:EulerCE-NG}).\\
$H$ & Bayes risk $H_{a,b,\lambda}$; shifted RDA potential $H_{a,b,\lambda,\varrho}$; Euler Hessian $\bH_{a,b,\lambda}$ and $\bH_t$ (Appendix~\ref{app:MMD-convergence}); cumulative expert losses $\bH_t$ (Algorithm~\ref{alg:moe-step}); perspective $\mathcal H_{a,b,\lambda}$.\\
$\mu$ & Sufficient $\ell_1$ modulus $\mu_{a,b,\lambda}(N)$; Bayes-risk modulus $\mu^{H}_{a,b,\lambda}$; local metric bounds $\mu_F$, $\mu_\delta$, $\mu_*$.\\
$\mathcal N$ & Exact simplex decision $\mathcal N_{a,b,\lambda}$; radial normalization $\mathcal N^{\rm rad}$ (Appendix~\ref{app:MMD-convergence}); normal cone $N_K$.\\
$\varepsilon$, $\epsilon$ & Elasticity $\varepsilon_{a,b,\lambda}$ and $\varepsilon_1=\varepsilon_{a,b,\lambda}(1)$; Fisher damping $\varepsilon_F$; application-level target $\varepsilon$; solver tolerances $\epsilon$, $\epsilon_t$; floor $\epsilon$ of $K_\epsilon$; objective tolerance $\epsilon_{\rm obj}$.\\
$L$, $\ell$ & {Losses $L$, $L(\bw)$, $\ell_t$; cumulative loss vector $\bL^j$; cumulative gradient $L_i$; statewise improvement test $L_s$; lower bracket end $\ell$ or log-scale $\ell=\ln\lambda$, depending on the subsection.}\\
$M$ & Number of blocks or players; curvature bound $M\ge\|\bA\|_2^2$ (Algorithm~\ref{alg:implicit-md}); momentum matrix $\mathbf M$.\\
$p$ & {$p=|a|$ in the inverse and scale calculations; $p_i$ is a probability coordinate of $\bp$. A superscript $j$ in $\bp^j$ denotes a player block. In Appendix~\ref{subsec:pnorm} only, $p$ is a $p$-norm exponent.}\\
$S$, $s$ & Squared half-range sum $S_t$; weight sum $S(\nu)$; support $S_j$; slope bound $s$; dual sum $\bs_t$; MDP state $s$; logits $\bs=\bz/\tau$; power index $s\in\{a,b\}$; proposal mass $s=\1^T\tilde\bw$ (Appendix~\ref{app:MMD-convergence}); AdaGrad scale $s_{t,i}$.\\
$u$, $v$ & Upper bracket end $u$; block dual coordinates $\bu^j$; upstream gradient $\bu_j$; squared-gradient sums $v_{t,i}$, $v_i$; expert prior $\bv_t$.\\
$y$ & Argument of $\exp_{a,b,\lambda}$; score vector $\by$; targets $\by$ (Algorithm~\ref{alg:implicit-md}); evaluation point $\by_t$ (Algorithm~\ref{alg:sf-ftrl}).\\
$z$ & Log coordinate $z=\ln x$ (Algorithm~\ref{alg:euler-exp}); FTRL dual state $\bz_t$, $z_{t,i}$; game profile $\bz$; online point $\bz_t$ (Algorithm~\ref{alg:sf-ftrl}); token scores $\bz_j$.\\\hline
\end{longtable}}

\subsection{Reading assumptions and interpreting guarantees}
\label{subsec:notation-reading}
Before using a result, check: (1) which exponents and scale (most optimization
results need $-1<a<0<b$; sparsity needs $a=0$ or a shift; constants depend on
$\lambda$ through $\alpha_\lambda,\beta_\lambda$); (2) which set and norm; (3) which
information (exact, stochastic, estimated with declared error bounds); (4) which
output (regret, value, averaged or last-iterate gap); and (5) conditional on what
(convergence of a flow, validity of error bounds, certified residuals).

\begingroup
A solver certificate $\hat\epsilon$ bounds a stationarity residual in the objective actually passed to the solver. It has dual-coordinate units. A strong-convexity modulus converts it to a primal error bound, as in Lemma~\ref{lem:certified-norm}. Compare this with the PMD objective gap $\zeta_s$, which uses advantage units. For an experiment, name the returned object explicitly: $V_t$ is a value iterate, $V^{\pi_t}$ is the value of a policy, $\bz_t$ is the current game profile, and $\overline\bz_T$ averages Mirror-Prox predictions.
\par\endgroup

\section{Further results on Euler losses, softmax and computation}
\label{app:supp-losses}
This appendix collects the statistical theory of Euler cross-entropy and secondary results on natural gradients, Fenchel--Young losses, closed-form inverses and parameter selection that are summarized in Sections~\ref{sec:euler-CE}--\ref{sec:inverse-computation}.

\subsection{Statistical characterization of Euler cross-entropy}
\label{subsec:EulerCE-statistics}

A loss for probabilistic classification is judged by three standard statistical criteria~\cite{Savage1971,Gneiting2007,Zhang2004,Tewari2007,ReidWilliamson2010,Williamson2016}. \emph{Properness} asks whether minimizing the expected loss returns the true class posterior. \emph{Calibration} asks whether it returns the Bayes decision or, more generally, the Bayes ranking of the classes. A \emph{surrogate-regret bound} quantifies how a small excess loss translates into a small excess classification error and an accurate probability estimate. This subsection answers all three questions for Euler cross-entropy.

In brief, a genuinely deformed Euler cross-entropy is \emph{not} a proper local scoring rule, and for three or more classes no differentiable increasing nonlogarithmic local score can be one (Theorem~\ref{thm:EulerCE-locality}). The loss is, however, a strictly proper \emph{composite} loss with an explicit invertible link (Theorem~\ref{thm:EulerCE-composite}). Its population minimizer is an order-preserving, invertible transformation of the posterior. Its regret is an exact Bregman divergence of a generalized entropy, the Bayes risk of the loss. For $a<0$ this entropy is uniformly strongly concave, which yields a dimension-free bound on the excess classification error and on the error of the recovered posterior (Theorem~\ref{thm:EulerCE-regret-bound}). The bounded GCE boundary with $b>1/2$ admits no such uniform quadratic bound. When the network output itself should be a calibrated probability, the corrected Euler score of Proposition~\ref{prop:proper-Euler-score} can be used instead.

Throughout this subsection, assume
\begin{equation}
a\le0\le b<1,\qquad \lambda>0,
\label{eq:stat-domain}
\end{equation}
with $(a,b)=(0,0)$ interpreted as $\ln$. In this domain $\log_{a,b,\lambda}$ is strictly concave, and $\log_{a,b,\lambda}'$ decreases continuously from infinity to zero on $(0,\infty)$. For the class posterior $p_i=\Pr(Y=i\mid\bx)$, the conditional risk and the conditional Bayes risk are
\begin{equation}
\mathcal R^{\rm CE}_{a,b,\lambda}(\bq;\bp)=-\sum_{i=1}^N p_i\log_{a,b,\lambda}(q_i),\qquad
H_{a,b,\lambda}(\bp)=\min_{\bq\in\Delta_N}\mathcal R^{\rm CE}_{a,b,\lambda}(\bq;\bp),
\qquad \bq\in\Delta_N,
\label{eq:population-risk}
\end{equation}
where $\Delta_N=\{\bq\ge\0:\1^T\bq=1\}$, $\Delta_N^\circ$ denotes its positive interior, and zero-posterior terms are omitted when taking boundary limits. As a minimum of functions that are linear in $\bp$, $H_{a,b,\lambda}$ is concave; it is the generalized entropy associated with the loss, and equals Shannon entropy in the ordinary-logarithm limit.

\begin{proposition}[Population prediction, order calibration, and inverse link]
\label{thm:EulerCE-population}
Assume Eq.~\eqref{eq:stat-domain}, $N\ge2$, and $\bp\in\Delta_N^\circ$.
\begin{enumerate}
\item The risk has a unique interior minimizer $\bq^*(\bp)$, determined by
\begin{equation}
p_i \log_{a,b,\lambda}'(q_i^*)=\nu(\bp),\qquad
q_i^*=(\log_{a,b,\lambda}')^{-1}\!\left(\nu(\bp)/p_i\right),\qquad
\sum_i (\log_{a,b,\lambda}')^{-1}\!\left(\nu(\bp)/p_i\right)=1.
\label{eq:population-KKT}
\end{equation}
The scalar multiplier $\nu(\bp)>0$ is unique.
\item \emph{Order calibration.} The minimizer preserves all class orderings: $p_i>p_j$ if and only if $q_i^*>q_j^*$. Moreover, let $C(\bp)=\{\bq\in\Delta_N:\ q_j\ge q_i\text{ for some }i,j\text{ with }p_i>p_j\}$ be the set of predictions that reverse at least one strict posterior ordering. If $C(\bp)\neq\emptyset$, then $\inf_{\bq\in C(\bp)}\mathcal R^{\rm CE}_{a,b,\lambda}(\bq;\bp)>H_{a,b,\lambda}(\bp)$. Consequently, the loss is classification-calibrated and top-$k$ calibrated for every $k=1,\ldots,N-1$~\cite{Tewari2007,YangKoyejo2020}.
\item The map $\mathcal T_{a,b,\lambda}:\bp\mapsto\bq^*(\bp)$ is a bijection of the open simplex with explicit inverse
\begin{equation}
p_i=\frac{m_{a,b,\lambda}(q_i^*)}{\sum_jm_{a,b,\lambda}(q_j^*)},\qquad
m_{a,b,\lambda}(x)=\frac{1}{\log_{a,b,\lambda}'(x)}
=\frac1{\alpha_\lambda x^{a-1}+\beta_\lambda x^{b-1}},
\label{eq:posterior-recovery}
\end{equation}
and $m_{a,b,\lambda}(x)=x$ in the ordinary-logarithm limit. Since $m_{a,b,\lambda}$ is strictly increasing, $\mathcal T_{a,b,\lambda}^{-1}$ also preserves all orderings.
\end{enumerate}
\end{proposition}
\begin{proof}
For interior $\bp$, the risk is lower semicontinuous on the compact simplex and strictly convex wherever finite. A minimizer exists. It is interior: when a missing coordinate has finite boundary loss, transferring mass to it has derivative $-p_i \log_{a,b,\lambda}'(t)\to-\infty$ as $t\downarrow0$; when its boundary loss is infinite, it cannot minimize. The KKT conditions give Eq.~\eqref{eq:population-KKT}. The sum in that equation decreases continuously from infinity to zero as $\nu$ increases, establishing uniqueness.

Since $(\log_{a,b,\lambda}')^{-1}$ is decreasing, Eq.~\eqref{eq:population-KKT} preserves class order, so $\bq^*\notin C(\bp)$. The set $C(\bp)$ is a finite union of closed sets and hence closed. The lower semicontinuous risk attains its infimum on the compact set $C(\bp)$, and uniqueness of the unrestricted minimizer makes that infimum strictly larger than $H_{a,b,\lambda}(\bp)$. A prediction whose $\arg\max$, or whose set of $k$ largest coordinates, is not Bayes optimal for $\bp$ reverses some strict ordering $p_i>p_j$ and therefore lies in $C(\bp)$; this gives classification and top-$k$ calibration.

The KKT equations yield $p_i=\nu m_{a,b,\lambda}(q_i^*)$, and normalization gives Eq.~\eqref{eq:posterior-recovery}. Conversely, any interior $\bq$ defines an interior $\bp$ by that formula and then satisfies the KKT conditions. This proves bijectivity. Finally, $m_{a,b,\lambda}'(x)=\varepsilon_{a,b,\lambda}(x)m_{a,b,\lambda}(x)/x>0$.
\end{proof}

For posteriors on the boundary, all optimal mass lies on their positive support: transferring mass from a zero-posterior class to a positive one strictly reduces the risk. Applying the same argument on that support establishes the Bayes ordering and calibration there as well, with $0\log_{a,b,\lambda}(0)$ interpreted as zero. Classification calibration supports Bayes consistency for learning procedures whose surrogate risk approaches the population optimum~\cite{Tewari2007}.

\begin{theorem}[Properness and locality]
\label{thm:EulerCE-locality}
\begin{enumerate}
\item Assume Eq.~\eqref{eq:stat-domain} and $N\ge2$. The Euler cross-entropy is strictly proper, that is, $\bq^*(\bp)=\bp$ for every $\bp\in\Delta_N^\circ$, if and only if $(a,b)=(0,0)$.
\item Let $N\ge3$ and let $g:(0,1)\to\mathbb R$ be differentiable and strictly increasing. If $\bq=\bp$ minimizes $-\sum_ip_ig(q_i)$ over $\Delta_N^\circ$ for every $\bp\in\Delta_N^\circ$, then $g(x)=\alpha\ln x+\beta$ with $\alpha>0$.
\end{enumerate}
\end{theorem}
\begin{proof}
(i) If $\bq^*(\bp)=\bp$ for every $\bp$, Eq.~\eqref{eq:population-KKT} requires $\omega_{a,b,\lambda}(p_i)=p_i\log_{a,b,\lambda}'(p_i)$ to be the same for all coordinates. If $|a|>0$, $\omega_{a,b,\lambda}(t)\to\infty$ as $t\downarrow0$; if $|a|=0<b$, $\omega_{a,b,\lambda}(t)=t^b\to0$. For $N=2$, both contradict $\omega_{a,b,\lambda}(t)=\omega_{a,b,\lambda}(1-t)$, since $\omega_{a,b,\lambda}(1-t)\to1$. For $N>2$, hold a second coordinate fixed and positive while taking the first to zero, and the same contradiction applies. For $|a|=b=0$, $\omega_{a,b,\lambda}\equiv1$ and the minimizer is $\bp$.

(ii) An interior minimizer satisfies the stationarity condition $p_ig'(p_i)=\nu(\bp)$ for all $i$. Given $x,y>0$ with $x+y<1$, choose $\bp$ with $p_1=x$, $p_2=y$ and the remaining mass spread over the other $N-2\ge1$ coordinates. Then $xg'(x)=yg'(y)$. For arbitrary $x,y\in(0,1)$ choose $z>0$ with $x+z<1$ and $y+z<1$; applying the preceding identity twice connects $x$ and $y$. Hence $xg'(x)=\alpha$ is constant on $(0,1)$, with $\alpha>0$ because $g$ is increasing, and integration gives the claim.
\end{proof}

Part~(ii) is Bernardo's characterization of proper local scoring rules~\cite{Bernardo1979}, specialized to the stationarity condition; for $N=2$ locality imposes no restriction, but part~(i) shows that binary Euler losses are not proper either. Hence the observation that ``$q=p$ for every interior multiclass posterior forces $x\log_{a,b,\lambda}'(x)$ to be constant, which, with the normalization $\log_{a,b,\lambda}'(1)=1$, gives the ordinary logarithm'' is correct, and for $N\ge3$ it applies to every differentiable increasing deformed logarithm, not only to the Euler family. Properness of a deformed loss must therefore be obtained either through an invertible link or through a non-local correction. The next theorem develops the first route; Proposition~\ref{prop:proper-Euler-score} gives the second.

\begin{theorem}[Proper composite structure, Bayes risk, and exact regret]
\label{thm:EulerCE-composite}
Assume Eq.~\eqref{eq:stat-domain}, $N\ge2$, and let $\mathcal T_{a,b,\lambda}$ be the bijection of Proposition~\ref{thm:EulerCE-population}.
\begin{enumerate}
\item $\mathcal T_{a,b,\lambda}$ and $H_{a,b,\lambda}$ are continuously differentiable on $\Delta_N^\circ$, and $H_{a,b,\lambda}$ is strictly concave with
\begin{equation}
\nabla H_{a,b,\lambda}(\bp)=-\log_{a,b,\lambda}\bigl(\mathcal T_{a,b,\lambda}(\bp)\bigr),
\label{eq:Bayes-risk-gradient}
\end{equation}
up to an additive multiple of $\1$, which does not act on the simplex.
\item \emph{Proper composite representation.} For every $\bp'\in\Delta_N^\circ$ and every label $y$,
\begin{equation}
-\log_{a,b,\lambda}\bigl(\mathcal T_{a,b,\lambda}(\bp')_y\bigr)=H_{a,b,\lambda}(\bp')+\bigl\langle\nabla H_{a,b,\lambda}(\bp'),\bolde_y-\bp'\bigr\rangle .
\label{eq:Savage-representation}
\end{equation}
Thus $\mathcal L^{\rm comp}_{a,b,\lambda}(\bp',y)=-\log_{a,b,\lambda}(\mathcal T_{a,b,\lambda}(\bp')_y)$, that is, Euler cross-entropy evaluated at the prediction $\bq=\mathcal T_{a,b,\lambda}(\bp')$, is the strictly proper loss with Bayes risk $H_{a,b,\lambda}$, and $\mathcal T_{a,b,\lambda}$ is its link~\cite{Savage1971,Williamson2016}. The proper loss $\mathcal L^{\rm comp}_{a,b,\lambda}$ is non-local: it depends on $\bp'$ through all coordinates of $\mathcal T_{a,b,\lambda}(\bp')$.
\item \emph{Exact regret.} For all interior $\bp,\bq$, with the recovered posterior $\br=\mathcal T_{a,b,\lambda}^{-1}(\bq)$ given by Eq.~\eqref{eq:posterior-recovery} (the symbol of Section~\ref{sec:canonical-softmax}),
\begin{equation}
\mathcal R^{\rm CE}_{a,b,\lambda}(\bq;\bp)-H_{a,b,\lambda}(\bp)
=\sum_{i=1}^Np_i\,D_{-\log_{a,b,\lambda}}(q_i,q_i^*)
=D_{-H_{a,b,\lambda}}(\bp\Vert\br),
\label{eq:EulerCE-exact-regret}
\end{equation}
where $D_{-\log_{a,b,\lambda}}(x,x')=\log_{a,b,\lambda}(x')+\log_{a,b,\lambda}'(x')(x-x')-\log_{a,b,\lambda}(x)\ge0$ is the scalar Bregman gap of the concave function $\log_{a,b,\lambda}$. Moreover,
\begin{equation}
\mathcal R^{\rm CE}_{a,b,\lambda}(\bq;\bp)-H_{a,b,\lambda}(\bp)\ge\frac{\varepsilon_{a,b,\lambda}(1)}{2}\sum_{i=1}^Np_i(q_i-q_i^*)^2 .
\label{eq:EulerCE-q-regret}
\end{equation}
\item \emph{Closed forms.} In the ordinary-logarithm limit, $H_{a,b,\lambda}$ is Shannon entropy and Eq.~\eqref{eq:EulerCE-exact-regret} is $D_{\rm KL}(\bp\Vert\bq)$. On the GCE boundary $a=0<b$, with $r_b=1/(1-b)$,
\begin{equation}
q_i^*=\frac{p_i^{r_b}}{\sum_jp_j^{r_b}},\qquad
H_{a,b,\lambda}(\bp)=\frac{1-\|\bp\|_{r_b}}{b}.
\label{eq:GCE-Bayes-risk}
\end{equation}
\end{enumerate}
\end{theorem}
\begin{proof}
(i) The KKT system~\eqref{eq:population-KKT} together with $\sum_iq_i=1$ has a nonsingular Jacobian in $(\bq,\nu)$ because $\log_{a,b,\lambda}''<0$, so $\mathcal T_{a,b,\lambda}$ and $\nu$ are continuously differentiable by the implicit function theorem; Eq.~\eqref{eq:population-sensitivity} below gives the derivative. Since $H_{a,b,\lambda}(\bp)=\mathcal R^{\rm CE}_{a,b,\lambda}(\mathcal T_{a,b,\lambda}(\bp);\bp)$,
\[
\frac{\partial H_{a,b,\lambda}}{\partial p_i}=-\log_{a,b,\lambda}(q_i^*)-\sum_jp_j\log_{a,b,\lambda}'(q_j^*)\frac{\partial q_j^*}{\partial p_i}
=-\log_{a,b,\lambda}(q_i^*)-\nu\frac{\partial}{\partial p_i}\sum_jq_j^*=-\log_{a,b,\lambda}(q_i^*).
\]
For a nonzero tangent perturbation $\mathrm d\bp$, put $s_i=\mathrm dp_i/p_i$, $t_i=q_i^*/\varepsilon_{a,b,\lambda}(q_i^*)>0$, and $\bar s=\sum_i t_i s_i/\sum_i t_i$. Differentiating the KKT equations gives $\mathrm dq_i^*=t_i(s_i-\bar s)$ and
\begin{equation}
\mathrm d\bp^T\nabla^2(-H_{a,b,\lambda})(\bp)\,\mathrm d\bp
=\nu\sum_i t_i(s_i-\bar s)^2>0.
\label{eq:Bayes-variance}
\end{equation}
The variance vanishes only when every $s_i$ is equal; $\sum_i p_i s_i=0$ then forces $\mathrm d\bp=0$. This proves strict concavity throughout Eq.~\eqref{eq:stat-domain}, including the GCE and logarithmic boundaries. The smooth KKT system also gives $H_{a,b,\lambda}\in C^2$ on the open simplex.

(ii) Substituting Eq.~\eqref{eq:Bayes-risk-gradient} gives $H_{a,b,\lambda}(\bp')+\langle\nabla H_{a,b,\lambda}(\bp'),\bolde_y-\bp'\rangle=-\sum_jp'_j\log_{a,b,\lambda}(q_j^*)-\log_{a,b,\lambda}(q_y^*)+\sum_jp'_j\log_{a,b,\lambda}(q_j^*)$, with $\bq^*=\mathcal T_{a,b,\lambda}(\bp')$, which is Eq.~\eqref{eq:Savage-representation}. Averaging over $Y\sim\bp$ gives the expected loss $H_{a,b,\lambda}(\bp')+\langle\nabla H_{a,b,\lambda}(\bp'),\bp-\bp'\rangle$, which exceeds $H_{a,b,\lambda}(\bp)$ by $D_{-H_{a,b,\lambda}}(\bp\Vert\bp')$. Strict concavity makes this gap positive unless $\bp'=\bp$.

(iii) Since $p_i\log_{a,b,\lambda}'(q_i^*)=\nu$ and $\sum_i(q_i-q_i^*)=0$, $\sum_ip_i\log_{a,b,\lambda}'(q_i^*)(q_i-q_i^*)=0$. Adding this zero to $\mathcal R^{\rm CE}_{a,b,\lambda}(\bq;\bp)-\mathcal R^{\rm CE}_{a,b,\lambda}(\bq^*;\bp)=\sum_ip_i[\log_{a,b,\lambda}(q_i^*)-\log_{a,b,\lambda}(q_i)]$ gives the first identity. Because $\bq=\mathcal T_{a,b,\lambda}(\br)$, part~(ii) gives the second. For the bound, $-\log_{a,b,\lambda}''(x)=\alpha_\lambda(1-a)x^{a-2}+\beta_\lambda(1-b)x^{b-2}$ is a sum of two nonnegative, nonincreasing terms, so $-\log_{a,b,\lambda}''(x)\ge-\log_{a,b,\lambda}''(1)=\varepsilon_{a,b,\lambda}(1)>0$ on $(0,1]$, and Taylor's theorem applies to each $D_{-\log_{a,b,\lambda}}$.

(iv) For GCE, $\log_{a,b,\lambda}'(x)=x^{b-1}$, and Eq.~\eqref{eq:population-KKT} gives $q_i^*\propto p_i^{1/(1-b)}$. With $Z=\sum_jp_j^{r_b}$, $\sum_ip_i(q_i^*)^b=Z/Z^b=Z^{1/r_b}=\|\bp\|_{r_b}$, so $H_{a,b,\lambda}=(1-\|\bp\|_{r_b})/b$. The logarithmic case is classical.
\end{proof}

The theorem explains what Euler cross-entropy estimates. Its population output is not the posterior $\bp$ but the deformed posterior $\mathcal T_{a,b,\lambda}(\bp)$, which preserves every class ranking. The posterior itself is recovered exactly by $\mathcal T_{a,b,\lambda}^{-1}$ in Eq.~\eqref{eq:posterior-recovery}. In the language of~\cite{ReidWilliamson2010,Williamson2016}, the deformation is absorbed into the link, and the statistical content of the loss is its Bayes risk $H_{a,b,\lambda}$. Equation~\eqref{eq:EulerCE-exact-regret} then measures the price of an imperfect prediction exactly: in prediction coordinates it is a $\bp$-weighted sum of scalar Bregman gaps, and in posterior coordinates it is the Bregman divergence of $-H_{a,b,\lambda}$ between the true posterior and the recovered one.

\begin{theorem}[Uniform curvature and a dimension-free surrogate-regret bound]
\label{thm:EulerCE-regret-bound}
Let $a<0<b<1$, let $c_{a,b,\lambda}=\min_{0<x\le1}\omega_{a,b,\lambda}(x)$, the constant that also appears in Eq.~\eqref{eq:entropy-comparison-constant} and equals the global minimum $\omega_{\min}$ of Eq.~\eqref{eq:omega-min} when $\lambda>(|a|/b)^{2/(b-a)}$, and put
\[
\mu^{H}_{a,b,\lambda}=\frac{c_{a,b,\lambda}}{1-a}.
\]
\begin{enumerate}
\item $-H_{a,b,\lambda}$ is $\mu^{H}_{a,b,\lambda}$-strongly convex on $\Delta_N^\circ$ with respect to $\|\cdot\|_1$, for every $N\ge2$:
$D_{-H_{a,b,\lambda}}(\bp\Vert\bp')\ge\tfrac12\mu^{H}_{a,b,\lambda}\|\bp-\bp'\|_1^2$.
\item For all interior $\bp,\bq$, with $\br=\mathcal T_{a,b,\lambda}^{-1}(\bq)$ and any $k\in\arg\max_iq_i$,
\begin{equation}
\max_ip_i-p_k\le\|\bp-\br\|_1\le
\sqrt{\frac{2}{\mu^{H}_{a,b,\lambda}}\bigl[\mathcal R^{\rm CE}_{a,b,\lambda}(\bq;\bp)-H_{a,b,\lambda}(\bp)\bigr]}.
\label{eq:EulerCE-conditional-bound}
\end{equation}
\item Let $\bq(\cdot)$ be any measurable predictor with values in $\Delta_N^\circ$, let $\hat y(\bx)\in\arg\max_iq_i(\bx)$, and assume $\bp(X)\in\Delta_N^\circ$ almost surely. Write $\mathrm{Err}$ for the misclassification risk, $\mathrm{Err}^*$ for the Bayes risk, $\overline{\mathcal R}_{a,b,\lambda}^{\rm CE}(\bq)=\mathbb E\,\mathcal L_{a,b,\lambda}(q_Y(X))$, and $\overline{\mathcal R}_{a,b,\lambda}^{\rm CE,*}=\mathbb E\,H_{a,b,\lambda}(\bp(X))$. Then
\begin{equation}
\mathrm{Err}(\hat y)-\mathrm{Err}^*\le\mathbb E\,\|\bp(X)-\br(X)\|_1\le
\sqrt{\frac{2(1-a)}{c_{a,b,\lambda}}\bigl[\overline{\mathcal R}_{a,b,\lambda}^{\rm CE}(\bq)-\overline{\mathcal R}_{a,b,\lambda}^{\rm CE,*}\bigr]}.
\label{eq:EulerCE-global-bound}
\end{equation}
\item On the GCE boundary $|a|=0$ with $1/2<b<1$, no such bound with a uniform constant exists: there is no $C<\infty$ with $\|\bp-\mathcal T_{a,b,\lambda}^{-1}(\bq)\|_1^2\le C[\mathcal R^{\rm CE}_{a,b,\lambda}(\bq;\bp)-H_{a,b,\lambda}(\bp)]$ for all interior $\bp,\bq$.
\end{enumerate}
\end{theorem}
\begin{proof}
(i) Use Eq.~\eqref{eq:Bayes-variance}. Since $\nu t_i/p_i=\omega_{a,b,\lambda}(q_i^*)/\varepsilon_{a,b,\lambda}(q_i^*)\ge c_{a,b,\lambda}/(1-a)=\mu^{H}_{a,b,\lambda}$,
\begin{align*}
\nu\sum_i t_i(s_i-\bar s)^2
&=\min_{c\in\mathbb R}\sum_i\nu t_i(s_i-c)^2\\
&\ge\mu^{H}_{a,b,\lambda}\min_{c\in\mathbb R}\sum_i p_i(s_i-c)^2
=\mu^{H}_{a,b,\lambda}\sum_i\frac{(\mathrm dp_i)^2}{p_i}.
\end{align*}
The last equality uses $\sum_i p_i s_i=0$. Integration along the segment between two interior posteriors yields the stronger comparison
\begin{equation}
D_{-H_{a,b,\lambda}}(\bp\Vert\bp')\ge\mu^{H}_{a,b,\lambda}D_{\rm KL}(\bp\Vert\bp')
\ge\frac{\mu^{H}_{a,b,\lambda}}2\|\bp-\bp'\|_1^2.
\label{eq:Bayes-KL-domination}
\end{equation}
The second inequality is Pinsker's inequality.

(ii) By Proposition~\ref{thm:EulerCE-population}(iii), $\arg\max_iq_i=\arg\max_ir_i$. For a Bayes class $y^*$ and $k\in\arg\max_iq_i$, $p_{y^*}-p_k\le(p_{y^*}-r_{y^*})+(r_k-p_k)\le\|\bp-\br\|_1$. The second inequality combines part~(i) with Eq.~\eqref{eq:EulerCE-exact-regret}.

(iii) The conditional excess misclassification risk of $\hat y$ at $\bx$ is $\max_ip_i(\bx)-p_{\hat y(\bx)}(\bx)$. Take expectations in Eq.~\eqref{eq:EulerCE-conditional-bound}, use Jensen's inequality for the square root, and note that $\overline{\mathcal R}_{a,b,\lambda}^{\rm CE}(\bq)-\overline{\mathcal R}_{a,b,\lambda}^{\rm CE,*}=\mathbb E[\mathcal R^{\rm CE}_{a,b,\lambda}(\bq(X);\bp(X))-H_{a,b,\lambda}(\bp(X))]$.

(iv) By Eq.~\eqref{eq:GCE-Bayes-risk}, $-H_{a,b,\lambda}=(\|\bp\|_{r_b}-1)/b$ with $r_b=1/(1-b)>2$. The norm is positively homogeneous of degree one, so its Hessian annihilates $\bp$. For $\mathbf d=\bolde_N-\bp$, which is zero-sum,
\[
\mathbf d^T\nabla^2(-H_{a,b,\lambda})\,\mathbf d=\bolde_N^T\nabla^2(-H_{a,b,\lambda})\,\bolde_N
=\frac{r_b-1}{b}\|\bp\|_{r_b}^{1-r_b}\Bigl(p_N^{r_b-2}-\frac{p_N^{2r_b-2}}{\|\bp\|_{r_b}^{r_b}}\Bigr)
\le\frac{r_b-1}{b}N^{r_b-1}p_N^{r_b-2},
\]
because $\|\bp\|_{r_b}\ge\max_ip_i\ge1/N$. Also $\|\mathbf d\|_1=2(1-p_N)\ge1$ for $p_N\le1/2$. The ratio therefore tends to zero as $p_N\downarrow0$. Taking $\br=\bp+\epsilon\mathbf d$ with $\epsilon\to0$, the second-order expansion $D_{-H_{a,b,\lambda}}(\bp\Vert\br)=\tfrac12\epsilon^2\mathbf d^T\nabla^2(-H_{a,b,\lambda})(\bp)\mathbf d+o(\epsilon^2)$ and surjectivity of $\mathcal T_{a,b,\lambda}^{-1}$ show that $\|\bp-\br\|_1^2/D_{-H_{a,b,\lambda}}(\bp\Vert\br)$ is unbounded.
\end{proof}
\begin{remark}[Interpretation and potential use]
This result makes posterior recovery quantitatively meaningful: a small excess surrogate risk controls an error in the recovered posterior, not necessarily the raw network output. Its dimension-free form is useful when the number of labels is large. The parameter-dependent constant must still be inspected; a uniform statement in dimension is not a uniform statement over all Euler parameters.
\end{remark}

Theorem~\ref{thm:EulerCE-regret-bound} gives the Euler loss a quantitative statistical guarantee that does not depend on the number of classes. Driving the excess Euler risk to zero drives both the excess misclassification risk and the $\ell_1$ error of the recovered posterior to zero, at the square-root rate familiar from Pinsker-type arguments~\cite{Zhang2004}. The constant is explicit and is governed by the minimum gradient weight. A smaller minimum gradient weight weakens this sufficient regret constant. Gradient attenuation alone does not establish robustness to label noise. For the ordinary logarithm, $H_{a,b,\lambda}$ is Shannon entropy and the exact Pinsker constant is $2$. The probability-domain minimum $c_{a,b,\lambda}$ need not approach one along every path to $(0,0)$, so the displayed sufficient bound need not converge sharply along every parameter path. Part~(iv) identifies a structural difference with the bounded GCE boundary: for $b>1/2$, including the common choice $b=0.7$, the GCE Bayes risk loses curvature near classes of small posterior probability. Any negative exponent $a<0$ restores a uniform modulus. This is a statistical counterpart of the gradient floor $\omega_{\min}>0$ in Proposition~\ref{prop:omega-profile}.

\begin{remark}[Dimension dependence on the GCE boundary]
For every fixed $0<b<1$, GCE has no positive quadratic $\ell_1$ curvature modulus valid for all class counts. At the uniform posterior, the tangent Hessian of $-H_{a,b,\lambda}$ is $N^{1-b}/(1-b)$ times the identity. For even $N$, choose a tangent vector with half its entries $1/N$ and half $-1/N$. Its $\ell_1$ norm is one and its curvature is $N^{-b}/(1-b)\to0$. For $b>1/2$, part~(iv) proves the stronger failure of a uniform modulus even at fixed $N$. These claims do not deny classification calibration or dimension-dependent bounds in other GCE regimes.
\end{remark}
The inverse link in Eq.~\eqref{eq:posterior-recovery} costs $\mathcal O(N)$ scalar operations and uses exactly the inverse-Hessian weight that appears in Euler optimization. At the population optimum it recovers the posterior exactly, and Theorem~\ref{thm:EulerCE-regret-bound} bounds its error for an imperfect predictor. For a fitted model, $r_i=m_{a,b,\lambda}(q_i)/\sum_jm_{a,b,\lambda}(q_j)$ is therefore a principled probability output whose finite-sample calibration can be assessed on held-out data.

Elasticity describes the sensitivity of this population prediction. To account for simplex normalization, put $t_i=q_i^*/\varepsilon_{a,b,\lambda}(q_i^*)$. Differentiating the KKT equations along a posterior perturbation with $\sum_i\mathrm dp_i=0$ gives
\begin{equation}
\mathrm dq_i^*=t_i\left(\mathrm d\ln p_i-
\frac{\sum_j t_j\,\mathrm d\ln p_j}{\sum_j t_j}\right).
\label{eq:population-sensitivity}
\end{equation}
Indeed, $\varepsilon_i\mathrm d\ln q_i^*=\mathrm d\ln p_i-\mathrm d\ln\nu$ and $\sum_i\mathrm dq_i^*=0$. The factor $1/\varepsilon_i$ sets the local response, while the weighted mean enforces the common normalization. This makes the role of curvature explicit without treating the class coordinates as independent.

On the GCE boundary, $q_i^*\propto p_i^{1/(1-b)}$, giving uniform power sharpening. On the Amari line, $q_i^*\propto p_i^{1/(1-a)}$, giving power flattening. Interior Euler choices permit more varied transformations. Table~\ref{tab:population-tilt} shows an example in which both the largest and smallest components increase. This illustrates flexibility beyond the particular CE, GCE, Amari, and Kaniadakis settings shown in the table.

\begin{table}[htbp]
\centering\small
\caption{Population minimizers at $\lambda=1$ for $\bp=(0.6,0.3,0.08,0.02)$. Values are rounded to three decimals and need not sum to one after rounding. They are computed from the KKT equations and checked by independent constrained minimization. Applying Eq.~\eqref{eq:posterior-recovery} recovers $\bp$ to numerical precision in every row.}
\label{tab:population-tilt}
\begin{tabular}{lcccc}
\hline
Loss & $q_1^*$ & $q_2^*$ & $q_3^*$ & $q_4^*$\\\hline
CE, $(0,0)$ & $0.600$ & $0.300$ & $0.080$ & $0.020$\\
GCE, $(0,0.5)$ & $0.788$ & $0.197$ & $0.014$ & $0.001$\\
Kaniadakis, $(-0.5,0.5)$ & $0.539$ & $0.304$ & $0.113$ & $0.043$\\
Amari, $(-0.5,0)$ & $0.501$ & $0.316$ & $0.131$ & $0.052$\\
Euler, $(-0.3,0.7)$ & $0.632$ & $0.268$ & $0.076$ & $0.024$\\
Euler, $(-0.2,0.5)$ & $0.646$ & $0.271$ & $0.066$ & $0.018$\\\hline
\end{tabular}
\end{table}

\begin{proposition}[Strictly proper Euler score]
\label{prop:proper-Euler-score}
Let $-1<a\le0\le b<1$ and let $F_{a,b,\lambda}$ be the Euler potential in Eq.~\eqref{eq:Euler-potential-vector}. Define
\begin{align}
S_{a,b,\lambda}(\bq;y)&=-\log_{a,b,\lambda}(q_y)+\sum_i\kappa_{a,b,\lambda}(q_i),\nonumber\\
\kappa_{a,b,\lambda}(x)&=\frac{\alpha_\lambda x^{a+1}}{a+1}+\frac{\beta_\lambda x^{b+1}}{b+1}.
\label{eq:proper-Euler-score}
\end{align}
Use $\kappa_{0,0}(x)=x$ by continuity. For interior $\bp,\bq$,
\begin{equation}
\mathbb E_{Y\sim\bp}S_{a,b,\lambda}(\bq;Y)-\mathbb E_{Y\sim\bp}S_{a,b,\lambda}(\bp;Y)
=D_{F_{a,b,\lambda}}(\bp\Vert\bq)\ge0,
\label{eq:proper-score-regret}
\end{equation}
with equality exactly when $\bq=\bp$. Thus $S_{a,b,\lambda}$ is strictly proper. Writing $\omega_i=q_i\log_{a,b,\lambda}'(q_i)$ as in Eq.~\eqref{eq:ri-Euler} and $\bar\omega=\sum_iq_i\omega_i$, its softmax gradient is
\begin{equation}
\frac{\partial S_{a,b,\lambda}}{\partial z_j}=\omega_y(q_j-\delta_{jy})+q_j(\omega_j-\bar\omega).
\label{eq:proper-score-gradient}
\end{equation}
The logarithmic limit is ordinary CE plus the constant $1$. On the boundary $a=0<b$, the score is the bounded power score
\begin{equation}
S_{a,b,\lambda}(\bq;y)=\frac{1-q_y^b}{b}+\frac{\sum_i q_i^{1+b}}{1+b}.
\end{equation}
\end{proposition}
\begin{proof}
For the displayed potential $\phi_{a,b,\lambda}$ in Eq.~\eqref{eq:Euler-potential-vector}, $\kappa_{a,b,\lambda}(x)=x\log_{a,b,\lambda}(x)-\phi_{a,b,\lambda}(x)+\phi_{a,b,\lambda}(0)$. Substituting this identity into the expected score yields Eq.~\eqref{eq:proper-score-regret}. Strict convexity of $F_{a,b,\lambda}$ proves strict propriety, consistent with the Bregman construction of scoring rules~\cite{Gneiting2007,Banerjee2005}. Since $a>-1$, $\phi_{a,b,\lambda}(0)$ is finite and the one-hot representation is also valid:
$D_{F_{a,b,\lambda}}(\bolde_y\Vert\bq)=S_{a,b,\lambda}(\bq;y)+\phi_{a,b,\lambda}(1)-\phi_{a,b,\lambda}(0)$.
Moreover, $\kappa_{a,b,\lambda}'(q_i)=q_i \log_{a,b,\lambda}'(q_i)=\omega_i$. Applying the softmax Jacobian to $\partial S_{a,b,\lambda}/\partial q_i=-\delta_{iy}\log_{a,b,\lambda}'(q_y)+\omega_i$ gives Eq.~\eqref{eq:proper-score-gradient}. The limits follow directly.
\end{proof}

This construction offers two useful training choices. Euler cross-entropy supplies a simple scalar reweighting of each one-hot CE gradient and an explicit probability-recovery map. The proper Euler score targets the posterior directly and adds the centered correction $q_j(\omega_j-\bar\omega)$ to the gradient. Both are computed in $\mathcal O(N)$ operations and share the Euler--Bregman potential used for optimization. Boundedness of the power-score boundary is a mathematical property; its resistance to a particular label-noise model requires a separate analysis or experiment.

\noindent\textbf{Which statistical route to use.} Table~\ref{tab:EulerCE-statistics} summarizes the results of this subsection. For class decisions, train with Euler cross-entropy and predict $\arg\max_iq_i$; order calibration makes this consistent. For probabilities, either apply the inverse link $\mathcal T_{a,b,\lambda}^{-1}$, whose error is controlled by Theorem~\ref{thm:EulerCE-regret-bound}, or train with the corrected score $S_{a,b,\lambda}$, whose output is itself the posterior at the population optimum. Posterior recovery uses all coordinates through its normalizer; the corrected score generally includes a sum over classes. The general restriction on nonlogarithmic proper local scores applies to $N\ge3$, as stated in Theorem~\ref{thm:EulerCE-locality}.

\begin{table}[ht]
\centering\small
\caption{Statistical properties of cross-entropy (CE), the GCE boundary $a=0<b$ (with $r_b=1/(1-b)$), and full-range Euler cross-entropy $a<0<b<1$. All entries are population statements proved in this subsection.}
\label{tab:EulerCE-statistics}
\begin{tabular}{L{.34\textwidth}L{.14\textwidth}L{.17\textwidth}L{.24\textwidth}}
\hline
Property & CE & GCE & Euler, $|a|>0$\\\hline
Proper local score (Theorem~\ref{thm:EulerCE-locality}) & yes & no & no\\
Order, classification and top-$k$ calibration & yes & yes & yes\\
Population output $\mathcal T_{a,b,\lambda}(\bp)$ & $\bp$ & $\bp^{r_b}/\|\bp\|_{r_b}^{r_b}$ & Eq.~\eqref{eq:population-KKT}\\
Closed-form inverse link & identity & power & Eq.~\eqref{eq:posterior-recovery}\\
Proper composite; Bayes risk $H_{a,b,\lambda}$ & Shannon & $(1-\|\bp\|_{r_b})/b$ & Eq.~\eqref{eq:population-risk}\\
Uniform strong concavity of $H_{a,b,\lambda}$; dimension-free regret bound & yes & no dimension-free modulus for $b>0$ & yes, $\mu^{H}_{a,b,\lambda}=c_{a,b,\lambda}/(1-a)$\\
Bounded loss & no & yes & no\\
Strictly proper corrected score & CE itself & power score & $S_{a,b,\lambda}$, Eq.~\eqref{eq:proper-Euler-score}\\\hline
\end{tabular}
\end{table}

\subsection{Natural-gradient backpropagation for Euler cross entropy}
\label{subsec:EulerCE-NG}

Section~\ref{subsec:EulerCE-backprop} gives the ordinary Euclidean backpropagation signal through the softmax layer. It can be extended to natural-gradient (NG) backpropagation by measuring steepest descent with respect to the Fisher information metric of the predictive distribution \cite{Amari1998,Amaribook,Martens2020}. Let
\begin{equation}
 h_{a,b,\lambda}(q_{i})=\frac{\partial}{\partial q_{i}}\log_{a,b,\lambda}(q_{i}) = \alpha_\lambda q_i^{a-1}+\beta_\lambda q_i^{b-1}, \qquad \omega_{i}=q_{i} h_{a,b,\lambda}(q_{i})=\omega_{a,b,\lambda}(q_i),
\end{equation}
as in Eq.~\eqref{eq:ri-Euler}, and define the probability-space gradient
\begin{equation}
 u_{i}=\frac{\partial\mathcal L_{a,b,\lambda}}{\partial q_{i}}=-p_{i} h_{a,b,\lambda}(q_{i}).
\end{equation}
The Fisher information matrix of a categorical distribution parameterized by logits is
\begin{equation}
 \bI_{\bz}(\bq)=\diag(\bq)-\bq\bq^{T}.
 \label{eq:softmax-FIM-EulerCE}
\end{equation}
For $q_i>0$, its nullspace is $\operatorname{span}\{\1\}$ and its restriction to $\1^\perp$ is positive definite. This is the logit-coordinate Fisher matrix; in probability coordinates the Fisher--Rao metric is $\diag(1/q_i)$ restricted to tangent vectors with zero sum.

\begin{remark}[Fisher--Rao versus Euler Hessian geometry]
\label{rem:fisher-vs-hessian}
The matrix in Eq.~\eqref{eq:softmax-FIM-EulerCE} is a genuine Fisher information matrix because it is derived from the categorical predictive model. By contrast, the matrix $\nabla^2F_{a,b,\lambda}(\bw)$ induced later by an Euler mirror potential is a Hessian metric. It defines a legitimate Riemannian preconditioner, but it should not be called a Fisher information matrix unless a statistical model is explicitly identified whose Fisher metric equals that Hessian. We therefore reserve ``Natural Gradient'' for the model-Fisher construction in this subsection and use ``Euler Hessian/Riemannian preconditioning'' for the mirror-potential geometry.
\end{remark}

The Euler logit gradient can be written as
\begin{equation}
 \bg_{\bz}^{a,b,\lambda}=\nabla_{\bz}\mathcal L_{a,b,\lambda}=
 \bI_{\bz}(\bq)\bu, \qquad \1^{T} \bg_{\bz}^{a,b,\lambda}=0,
 \label{eq:Euler-logit-gradient-Fisher}
\end{equation}

which is exactly Eq.~\eqref{eq:general-logit-gradient}. More precisely, if $\bu=\nabla_{\bq}\mathcal L_{a,b,\lambda}$ is viewed as a covector on the probability simplex, then $\bI_{\bz}(\bq)\bu$ is the corresponding Fisher--Rao natural-gradient vector in probability coordinates. The same numerical vector appears here as the Euclidean gradient with respect to logits because the softmax Jacobian is $\bI_{\bz}$. This identity should not be confused with the natural gradient \emph{in logit coordinates} or in the full network-parameter space.

At the output level, the logit-space NG direction is an equivalence class modulo the null direction $\1$. A convenient $\bq$-centered representative solves
\begin{equation}
 \bI_{\bz}(\bq) \bd_{\bz}^{a,b,\lambda}= \bg_{\bz}^{a,b,\lambda}, \qquad \bq^{T} \bd_{\bz}^{a,b,\lambda}=0,
 \label{eq:logit-NG-linear-system}
\end{equation}
and, because $\1^{T} \bg_{\bz}^{a,b,\lambda}=0$, is
\begin{equation}
 d_{z,j}^{a,b,\lambda}=\frac{g_{z,j}^{a,b,\lambda}}{q_{j}}=\sum_{i}p_{i}\omega_{i}-p_{j}h_{a,b,\lambda}(q_{j}), \qquad j=1,\ldots,N.
 \label{eq:exact-logit-NG-Euler}
\end{equation}
Indeed, $\bq^{T} \bd_{\bz}^{a,b,\lambda}=\sum_jg_{z,j}^{a,b,\lambda}=0$ and $\bI_{\bz} \bd_{\bz}^{a,b,\lambda}= \bg_{\bz}^{a,b,\lambda}$. The Euclidean Moore--Penrose representative uses the different gauge $\1^T\bd^{\mathrm{MP}}_{\bz}=0$ and is
\begin{equation}
 \bd^{\mathrm{MP}}_{\bz}=\bI_{\bz}^{+} \bg_{\bz}^{a,b,\lambda}
 =\bd_{\bz}^{a,b,\lambda}-\frac{\1^T \bd_{\bz}^{a,b,\lambda}}{N}\1.
\label{eq:MP-logit-NG-gauge}
\end{equation}
Thus $\bd_{\bz}^{a,b,\lambda}$ and $\bI_{\bz}^{+} \bg_{\bz}^{a,b,\lambda}$ represent the same quotient-space direction but are not generally identical vectors. For one-hot labels with true class $k$ ($p_k=1$), Eq.~\eqref{eq:onehot-logit-gradient} gives
\begin{equation}
 d_{z,j}^{a,b,\lambda} = \omega_{a,b,\lambda}(q_k)\left(1-\frac{p_{j}}{q_{j}}\right), \qquad \omega_{a,b,\lambda}(q_k)=\alpha_\lambda q_k^a+\beta_\lambda q_k^b.
\label{eq:onehot-logit-NG-Euler}
\end{equation}
The logit-space NG representative removes one factor of softmax probability from the Euclidean logit gradient and therefore changes the conditioning and tail scaling. Whether this accelerates training is problem-dependent and should be established empirically. In the bounded robust subfamily $a=0$, $0<b\le1$, the true-class scaling changes from $q_k^b$ in \eqref{eq:onehot-logit-gradient} to order $q_k^{b-1}$ in \eqref{eq:onehot-logit-NG-Euler}; undamped output-level NG can therefore re-amplify very small $q_k$ and should be combined with damping, clipping, or trust-region control when robustness is required.

A regularized output-level direction, with operator bound $\|\bd_{\bz}^{\rm damp}\|_2\le\|\bg_{\bz}^{a,b,\lambda}\|_2/\lambda_F$, can be computed from
\begin{equation}
 (\bI_{\bz}(\bq)+\lambda_F \bI) \bd_{\bz}^{\text{damp}}= \bg_{\bz}^{a,b,\lambda}, \qquad \lambda_F>0.
 \label{eq:damped-output-NG-EulerCE}
\end{equation}

\medskip\noindent\textbf{Diagonal-Fisher approximation.}\ A computationally cheaper approximation uses only the diagonal of the FIM. At the logit level, using the literal diagonal of the logit Fisher matrix $[\bI_{\bz}(\bq)]_{jj} = q_j(1-q_j)$, the diagonal natural gradient direction becomes:
\begin{equation}
 d_{z,j}^{\mathrm{diag}} = \frac{g_{z,j}^{a,b,\lambda}}{q_j(1-q_j) + \varepsilon_F},
 \label{eq:diag-ng-logit}
\end{equation}
where $\varepsilon_F>0$ prevents division by very small probabilities.  Substituting the Euclidean gradient $g_{z,j}^{a,b,\lambda} = \omega_{a,b,\lambda}(q_k)(q_j - \delta_{jk})$, we obtain:

\begin{empheq}[box=\fbox]{equation}
 d_{z,j}^{\mathrm{diag}}=
 \begin{cases}
 \displaystyle -\frac{\omega_{a,b,\lambda}(q_k)(1-q_k)}{q_k(1-q_k)+\varepsilon_F}, & j=k,\\[12pt]
 \displaystyle \frac{\omega_{a,b,\lambda}(q_k)q_j}{q_j(1-q_j)+\varepsilon_F}, & j\ne k.
 \end{cases}
\label{eq:diag-ng-onehot-damped}
\end{empheq}
In the special undamped case $\varepsilon_F=0$, when the factors cancel, this leads to a striking analytical simplification:
\begin{equation}
 d_{z,k}^{\mathrm{diag}}=-\frac{\omega_{a,b,\lambda}(q_k)}{q_k}=-h_{a,b,\lambda}(q_k),\qquad
 d_{z,j}^{\mathrm{diag}}=\frac{\omega_{a,b,\lambda}(q_k)}{1-q_j}\quad(j\ne k).
\label{eq:diag-ng-onehot-undamped}
\end{equation}
Note that damping changes the scaling quantitatively and cannot, in general, be absorbed by replacing $q_k(1-q_k)+\varepsilon_F$ with $(q_k+\varepsilon_F)(1-q_k)$, or analogously for false classes. 

Since diagonal preconditioning may also introduce an irrelevant constant logit component, it is useful to recenter the direction by

\begin{equation}
 \bd_{\bz}\leftarrow \bd_{\bz}-(\bq^{T} \bd_{\bz})\1,
 \label{eq:logit-NG-recentering}
\end{equation}
which fixes the gauge $\bq^{T} \bd_{\bz}=0$.

For the full network, let $\btheta$ denote the parameters that determine the predictive logits and let $\bJ_{n}=\partial \bz_{n}/\partial\btheta$ be the logit Jacobian for sample $n$ in a mini-batch $\mathcal{B}$. The Euclidean Euler-gradient and the model Fisher matrix are
\begin{align}
\bg_{\btheta}^{a,b,\lambda} &= \frac{1}{|\mathcal{B}|}\sum_{n\in\mathcal{B}} \bJ_{n}^{T} \bg_{\bz,n}^{a,b,\lambda}, \label{eq:theta-gradient-EulerCE}\\
\bI_{\btheta} &= \frac{1}{|\mathcal{B}|}\sum_{n\in\mathcal{B}} \bJ_{n}^{T}[\diag(\bq_{n})-\bq_{n} \bq_{n}^{T}] \bJ_{n}. \label{eq:theta-FIM-EulerCE}
\end{align}
The damped parameter-space NG update is
\begin{empheq}[box=\fbox]{equation}
 \btheta_{t+1}=\btheta_{t}-\eta_{t}(\bI_{\btheta}+\lambda_F \bI)^{-1} \bg_{\btheta}^{a,b,\lambda}.
 \label{eq:EulerCE-NG-param}
\end{empheq}

If $a,b,\lambda$ occur only in the loss and not in the predictive probabilities, they have zero model-Fisher information and must not be included as identifiable statistical parameters in an undamped Fisher inverse. Their direct loss derivatives require a separately specified update. Backpropagating an output-preconditioned signal $\bJ_n^T\bd_{\bz,n}$ is not generally equivalent to solving the parameter-space Fisher system. Euclidean damping and diagonal-Fisher approximations also sacrifice general reparameterization invariance. The full inverse in \eqref{eq:EulerCE-NG-param} is usually too expensive for large neural networks. A scalable diagonal-NG backpropagation rule is
\begin{empheq}[box=\fbox]{equation}
\theta_{m,t+1}=\theta_{m,t}-\eta_{t}\frac{g_{m,t}^{a,b,\lambda}}{\widehat I_{m,t}+\varepsilon_F}, \qquad g_{m,t}^{a,b,\lambda}=\frac{\partial\mathcal L_{a,b,\lambda}}{\partial\theta_{m}},
\label{eq:diag-NG-Euler-backprop}
\end{empheq}
with the exact mini-batch diagonal Fisher estimate

\begin{equation}
\widetilde{I}_{m,t}=\frac{1}{|\mathcal{B}|}\sum_{n\in\mathcal{B}}\left[\sum_{j}q_{n,j}\left(\frac{\partial z_{n,j}}{\partial\theta_{m}}\right)^{2}-\left(\sum_{j}q_{n,j}\frac{\partial z_{n,j}}{\partial\theta_{m}}\right)^{2}\right].
\label{eq:diag-Fisher-exact-EulerCE}
\end{equation}

A separate exponentially smoothed estimate can be maintained as $\widehat I_{m,t}=\beta_F\widehat I_{m,t-1}+(1-\beta_F)\widetilde I_{m,t}$ with $0\le\beta_F<1$. The raw and smoothed quantities are therefore not denoted by the same symbol. For a final affine layer $z_{n,j}=\bw_{j}^{T}\bh_{n}+c_{j}$, the model-Fisher diagonal becomes especially simple:

\begin{align}
\widetilde{I}_{W_{jr},t} &= \frac{1}{|\mathcal{B}|}\sum_{n\in\mathcal{B}}q_{n,j}(1-q_{n,j})h_{n,r}^{2}, \label{eq:diag-Fisher-W}\\
\widetilde{I}_{c_{j},t} &= \frac{1}{|\mathcal{B}|}\sum_{n\in\mathcal{B}} q_{n,j}(1-q_{n,j}). \label{eq:diag-Fisher-c}
\end{align}

The Euler deformation affects $g_{z,n}^{a,b,\lambda}$ through $\omega_{i}=\omega_{a,b,\lambda}(q_i)$ and $h_{a,b,\lambda}(q_{i})$, whereas the Fisher matrix in \eqref{eq:theta-FIM-EulerCE} is determined purely by the categorical predictive model. This separation is useful in implementation: first compute the Euler CE backward pass, and then rescale each parameter-gradient component by the chosen inverse Fisher or diagonal-Fisher estimate.

\subsection{The convex conjugate and a proper Fenchel--Young alternative}\label{subsec:FY-conjugate}
Define the regularized maximum
\begin{equation}
A_{a,b,\lambda}(\bs)=\max_{\bq\in\Delta_N}\{\langle\bs,\bq\rangle-F_{a,b,\lambda}(\bq)\},
\qquad \nabla A_{a,b,\lambda}(\bs)=\bq,\qquad \nabla^2A_{a,b,\lambda}(\bs)=\bJ.
\label{eq:Euler-conjugate}
\end{equation}
Uniqueness of the maximizer gives the gradient identity. Generally $A_{a,b,\lambda}$ and $\nu$ are different functions: their gradients are respectively $\bq$ and $\br$. For the stated logarithmic potential, $A_{a,b,\lambda}=1-N+\ln\sum_i e^{s_i}$ and $\nu=\ln\sum_i e^{s_i}$; the additive constant depends on the chosen potential.

The Fenchel--Young construction~\cite{Blondel2020} yields
\begin{equation}
\mathcal L^{\rm FY}_{a,b,\lambda}(\bs;y)=A_{a,b,\lambda}(\bs)-s_y+F_{a,b,\lambda}(\bolde_y),\qquad
\nabla_{\bs}\mathcal L^{\rm FY}_{a,b,\lambda}=\bq-\bolde_y.
\label{eq:Euler-FY}
\end{equation}
It is convex and nonnegative. For interior $\bq$, use $\bs=\log_{a,b,\lambda}(\bq)+\nu\1$ to obtain
\begin{equation}
\mathcal L^{\rm FY}_{a,b,\lambda}(\bs;y)=D_{F_{a,b,\lambda}}(\bolde_y\Vert\bq)
=S_{a,b,\lambda}(\bq;y)+\phi_{a,b,\lambda}(1)-\phi_{a,b,\lambda}(0),
\label{eq:Euler-FY-proper}
\end{equation}
where $S_{a,b,\lambda}$ is the corrected Euler score of Proposition~\ref{prop:proper-Euler-score}. Consequently,
\begin{equation}
\mathbb E_{Y\sim\bp}\mathcal L^{\rm FY}_{a,b,\lambda}(\bs;Y)
-\inf_{\bs'}\mathbb E_{Y\sim\bp}\mathcal L^{\rm FY}_{a,b,\lambda}(\bs';Y)
=D_{F_{a,b,\lambda}}(\bp\Vert\bq).
\label{eq:Euler-FY-excess}
\end{equation}
To verify this, the expected loss is $A_{a,b,\lambda}(\bs)-\langle\bs,\bp\rangle+F_{a,b,\lambda}(\bolde_1)$, whose infimum is $-F_{a,b,\lambda}(\bp)+F_{a,b,\lambda}(\bolde_1)$ by conjugacy. Substitution of the KKT identity gives the displayed Bregman divergence. The unique optimal prediction is $\bq=\bp$ for interior $\bp$; boundary targets are obtained as limits in the full-range regime. Thus both logit losses are convex, but they have different statistical targets and different residuals.

\begin{table}[ht]
\centering\small
\caption{Two losses using the same full-range Euler softmax. Here $\br=\mathcal T_{a,b,\lambda}^{-1}(\bq)$; gradients are with respect to dimensionless logits $\bs=\bz/\tau$. Excess risks are conditional on an interior true posterior $\bp$.}
\begin{tabular}{L{.22\textwidth}L{.30\textwidth}L{.32\textwidth}}
\hline
Property & Euler cross-entropy & Euler Fenchel--Young loss\\\hline
Single-label loss & $\nu-s_y$ & $A_{a,b,\lambda}-s_y+F_{a,b,\lambda}(\bolde_y)$\\
Logit gradient & $\br-\bolde_y$ & $\bq-\bolde_y$\\
Population output & $\mathcal T_{a,b,\lambda}(\bp)$ & $\bp$\\
Probability estimate & $\br$ & $\bq$\\
Exact excess risk & $D_{-H_{a,b,\lambda}}(\bp\Vert\br)$ & $D_{F_{a,b,\lambda}}(\bp\Vert\bq)$\\
Convex in logits & yes & yes\\\hline
\end{tabular}
\end{table}

\subsection{Closed forms, local series, and integral identities}
\label{subsec:exp-closed}
The rescaling identity gives
\[
 \exp_{a,b,\lambda}(y)=\lambda^{-1}\exp_{a,b}(Y),\qquad
 Y=\omega_0(\lambda)y+\log_{a,b}(\lambda).
\]
Consequently an unscaled explicit inverse, where available, immediately
gives a scaled one. For $a=-b=-\kappa$, $\kappa>0$,
\[
 \ln \exp_{a,b,\lambda}(y)
 =\kappa^{-1}\operatorname{arsinh}
 \left(\kappa\cosh(\kappa\ln\lambda)y+\sinh(\kappa\ln\lambda)\right)
 -\ln\lambda.
\]
This expression avoids cancellation in
$\kappa Y+\sqrt{1+\kappa^2Y^2}$ for negative $Y$.
The $\gamma$ subfamily similarly uses its unscaled cubic inverse with
input $Y$ and output divided by $\lambda$.

For completeness, the Lambert--Tsallis representation~\cite{lambert_tsallis}
becomes
\begin{equation}
 \exp_{a,b,\lambda}(y)=\lambda^{-1}
 \left[1+\frac{a-b}{b}W_{2-a/b}(bY)\right]^{1/(a-b)},
 \qquad Y=\omega_0(\lambda)y+\log_{a,b}(\lambda).
 \label{eq:LambertTsallis-expab}
\end{equation}
Use the real branch with positive bracket corresponding to the unique
positive Euler inverse. This identity does not eliminate numerical
inversion of the special function.

Set $\varepsilon_1=1-\alpha_\lambda a-\beta_\lambda b$ and
$J_1=\alpha_\lambda(a-1)(a-2)+\beta_\lambda(b-1)(b-2)$.
Inverse differentiation at $x=1$ gives
\begin{equation}
 \exp_{a,b,\lambda}(y)=1+y+\frac{\varepsilon_1}{2}y^2
 +\frac{3\varepsilon_1^2-J_1}{6}y^3+\mathcal O(y^4).
 \label{eq:expab-series}
\end{equation}
Thus $\lambda$ changes even the local quadratic response while preserving
$\exp_{a,b,\lambda}(0)=\exp_{a,b,\lambda}'(0)=1$.
Substituting $y=\log_{a,b,\lambda}(x)$ also yields
\begin{align}
 \int_{-\infty}^0 \exp_{a,b,\lambda}(y)\,\mathrm dy
 &=\frac{\alpha_\lambda}{1+a}+\frac{\beta_\lambda}{1+b}
 =\phi_{a,b,\lambda}(0),\qquad a>-1,
 \label{eq:int-left-expab}\\
 \int_0^\infty\frac{\mathrm dy}{\exp_{a,b,\lambda}(y)}
 &=\frac{\alpha_\lambda}{1-a}+\frac{\beta_\lambda}{1-b},
 \qquad b<1.
 \label{eq:int-right-expab}
\end{align}
For interior parameters these integrability conditions are necessary
as well as sufficient.

\subsection{Why raw loss minimization can select a degenerate deformation}
\label{subsec:ab-evolution}
For fixed $q\in(0,1)$,
\begin{equation}
 -\log_{a,b,\lambda}(q)=
 -\ln q-\frac{\alpha_\lambda a+\beta_\lambda b}{2}(\ln q)^2
 -\frac{\alpha_\lambda a^2+\beta_\lambda b^2}{6}(\ln q)^3-\cdots .
 \label{eq:local-loss-ab}
\end{equation}
This is the convergent log-coordinate series; truncation is a local
approximation near $q=1$, rather than a uniform approximation at tiny
probabilities. At $\lambda=1$ it reduces to the previous symmetric
two-exponent expansion.
More directly, with the log-scale $\ell=\ln\lambda$, $\partial_\ell[-\log_{a,b,\lambda}(q)]<0$ in the interior:
minimizing this loss at fixed predictions always favors increasing
$\lambda$, approaching the positive-power boundary link on a fixed
operating interval. This preference is not evidence of better prediction.

On the GCE boundary, $\lambda$ cancels and for $q=e^{-t}$, $t>0$,
\begin{equation}
 \frac{\mathrm d}{\mathrm db}\frac{1-e^{-bt}}b
 =\frac{e^{-bt}(1+bt)-1}{b^2}<0.
 \label{eq:bounded-loss-db}
\end{equation}
Thus raw fixed-prediction training loss does not automatically choose the
ordinary logarithm even on clean data.
Use a separately defined predictive/validation objective, or explicitly
specified regularization, for example
\begin{equation}
 \mathcal J=\mathcal L_{\rm task}
 +\frac{\rho_{ab}}2(a^2+b^2)
 +\frac{\rho_\ell}2(\ell-\ell_0)^2,\qquad \rho_{ab},\rho_\ell>0.
 \label{eq:ab-regularization}
\end{equation}
Such penalties encode preferences, not a theorem of optimal deformation.
Quadratic exponent shrinkage need not select $(0,0)$ exactly, and the
open sigmoid parameterization cannot attain that point at finite
training parameters. On a closed GCE branch, an $\ell_1$ penalty
$\rho_b b$ selects $b=0$ for a single fixed prediction if
$\rho_b\ge t^2/2$, since the loss is $\int_0^t e^{-bs}\,\mathrm ds$.
Joint training requires its own analysis.
At $(a,b)=(0,0)$, and on the single-power boundaries,
$\lambda$ is unidentifiable from the link, so learning it adds no
functional freedom there.


\section{Further mirror-descent and FTRL results}
\label{app:supp-online}
This appendix contains the power-metric decomposition, the discounted and Adam-like FTRL variants, mirror descent with a guaranteed probability floor, and a discussion of parameter-free learning, all summarized in Sections~\ref{sec:FTRL-3p}--\ref{sec:extensions}.

\subsection{Adaptive online MD with an explicit exploration floor}
\label{subsec:FTRL-floor}
Applications that prescribe a minimum probability need more than the
decreasing floor of Corollary~\ref{cor:adaptive-FTRL-floor}. Fix
$0<\epsilon<1/N$, $K_\epsilon=\{\bw\in\Delta_N:w_i\ge\epsilon\}$, and
\begin{equation}
 B_\epsilon:=\sup_{\bx,\by\in K_\epsilon}D_{F_{a,b,\lambda}}(\bx\Vert\by)
 \le h_{a,b,\lambda}(\epsilon)(1-N\epsilon)^2<\infty,
 \label{eq:floor-Bregman-diameter}
\end{equation}
using the nonincreasing Hessian entries and the Euclidean diameter
$\sqrt2(1-N\epsilon)$ of $K_\epsilon$.

\begin{theorem}[Adaptive Euler OMD on a controlled simplex]
\label{thm:adaptive-Euler-OMD}
For $\delta,\eta_0>0$, $\bw_1\in K_\epsilon$, $\eta_t=\eta_0/\sqrt{\delta+S_t}$ and
\begin{empheq}[box=\fbox]{equation}
\begingroup
\begin{gathered}
\bw_{t+1}=\operatorname*{arg\,min}_{\bw\in K_\epsilon}
 \{\eta_t\langle\bar\bg_t,\bw\rangle+D_{F_{a,b,\lambda}}(\bw\Vert\bw_t)\},\\
w_{i,t+1}=\max\{\epsilon,\exp_{a,b,\lambda}(\log_{a,b,\lambda}(w_{i,t})-\eta_t\bar g_{i,t}-\nu_t)\},
\end{gathered}
\endgroup
 \label{eq:floor-OMD-solve}
\end{empheq}
every $\bx\in K_\epsilon$ satisfies
\begin{equation}
 \mathrm{Reg}_T(\bx)\le\frac{B_\epsilon}{\eta_0}\sqrt{\delta+S_T}
 +\frac{\eta_0}{\mu}\bigl(\sqrt{\delta+S_T}-\sqrt\delta\bigr).
 \label{eq:adaptive-OMD-bound}
\end{equation}
Thus $\eta_0=\sqrt{\mu B_\epsilon}$ gives $2\sqrt{B_\epsilon(\delta+S_T)/\mu}$;
for an arbitrary $\bx\in\Delta_N$ add at most $2(N-1)\epsilon\sum_t\sigma_t$.
\end{theorem}
\begin{proof}
The MD variational inequality gives
$\langle\bar\bg_t,\bw_t-\bx\rangle\le[D(\bx\Vert\bw_t)-D(\bx\Vert\bw_{t+1})]/\eta_t
+\eta_t\sigma_t^2/(2\mu)$. Since $\eta_t$ is nonincreasing, the divergence terms
sum to at most $B_\epsilon/\eta_T$, and
$\sum_t\sigma_t^2/\sqrt{\delta+S_t}\le2(\sqrt{\delta+S_T}-\sqrt\delta)$. Compare
an arbitrary $\bx$ with $(1-N\epsilon)\bx+\epsilon\1$.
\end{proof}
Full-simplex FTRL allows arbitrarily small positive coordinates and pays the
finite radius $R$; OMD on $K_\epsilon$ enforces a user-selected floor and pays
twice, through $B_\epsilon$, which grows as $\epsilon\to0$, and through the
comparator bias. Recursive OMD needs distances between arbitrary points, which
are infinite on the full simplex; this turns a boundary issue into an explicit
choice.

\subsection{Scale-free, schedule-free, and parameter-free are different claims}
\label{subsec:new-parameterfree}
With $\delta=0$ the adaptive learner is scale-free; the averaging construction
is schedule-free; the triple $(a,b,\lambda)$ remains a selected
hyperparameter. Unconstrained \emph{parameter-free} learning adapts to unknown
comparator norms, a stronger requirement~\cite{Zhao2026Variation}. The conic
lifting of~\cite{GrandClement2026} suggests the perspective
\begin{equation}
 \mathcal H_{a,b,\lambda}(\varsigma,\bv)=\varsigma\,\mathcal R_{a,b,\lambda}(\bv/\varsigma)
 \quad\text{on}\quad\{(\varsigma,\bv):\varsigma>0,\ \bv/\varsigma\in\Delta_N\},
 \label{eq:new-cone}
\end{equation}
which is convex but linear along rays, hence never jointly strongly convex: a
separate radial mechanism is necessary. A safer way to reduce sensitivity to
the triple is a portfolio of $M$ fixed geometries combined by Hedge with rate
$\eta_{\rm meta}>0$ on losses in $[0,1]$:
\begin{equation}
 \mathrm{Reg}_T(\bp^*)\le\min_{1\le j\le M}\mathrm{Reg}^{(j)}_T(\bp^*)
 +\frac{\ln M}{\eta_{\rm meta}}+\frac{\eta_{\rm meta} T}{8},
 \label{eq:new-meta-bound}
\end{equation}
which competes with the best fixed grid geometry in hindsight, not with a
continuously optimal one.

\subsection{\texorpdfstring{$p$}{p}-norm descent and the two-component Euler metric}
\label{subsec:pnorm}
Classical $p$-norm and power-potential descent uses the scalar potential $w^p/[p(p-1)]$, $p\notin\{0,1\}$, whose second derivative is $w^{p-2}$; the smaller $p$, the more strongly small weights resist being pushed down. Up to affine terms this is the power potential $\Phi_s$ of Eq.~\eqref{eq:scaled-potential} with $p=1+s$, and $\Phi_s''(w)=w^{s-1}$. (In this subsection only, $p$ denotes this norm exponent rather than $|a|$.)

\begin{theorem}[Euler Hessians as mixtures of separable power Hessians]
\label{thm:pnorm-mixture}
Let $-1<a<0<b<1$ and $\lambda>0$. The single-power boundary links generate the
separable power geometries $\Phi_s$ with exponent $p=1+s$, $s\in\{a,b\}$, up to affine
changes, and
\begin{equation}
 h_{a,b,\lambda}(w)=\alpha_\lambda \Phi_a''(w)+\beta_\lambda \Phi_b''(w).
 \label{eq:metric-mixture}
\end{equation}
Both components diverge at zero; the component with $0<1+a<1$ also has
derivative $-\infty$ there, while the component with $1<1+b<2$ has a finite
boundary derivative. Both potentials are finite at zero.
\end{theorem}
\begin{proof}
Differentiate Eq.~\eqref{eq:scaled-power-mixture}: $\log_{s,0}'(w)=w^{s-1}=\Phi_s''(w)$.
\end{proof}
The component $p=1+a<1$, steeper than entropy near zero, produces the
protection of small weights and the polynomial floors; the component
$p=1+b\in(1,2)$ governs large weights. At $\lambda=1$ the weights
$|a|:b$ are fixed by the exponents; the scale is exactly what makes their
ratio $\beta_\lambda/\alpha_\lambda=(b/|a|)\lambda^{b-a}$ an independent
design variable. By contrast, on the positive orthant with $S=\sum_iw_i^p$,
$\nabla^2(\tfrac12\|\bw\|_p^2)=(p-1)S^{2/p-1}\diag(w_i^{p-2})+(2-p)S^{2/p-2}\bw^{p-1}(\bw^{p-1})^T$;
its rank-one term distinguishes the squared $p$-norm from any separable Euler
Hessian, while the separable power geometry with $1<p<2$ is exactly the Euler
boundary $(0,p-1)$.

\subsection{Scope of the extensions, and a reusable constrained solver}
We distinguish an \emph{established framework}, attributed to its source; an
\emph{Euler specialization}, obtained by substituting $F_{a,b,\lambda}$ and
checking its assumptions; and a \emph{proposed extension}, whose comparative
performance is untested. The triple $(a,b,\lambda)$ determines how an
algorithm penalizes changes of small and large coordinates; it does not by
itself identify a data distribution, select an optimal geometry, or remove
learning-rate and statistical assumptions. Unless stated otherwise,
$-1<a\le0\le b\le1$, $a<b$, $\lambda>0$ is fixed, and
$d_{a,b,\lambda}(x,y)=\phi_{a,b,\lambda}(x)-\phi_{a,b,\lambda}(y)-\log_{a,b,\lambda}(y)(x-y)$
denotes the scalar Bregman gap.

By Eq.~\eqref{eq:scaled-power-mixture} the potential, its divergence and its
curvature are the same convex combination of a \emph{sparse} power geometry of
order $1+b>1$ and a \emph{retaining} one of order $1+a\in(0,1)$:
\begin{equation}
 F_{a,b,\lambda}=\alpha_\lambda F_{a,0}+\beta_\lambda F_{0,b},\qquad
 h_{a,b,\lambda}(x)=\alpha_\lambda x^{a-1}+\beta_\lambda x^{b-1},
 \label{eq:new-mixture}
\end{equation}
with weights fixed by $(a,b,\lambda)$ rather than an independent coefficient;
the scale is precisely the parameter that moves this mixture. The $1+a$
boundary with $|a|=1/2$ connects, up to scale and affine terms, to the
regularizer of Tsallis-INF~\cite{Zimmert2021}, whose best-of-both-worlds
bandit guarantees rely on loss estimation and a dedicated stability analysis
and do not transfer to every mixture; an adversarial bound for Euler mixtures is Proposition~\ref{prop:euler-bandit}. For a score $\bz$, a reference
$\bq\in\Delta_N^\circ$ and $\eta>0$, the shared prox solver is
\begin{equation}
 \mathcal P_{a,b,\lambda}(\bq,\bz;\eta)
 =\operatorname*{arg\,min}_{\bp\in\Delta_N}
 \{\eta\langle\bz,\bp\rangle+D_{F_{a,b,\lambda}}(\bp\Vert\bq)\},\qquad
 p_i=\exp_{a,b,\lambda}\bigl(\log_{a,b,\lambda}(q_i)-\eta z_i-\nu\bigr),
 \label{eq:new-prox}
\end{equation}
that is, one call of Algorithm~\ref{alg:euler-norm} with scores
$\log_{a,b,\lambda}(\bq)-\eta\bz$. It is used by OMD, optimistic OMD, PMD and the policy
step of VMD; FTRL changes its input to cumulative gradients.

\subsection{The online mirror map: further formulas}

\label{subsec:ftrl3-foundations}
Let $\mathcal W\subseteq\mathbb R^N$ be a convex decision set, with minimizers
assumed to exist. Typical choices are the open positive orthant
$\mathbb R_+^N=\{\bw:w_i>0\}$, the closed nonnegative orthant
$\mathbb R_{\ge0}^N$, and the closed simplex $\Delta_N$. At each round
$t=1,\ldots,T$ the learner selects $\bw_t\in\mathcal W$, nature reveals a
convex loss $L_t$, and the learner observes $\bg_t\in\partial L_t(\bw_t)$.
With $\bg_{1:t}=\sum_{s\le t}\bg_s$, the regret against a fixed comparator is
\begin{equation}
 \operatorname{Regret}_T(\bw^*)=\sum_{t=1}^T\bigl(L_t(\bw_t)-L_t(\bw^*)\bigr)
 \le\sum_{t=1}^T\langle\bg_t,\bw_t-\bw^*\rangle.
 \label{eq:regret-def}
\end{equation}

The geometry is induced by the separable potential
$F_{a,b,\lambda}(\bw)=\sum_i\phi_{a,b,\lambda}(w_i)$ of
Eqs.~\eqref{eq:Euler-potential-vector}--\eqref{eq:scaled-potential}, whose componentwise derivative is the link
$\log_{a,b,\lambda}$ of Eq.~\eqref{logab}, a normalized mixture
$\alpha_\lambda\log_{a,0}+\beta_\lambda\log_{0,b}$ of two power links
(Eq.~\eqref{eq:scaled-power-mixture}). Lemma~\ref{lem:audit-domains} gives
monotonicity, concavity for $b\le1$ and the full range $\mathbb R$ for
$a<0<b$; Theorem~\ref{thm:sharp-simplex-geometry} gives the finite boundary
value $\phi_{a,b,\lambda}(0)=\alpha_\lambda/(1+a)+\beta_\lambda/(1+b)$, so the
lower-semicontinuous extension of $F_{a,b,\lambda}$ is a Legendre potential
with positive interior. In the coordinates $p=|a|$, $q=b$ and with
$D_\lambda=p\lambda^{-p}+q\lambda^q$, the gradient weight and elasticity of
Eq.~\eqref{eq:elasticity-def} take the form
\begin{equation}
 \omega_{a,b,\lambda}(w)=wh_{a,b,\lambda}(w)
 =\frac{p(\lambda w)^{-p}+q(\lambda w)^q}{D_\lambda},\qquad
 \varepsilon_{a,b,\lambda}(w)
 =1+\frac{p^2(\lambda w)^{-p}-q^2(\lambda w)^q}
 {p(\lambda w)^{-p}+q(\lambda w)^q},
 \label{eq:ftrl3-omega}
\end{equation}
so that $\varepsilon_{a,b,\lambda}(0^+)=1+p$ and
$\varepsilon_{a,b,\lambda}(\infty)=1-q$. The Bregman divergence is explicit:
\begin{equation}
 D_{F_{a,b,\lambda}}(\bw\Vert\bv)
 =\sum_{i=1}^N\left\{\frac{\alpha_\lambda}{a}
 \left[\frac{w_i^{a+1}-v_i^{a+1}}{a+1}-v_i^{a}(w_i-v_i)\right]
 +\frac{\beta_\lambda}{b}
 \left[\frac{w_i^{b+1}-v_i^{b+1}}{b+1}-v_i^{b}(w_i-v_i)\right]\right\},
 \label{eq:euler-bregman-explicit}
\end{equation}
Comparators may lie on the simplex boundary, but unshifted Bregman anchors
must be positive. The inverse $\exp_{a,b,\lambda}$ is evaluated by
Algorithm~\ref{alg:euler-exp}; its rescaling identity~\eqref{defexpEuler}
relates it to the unscaled inverse, and its tails are
$\exp_{a,b,\lambda}(y)\sim(p|y|/\alpha_\lambda)^{-1/p}$ as $y\to-\infty$ and
$\exp_{a,b,\lambda}(y)\sim(qy/\beta_\lambda)^{1/q}$ as $y\to\infty$
(Proposition~\ref{prop:euler-tails}).

\paragraph{Local third-derivative control.}
For $[\delta,W]\subset(0,\infty)$,
\[
 M_\lambda(\delta,W)=\sup_{\delta\le w\le W}
 \frac{|\phi_{a,b,\lambda}'''(w)|}{2\phi_{a,b,\lambda}''(w)^{3/2}}
 =\sup_{\delta\le w\le W}
 \frac{|\varepsilon_{a,b,\lambda}(w)|}{2\sqrt{w\,\omega_{a,b,\lambda}(w)}}<\infty .
\]
For $p<1$ the ratio diverges at zero like $w^{(p-1)/2}$, so
$\phi_{a,b,\lambda}$ is not globally self-concordant with any finite constant
(Section~\ref{sec:phase2-barrier}). There is no general simultaneous minimizer
of this constant and of $\max_{[\delta,1]}\omega/\min_{[\delta,1]}\omega$;
placing $w_{\min}$ at $\sqrt\delta$ by $\lambda=r/\sqrt\delta$ is only a
geometric placement rule.

\paragraph{Covariance under changes of coordinate units.}
If $x=\varsigma w$ for a unit factor $\varsigma>0$, $\lambda'=\lambda/\varsigma$,
$L'_t(x)=L_t(x/\varsigma)$ and $\mathcal W'=\varsigma\mathcal W$, put
$\omega_\varsigma=\omega_{a,b,\lambda}(1/\varsigma)>0$. Then
\[
 \log_{a,b,\lambda'}(\varsigma w)
 =\frac{\log_{a,b,\lambda}(w)-\log_{a,b,\lambda}(1/\varsigma)}{\omega_\varsigma},\qquad
 D_{F_{a,b,\lambda'}}(\varsigma u\Vert\varsigma v)
 =\frac{\varsigma}{\omega_\varsigma}D_{F_{a,b,\lambda}}(u\Vert v).
\]
Thus corresponding MD trajectories use
$\eta'_t=\varsigma\eta_t/\omega_\varsigma$ and correspondingly rescaled initial
points. This is covariance under transformed coordinates, losses and
domains; it is not invariance on a fixed unit simplex.

\begin{remark}[Why the Euler potential is an admissible FTRL regularizer]
\label{rem:why-Euler-regularizer}
(i) Since $a>-1$, $\phi_{a,b,\lambda}(0)$ is finite and boundary comparators are
admissible. (ii) Theorem~\ref{thm:sharp-simplex-geometry} supplies the $\ell_1$
modulus $\mu$. (iii) Since $\phi_{a,b,\lambda}'(0^+)=-\infty$, finite cumulative
gradients give an interior minimizer, so only the equality multiplier is
needed. (iv) Symmetry and convexity place the minimum at $\1/N$ and the maximum
at the vertices; the range is $R$. (v) The integral-generated entropy
$-F_{a,b,\lambda}$~\cite{Naudts2011} tends to Shannon entropy plus $1-N$ as
$(a,b)\to(0,0)$ and differs from the trace entropy~\eqref{eq:traceform-general}.
By Danskin's theorem the simplex conjugate
$A_{a,b,\lambda}(\bs)=\max_{\Delta_N}\{\langle\bs,\bw\rangle-F_{a,b,\lambda}(\bw)\}$ of Eq.~\eqref{eq:Euler-conjugate}
has gradient $\sigma_{a,b,\lambda,1}(\bs)$, the Euler softmax of
Definition~\ref{def:Euler-softmax}; in the logarithmic limit FTRL becomes Hedge.
\end{remark}

\subsection{Normalization certificates in finite precision}
\label{subsec:supp-oracle}
\paragraph{Full residual with inexact inner solves.}
Algorithm~\ref{alg:euler-exp} with positive tolerance does not supply exact
$w_i$, so its errors must enter the outer certificate. From inner log
enclosures form outward-rounded weight bounds and a mass enclosure
$[S_-,S_+]$; raise $\ell$ only when $S_->1$ and lower $u$ only when $S_+<1$.
For a feasible candidate $\bp$ (strictly positive when $a<0$) set
$d_i=D_i\log_{a,b,\lambda}(p_i)-y_i+\nu$, with $r_i=d_i$ on positive coordinates and
$r_i=\min\{d_i,0\}$ at zero coordinates when $a=0$ (where
$\log_{a,b,\lambda}(0)=-1/b$). Then
\begin{equation}
 \hat\epsilon_{\rm all}=\tfrac12\bigl(\max_ir_i-\min_ir_i\bigr)
 \label{eq:normalization-full-residual}
\end{equation}
is a valid stationarity residual modulo the simplex normal cone; for interval
enclosures use $\frac12(\max_ir_i^+-\min_ir_i^-)$. Return success only when
feasibility and this bound are verified; otherwise increase precision or
return \textsc{precision-limit}.

\paragraph{A worked example.}
For $(a,b,\lambda)=(-\frac12,\frac12,1)$, $\log_{-1/2,1/2,1}(x)=\sqrt x-1/\sqrt x$ and
$\ln\exp_{-1/2,1/2,1}(y)=2\operatorname{arsinh}(y/2)$. For $\by=(0,-1,-2)$ the bracket is
$[0,2/\sqrt3]$, and
\[
 \nu^*\simeq0.4802363792,\qquad
 \bp^*\simeq(0.6214266802,\ 0.2539936647,\ 0.1245796551),
\]
against the ordinary
softmax $(0.66524,0.24473,0.09003)$: the polynomial tail retains more mass on
low-scoring coordinates, but positivity alone gives no score-independent lower
bound on a probability.

\begin{lemma}[From a normalization certificate to online regret]
\label{lem:stage2-oracle-error}
Let the implemented feasible decision $\widetilde\bp_t$ for score $\by_t$ have
a verified full certificate $\epsilon_t$ (Eq.~\eqref{eq:normalization-full-residual}),
and let $\bp_t^*=\mathcal N_{a,b,\lambda}(\by_t)$. Then
\begin{equation}
 \|\widetilde\bp_t-\bp_t^*\|_1\le\min\{2,\epsilon_t/\mu\},\qquad
 \langle\bg_t,\widetilde\bp_t-\bp_t^*\rangle\le
 \operatorname{hr}(\bg_t)\min\{2,\epsilon_t/\mu\}.
 \label{eq:stage2-oracle-cost}
\end{equation}
Hence the regret bounds of Algorithms~\ref{alg:adaptive-ftrl}
and~\ref{alg:aoftrl} acquire the term
$\sum_t\operatorname{hr}(\bg_t)\min\{2,\epsilon_t/\mu\}$, and that of
Algorithm~\ref{alg:sf-ftrl} its expectation divided by $T$, evaluated on the
actually observed history.
\end{lemma}
\begin{proof}
The certificate gives $\nabla F(\widetilde\bp_t)-\by_t+\bn_t=\mathbf e_t$ with
$\bn_t$ in the normal cone and $\|\mathbf e_t\|_\infty\le\epsilon_t$. Strong
monotonicity gives
$\mu\|\widetilde\bp_t-\bp_t^*\|_1^2\le\langle\mathbf e_t,\widetilde\bp_t-\bp_t^*\rangle$;
use the simplex diameter two and center $\bg_t$.
\end{proof}
With $\operatorname{hr}(\bg_t)\le G$ and $\epsilon_t\le\epsilon_0/t^2$ the extra
regret is at most $G\epsilon_0\pi^2/(6\mu)$; without a gradient bound, retain
the weighted sum.

\subsection{Further FTRL constructions and their mirror-descent counterparts}
\label{subsec:ftrl3-overview}
Table~\ref{tab:conceptual} distinguishes identical algorithms from related
constructions; the word \emph{counterpart} does not imply identical iterates or
transferred guarantees. Below,
$\nu_t$ denotes a normalization multiplier, zero on the
positive orthant.

\begin{table}[ht]
\centering\small
\caption{FTRL constructions and their relationship to mirror descent.}
\label{tab:conceptual}
\begin{tabular}{L{.27\textwidth}L{.65\textwidth}}
\hline Construction & Relationship\\\hline
Linearized FTRL & Exactly dual averaging (lazy MD), with matching regularizer and initialization.\\
Proximal FTRL & Exactly scalar-step active MD on the positive orthant and interior simplex.\\
Shifted FTRL-Proximal & Related to composite MD (FOBOS, forward--backward splitting); cumulative and single-step penalties generally differ.\\
RDA & Composite dual averaging with an origin-centered regularizer.\\
Adaptive FTRL & Related to adaptive MD; current scaling of all past gradients differs from incremental scaling.\\
Optimistic FTRL & Related to two-step optimistic MD; hints precede the predicted loss.\\
Euler-Hedge FTRL & Exactly fixed-step Euler-EG on the simplex with matching initialization.\\
Time-varying FTRL & Generally different from decaying-step active MD.\\\hline
\end{tabular}
\end{table}

\subsubsection{Linearized FTRL and lazy mirror descent}
Linearized FTRL replaces each loss by the linear term $\langle\bg_s,\bw\rangle$
of its affine lower bound and regularizes by $F/\eta$, with
$\bw_1=\argmin_{\mathcal W}F$ ($\1$ on the orthant, $\1/N$ on the simplex):
\begin{equation}
 \bw_{t+1}^{\text{Lin-FTRL}}=\argmin_{\bw\in\mathcal W}
 \Bigl\{\langle\bg_{1:t},\bw\rangle+\tfrac1\eta F(\bw)\Bigr\}.
 \label{eq:lin-ftrl}
\end{equation}
Lazy MD (Nesterov's dual averaging) accumulates $\bz_t=\bz_{t-1}-\eta\bg_t$,
$\bz_0=\0$, and maps back by the constrained conjugate,
$\bw_{t+1}=\argmin_{\bw\in\mathcal W}\{-\langle\bz_t,\bw\rangle+F(\bw)\}
=\nabla F^*_{\mathcal W}(\bz_t)$. On $\mathbb R_+^N$ or $\Delta_N$ both have the
KKT condition $\log_{a,b,\lambda}(w_{t+1,i})+\eta g_{1:t,i}+\nu_t=0$, hence the same update
\begin{equation}
 w_{t+1,i}^{\text{Lin-FTRL}}=w_{t+1,i}^{\text{Lazy-MD}}
 =\exp_{a,b,\lambda}\bigl(-\eta g_{1:t,i}-\nu_t\bigr),
 \label{eq:euler-lin-ftrl}
\end{equation}
with $\sum_i\exp_{a,b,\lambda}(-\eta g_{1:t,i}-\nu_t)=1$ on $\Delta_N$.

\subsubsection{Proximal FTRL and active online mirror descent}

Active OMD steps from the previous iterate,
\begin{equation}
 \bw_{t+1}^{\text{OMD}}=\argmin_{\bw\in\mathcal W}
 \Bigl\{\langle\bg_t,\bw\rangle+\tfrac1{\eta_t}D_F(\bw\Vert\bw_t)\Bigr\}.
 \label{eq:active-omd}
\end{equation}
Proximal FTRL places incremental Bregman regularizers at all past iterates,
with $\sigma_1=\eta_1^{-1}$, $\sigma_s=\eta_s^{-1}-\eta_{s-1}^{-1}\ge0$, so that
$\sum_{s\le t}\sigma_s=\eta_t^{-1}$:
\begin{equation}
 \bw_{t+1}^{\text{Prox-FTRL}}=\argmin_{\bw\in\mathcal W}
 \Bigl\{\langle\bg_{1:t},\bw\rangle+\sum_{s=1}^t\sigma_sD_F(\bw\Vert\bw_s)\Bigr\}.
 \label{eq:prox-ftrl}
\end{equation}
Since $\partial_{w_i}D_F(\bw\Vert\bw_s)=\log_{a,b,\lambda}(w_i)-\log_{a,b,\lambda}(w_{s,i})$, its optimality
condition gives, with $\tilde z_{t,i}=\sum_{s\le t}\sigma_s\log_{a,b,\lambda}(w_{s,i})-g_{1:t,i}$,
\begin{equation}
 w_{t+1,i}^{\text{Prox-FTRL}}=\exp_{a,b,\lambda}\bigl(\eta_t(\tilde z_{t,i}-\nu_t)\bigr).
 \label{eq:euler-prox-ftrl}
\end{equation}
The scalar-step equivalence holds for every nonincreasing positive $\eta_t$.
Subtracting successive optimality equations gives,
for $t\ge2$,
$\eta_t^{-1}[\log_{a,b,\lambda}(w_{t+1,i})-\log_{a,b,\lambda}(w_{t,i})]=-g_{t,i}-(\nu_t-\nu_{t-1})$, and the first update
follows from $\sigma_1=\eta_1^{-1}$. On the positive orthant multipliers vanish; on
the simplex their difference is a common scalar, exactly the MD normalization
shift after multiplication by $\eta_t$. Strict convexity proves equality of the
iterates. This does not extend unchanged to general inequality constraints,
whose active normal vectors can change.

\subsubsection{Regularized dual averaging}
RDA keeps the regularizer origin-centered:
\begin{empheq}[box=\fbox]{equation}
 \bw_{t+1}^{\text{RDA}}=\argmin_{\bw\in\mathbb R^N_{\ge0}}
 \Bigl\{\langle\bg_{1:t},\bw\rangle+t\rho_1\|\bw\|_1+\tfrac1{\eta_t}\tilde F_{a,b,\lambda}(\bw)\Bigr\},
 \qquad
 w_{t+1,i}^{\text{RDA}}=\bigl[\exp_{a,b,\lambda}\bigl(-\eta_t(g_{1:t,i}+t\rho_1)\bigr)-1\bigr]_+ .
 \label{eq:euler-rda}
\end{empheq}
Removing the proximal terms and setting the origin coefficient to
$\eta_t^{-1}$ gives RDA; setting $\rho_1=0$ alone does not yield FOBOS, since
the origin penalty, domain and schedule still matter.

\subsubsection{Adaptive FTRL and Euler-AdaGrad}

 By setting the regularization from the square root of past squared gradients, the FTRL form of AdaGrad adapts automatically to sparse gradients; it is the theoretical backbone of many adaptive deep-learning optimizers.

With $s_{t,i}=\sqrt{\delta+\sum_{k\le t}g_{k,i}^2}$, $\delta>0$, and the
diagonally weighted potential $F_t(\bw)=\sum_is_{t,i}\phi_{a,b,\lambda}(w_i)$,
\begin{align}
 \bw_{t+1}^{\text{AdaFTRL}}&=\argmin_{\bw\in\mathcal W}
 \Bigl\{\langle\bg_{1:t},\bw\rangle+\tfrac1\eta F_t(\bw)\Bigr\},\quad
 w_{t+1,i}^{\text{AdaFTRL}}=\exp_{a,b,\lambda}\Bigl(-\frac{\eta(g_{1:t,i}+\nu_t)}{s_{t,i}}\Bigr),
 \label{eq:euler-adaftrl}\\
 \bw_{t+1}^{\text{AdaGrad}}&=\argmin_{\bw\in\mathcal W}
 \Bigl\{\langle\bg_t,\bw\rangle+\tfrac1\eta D_{F_t}(\bw\Vert\bw_t)\Bigr\},\notag\\
 w_{t+1,i}^{\text{AdaGrad}}&=\exp_{a,b,\lambda}\Bigl(\log_{a,b,\lambda}(w_{t,i})-\frac{\eta(g_{t,i}+\nu_t)}{s_{t,i}}\Bigr).
 \label{eq:euler-adagrad-md}
\end{align}
AdaFTRL scales the whole cumulative sum by the latest $s_{t,i}^{-1}$, whereas
the active method accumulates incrementally scaled gradients; on the simplex,
coordinate-weighted multiplier terms accumulate as well. These inverse
formulas apply to the positive orthant or the simplex, not to an arbitrary
convex set. A signed, regret-certified variant is Algorithm~\ref{alg:euler-adagrad}.

\subsubsection{Optimistic FTRL and optimistic mirror descent}
\label{subsubsec:ftrl3-optimistic}

Optimistic FTRL (OFTRL) uses a forecast (``hint'') of the next loss vector
before acting. Its regret scales with the accumulated forecast \emph{errors}
rather than with the gradients themselves, so accurate forecasts reduce it
substantially, to a constant when the total forecast error stays bounded. This
makes optimism useful in games and for smoothly varying, predictable
sequences.

Given a predictable hint $\vm_{t+1}$ of the next gradient (e.g.\
$\vm_{t+1}=\bg_t$ in smooth games),
\begin{equation}
 \bw_{t+1}^{\text{OFTRL}}=\argmin_{\bw\in\mathcal W}
 \Bigl\{\langle\bg_{1:t}+\vm_{t+1},\bw\rangle+\tfrac1\eta F(\bw)\Bigr\},
 \qquad
 w_{t+1,i}^{\text{OFTRL}}=\exp_{a,b,\lambda}\bigl(-\eta(g_{1:t,i}+m_{t+1,i})-\nu_t\bigr).
 \label{eq:euler-oftrl}
\end{equation}
Optimistic MD keeps an auxiliary sequence and performs two Euler steps,
\begin{empheq}[box=\fbox]{equation}
 \begin{aligned}
 \hat w_{t+1,i}&=\exp_{a,b,\lambda}\bigl(\log_{a,b,\lambda}(\hat w_{t,i})-\eta g_{t,i}-\hat\nu_t\bigr),\\
 w_{t+1,i}^{\text{OMD}}&=\exp_{a,b,\lambda}\bigl(\log_{a,b,\lambda}(\hat w_{t+1,i})-\eta m_{t+1,i}-\nu_t\bigr),
 \end{aligned}
 \label{eq:euler-omd}
\end{empheq}
initialized at $\hat\bw_1=\argmin F$ and
$\bw_1=\argmin_{\bw\in\mathcal W}\{\eta\langle\vm_1,\bw\rangle+D_F(\bw\Vert\hat\bw_1)\}$
(or $\vm_1=\0$). On the simplex its regret satisfies
\begin{equation}
 \operatorname{Regret}_T(\bu)\le\frac{D_F(\bu\Vert\hat\bw_1)}{\eta}
 +\frac{\eta}{2\mu}\sum_{t=1}^T\|\bg_t-\vm_t\|_\infty^2,
 \label{eq:ftrl3-oomd}
\end{equation}
with $\mu=\mu_{a,b,\lambda}(N)$ and $D_F(\bu\Vert\hat\bw_1)\le R_{a,b,\lambda}(N)$ at uniform initialization
(Theorem~\ref{thm:new-oomd} gives the statement and proof). Fixed-geometry
optimistic FTRL has the same iterates on the interior simplex, since the
auxiliary dual states telescope up to scalar shifts. Tuning $\eta$ to a known
prediction-error sum gives a square-root bound; an unknown sum requires
Theorem~\ref{thm:new-adaptive-oftrl}. Smoothness alone does not imply bounded
total prediction error, and hence no unconditional $\mathcal O(1)$ regret.

\subsubsection{Euler-Hedge FTRL and Euler exponentiated gradient}
\label{subsubsec:ftrl3-hedge}
On $\Delta_N$, centering the potential at $\1/N$ gives Euler-entropic FTRL and
its active counterpart,
\begin{equation}
\begingroup
\begin{gathered}
w_{t+1,i}^{\text{Euler-FTRL}}=\exp_{a,b,\lambda}\bigl(\log_{a,b,\lambda}(1/N)-\eta g_{1:t,i}-\nu_t^{\rm F}\bigr),\\
w_{t+1,i}^{\text{Euler-EG}}=\exp_{a,b,\lambda}\bigl(\log_{a,b,\lambda}(w_{t,i})-\eta g_{t,i}-\nu_t^{\rm M}\bigr),
\end{gathered}
\endgroup
 \label{eq:euler-hedge-ftrl}
\end{equation}
{The FTRL multiplier $\nu_t^{\rm F}$ normalizes cumulative scores, whereas the mirror-step multiplier $\nu_t^{\rm M}$ normalizes one update. Each enforces unit mass in its own equation.} For fixed $\eta$ and uniform initialization the iterates
coincide: substitute the EG dual equations successively and absorb the sum of
normalization shifts into one multiplier. Their regret obeys
Eq.~\eqref{eq:verified-regret}.

\subsection{Weighted simplex normalization}

Every coordinate update calls the certified inverse of
Algorithm~\ref{alg:euler-exp}; the scale enters only through
$\alpha_\lambda,\beta_\lambda$, whose log-odds
$\ln(q/p)+(p+q)\ln\lambda$ (with $p=|a|$, $q=b$) should be used to avoid extreme powers of
$\lambda$. A returned log value must not be substituted for a primal
coordinate; use log-domain arithmetic or higher precision when needed. The
simplex rows require one multiplier for weighted scores: solve
\begin{equation}
 S(\nu)=\sum_i\exp_{a,b,\lambda}\bigl((y_i-\nu)/D_i\bigr)-1=0,\qquad
 S'(\nu)=-\sum_i\frac{m_{a,b,\lambda}(w_i(\nu))}{D_i}<0 .
 \label{eq:weighted-normalization}
\end{equation}
At $\nu_{\rm lo}=\max_iy_i$ some coordinate equals one, so
$S(\nu_{\rm lo})\ge0$; at $\nu_{\rm hi}=\max_i(y_i-D_i\log_{a,b,\lambda}(1/N))$ every
coordinate is at most $1/N$, so $S(\nu_{\rm hi})\le0$. Strict decrease gives a
unique root, which Algorithm~\ref{alg:euler-norm} computes with a residual
certificate. If inverse evaluations have log-error at most $\epsilon$, the exact
mass lies between $e^{-\epsilon}\sum_i\hat w_i$ and $e^{\epsilon}\sum_i\hat w_i$, apart
from arithmetic roundoff; refine inner tolerances before deciding a sign when
this interval straddles one, and let an iteration cap return an explicit
failure status.

\subsection{Weighted dual averaging and composite FTRL}
Let $\omega_t>0$ be predictable weights, $\bs_t=\sum_{s\le t}\omega_s\bg_s$,
$A_t=\sum_{s\le t}\omega_s$, $\beta_t=1/\eta_t>0$ and $\Psi$ a convex penalty
with weight $\rho_1\ge0$:
\begin{equation}
 \bp_{t+1}=\operatorname*{arg\,min}_{\bp\in K}
 \{\langle\bs_t,\bp\rangle+\beta_t\mathcal R_{a,b,\lambda}(\bp)+A_t\rho_1\Psi(\bp)\},
 \qquad\mathcal R_{a,b,\lambda}=F_{a,b,\lambda}-\min_KF_{a,b,\lambda}.
 \label{eq:new-da}
\end{equation}
On the simplex with $\Psi=0$, $p_{i,t+1}=\exp_{a,b,\lambda}(-s_{i,t}/\beta_t-\nu_t)$; the
effective coefficient of the average gradient $\bs_t/A_t$ is $\beta_t/A_t$,
which should not be confused with an MD step size, and a decreasing rate needs
the stabilization of Lemma~\ref{lem:audit-exact-ftrl}. On $\Delta_N$ an
$\ell_1$ penalty is constant; sparsity must come from finite boundary slope.
At $a=0<b$ (where $\lambda$ cancels)
\begin{equation}
 p_{i,t+1}=[1-b(s_{i,t}/\beta_t+\nu_t)]_+^{1/b},
 \label{eq:new-da-sparse}
\end{equation}
the existing entmax boundary of the family~\cite{Martins2016,Peters2019}; with
$a<0$ and finite inputs the solution is dense.

\begin{proposition}[Support re-entry differs between sparse MD and sparse FTRL]
\label{prop:new-support}
At $a=0<b$ both sparse FTRL and sparse Euler MD can reactivate a zero
coordinate: if $q_i=0$, then
$[\mathcal P_{0,b,\lambda}(\bq,\bg;\eta)]_i=[-b(\eta g_i+\nu)]_+^{1/b}$, positive exactly
when $\eta g_i+\nu<0$. The two algorithms nevertheless have generally
different trajectories.
\end{proposition}
\begin{proof}
Use $\log_{0,b}(0)=-1/b$ in the MD KKT equation. FTRL evaluates
Eq.~\eqref{eq:new-da-sparse} from the whole history, independently of the
previous support; boundary normal-cone multipliers in MD do not telescope.
\end{proof}
Unlike KL-based multiplicative updates that trap coordinates at zero, Euler
sparse MD can reactivate them; an implementation that permanently masks zeros
alters the algorithm.

\subsection{Schedule-free Euler FTRL through an online-to-batch conversion}
\label{subsec:new-schedulefree}
Schedule-free learning removes dependence on a terminal horizon or decay
schedule, not the base rate or geometry. Defazio et al.~\cite{Defazio2024SF}
give a conversion that accepts an arbitrary online learner. Let
$L(\bp)=\mathbb E_\xi\mathcal L(\bp;\xi)$ be convex on $\Delta_N$, let the
online iterate $\bz_t$ be measurable before the sample $\xi_t$, and require
$\mathbb E[\bg_t\mid\mathcal F_{t-1}]\in\partial L(\by_t)$. Define
\begin{equation}
\begingroup
\begin{gathered}
\bx_t=\frac1t\sum_{s\le t}\bz_s,\quad
 \by_t=\vartheta\bx_t+(1-\vartheta)\bz_t,\\
 \bg_t\in\partial\mathcal L(\by_t;\xi_t),\quad
 \bz_{t+1}=\operatorname*{arg\,min}_{\bp\in\Delta_N}
 \Bigl\{\Bigl\langle\sum_{s\le t}\bg_s,\bp\Bigr\rangle+\beta_t\mathcal R_{a,b,\lambda}(\bp)\Bigr\},
\end{gathered}
\endgroup
 \label{eq:new-sf-ftrl}
\end{equation}
with $\vartheta\in[0,1]$; averaging is in primal coordinates.

\begin{theorem}[Convex schedule-free Euler conversion]
\label{thm:new-sf}
For each deterministic $T\ge1$ and comparator $\bp^*\in\Delta_N$,
\begin{equation}
 \mathbb E[L(\bx_T)-L(\bp^*)]\le\frac1T\mathbb E\sum_{t=1}^T\langle\bg_t,\bz_t-\bp^*\rangle .
 \label{eq:new-sf-conversion}
\end{equation}
With $\beta_t=1/\eta_t$ from Theorem~\ref{thm:Euler-adaptive}, $\delta=0$ and
$\eta_0=\sqrt{2\mu R}$, the right side is at most
\begin{equation}
 \frac{C_{a,b,\lambda}}{T}\mathbb E\sqrt{\sum_{t\le T}\sigma_t^2},\qquad
 C_{a,b,\lambda}=\sqrt{2R/\mu}+\sqrt{R/(2\mu)+4},
 \label{eq:new-sf-rate}
\end{equation}
which is $C_{a,b,\lambda}G/\sqrt T$ if $\sigma_t\le G$ almost surely, without
knowing $T$ or $G$.
\end{theorem}
\noindent\textit{The proof is given in Appendix~\ref{app:proofs}, p.~\pageref{prf:thm:new-sf}.}
For $\vartheta=0$ this is primal averaging of an FTRL learner; for $\vartheta=1$
it is the anytime conversion of~\cite{Cutkosky2019}. The theorem separates two
decisions: the learner chooses the geometry, while averaging and interpolation
convert regret into an optimization bound. It does not establish schedule-free
performance for nonconvex networks; a 2026 large-batch
report~\cite{Defazio2026Plus} motivates, but does not support, transfer to
transformer training.

\begin{algorithm}[ht]
\caption{\textsc{SF-EulerFTRL}: schedule-free Euler FTRL with online, average and evaluation points}
\label{alg:sf-ftrl}
\begingroup\small
\begin{algorithmic}[1]
\Require $N\ge2$; $-1<a<0<\rev{b\le1}$, $\lambda>0$; interpolation $\vartheta\in[0,1]$, base rate $\eta_0>0$, offset $\delta\ge0$; deterministic horizon $T\ge1$; stochastic oracle for convex expected loss $L$
\Ensure Primal average $\bx_T$ of the $T$ online decisions
\State $\bz_1\gets\1/N$; $\bx_0\gets\bz_1$; $\bh\gets\0$; $S\gets0$.
\For{$t=1,\ldots,T$}
 \State $\bx_t\gets\bx_{t-1}+(\bz_t-\bx_{t-1})/t$.
 \State $\by_t\gets\vartheta\bx_t+(1-\vartheta)\bz_t$.
 \State Query at $\by_t$ to obtain $\bg_t$ whose conditional mean, given the pre-query history, belongs to $\partial L(\by_t)$.
 \If{$t<T$}
  \State $\bh\gets\bh+\operatorname{ctr}(\bg_t)$; $S\gets S+\operatorname{hr}(\bg_t)^2$.
  \State Set $\bz_{t+1}\gets\1/N$ if $\delta+S=0$; otherwise use the exact map $\mathcal N_{a,b,\lambda}(-\eta_0\bh/\sqrt{\delta+S})$.
 \EndIf
\EndFor
\State \Return $\bx_T$
\Guarantee Theorem~\ref{thm:new-sf} gives the online-to-batch bound. Its displayed scale-free rate uses $\delta=0$ and $\eta_0=\sqrt{2\mu R}$.
\end{algorithmic}
\endgroup
\end{algorithm}

\paragraph{Three different claims.} Scale-free (invariance to the loss unit), schedule-free (no horizon) and parameter-free (adaptation to unknown comparator norms) are different properties; the last is not obtained by choosing a kernel. A portfolio of fixed geometries is a simple, guaranteed way to reduce sensitivity to $(a,b,\lambda)$ (Appendix~\ref{subsec:new-parameterfree}).

\subsection{Learning the geometry incurs a drift term}
\label{subsec:new-drift}
Section~\ref{sec:learn-parameters} gives exact derivatives in all three
parameters. In an online algorithm, changing them changes both the dual
coordinates and the potential used in telescoping.

\begin{theorem}[OMD with a changing Euler potential]
\label{thm:new-drift}
Let $K=K_\epsilon$ and let $\theta_t=(a_t,b_t,\lambda_t)$ be chosen before round
$t$ from a compact admissible set on which $F_{\theta_t}$ is differentiable on $K$
and uniformly $\mu_*$-strongly convex in $\|\cdot\|_1$. For
$\bp_{t+1}=\argmin_{\bp\in K}\{\eta\langle\bg_t,\bp\rangle+D_{F_{\theta_t}}(\bp\Vert\bp_t)\}$
and
\begin{equation}
 \Delta_t=\sup_{\bp,\bq\in K}\bigl|D_{F_{\theta_t}}(\bp\Vert\bq)-D_{F_{\theta_{t-1}}}(\bp\Vert\bq)\bigr|,
 \label{eq:new-drift-definition}
\end{equation}
every $\bp^*\in K$ satisfies
\begin{equation}
 \mathrm{Reg}_T(\bp^*)\le\frac{D_{F_{\theta_1}}(\bp^*\Vert\bp_1)+\sum_{t=2}^T\Delta_t}{\eta}
 +\frac{\eta}{2\mu_*}\sum_{t=1}^T\|\bg_t\|_\infty^2 .
 \label{eq:new-drift-bound}
\end{equation}
If the divergence is uniformly $L_\theta$-Lipschitz in $\theta$ on this set,
$\Delta_t\le L_\theta\|\theta_t-\theta_{t-1}\|_2$.
\end{theorem}
\begin{proof}
Sum the one-step MD inequality for $F_{\theta_t}$ and bound each mismatch
$D_{F_{\theta_t}}(\bp^*\Vert\bp_t)-D_{F_{\theta_{t-1}}}(\bp^*\Vert\bp_t)$ by $\Delta_t$.
\end{proof}
Slow or infrequent shape updates therefore preserve a sublinear bound when
$\sum_t\Delta_t=O(1)$. A shape update must recompute the dual coordinate
$\log_{\theta_{t+1}}(p_i)$; retaining the old number silently changes the
represented distribution. A practical, proposed criterion selects the
geometry on validation data with a proximity penalty and optional curvature
targets $\widehat h_x$ at representative probabilities $x$,
\begin{equation}
 \min_{\theta\in\Theta}\widehat L_{\rm val}(\theta)+\rho_\theta\|\theta-\theta_{\rm old}\|_2^2
 +\rho_h\sum_{x\in\{x_{\rm tail},x_{\rm head}\}}[\ln h_\theta(x)-\ln\widehat h_x]^2,
 \label{eq:new-shape-fit}
\end{equation}
updated between epochs; the portfolio bound~\eqref{eq:new-meta-bound} is a
simpler alternative with a full-information guarantee.

\subsection{A reader's guide to the algorithms}
\label{subsec:alg-guide}
The paper contains sixteen pseudocodes, all written in one format. Each has a
name in small capitals, an \textbf{Input} line (data, Euler parameters
$(a,b,\lambda)$ with their admissible ranges, step sizes and tolerances), an
\textbf{Output} line (the returned objects and, where a run can stop early, an
explicit status such as \textsc{success}, \textsc{precision-limit} or
\textsc{budget-exhausted}), and a \textbf{Guarantee} line naming the result that
supports it. Online methods output one decision per round. Algorithm~\ref{alg:euler-exp} evaluates
the scalar inverse and Algorithm~\ref{alg:euler-norm} the simplex normalization;
almost all others apply one template: \emph{form a score vector} (what the
learner remembers minus a step times the newest loss information), \emph{turn
scores into probabilities} by $\mathcal N_{a,b,\lambda}$ or a certified call of
Algorithm~\ref{alg:euler-norm}, and \emph{update the memory}. Signed
box-constrained updates and shifted sparse gates have their own KKT
conditions. Exact-map pseudocode specifies a mathematical update; a
finite-tolerance implementation must use the corresponding residual term, return
an explicit unsuccessful status when a certificate is unavailable, keep every
stored full-range probability strictly positive, and never use underflow to
switch between the regimes $a<0$ and $a=0$. In every pseudocode ``$x\gets v$''
assigns, text after $\triangleright$ explains a line, $\log_{a,b,\lambda}$, $\exp_{a,b,\lambda}$,
$\min$ and $[\cdot]_+$ act coordinatewise, and ``play $\bp_t$'' commits before
$\bg_t$ is revealed. Table~\ref{tab:alg-overview} records the precise scope of
each guarantee.

{\footnotesize\setlength{\tabcolsep}{4pt}\renewcommand{\arraystretch}{1.15}
\begin{longtable}{@{}r>{\raggedright\arraybackslash}p{0.33\textwidth}>{\raggedright\arraybackslash}p{0.56\textwidth}@{}}
\caption{Algorithms and the precise scope of their guarantees.}\label{tab:alg-overview}\\
\hline & Method & Guarantee and supporting result\\\hline
\endfirsthead
\hline & Method & Guarantee and supporting result\\\hline
\endhead
\ref{alg:euler-exp} & \textsc{EulerExp}: Scalar inverse & Log-error bound; rounded certification needs enclosures (Proposition~\ref{prop:exp-newton}).\\
\ref{alg:euler-norm} & \textsc{EulerNorm}: Weighted simplex normalization & Feasibility and full KKT residual (Lemma~\ref{lem:certified-norm}).\\
\ref{alg:adaptive-ftrl} & \textsc{AdaEulerFTRL}: Adaptive FTRL & Theorem~\ref{thm:Euler-adaptive}; numerical cost Lemma~\ref{lem:stage2-oracle-error}.\\
\ref{alg:ao-ftrl-prox} & \textsc{AO-EulerProx}: Adaptive optimistic shifted FTRL-Proximal & Exact minimizer of its cumulative objective; no regret claim beyond the cited structure (Section~\ref{sec:pseudocode}).\\
\ref{alg:euler-rda} & \textsc{EulerRDA}: Signed sparse dual averaging & Finite-box composite regret (Proposition~\ref{prop:stage2-rda-regret}).\\
\ref{alg:aoftrl} & \textsc{AO-EulerFTRL}: Optimistic FTRL & Forecast-error regret plus solver errors (Theorem~\ref{thm:new-adaptive-oftrl}).\\
\ref{alg:euler-pmd} & \textsc{EulerPMD}: Tabular policy improvement & Accepted updates preserve value when advantage-error bounds hold (Proposition~\ref{prop:approximate-Euler-policy}).\\
\ref{alg:euler-vmd} & \textsc{EulerVMD}: Known-model planning & Geometric rate under its schedule; Bellman stopping certificate (Theorem~\ref{thm:new-vmd}, Proposition~\ref{prop:stage3-vmd-certificate}).\\
\ref{alg:euler-mp} & \textsc{EulerMP}: Averaged Mirror-Prox & Monotone Lipschitz operator; residual cost (Theorem~\ref{thm:p2-mirrorprox}, Lemma~\ref{lem:stage3-mp-error}).\\
\ref{alg:anchored-mp} & \textsc{AnchoredEulerMP}: Anchored last iterate & Exact contraction; inexact recurrence and anchor bias (Theorem~\ref{thm:anchored-mp}).\\
\ref{alg:selfplay} & \textsc{EulerSelfPlay}: Full-information self-play & Exact additive CCE violation; solver errors enter the bound (Proposition~\ref{prop:selfplay}).\\
\ref{alg:implicit-md} & \textsc{EulerMD-Fit}: Positive interpolation & Capped $O(1/T)$ fitting rate; projection selection needs uncapped steps (Proposition~\ref{prop:implicit-md}).\\
\ref{alg:euler-adagrad} & \textsc{EulerAdaGrad}: Signed Euler-AdaGrad & Finite box, convex losses, declared gradient bounds, residual cost (Theorem~\ref{thm:euler-adagrad}).\\
\ref{alg:sf-ftrl} & \textsc{SF-EulerFTRL}: Schedule-free averaging & Convex expected excess loss at a deterministic horizon (Theorem~\ref{thm:new-sf}).\\
\ref{alg:certified-md} & \textsc{CertEulerMD}: Residual-controlled MD & Regret plus $\pi^2\epsilon_0/3$ (Corollary~\ref{cor:certified-md}).\\
\ref{alg:moe-step} & \textsc{EulerMoE}: Shifted sparse routing & Gate bracket and fixed-support derivatives; joint convergence open (Lemma~\ref{lem:gate-bracket}).\\
\hline
\end{longtable}
}

\section{Further results for policy optimization, games and applications}
\label{app:supp-apps}
This appendix contains the proposed LLM-scale policy constructions, regularized responses in games, the barrier bridge, inexact mirror steps, and the two-level mixture-of-experts router summarized in Sections~\ref{sec:new-policy-scale}--\ref{sec:phase2-moe}.

\subsection{Online policy MD and a deformed-log regression}
\label{subsec:new-opmd}
Kimi k1.5~\cite{Kimi2025} uses online policy mirror descent (OPMD, not
optimistic OMD) with a KL-regularized response update and a squared regression
surrogate. For a finite response set, a positive frozen reference $\bq$ and
rewards $\br$, the proposed Euler response update
$\bp^*=\argmax_{\bp\in\Delta_N}\{\langle\br,\bp\rangle-\tau D_{F_{a,b,\lambda}}(\bp\Vert\bq)\}$
has the interior KKT condition
\begin{equation}
 r_i-c=\tau\bigl[\log_{a,b,\lambda}(p_i^*)-\log_{a,b,\lambda}(q_i)\bigr],
 \label{eq:new-opmd-kkt}
\end{equation}
a \emph{difference of deformed logarithms}, not $\log_{a,b,\lambda}(p_i^*/q_i)$. This
suggests the frozen-sampling regression
\begin{equation}
 \min_{\theta,c}\mathbb E_{i\sim\bq}\bigl[r_i-c-\tau\{\log_{a,b,\lambda}(p_{\theta,i})-\log_{a,b,\lambda}(q_i)\}\bigr]^2,
 \label{eq:new-opmd-regression}
\end{equation}
whose zero, when attainable with $q_i>0$ for every response, is the exact
proximal optimizer. Treating $c$ as a nuisance scalar is more faithful than
replacing it by a reward mean; large $|a|$ may amplify errors in tiny sequence
probabilities.

\subsection{A ratio-based Euler alternative}
\label{subsec:new-ratio}
With a fixed positive reference, the reference-weighted potential
\begin{equation}
 \Phi^{\bq}_{a,b,\lambda}(\bp)=\sum_iq_i\phi_{a,b,\lambda}(p_i/q_i),\qquad
 D_{\Phi^{\bq}}(\bp\Vert\bq)=\sum_iq_id_{a,b,\lambda}(p_i/q_i,1),
 \label{eq:new-perspective}
\end{equation}
yields the ratio update $p_i^*=q_i\exp_{a,b,\lambda}((r_i-c)/\tau)$ with
$\sum_iq_i\exp_{a,b,\lambda}((r_i-c)/\tau)=1$, which justifies a deformed-log-ratio
regression for this potential, not for $F_{a,b,\lambda}$.

\begin{proposition}[Two Euler penalties with different local metrics]
\label{prop:new-two-metrics}
$[\nabla^2F_{a,b,\lambda}(\bp)]_{ii}=h_{a,b,\lambda}(p_i)$ and
$[\nabla^2\Phi^{\bq}_{a,b,\lambda}(\bp)]_{ii}=h_{a,b,\lambda}(p_i/q_i)/q_i$. Since
$h_{a,b,\lambda}(1)=1$ for every admissible triple, the ratio metric equals
$\diag(1/q_i)$ at $\bp=\bq$: it agrees with KL to second order at the reference,
whereas the probability-space penalty generally changes the local metric.
\end{proposition}
\begin{proof}
Differentiate twice and use the normalization $\log_{a,b,\lambda}'(1)=1$ of
Eq.~\eqref{eq:scaled-Euler-identity}.
\end{proof}
To change the trust-region sensitivity to small \emph{absolute} probabilities,
use $D_{F_{a,b,\lambda}}$; to keep the categorical local metric but change
finite \emph{relative} changes, use $D_{\Phi^{\bq}}$. These are experimentally
distinguishable hypotheses.

\subsection{An analytic Euler instance of GBMPO, and sampled penalties}
\label{subsec:new-gbmpo}\label{subsec:new-estimator}
GBMPO~\cite{Yuan2026GBMPO} explores flexible Bregman regularization in
group-based policy optimization. A proposed Euler surrogate freezes the behavior
policy $\pi_k$, uses the leave-one-out advantage
$\widehat A_j=r_j-\frac1{K-1}\sum_{l\ne j}r_l$ over $K\ge2$ responses, and
maximizes
\begin{equation}
 \mathcal J(\theta)=\widehat{\mathbb E}_{(s,\alpha)\sim\pi_k}[\rho_\theta(s,\alpha)\widehat A(s,\alpha)]
 -\tau\widehat{\mathbb E}_{s\sim d^{\pi_k}}D_{F_{a,b,\lambda}}(\pi_\theta(\cdot|s)\Vert\pi_k(\cdot|s)),
 \qquad\rho_\theta=\pi_\theta/\pi_k,
 \label{eq:new-gbmpo-surrogate}
\end{equation}
where $d^{\pi_k}$ is the state distribution of the behavior policy, with the exact penalty gradient
$\sum_i[\log_{a,b,\lambda}(p_{\theta,i})-\log_{a,b,\lambda}(q_i)]\nabla_\theta p_{\theta,i}$, computable
through an ordinary softmax network. Improved LLM reasoning for this kernel has
not been established. A full divergence sums over the vocabulary; for a frozen
proposal $\boldsymbol\varsigma\in\Delta_N^\circ$ and $I\sim\boldsymbol\varsigma$,
\begin{equation}
 \mathbb E_{I\sim\boldsymbol\varsigma}\frac{d_{a,b,\lambda}(p_I,q_I)}{\varsigma_I}=D_{F_{a,b,\lambda}}(\bp\Vert\bq),\qquad
 \mathbb E_{I\sim\boldsymbol\varsigma}\frac{[\log_{a,b,\lambda}(p_I)-\log_{a,b,\lambda}(q_I)]\nabla_\theta p_I}{\varsigma_I}
 =\nabla_\theta D_{F_{a,b,\lambda}}(\bp\Vert\bq),
 \label{eq:new-is-divergence}
\end{equation}
unbiased but without a variance guarantee; sampling from $\bp_\theta$ instead
omits a score-function term. For the ratio penalty with $I\sim\bq$,
$\phi_{a,b,\lambda}(p_I/q_I)$ is unbiased, deforming the forward-KL estimator
$x\ln x-x+1$. Because $\phi_{a,b,\lambda}$ lacks logarithmic additivity, a sum of
tokenwise Euler divergences under chosen prefix distributions is a well-defined
surrogate but generally not the Euler divergence between sequence
distributions.

\subsection{Reference-regularized responses and dissipation}\label{subsec:supp-games}
A prior-aware regularized best response,
$\operatorname{BR}(\br)=\argmax_{\bp\in\Delta_n}\{\langle\br,\bp\rangle-\tau D_{F_{a,b,\lambda}}(\bp\Vert\bq)\}$,
is $p_i=\exp_{a,b,\lambda}(\log_{a,b,\lambda}(q_i)+r_i/\tau-\nu)$, i.e.\ $p_i\propto q_ie^{r_i/\tau}$
in the logarithmic limit. Fixed points of such responses are generalized
regularized-response equilibria; classical logit QRE is the entropic, uniform
case, and existence follows from Brouwer's theorem, uniqueness in general-sum
games does not.

\begin{theorem}[Dissipation in a reference-regularized monotone Euler flow]
\label{thm:games-dissipation}
Let $\mathcal G$ be monotone, $\tau>0$, $\bq$ positive, and suppose an interior
equilibrium $\bp^*$ of $\mathcal G_\tau$ and an interior solution of
$\frac{\mathrm d}{\mathrm dt}\nabla F(\bp)=-\mathcal G_\tau(\bp)+\boldsymbol\nu(t)$, with blockwise
constant multipliers, exist for $t\ge0$. Then
\begin{equation}
 D(\bp^*\Vert\bp(t))\le e^{-\tau t}D(\bp^*\Vert\bp(0)).
 \label{eq:games-decay}
\end{equation}
\end{theorem}
\begin{proof}
Differentiate, use monotonicity and
$\langle\nabla F(\bp)-\nabla F(\bp^*),\bp-\bp^*\rangle=D(\bp\Vert\bp^*)+D(\bp^*\Vert\bp)$,
and apply Gr\"onwall's inequality.
\end{proof}
Persistent regularization introduces dissipation absent from the conservative
flow; larger $\tau$ contracts faster but moves the equilibrium more. In
preference games~\cite{Munos2024Nash,Zhang2025Optimistic,Tiapkin2026Prox}, reasoning
self-play~\cite{Liu2025Spiral,Yuan2025Mars}, heterogeneous cooperative
MARL~\cite{Nasiri2023Hamdpo} and PSRO-style meta-solvers, an Euler Bregman
divergence can take the role of a KL penalty, an opponent-mixture regularizer or a response map; these
are separate interventions whose assumptions, advantage estimators and
constants must be rechecked. With bandit feedback the importance-weighted
estimator has conditional second moment $c_{i,t}^2/p_{i,t}$, so positivity alone
does not control variance. Proposition~\ref{prop:euler-bandit} gives a two-exponent bound whose variance term is $\sum_i1/\omega_{a,b,\lambda}(p_{i,t})$; the best-of-both-worlds guarantees of Tsallis-INF~\cite{Zimmert2021} are not transferred to $b>0$.

\paragraph{Natural generalizations.}
For structured actions, a fixed convex mixture of Euler potentials with
coordinate weights $d_i>0$,
\begin{equation}
 F^{\rm mix}(\bp)=\sum_id_i\sum_k\omega_k\phi_{a_k,b_k,\lambda_k}(p_i),\qquad\omega_k\ge0,\ \sum_k\omega_k=1,
 \label{eq:games-mixture}
\end{equation}
{uses a strictly increasing scalar link $\log^{\rm mix}$ with $(\log^{\rm mix})'(1)=1$. Its interior prox step is} $p_i=(\log^{\rm mix})^{-1}(\log^{\rm mix}(q_i)-(\eta g_i+\nu)/d_i)$; {if the unweighted separable component potentials have moduli $\mu_k$ in a common norm on the same domain, a valid modulus for $F^{\rm mix}$ is $(\min_i d_i)\sum_k\omega_k\mu_k$. This follows by summing the nonnegative coordinate Bregman divergences. The mixture generally lies outside the three-parameter family.}

\begin{proposition}[Convex mixtures preserve the Euler mirror construction]
\label{prop:games-mixture}
Under the stated assumptions $\log^{\rm mix}=\sum_k\omega_k\log_{a_k,b_k,\lambda_k}$ is
strictly increasing with $\log^{\rm mix}(1)=0$, $(\log^{\rm mix})'(1)=1$, its
curvature is the weighted sum of component curvatures, and the interior prox
step has the displayed inverse form.
\end{proposition}
\begin{proof}
Differentiate the finite sum and the constrained prox objective.
\end{proof}
For sequential games, a dilated analogue
$\mathcal F(\bx)=\sum_Id_I\chi_IF_{a_I,b_I,\lambda_I}(\bx_I/\chi_I)$ over information
sets $I$, with reach variables $\chi_I$, is convex as a sum of perspectives~\cite{Farina2019Dilated}; its
strong-convexity modulus on the treeplex and an efficient recursive solver
remain to be established against the near-optimal entropic
benchmark~\cite{Fan2024Dilated}.

\subsection{An infinite slope is not an infinite barrier value}\label{subsec:supp-barrier}
For $-1<a<0$ the potential $\phi_{a,b,\lambda}$ is finite at zero although its
slope diverges: a singular slope prevents a finite dual update from reaching
zero, whereas a barrier potential is infinite at the boundary. The family
crosses this distinction at $|a|=1$. In this subsection only, allow $|a|\ge1$
with $b\in[0,1]$. At the endpoint $(a,b)=(-1,0)$ the scale cancels and
\begin{equation}
 \begin{gathered}
 \log_{-1,0}(x)=1-\frac1x,\qquad \exp_{-1,0}(z)=\frac1{1-z}\ (z<1),\qquad\phi(x)=x-1-\ln x,\\
 D_F(\bp\Vert\bq)=\sum_i\Bigl[\frac{p_i}{q_i}-1-\ln\frac{p_i}{q_i}\Bigr],
 \end{gathered}
 \label{eq:p2-burg}
\end{equation}
the Burg potential and the Itakura--Saito divergence, infinite at sparse
comparators. Logarithmic-barrier MD has its own tight $O(\log k/k)$
theory~\cite{DeMarchi2026Barrier}; the finite-radius constants of this paper do
not extend to $|a|=1$. Table~\ref{tab:p2-boundary} summarizes the regimes.

\begin{table}[ht]
\centering\small
\caption{Euler boundary regimes and the claims they support.}
\label{tab:p2-boundary}
\begin{tabular}{L{.17\textwidth}L{.25\textwidth}L{.23\textwidth}L{.24\textwidth}}
\hline
Regime & {$\phi(0)$ and $\phi'(0^+)$} & Finite-score output & Main limitation\\\hline
$a=0<b\le1$ & finite, finite & zeros can occur & support changes, lost exploration\\
$a=b=0$ & finite, $-\infty$ & dense entropy map & exponential small-probability scale\\
$0<|a|<1$, $\lambda>0$ & finite, $-\infty$ & dense polynomial tails & no global self-concordance\\
$(a,b)=(-1,0)$ & {$+\infty$, $-\infty$ (log barrier)} & dense rational inverse & boundary radius infinite\\
{$a=-1$, $b>0$} & {$+\infty$, $-\infty$} & {dense, full-range inverse} & {self-concordant, $M=\sqrt{1+b\lambda^{b+1}}$ for $b\le1$ (Prop.~\ref{prop:barrier-sc}); infinite radius}\\
$|a|>1$ & {$+\infty$, $-\infty$ (power barrier)} & dense & new domain and complexity analysis\\\hline
\end{tabular}
\end{table}

For the regularized linear program
$\min_{\bp\in\Delta_N^\circ}\bc^T\bp+\tau F_{a,b,\lambda}(\bp)$, interior
stationarity gives $p_i=\exp_{a,b,\lambda}(-(c_i+\nu)/\tau)$. At $(-1,0)$, with
$s_i=c_i+\nu+\tau$,
\begin{equation}
 p_i>0,\qquad s_i>0,\qquad p_is_i=\tau,\qquad\sum_ip_i=1,
 \label{eq:p2-complementarity}
\end{equation}
the perturbed complementarity equations of a simplex LP: the family has an
exact log-barrier central-path endpoint. For $b=0<|a|<1$ one obtains instead
$(c_i+\nu+\tau/|a|)p_i^{|a|}=\tau/|a|$, a power-weighted relation. An FTRL
decision at fixed cumulative cost lies on such a path with $\tau=\beta_t$, but
online FTRL and interior-point methods have different trajectories and
complexity guarantees. Near zero, for $|a|>0$,
\begin{equation}
 \frac{|\phi_{a,b,\lambda}'''(x)|}{\phi_{a,b,\lambda}''(x)^{3/2}}
 \sim\frac{1-a}{\sqrt{\alpha_\lambda}}\,x^{(|a|-1)/2},\qquad x\downarrow0,
 \label{eq:p2-selfconcordance}
\end{equation}
which diverges for $0<|a|<1$, equals $2$ at the Burg endpoint, and tends to zero
for $|a|>1$ without, by itself, proving self-concordance or a barrier parameter.
Use $0<|a|<1$ for finite boundary costs and polynomial retention, the endpoint
when a genuine barrier is required, and a floored domain when a controlled
numerical bridge suffices.

\paragraph{Global self-concordance of the barrier hybrids.}
For $a=-1$ the ratio in Eq.~\eqref{eq:p2-selfconcordance} tends to $2/\sqrt{\alpha_\lambda}$ at zero, and the bulk term does not destroy the bound elsewhere. Following~\cite{NesterovNemirovski1994}, call $\phi$ $M$-self-concordant if $|\phi'''|\le2M(\phi'')^{3/2}$; the ratio in Eq.~\eqref{eq:p2-selfconcordance} is then $2M$.

\begin{proposition}[Exact self-concordance constant for $a=-1$]
\label{prop:barrier-sc}
For $a=-1$, $0\le b\le1$ and $\lambda>0$, $\phi_{-1,b,\lambda}$ is self-concordant on $(0,\infty)$ with the exact constant
\begin{equation}
 M_{\rm sc}(\lambda)=\sup_{x>0}\frac{|\phi_{-1,b,\lambda}'''(x)|}{2\,\phi_{-1,b,\lambda}''(x)^{3/2}}
 =\alpha_\lambda^{-1/2}=\sqrt{1+b\lambda^{b+1}},
 \label{eq:barrier-sc}
\end{equation}
attained as $x\downarrow0$. Consequently, a step $\Delta$ with $|\Delta|\sqrt{h_{-1,b,\lambda}(x)}\le\rho\sqrt{\alpha_\lambda}$, $\rho<1$, changes the curvature by a factor in $[(1-\rho)^2,(1-\rho)^{-2}]$; with $\rho=\frac12$ the local radius is $\sqrt{\alpha_\lambda}/2$ and the factor is at most four.
\end{proposition}
\begin{proof}
With $A=\alpha_\lambda x^{-2}$, $B=\beta_\lambda x^{b-1}$ and $u=B/A$, $|\phi'''|=(2A+(1-b)B)/x$ and $\phi''=A(1+u)$, so the ratio equals $\alpha_\lambda^{-1/2}(1+cu)(1+u)^{-3/2}$ with $c=(1-b)/2\in[0,\frac12]$. Its derivative in $u$ is $(1+u)^{-5/2}[(c-\frac32)-\frac c2u]<0$, so the supremum is the limit $u\downarrow0$. At $a=-1$, $\alpha_\lambda=1/(1+b\lambda^{b+1})$. The curvature-change statement is the standard consequence of self-concordance~\cite{NesterovNemirovski1994}.
\end{proof}
The scale therefore trades two properties: a small $\lambda$ keeps $M_{\rm sc}\approx1$ and gives the best bandit constant (Corollary~\ref{cor:bandit-shape}), whereas a large $\lambda$ shrinks the comparator divergence and the uniform-centre curvature but raises $M_{\rm sc}$ and the variance constant $1/\alpha_\lambda$ by the same factor. No barrier parameter is claimed: the bulk term makes $\phi'$ unbounded at infinity.

\subsection{Robustness to approximate prox solutions needs an error model}
\label{subsec:p2-inexact}
Objective gaps, first-order (KKT) residuals and coordinate errors are not
equivalent near a singular boundary. For objective gaps, Schlisselberg et
al.~\cite{Schlisselberg2026} require a power-curvature condition, which Euler
kernels satisfy explicitly:
\begin{equation}
 \alpha_\lambda x^{a-1}\le h_{a,b,\lambda}(x)\le x^{a-1},\qquad0<x\le1,\ a<0\le b,
 \label{eq:p2-curvature-sandwich}
\end{equation}
since $x^{1-a}h_{a,b,\lambda}(x)=\alpha_\lambda+\beta_\lambda x^{b-a}$ increases
from $\alpha_\lambda$ to one; the boundary order $1-a$ is independent of $b$ and
$\lambda$, which change only constants.

\begin{corollary}[Euler transfer of an objective-error robustness result]
\label{cor:p2-objective-errors}
Fix $0<|a|<1$, $0\le b\le1$, $\lambda>0$, a valid $\ell_1$ modulus $\mu$, and set
$c_1=\alpha_\lambda/\mu$, $c_2=1/\mu$. On the simplex $\Delta_N$ with uniform
initialization, losses in $[-1,1]^N$ and $\eta\le1/(16c_1)$, if each normalized
prox objective is minimized to additive error at most
\begin{equation}
 \epsilon_{\rm obj}\le\eta^4\min\{c_2^{-1},c_2\}
 \Bigl[\frac{16\eta TN}{c_1}+2(2N)^{|a|}\Bigr]^{-(1-a)/|a|},
 \label{eq:p2-objective-tolerance}
\end{equation}
then $\mathrm{Reg}_T(\bp^*)\le D_{F_{a,b,\lambda}/\mu}(\bp^*\Vert\1/N)/\eta+O(\eta T)$, with the
dependencies of~\cite[Theorem~6]{Schlisselberg2026}.
\end{corollary}
\begin{proof}
Eq.~\eqref{eq:p2-curvature-sandwich} is their separable power-curvature
condition with exponent $1-a$ after normalization by $\mu$.
\end{proof}
\begingroup
To apply the regret bound through round $t$, certify the successor of the played decision in that round, even at the terminal horizon. If that call fails, the bound applies through the preceding round. For any comparator in the simplex the additional linearized regret of the final played round is at most $2\operatorname{hr}(\bg_t)$, since the displacement has zero sum and $\ell_1$ norm at most two.
\par\endgroup
For first-order residuals, let the feasible approximate update satisfy
\begin{equation}
 \eta\bg_t+\nabla F_{a,b,\lambda}(\bp_{t+1})-\nabla F_{a,b,\lambda}(\bp_t)+\bn_t=\br_t,\qquad
 \bn_t\in N_K(\bp_{t+1}),\ \|\br_t\|_*\le\epsilon_t,
 \label{eq:p2-prox-residual}
\end{equation}
on a set $K$ of diameter $D_K$; on the interior simplex the residual is the
spread of the coordinate KKT scores.

\begin{proposition}[A residual-based inexact Euler MD bound]
\label{prop:p2-inexact}
If $F_{a,b,\lambda}$ is $\mu$-strongly convex on $K$ and $\eta>0$ is constant, then
\begin{equation}
 \sum_{t=1}^T\langle\bg_t,\bp_t-\bp^*\rangle\le\frac{D_{F_{a,b,\lambda}}(\bp^*\Vert\bp_1)}{\eta}
 +\frac{\eta}{2\mu}\sum_{t=1}^T\|\bg_t\|_*^2+\frac{D_K}{\eta}\sum_{t=1}^T\epsilon_t .
 \label{eq:p2-inexact-bound}
\end{equation}
\end{proposition}
\begin{proof}
Pair Eq.~\eqref{eq:p2-prox-residual} with $\bp_{t+1}-\bp^*$, use the normal-cone
sign and $\langle\br_t,\bp_{t+1}-\bp^*\rangle\le D_K\epsilon_t$, the three-point
identity and Young's inequality, and telescope.
\end{proof}
A fixed nonzero tolerance contributes a term linear in $T$, whereas a summable
schedule costs a constant; on the simplex $D_K=2$ and the certificate of
Algorithm~\ref{alg:euler-norm} is exactly such a residual.

\begin{corollary}[Residual-certified inexact Euler MD]
\label{cor:certified-md}
For $-1<a<0<b\le1$, $\lambda>0$, $\mu=\mu_{a,b,\lambda}(N)$ and tolerances
{$\epsilon_0\eta/t^2$, suppose Algorithm~\ref{alg:certified-md} certifies the first $T$ successor updates, including the update after the last played round. Then, for every}
$\bp^*\in\Delta_N$,
\begin{equation}
 \sum_{t=1}^T\langle\bg_t,\bp_t-\bp^*\rangle\le\frac{D_{F_{a,b,\lambda}}(\bp^*\Vert\1/N)}{\eta}
 +\frac{\eta}{2\mu}\sum_{t=1}^T\operatorname{hr}(\bg_t)^2+\frac{\pi^2\epsilon_0}{3}.
 \label{eq:certified-md}
\end{equation}
\end{corollary}
\begin{proof}
{Parts (i) and (ii) of Lemma~\ref{lem:certified-norm}, together with the acceptance test, give interior feasible outputs with residual}
half-range $\hat\epsilon_t$; apply Proposition~\ref{prop:p2-inexact} with
$D_K=2$ and centered gradients.
\end{proof}

\begin{algorithm}[ht]
\caption{\textsc{CertEulerMD}: residual-certified inexact Euler mirror descent}
\label{alg:certified-md}
\begingroup\small
\begin{algorithmic}[1]
\Require $N\ge2$; $-1<a<0<b\le1$, $\lambda>0$; step $\eta>0$; tolerance scale $\epsilon_0>0$; horizon $T\ge1$; online loss vectors $\bg_t$
\Ensure Played decisions, number of certified updates, and status
\State $\bp_1\gets\1/N$.
\For{$t=1,\ldots,T$}
 \State Play $\bp_t$; observe $\bg_t$.
 \State $\epsilon_t\gets\epsilon_0\eta/t^2$; $\by_t\gets\log_{a,b,\lambda}(\bp_t)-\eta\bg_t$.
 \State Use $\mathsf{Norm}(\by_t;\epsilon_t)$ to certify the successor $\bp_{t+1}$.
 \State If the call fails, \Return $(\bp_1,\ldots,\bp_t)$, $t-1$, and \textsc{precision-limit}.
\EndFor
\State \Return $(\bp_1,\ldots,\bp_T)$, $T$, and \textsc{success}.
\Guarantee Eq.~\eqref{eq:certified-md} holds through the last certified update. On failure, the final played round adds at most $2\operatorname{hr}(\bg_t)$ to the linearized regret of the certified prefix. The terminal successor is certified for the bound and is not played.
\end{algorithmic}
\endgroup
\end{algorithm}
{The residual tolerance decreases as $t^{-2}$. For fixed bracket width and geometry, the outer iteration bound therefore grows only logarithmically with $t$. Total work also includes inner inversions and the precision needed to verify feasibility.} For example, a
primal perturbation of $10^{-8}$ can destroy a coordinate whose exact value is
$10^{-20}$, so log-domain computation and explicit support conventions are part
of the algorithm.

\subsection{A two-level dense-retention and sparse-dispatch design}
\label{subsec:p2-twolevel}
A complementary proposal keeps a dense availability state
$\bv_t\in\Delta_N^\circ$ learned by full-range Euler FTRL, never zero but not
deciding dispatch, and adjusts token scores by
\begin{equation}
 \widetilde z_i(x)=z_i(x)+c_v\ln v_{i,t}-\xi_{i,t},\qquad c_v\ge0,
 \label{eq:p2-twolevel-routing}
\end{equation}
before a sparse gate, where $\xi_{i,t}\ge0$ are capacity prices. For a batch of
$B$ tokens with capacities $C_i>0$, $\sum_iC_i>B$, prices can be updated by
projected dual ascent $\xi_{i,t+1}=[\xi_{i,t}+\eta_\xi(\widehat L_{i,t}-C_i)]_+$.

\begin{proposition}[The price update is dual ascent for frozen scores]
\label{prop:price-ascent}
{Let $-1<a\le0<b\le1$, $\lambda,\delta,\tau>0$, and $C_i>0$ with $\sum_i C_i>B$. For fixed scores $\bz_1,\ldots,\bz_B$ consider}
\begin{equation}
 \min_{\bp_j\in\Delta_N}\sum_{j=1}^B\bigl[-\langle\bz_j,\bp_j\rangle+\tau F^\delta_{a,b,\lambda}(\bp_j)\bigr]
 \quad\text{s.t.}\quad\sum_jp_{j,i}\le C_i .
 \label{eq:capacity-problem}
\end{equation}
With $\mu_\delta=h_{a,b,\lambda}(1+\delta)$ and loads $\bL(\boldsymbol\xi)$ of the
shifted gate applied to $\bz_j-\boldsymbol\xi$, the dual function $g$ is concave
with $\nabla g=\bL-\bC$, $(B/\tau\mu_\delta)$-Lipschitz; for
$0<\eta_\xi\le\tau\mu_\delta/B$ the projected ascent increases $g$ monotonically,
$g(\boldsymbol\xi^*)-g(\boldsymbol\xi_T)\le\|\boldsymbol\xi_0-\boldsymbol\xi^*\|_2^2/(2\eta_\xi T)$,
and $\sum_j\|\bp_j(\boldsymbol\xi_T)-\bp_j^*\|_2^2\le2[g(\boldsymbol\xi^*)-g(\boldsymbol\xi_T)]/(\tau\mu_\delta)$.
\end{proposition}
\begin{proof}
$h_{a,b,\lambda}$ is nonincreasing, so the primal objective is
$\tau\mu_\delta$-strongly convex in $\ell_2$. The Lagrangian separates over
tokens with minimizers given by the shifted gate; Danskin's theorem and the
$1/\sigma$-Lipschitz minimizer of a $\sigma$-strongly convex function \rev{give the $(B/\tau\mu_\delta)$-Lipschitz gradient of $g$};
{assigning every token the vector $\bC/\sum_i C_i$ gives strict capacity feasibility since $B<\sum_i C_i$. This yields strong duality and an optimal price. For each fixed price $\boldsymbol\xi\ge0$, the constrained primal optimum evaluated in the Lagrangian has value at most the primal optimum value $g(\boldsymbol\xi^*)$. Strong convexity of that Lagrangian therefore bounds the squared distance from its minimizer by $2[g(\boldsymbol\xi^*)-g(\boldsymbol\xi)]/(\tau\mu_\delta)$. The dual objective estimate follows from projected ascent with the stated step size.}
\end{proof}
During training the scores move and batches are random, so joint convergence is
not covered; a mass bound also does not enforce an integer dispatch limit.

\begin{lemma}[Bracket and candidate set for the shifted Euler gate]
\label{lem:gate-bracket}
The multiplier satisfies
$z_{(1)}-\tau\kappa_\delta\le\nu^*\le z_{(1)}-\tau\kappa_\delta(1/N)<z_{(1)}$,
every inverse evaluated on this bracket has argument in
$(\log_{a,b,\lambda}(\delta),\log_{a,b,\lambda}(1+\delta)]$, every expert with
$z_i\le z_{(1)}-\tau\kappa_\delta$ is inactive, and on the active set
$\partial_\nu\sum_ip_i=-\tau^{-1}\sum_{i\in S}m_{a,b,\lambda}(p_i+\delta)$.
\end{lemma}
\begin{proof}
The mass is continuous and nonincreasing; it is $0$ at $\nu=z_{(1)}$ and at least
one at $\nu=z_{(1)}-\tau\kappa_\delta$, where the top term equals one. Both
argument endpoints lie in the range of $\log_{a,b,\lambda}$, also on the sparse branch.
\end{proof}

\begin{algorithm}[ht]
\caption{\textsc{EulerMoE}: two-level Euler routing with an ideal forward map and a support-aware backward pass}
\label{alg:moe-step}
\begingroup\small
\begin{algorithmic}[1]
\Require $B$ token score vectors $\bz_j\in\mathbb R^N$, $N\ge2$; gate triple $-1<a\le0<b\le1$, $\lambda>0$, shift $\delta>0$, temperature $\tau>0$; capacity prices $\boldsymbol\xi_t\ge0$, capacities $C_i>0$ and step $\eta_\xi>0$
\Require Positive simplex prior $\bv_t$ with full-range triple $a_v<0<b_v\le1$, $\lambda_v>0$; coupling $c_v\ge0$, rate $\eta_v>0$, prior tolerance $\epsilon_v>0$, cumulative losses $\bH_{t-1}$; expert loss vector $\boldsymbol\ell_t$ supplied after routing
\Ensure Routing vectors, adjusted-score gradients, new prices and prior state
\State Set $\kappa(d)=\log_{a,b,\lambda}(\delta+d)-\log_{a,b,\lambda}(\delta)$ for $d\in[0,1]$.
\For{token $j=1,\ldots,B$}
 \State $\widetilde\bz_j\gets\bz_j+c_v\ln\bv_t-\boldsymbol\xi_t$; $z_{\max}\gets\max_i\widetilde z_{j,i}$.
 \State Keep candidates with $\widetilde z_{j,i}>z_{\max}-\tau\kappa(1)$.
 \State For trial $\nu$, set $p_i(\nu)=\exp_{a,b,\lambda}(\log_{a,b,\lambda}(\delta)+(\widetilde z_{j,i}-\nu)/\tau)-\delta$ for candidates with $\widetilde z_{j,i}>\nu$, and set all other entries to zero.
 \State Find the unique root of $\sum_i p_i(\nu)=1$ on $[z_{\max}-\tau\kappa(1),z_{\max}-\tau\kappa(1/N)]$ by bisection; use the ideal values $p_{j,i}=p_i(\nu)$.
 \State Dispatch token $j$ to $S_j=\{i:p_{j,i}>0\}$ under the chosen routing policy.
\EndFor
\For{token $j$ with upstream gradient $\bu_j=\nabla_{\bp_j}\mathrm{loss}$}
 \State For $i\in S_j$, $\widetilde m_i\gets \rev{m_{a,b,\lambda}(\delta+p_{j,i})}$; $\bar u_j\gets\sum_{i\in S_j}\widetilde m_i u_{j,i}/\sum_{i\in S_j}\widetilde m_i$.
 \State Set score gradients to $\widetilde m_i(u_{j,i}-\bar u_j)/\tau$ on $S_j$ and zero elsewhere, treating threshold coordinates as inactive.
\EndFor
\State $\xi_{t+1,i}\gets[\xi_{t,i}+\eta_\xi(\sum_j p_{j,i}-C_i)]_+$ for every $i$.
\State $\bH_t\gets\bH_{t-1}+\boldsymbol\ell_t$; call $\mathsf{Norm}_v(-\eta_v\bH_t;\epsilon_v)$.
\State Set $\bv_{t+1}$ to the returned decision if certified; otherwise retain $\bv_t$.
\State \Return all routing vectors, score gradients, and $(\boldsymbol\xi_{t+1},\bH_t,\bv_{t+1})$.
\Guarantee The ideal gate solves the stated simplex problem. The backward formula is exact on fixed support. The price update targets routing mass.
\end{algorithmic}
\endgroup
\end{algorithm}
{On a round when a new prior is certified and accepted, the updated state satisfies}
$v_{i,t+1}\ge \exp_{a_v,b_v,\lambda_v}(\log_{a_v,b_v,\lambda_v}(1/N)-\eta_v\operatorname{range}(\bH_t)-2\hat\epsilon_v)>0$,
which is not an exploration guarantee. For a feasible gate candidate, the
primal KKT residual $\epsilon_{\rm gate}$ (equalities on the support, one-sided
inequalities off it) gives
$\|\bp-\bp^*\|_2\le\epsilon_{\rm gate}/[\tau h_{a,b,\lambda}(1+\delta)]$.

\section{Proofs of results in the main text}
\label{app:proofs}
Proofs that are short enough to aid understanding are kept beside their statements; the longer ones are collected here in the order of the main text.

\subsection*{Section~\ref{sec:logab-properties}: The three-parameter scale-free Euler logarithm and its geometry}
\paragraph{Proof of Lemma~\ref{lem:audit-domains}.}\label{prf:lem:audit-domains}
The derivative is $\alpha_\lambda x^{a-1}+\beta_\lambda x^{b-1}$.
In the admissible regime its coefficients are nonnegative. For exponents
of the same nonzero sign, the derivative changes sign at
$x=\lambda^{-1}(b/a)^{1/(a-b)}$.
The second derivative is
$\alpha_\lambda(a-1)x^{a-2}+\beta_\lambda(b-1)x^{b-2}$.
If $b>1$, its positive term dominates at infinity; otherwise both terms
are nonpositive. Endpoint limits prove the ranges. The mixture of
$\log_{a,0},\log_{0,b}$ tends to $\ln x$ as both exponents tend to zero.
The coincident-exponent formula follows by differentiating powers in
the rescaling identity; its derivative changes sign for $c\ne0$. In detail, $\log_{c,c}(u)=u^c\ln u$ and $u\log_{c,c}'(u)=u^c(1+c\ln u)$, so Eq.~\eqref{eq:scaled-Euler-identity} gives $[x^c\ln x+(x^c-1)\ln\lambda]/(1+c\ln\lambda)$, whose derivative $x^{c-1}[1+c\ln(\lambda x)]/(1+c\ln\lambda)$ changes sign at $x=\lambda^{-1}e^{-1/c}$.\hfill$\square$

\subsection*{Section~\ref{sec:canonical-softmax}: Euler softmax: normalization, posterior recovery, and convex logit losses}
\paragraph{Proof of Proposition~\ref{prop:Euler-softmax}.}\label{prf:prop:Euler-softmax}
(i) Uniqueness and positivity follow from Remark~\ref{rem:why-Euler-regularizer}(ii)--(iii), and the closed form from the KKT conditions. Adding $c\1$ shifts $\nu$ by $c/\tau$. Order is preserved because $\exp_{a,b,\lambda}$ is increasing. (ii) Differentiating $\log_{a,b,\lambda}(\rev{\rho_\alpha})=\theta_\alpha/\tau-\nu$ gives $\mathrm d\rev{\rho_\alpha}=m_\alpha(\mathrm d\theta_\alpha/\tau-\mathrm d\nu)$, and $\sum_\alpha\mathrm d\rev{\rho_\alpha}=0$ yields $\mathrm d\nu=\sum_\beta m_\beta\mathrm d\theta_\beta/(\tau\1^T\mathbf{m})$. (iii) For large $\tau$, divide the objective by $\tau$; compactness and the unique minimum of $F_{a,b,\lambda}$ give the uniform limit. For small $\tau$, a maximizing coordinate has inverse argument at most $\log_{a,b,\lambda}(1)=0$. Every nonmaximizing argument is smaller by a quantity diverging to $+\infty$, so its probability vanishes. Tied maximum logits have equal probabilities, proving the stated limit. (iv) Apply Proposition~\ref{prop:sparse-FTRL}. For $0<b<1$, $[z]_+^{1/b}$ is continuously differentiable, with zero derivative at $z=0$; implicit normalization therefore gives the Jacobian even at support changes. For $b=1$, differentiate on each fixed-support region~\cite{Peters2019}.\hfill$\square$

\paragraph{Proof of Proposition~\ref{prop:matched-Euler-loss}.}\label{prf:prop:matched-Euler-loss}
The KKT identity $\log_{a,b,\lambda}(q_y)=s_y-\nu$ proves the loss identity, and Eq.~\eqref{eq:normalizer-gradient} proves the gradient. Let $m_i'=m_{a,b,\lambda}'(q_i)=-\log_{a,b,\lambda}''(q_i)/\log_{a,b,\lambda}'(q_i)^2>0$. Differentiating $\br=\mathbf{m}/Z_m$ gives
\begin{equation}
\nabla^2\nu(\bs)=\frac1{Z_m}
(\bI-\br\1^T)\diag(m_i m_i')(\bI-\1\br^T)\succeq0.
\label{eq:normalizer-Hessian}
\end{equation}
The two exterior matrices are transposes of each other, and all diagonal weights are positive. Its kernel consists precisely of common shifts. The gradient vanishes in expectation exactly when $\br=\bp$, equivalently $\bq=\mathcal T_{a,b,\lambda}(\bp)$. Every interior $\bq$ is attained by logits $\bs=\log_{a,b,\lambda}(\bq)$, so such minimizers exist.\hfill$\square$

\subsection*{Section~\ref{sec:inverse-computation}: Computing and differentiating the scaled Euler exponential}
\paragraph{Proof of Proposition~\ref{prop:exp-solver-shape}.}\label{prf:prop:exp-solver-shape}
The global slope minimum is Proposition~\ref{prop:omega-profile}.
The curvature inequality follows by bounding
$|\alpha_\lambda a e^{az}+\beta_\lambda b e^{bz}|$
by $c(\alpha_\lambda e^{az}+\beta_\lambda e^{bz})$.
For $z\ge0$, $(1-e^{-pz})/p\le z\le(e^{bz}-1)/b$.
The mixture therefore lies between
$\beta_\lambda(e^{bz}-1)/b$ and $(e^{bz}-1)/b$.
For $z\le0$, applying the same argument to $-z$ bounds
$-\log_{a,b,\lambda}(e^z)$ between
$\alpha_\lambda(e^{-pz}-1)/p$ and $(e^{-pz}-1)/p$.
Inversion proves the brackets. Taking the largest possible endpoint
ratio proves their width bounds. Explicitly, for $y>0$ the root satisfies $1+by\le e^{bz^*}\le1+by/\beta_\lambda$, so the width is $b^{-1}\ln\frac{1+by/\beta_\lambda}{1+by}\le-\ln\beta_\lambda/b$; the case $y<0$ follows by exchanging $(b,\beta_\lambda)$ with $(p,\alpha_\lambda)$.\hfill$\square$

\paragraph{Proof of Lemma~\ref{lem:exp-residual}.}\label{prf:lem:exp-residual}
Because $G_{a,b,\lambda}(z^*)=0$, the mean-value theorem gives $g(\hat z)=G_{a,b,\lambda}(\hat z)-G_{a,b,\lambda}(z^*)=G_{a,b,\lambda}'(\xi)(\hat z-z^*)$ for some $\xi$ between $\hat z$ and $z^*$, hence $\xi\in J$. By Proposition~\ref{prop:exp-solver-shape}(i), $G_{a,b,\lambda}'(\xi)=\omega_{a,b,\lambda}(e^\xi)\in[s_J,S_J]$, which proves~\eqref{eq:exp-residual}. By part~(ii) of the same proposition, $t\mapsto\omega_{a,b,\lambda}(e^t)=G_{a,b,\lambda}'(t)$ decreases for $t<z_q$ and increases for $t>z_q$; its minimum over $J$ is therefore attained at the right endpoint, at the left endpoint, or at $z_q$, as stated.\hfill$\square$

\paragraph{Proof of Proposition~\ref{prop:exp-newton}.}\label{prf:prop:exp-newton}
Monotonicity of $G_{a,b,\lambda}$ preserves the root in the bracket. Its midpoint has
error at most half its width. At the residual test both $z$ and $z^*$ belong to
the updated bracket, so Lemma~\ref{lem:exp-residual} gives
$|z-z^*|\le |g|/s$. The initial midpoint, and every later trial, lie in the
central fraction $[\ell+\theta(u-\ell),u-\theta(u-\ell)]$ of the bracket
on which they were selected. Whichever endpoint is replaced, the retained
width is at most $(1-\theta)(u-\ell)$. Iterating this inequality proves (ii).
For (iii), Taylor expansion between $z_k$ and $z^*$ gives
$|e_{k+1}|=|G''(\xi)|e_k^2/(2G'(z_k))$.
Proposition~\ref{prop:exp-solver-shape} gives $|G''|\le cG'$ and
$|\mathrm d\ln G'/\mathrm dz|\le c$, hence
$G'(\xi)\le G'(z_k)e^{c|e_k|}$.\hfill$\square$

\paragraph{Proof of Corollary~\ref{cor:exp-monotone-ab}.}\label{prf:cor:exp-monotone-ab}
$J_s(x)>0$ for $x\ne1$, as follows by differentiating
$\log_{s,0}(x)=\int_0^{\ln x}e^{su}\,\mathrm du$.
Hence $\log_{0,b}(x)>\log_{a,0}(x)$ for $a<b$, and the third partial in
\eqref{eq:dlog-dab} is positive. Equation~\eqref{eq:du_da} proves
scale monotonicity.
At $\lambda=1$ the unscaled divided-difference representation proves
positive exponent partials of the link.
For general scale, the mixture-weight derivative terms can reverse their
signs. For example, at fixed $x\ne1$, as $\ell\to+\infty$,
$\partial_a \log_{a,b,\lambda}/\alpha_\lambda
=J_a+\beta_\lambda(1/a+\ell)(\log_{a,0}-\log_{0,b})$ becomes negative.\hfill$\square$

\subsection*{Section~\ref{sec:mirror-EG}: Mirror Descent and Exponentiated Gradient (EG) Update}
\paragraph{Proof of Theorem~\ref{thm:Euler-simplex-wellposed}.}\label{prf:thm:Euler-simplex-wellposed}
$\Phi_t$ is continuous and strictly decreasing because each $\exp_{a,b,\lambda}(s_{i,t}-\nu)$ is strictly decreasing in $\nu$. In the full-range regime, $\exp_{a,b,\lambda}(y)\to\infty$ as $y\to\infty$ and $\exp_{a,b,\lambda}(y)\to0$ as $y\to-\infty$. Hence $\Phi_t(\nu)\to\infty$ as $\nu\to-\infty$ and $\Phi_t(\nu)\to-1$ as $\nu\to\infty$. The intermediate value theorem and strict monotonicity give existence and uniqueness. Positivity is automatic and Eq.~\eqref{eq:lambda-root} enforces unit sum. The resulting point satisfies the equality-constrained first-order conditions of a strictly convex objective, so it is the unique minimizer of the stated mirror subproblem. For all full-range exponents, define the potential by its integral if the displayed antiderivative has an exceptional exponent $-1$.\hfill$\square$

\paragraph{Proof of Theorem~\ref{thm:Euler-MD-convergence}.}\label{prf:thm:Euler-MD-convergence}
For $0<w\le1$ in the stated parameter range,
\[
\phi_{a,b,\lambda}''(w)=(\log_{a,b,\lambda})'(w)
=\alpha_\lambda w^{a-1}+\beta_\lambda w^{b-1}\ge1,
\]
because $\alpha_\lambda+\beta_\lambda=1$ and $w^{b-1}\ge1$, $w^{a-1}\ge1$ on $(0,1]$ under
$-1<a<0<\rev{b\le 1}$. Thus $F_{a,b,\lambda}$ is $1$-strongly convex on
$\overline{\Delta_N}$ in the Euclidean norm. The standard three-point
identity for the exact mirror step gives, for every $t$,
\[
\eta\langle\bg_t,\bw_t-\bw^*\rangle
\le D_{F_{a,b,\lambda}}(\bw^*\Vert\bw_t)-D_{F_{a,b,\lambda}}(\bw^*\Vert\bw_{t+1})+\frac{\eta^2}{2}\|\bg_t\|_2^2.
\]
Since $L$ is convex on $\Delta_N\supseteq[\bw_t,\bw^*]$ for every
$t$, the segment joining the interior iterate $\bw_t$ and the
(possibly boundary) comparator $\bw^*$ lies in the domain of
convexity, and the standard subgradient inequality gives
$L(\bw_t)-L(\bw^*)\le\langle\bg_t,\bw_t-\bw^*\rangle$. Summing over
$t$, telescoping the Bregman terms, using $\|\bg_t\|_2\le G$,
dividing by $\eta T$, and applying Jensen's inequality (using
convexity of $L$ on $\Delta_N$, which contains $\bar{\bw}_T$) yields
Eq.~\eqref{eq:Euler-MD-bound}; optimizing the bound over $\eta$ gives
Eq.~\eqref{eq:Euler-MD-rate}.\hfill$\square$

\paragraph{Proof of Theorem~\ref{thm:sharp-simplex-geometry}.}\label{prf:thm:sharp-simplex-geometry}
The formula for $c$ follows from the minimum location $r/\lambda$.
Since $h(x)\ge c/x$, the potential
$F-c\sum_i x_i\ln x_i$ is convex; this proves the KL comparison
and the first sufficient modulus.
For a tangent vector $\bz$, $\1^T\bz=0$, Cauchy--Schwarz gives
\[
 \sum_i z_i^2w_i^{b-1}
 \ge\frac{\|\bz\|_1^2}{\sum_iw_i^{1-b}}
 \ge N^{-b}\|\bz\|_1^2.
\]
If $\bz\ne0$, its positive and negative parts each have mass
$d=\|\bz\|_1/2$. Write $P=\sum_{z_i>0}w_i$ and
$Q=\sum_{z_i<0}w_i$, so $P+Q\le1$. Then
\[
 \sum_i z_i^2w_i^{a-1}
 \ge d^2(P^{a-1}+Q^{a-1})
 \ge 2^{|a|}\|\bz\|_1^2.
\]
The first inequality uses $\sum_{i\in S}w_i^{1-a}\le
(\sum_{i\in S}w_i)^{1-a}$, and the last minimum is at $P=Q=1/2$.
Multiply by $\alpha_\lambda,\beta_\lambda$ and add.
Integrating along simplex segments proves strong convexity, with
continuity extending the first argument to boundary comparators.
The mixture exceeds $N^{-b}$ because $\alpha_\lambda>0$.

At the uniform center the affine Bregman term vanishes.
Symmetry and convexity minimize $F$ at $\1/N$ and maximize it at
vertices. Evaluate $(N-1)\phi(0)-N\phi(1/N)$ using the power
primitive mixture. This gives the range and its limits.
The mixture part of $\mu$ tends to one, and $c\le1$, proving the
modulus limit.\hfill$\square$

\paragraph{Proof of Theorem~\ref{prop:weight-floor}.}\label{prf:prop:weight-floor}
Subtracting two coordinates of Eq.~\eqref{eq:Euler-simplex-KKT} eliminates $\nu_t$. Summing over the steps yields
\[
\log_{a,b,\lambda}(w_{i,T+1})\ge
\log_{a,b,\lambda}(w_{j,T+1})-C_1-2GA_T.
\]
Some coordinate $j$ is at least $1/N$. Monotonicity of the logarithm and its inverse gives Eq.~\eqref{eq:weight-floor}. The inverse left-tail asymptotic gives Eq.~\eqref{eq:weight-floor-asymptotic}. For EG, use the same argument with $\ln$. From uniform initialization and $\bg_t=(G,-G,\ldots,-G)^T$, its first coordinate is $[1+(N-1)e^{2GA_T}]^{-1}$, demonstrating exponential decay.\hfill$\square$

\subsection*{Section~\ref{sec:FTRL}: Certified normalization and scale-free adaptive Euler FTRL}
\paragraph{Proof of Lemma~\ref{lem:certified-norm}.}\label{prf:lem:certified-norm}
\begingroup
On the bracket $\nu\ge\max_i y_i$, each $w_i\le1$. On $(0,1]$, $h_{a,b,\lambda}(w_i)\ge1$, so coordinate $i$ is $D_i^{-1}$-Lipschitz in $\nu$ on its positive branch. At the sparse boundary the coordinate is continuous and constant on its zero branch, so the same bound holds globally on the bracket. Thus $S$ is $L_D$-Lipschitz. For an index attaining $\nu_{\rm hi}=\max_i(y_i-D_i\log_{a,b,\lambda}(1/N))$, its weight at that endpoint is $1/N$ and is at least that at any lower multiplier. For a tightened uniform bracket this remains true. Therefore $w_j\ge1/N$, and (i) follows.

For (ii), the derivative of the weighted objective at positive coordinate $i$, plus $\nu$, is $D_i\log_{a,b,\lambda}(p_i)-y_i+\nu$. This is zero except at $j$. At any zero coordinate on the sparse boundary it is nonnegative and is canceled by a nonpositive normal-cone component. The remaining vector has only one possibly nonzero entry, and subtracting its midrange by adjusting the common multiplier gives the claimed residual. For (iii), apply the mean-value theorem on $[1/(2N),1]$ and use $h_{a,b,\lambda}(x)\le x^{a-1}$. The central-interval safeguard reduces the bracket by at least the factor $1-\theta$ on each nonterminal update. This proves the count.
\endgroup\hfill$\square$

\paragraph{Proof of Theorem~\ref{thm:Euler-FTRL-regret}.}\label{prf:thm:Euler-FTRL-regret}
Let $\mathcal R_{a,b,\lambda}=F_{a,b,\lambda}-F_{a,b,\lambda}(\1/N)\in[0,R]$ and
$V_\eta(\bh)=\min_{\Delta_N}\{\langle\bh,\bw\rangle+\mathcal R_{a,b,\lambda}(\bw)/\eta\}$
with minimizer $\bw_\eta(\bh)$. Strong convexity gives
\begin{equation}
 V_\eta(\bh+\bar\bg)\ge V_\eta(\bh)+\langle\bar\bg,\bw_\eta(\bh)\rangle
 -\frac{\eta}{2\mu}\|\bar\bg\|_\infty^2,
 \label{eq:audit-value-smooth}
\end{equation}
because the old objective grows by at least
$\mu\|\bw-\bw_\eta(\bh)\|_1^2/(2\eta)$. Telescope from $V_\eta(\0)=0$ and use
$V_\eta(\bh_T)\le\langle\bh_T,\bx\rangle+R/\eta$.\hfill$\square$

\paragraph{Proof of Theorem~\ref{thm:Euler-adaptive}.}\label{prf:thm:Euler-adaptive}
Write $V_t=V_{\eta_t}$. As $\mathcal R_{a,b,\lambda}\ge0$ and $\eta_t$ is
{nonincreasing, $V_t\ge V_{t-1}$ pointwise. When $\delta+S_{t-1}=0$, use the zero-regularization value $V_{\infty}(\bh)=\min_{\bw\in\Delta_N}\langle\bh,\bw\rangle$. The centered history is then zero and the uniform decision is among its minimizers. On that round use the range cap alone. The stability gap}
$d_t=\langle\bar\bg_t,\bw_t\rangle-[V_{t-1}(\bh_{t-1}+\bar\bg_t)-V_{t-1}(\bh_{t-1})]$
is at most $\eta_{t-1}\sigma_t^2/(2\mu)$ by Eq.~\eqref{eq:audit-value-smooth},
and at most $2\sigma_t$ because the value increment is at least
$\min_i\bar g_{i,t}$ while $\langle\bar\bg_t,\bw_t\rangle\le\max_i\bar g_{i,t}$.
Telescoping and the comparator bound give
Eq.~\eqref{eq:audit-adaptive-predictable}. With $k=\eta_0/(2\mu)$, for $A>0$,
$x\ge0$,
\begin{equation}
 \min\{kx^2/\sqrt A,2x\}\le\bigl(k+\sqrt{k^2+4}\bigr)(\sqrt{A+x^2}-\sqrt A):
 \label{eq:capped-stability-telescope}
\end{equation}
{For $x>0$, divide the left-hand side by $\sqrt{A+x^2}-\sqrt A$ and set $r=x/\sqrt A$. The resulting ratio is the minimum of the increasing $k(\sqrt{1+r^2}+1)$ and the decreasing $2(\sqrt{1+r^2}+1)/r$. They intersect at $r=2/k$, where the ratio is $k+\sqrt{k^2+4}$. For $x=0$, both sides vanish. For $A=0$ the capped term is $2x$, which obeys the same inequality. Taking $A=\delta+S_{t-1}$ and $x=\sigma_t$ makes the square-root increments telescope, proving Eq.~\eqref{eq:Euler-adaptive-bound}.} For $\delta=0$,
scaling gradients by $c>0$ multiplies the whole FTRL objective by $c$.\hfill$\square$

\paragraph{Proof of Proposition~\ref{prop:sparse-FTRL}.}\label{prf:prop:sparse-FTRL}
KKT with multipliers $\nu_i\ge0$ for $w_i\ge0$ gives
$\log_{0,b}(w_i)=-\eta h_i-\nu+\nu_i$, $\nu_iw_i=0$, hence Eq.~\eqref{eq:sparse-FTRL}; the
normalization sum is strictly decreasing where positive. Subtracting the power
base of a minimum-loss coordinate gives (i). {For (ii), on the simplex $F_{0,b}$ is $1/(1+b)$ times the negative Tsallis entropy of order $1+b$, plus a constant, using the convention $H_\alpha(\bp)=(1-\sum_i p_i^\alpha)/(\alpha-1)$.} For (iii),
$\sum_iz_i^2w_i^{b-1}\ge\|\bz\|_1^2/\sum_iw_i^{1-b}\ge N^{-b}\|\bz\|_1^2$; the
regret proofs use only the modulus, the range and minimality at $\1/N$.\hfill$\square$

\subsection*{Section~\ref{sec:recent-euler}: Dual averaging, schedule-free FTRL, and sparse learning}
\paragraph{Proof of Proposition~\ref{prop:stage2-rda-regret}.}\label{prf:prop:stage2-rda-regret}
$h_{a,b,\lambda}$ is nonincreasing, so $H_{a,b,\lambda,\varrho}$ is $h_{\min}$-strongly convex in
$\ell_2$ on $K_{\max}$. Let $f_t(\bw)=\omega_t(\langle\bg_t,\bw\rangle+\rho_1\|\bw\|_1)$,
which is $J_t$-Lipschitz, and $Q_t=\sum_{s\le t}f_s+\beta_tH_{a,b,\lambda,\varrho}$; $\bw_t$ minimizes
$Q_{t-1}$. Strong convexity gives the one-step stability term
$J_t^2/(2\beta_{t-1}h_{\min})$; since $H_{a,b,\lambda,\varrho}\ge0$ and $\beta_t$ is nondecreasing,
$\min Q_t\ge\min(Q_{t-1}+f_t)$. Telescope from $\min Q_0=0$ and use convexity.\hfill$\square$

\paragraph{Proof of Theorem~\ref{thm:new-sf}.}\label{prf:thm:new-sf}
It suffices to show
$t[L(\bx_t)-L^*]-(t-1)[L(\bx_{t-1})-L^*]\le\langle\bar\bg_t,\bz_t-\bp^*\rangle$
with $\bar\bg_t=\mathbb E[\bg_t\mid\mathcal F_{t-1}]$, since the left side
telescopes. For $t=1$ it is the subgradient inequality. For $t\ge2$ the points
$\bx_{t-1},\bx_t,\by_t,\bz_t$ lie on the line
$\bx(s)=\bx_{t-1}+s(\bx_t-\bx_{t-1})$ at $s=0,1,\sigma,t$ with
$\sigma=\vartheta+(1-\vartheta)t\in[1,t]$. With $f(s)=L(\bx(s))$ and the
subgradient $\kappa=\langle\bar\bg_t,\bx_t-\bx_{t-1}\rangle$ at $\sigma$, the
subgradient inequality at $\by_t$ gives
$\langle\bar\bg_t,\bz_t-\bp^*\rangle\ge f(\sigma)-L^*+(t-\sigma)\kappa$, while
convexity gives $f(0)\ge f(\sigma)-\sigma\kappa$ and
$f(1)\le(1-1/\sigma)f(0)+f(\sigma)/\sigma$; since $1-t/\sigma\le0$,
$tf(1)-(t-1)f(0)\le f(\sigma)+(t-\sigma)\kappa$. The Euler regret bound is
pathwise; substitute Theorem~\ref{thm:Euler-adaptive}.\hfill$\square$

\subsection*{Section~\ref{sec:new-optimism}: Optimistic Euler learning and changing geometries}

\paragraph{Proof of Lemma~\ref{lem:bandit-local-norm}.}\label{prf:lem:bandit-local-norm}
The left side is at most $\sup_{\bx>0}\{\langle\bg,\bw_t-\bx\rangle-D_{F_{a,b,\lambda}}(\bx\Vert\bw_t)/\eta\}$. Because $a<0$, the link range is unbounded below (Lemma~\ref{lem:audit-domains}), so the supremum is attained at the unconstrained proposal $\tilde\bw$ with $\log_{a,b,\lambda}(\tilde\bw)=\log_{a,b,\lambda}(\bw_t)-\eta\bg$; since $\bg\ge0$ the proposal stays in range also when $b=0$. The three-point identity gives the value $D_{F_{a,b,\lambda}}(\bw_t\Vert\tilde\bw)/\eta$. In dual coordinates $z_i=\log_{a,b,\lambda}(w_{t,i})$ this divergence is $\sum_i[\phi^*(z_i-\eta g_i)-\phi^*(z_i)+\eta g_i(\phi^*)'(z_i)]$ with $(\phi^*)''=m_{a,b,\lambda}\circ\exp_{a,b,\lambda}$ (Proposition~\ref{prop:euler-tails}); it therefore equals $\frac{\eta^2}{2}\sum_ig_i^2m_{a,b,\lambda}(\xi_i)$ for some $\xi_i\in[\tilde w_i,w_{t,i}]$. For $b\le1$, $h_{a,b,\lambda}$ is nonincreasing, so $m_{a,b,\lambda}(\xi_i)\le m_{a,b,\lambda}(w_{t,i})$. This is the stability argument behind Exp3 and Tsallis-INF~\cite{Audibert2009,Zimmert2021,LattimoreSzepesvari2020}; it uses only full range below and monotone mobility.\hfill$\square$

\paragraph{Proof of Proposition~\ref{prop:euler-bandit}.}\label{prf:prop:euler-bandit}
With constant $\eta$ and uniform initialization, FTRL coincides with exact simplex Euler mirror descent (Lemma~\ref{lem:audit-exact-ftrl}), whose standard decomposition gives
$\sum_t\langle\hat\bg_t,\bw_t-\bu\rangle\le D_{F_{a,b,\lambda}}(\bu\Vert\1/N)/\eta+\sum_t[\langle\hat\bg_t,\bw_t-\bw_{t+1}\rangle-D_{F_{a,b,\lambda}}(\bw_{t+1}\Vert\bw_t)/\eta]$, with $D_{F_{a,b,\lambda}}(\bu\Vert\1/N)\le R_{a,b,\lambda}(N)$ (Theorem~\ref{thm:sharp-simplex-geometry}). Lemma~\ref{lem:bandit-local-norm} bounds each stability term by $\frac\eta2\sum_i\hat g_{t,i}^2m_{a,b,\lambda}(w_{t,i})$. Since $\bw_t$ is fixed given the past, unbiasedness gives $\mathbb E_t\langle\hat\bg_t,\bw_t-\bu\rangle=\langle\boldsymbol\ell_t,\bw_t-\bu\rangle$, and Eq.~\eqref{eq:bandit-m-over-x} gives $\mathbb E_t[\hat g_{t,i}^2]m_{a,b,\lambda}(w_{t,i})=\ell_{t,i}^2/\omega_{a,b,\lambda}(w_{t,i})$. Eq.~\eqref{eq:bandit-variance} gives the last inequality. For $a=-1$, $\langle\boldsymbol\ell_t,\tilde\bu_\gamma-\bu\rangle\le\gamma$.\hfill$\square$

\paragraph{Proof of Theorem~\ref{thm:new-oomd}.}\label{prf:thm:new-oomd}
With $D=D_{F_{a,b,\lambda}}$, the three-point inequality for the correction gives
$\eta\langle\bg_t,\bz_{t+1}-\bp^*\rangle\le D(\bp^*\Vert\bz_t)-D(\bp^*\Vert\bz_{t+1})-D(\bz_{t+1}\Vert\bz_t)$,
and the prediction step tested at $\bz_{t+1}$ gives
$\eta\langle\widehat\bg_t,\bp_t-\bz_{t+1}\rangle\le D(\bz_{t+1}\Vert\bz_t)-D(\bz_{t+1}\Vert\bp_t)-D(\bp_t\Vert\bz_t)$.
Adding them and $\eta\langle\bg_t-\widehat\bg_t,\bp_t-\bz_{t+1}\rangle$, Young's
inequality against $-\frac\mu2\|\bp_t-\bz_{t+1}\|_1^2$ bounds the remainder by
$\eta^2\|\bg_t-\widehat\bg_t\|_\infty^2/(2\mu)$. Telescope.\hfill$\square$

\paragraph{Proof of Theorem~\ref{thm:new-adaptive-oftrl}.}\label{prf:thm:new-adaptive-oftrl}
Let $V_t(\bh)=\min_{\Delta_N}\{\langle\bh,\bp\rangle+\mathcal R_{a,b,\lambda}(\bp)/\eta_t\}$,
concave with supergradient $\bp^{(t)}(\bh)$, and satisfying the smoothness
bound~\eqref{eq:audit-value-smooth}; $V_t\ge V_{t-1}$. The comparator bound and
telescoping give regret at most $R/\eta_T+\sum_td_t$ with
$d_t=\langle\bg_t,\bp_t\rangle-[V_t(\bL_{t-1}+\bg_t)-V_t(\bL_{t-1})]$. With
$\hat\bh=\bL_{t-1}+\widehat\bg_t$ and $\boldsymbol\epsilon_t=\bg_t-\widehat\bg_t$,
concavity at $\hat\bh$ and smoothness give
$d_t\le\eta_t\|\boldsymbol\epsilon_t\|_\infty^2/(2\mu)$, while concavity at
$\bL_{t-1}+\bg_t$ gives
$d_t\le\langle\boldsymbol\epsilon_t,\bp_t-\bp^{(t)}(\bL_{t-1}+\bg_t)\rangle\le2\widetilde\sigma_t$;
centering is valid because $V_t(\bh+c\1)=V_t(\bh)+c$. Sum with
inequality~\eqref{eq:capped-stability-telescope}.\hfill$\square$

\subsection*{Section~\ref{sec:phase2-games}: Learning in games: Euler self-play, Mirror-Prox, and equilibrium certificates}
\paragraph{Proof of Theorem~\ref{thm:anchored-mp}.}\label{prf:thm:anchored-mp}
With $r(\bz)=\tau D(\bz\Vert\bq)$, affine equivalence to $\tau F$ gives
$\langle\nabla r(\bz),\bz-\bw\rangle=r(\bz)-r(\bw)+\tau D(\bw\Vert\bz)$.
{For existence, apply Brouwer's theorem on $K$ to the continuous single-valued map $\bz\mapsto\argmin_{\bw\in K}\{\tfrac12\|\bw-\bz\|_2^2+\langle\mathcal G(\bz),\bw\rangle+r(\bw)\}$. At any fixed point the minimizing profile is interior, since transferring a small mass to a zero coordinate has directional derivative $-\infty$ from the anchor regularizer. The fixed-point condition then gives the stated variational inequality,} and uniqueness from monotonicity, which gives
$\tau[D(\bp\Vert\bp')+D(\bp'\Vert\bp)]\le0$ for two solutions. Adding the
optimality conditions of the two steps, the anchor terms contribute
$\eta[r(\widetilde\bz_t)-r(\bw)+\tau D(\bw\Vert\bz_{t+1})+\tau D(\bz_{t+1}\Vert\widetilde\bz_t)]$,
and the Lipschitz cross term is absorbed as in Theorem~\ref{thm:p2-mirrorprox}, so
$\eta[\langle\mathcal G(\widetilde\bz_t),\widetilde\bz_t-\bw\rangle+r(\widetilde\bz_t)-r(\bw)]
+(1+\eta\tau)D(\bw\Vert\bz_{t+1})\le D(\bw\Vert\bz_t)$. At $\bw=\bp^*_\tau$ the
bracket is nonnegative. For the gaps use the variational inequality, the
identity above, monotonicity, and $D\ge\frac12\|\cdot\|_2^2$.\hfill$\square$

\subsection*{Section~\ref{sec:phase2-barrier}: Implicit bias, damping, and approximation robustness}
\paragraph{Proof of Proposition~\ref{prop:implicit-md}.}\label{prf:prop:implicit-md}
$h_{a,b,\lambda}$ is nonincreasing, so $D(\bw'\Vert\bw)\ge\frac12h_{a,b,\lambda}(w_{\max})\|\bw'-\bw\|_2^2$
on the box and $L$ is $(1/\eta)$-smooth relative to $F_{a,b,\lambda}$. The update
is the exact separable prox step on the box; the three-point inequality,
relative smoothness and convexity give
$\eta L(\bw_{t+1})\le\eta L(\bar\bw)+D(\bar\bw\Vert\bw_t)-D(\bar\bw\Vert\bw_{t+1})$.
Take $\bar\bw=\bw_t$ for monotonicity and sum for (i); (ii) telescopes the
uncapped dual updates, and (iii) follows as before.\hfill$\square$

\paragraph{Proof of Proposition~\ref{prop:decay-vs-damping}.}\label{prf:prop:decay-vs-damping}
(i) $\dot w_i=-m_{a,b,\lambda}(w_i)\partial_iL-\kappa w_i\le-\kappa w_i$. (ii) With
$z_i=\log_{a,b,\lambda}(w_i)-\log_{a,b,\lambda}(w_{0,i})$, $\dot z_i=-\partial_iL-\kappa z_i$, so
$|z_i|\le G/\kappa$; $\exp_{a,b,\lambda}$ is defined on all of $\mathbb R$. (iii)
$\Phi=L+\kappa D(\cdot\Vert\bw_0)$ satisfies
$\Phi(\bu)\ge\Phi(\bw)+\langle\nabla\Phi(\bw),\bu-\bw\rangle+\kappa D(\bu\Vert\bw)$, so
$\frac{\mathrm d}{\mathrm dt}D(\bw^*_\kappa\Vert\bw)\le-\kappa D(\bw^*_\kappa\Vert\bw)$.\hfill$\square$

\subsection*{Section~\ref{sec:design-geometry}: Euler potentials as designed Bregman geometries and comparison with Adam, Shampoo and Muon}
\paragraph{Proof of Theorem~\ref{thm:design-criteria}.}\label{prf:thm:design-criteria}
(i) Lemma~\ref{lem:audit-domains}. (ii)--(iii) are the two power terms of
$h_{a,b,\lambda}$, each nonincreasing for $b\le1$. (iv) $\ln$ of a sum of two
exponentials in $t$ is convex, with derivative $-\varepsilon_{a,b,\lambda}$
(Proposition~\ref{prop:elasticity}). (v) $\mathrm d\ln x/\mathrm dz=1/\omega_{a,b,\lambda}(x)$
and $\mathrm d\ln h/\mathrm dz=-\varepsilon/\omega$; use $\omega\ge\omega_{\min}$
and $\varepsilon\le1-a$. (vi)
$|h'|/h^{3/2}=\varepsilon/\sqrt{x\omega}$.\hfill$\square$

\section{Special cases and related deformed logarithms}
\label{app:special-cases}
This appendix records, for reference, the classical deformed logarithms contained in the Euler family, the entropies they generate, and three related two-parameter logarithms. None of this material is needed for the algorithms of the paper.

The Euler logarithm is a general form of deformed logarithm and associated entropy. Two-parameter deformations of this kind, and the entropies they generate, have been studied extensively in information theory and generalized statistical mechanics \cite{taneja1989,kaniadakiseditorial2004,furuichi2010,wada2010,Cantruk2018}. In particular cases it gives the following well-known deformed logarithms \cite{Cichocki2025}; Table~\ref{tab:special-cases} summarizes them with their inverses.

\begin{itemize}
\item For $a \rightarrow 0$ and $b\rightarrow 0$, we obtain the classical natural logarithm $\log_{0,0}(x)= \ln(x)$ and Boltzmann-Gibbs-Shannon (BGS) entropy $S^{BGS}(\bp) = \sum_i p_i \ln (1/p_i)$ \cite{shannon1948,tsallis1988,kaniadakis2002}.

\item For $a=r-\kappa$ and $b=r+\kappa$, with $\kappa>0$ and $-\kappa<r\le\kappa$, we obtain the KLS (Kaniadakis--Lissia--Scarfone) logarithm \cite{kaniadakis2004,kaniadakis2005}:
\begin{equation}
\log^{KLS}_{\kappa,r} (x)  = x^r \frac{x^{\kappa} - x^{-\kappa}}{2 \kappa}.
\end{equation}

\item For $a=1-\max(\sigma,\sigma^{-1})$ and $b=1-\min(\sigma,\sigma^{-1})$, we obtain the Abe logarithm \cite{abe1997}:
\begin{equation}
\log^{Abe}_{\sigma} (x) = \frac{x^{1-\sigma} - x^{1-\sigma^{-1}}}{\sigma^{-1} - \sigma}, \quad 1/2 < \sigma < 2, \; \sigma \neq 1,
\end{equation}
and the associated Abe entropy \cite{abe1997}:
\begin{equation}
S^{Abe}(\bp) = \sum_i p_i \frac{p_i^{\sigma-1} - p_i^{\sigma^{-1}-1}}{\sigma^{-1} - \sigma}  = \sum_i p_i \log^{Abe}_{\sigma}(1/p_i)
\;=\;\sum_i \frac{p_i^{\sigma}-p_i^{\sigma^{-1}}}{\sigma^{-1}-\sigma} .
\end{equation}
These are the exponents for which the trace form of the Euler kernel,
\begin{equation}
S^{\rm tr}_{a,b}(\bp)=\sum_i p_i\log_{a,b}(1/p_i)=\sum_i\frac{p_i^{\,1-a}-p_i^{\,1-b}}{a-b},
\label{eq:traceform-general}
\end{equation}
returns Abe's entropy, in the same way that $(a,b)=(\min(0,1-q),\max(0,1-q))$ returns the Tsallis entropy $\sum_i(p_i^{q}-p_i)/(1-q)$ and {the general pair gives the Borges--Roditi trace entropy}. The complementary ordered pair $a=\min(\sigma,\sigma^{-1})-1$, $b=\max(\sigma,\sigma^{-1})-1$ generates the dual logarithm $x\mapsto-\log_{1-\max(\sigma,\sigma^{-1}),1-\min(\sigma,\sigma^{-1})}(1/x)$ and the exponents $2-\sigma^{-1}$, $2-\sigma$ in the trace form.
The point $\sigma=1$ is a removable singularity of the displayed Abe formula; its continuous limit is the ordinary logarithm $\ln x$.

\item For $a=-\alpha<0=b$, we obtain the Amari $\alpha$-logarithm defined as \cite{Amari2009, Amari-PAN}:
\begin{equation}
\log_{-\alpha,0}(x)= \log_{\alpha}^A (x)= \frac{1-x^{-\alpha}}{\alpha},
\end{equation}
which is equivalent to Tsallis $q$-logarithm for $\alpha =q-1$. Alternatively, for $(a,b)=(\min(0,1-q),\max(0,1-q))$, we obtain the Tsallis $q$-logarithm $\log_q^T(x)=(x^{1-q}-1)/(1-q)$ and the Havrda--Charvat \cite{harvda1967} and Tsallis entropy \cite{tsallis1988,Tsallis1994}.

\item For $a=-\kappa$ and $b=\kappa$, with $\kappa>0$, we obtain the $\kappa$-logarithm $\log_{-|\kappa|,|\kappa|}(x)= \log_{\kappa}^K(x)$ and the Kaniadakis entropy \cite{kaniadakis2002}.

\item For $a=\min(2\gamma,-\gamma)$ and $b=\max(2\gamma,-\gamma)$, we obtain the so-called $\gamma$-logarithm $\log_{\gamma}(x)= \frac{x^{2 \gamma}- x^{-\gamma}}{3  \gamma}$ for $x >0$ and $-1/2 < \gamma < 1/2$ \cite{kaniadakis2005}.

\end{itemize}

\subsection*{Related two-parameter logarithms}
{The examples below distinguish composition from rescaling. Composing two Tsallis links through the ordinary exponential gives the Schw\"ammle--Tsallis logarithm, which can be written as follows \cite{schwammle2007}.}
\begin{equation}
 {\log^{ST}_{q,q'}(x)=\log^T_{q'}\!\left(\exp(\log^T_q(x))\right).}
\end{equation}
\begin{align}
\log_{q,q'}^{ST}(x)
&=\frac{1}{1-q'}\left[\exp\left((1-q')\log^T_q(x)\right)-1\right] \nonumber\\
&=\frac{1}{1-q'}\left[\exp\left(\frac{1-q'}{1-q}(x^{1-q}-1)\right)-1\right], \nonumber\\
&\hspace{2em} x>0,\quad q\neq1,\quad q'\neq1,\quad \text{typically }q>1\text{ and }q'<1.
\label{STlog2}
\end{align}

Its inverse function can be easily computed as a deformed $q,q'$-exponential:
\begin{equation}
\exp_{q,q'}^{ST} (x) = \left[ 1+ \frac{1-q}{1-q'} \ln\big(1+(1-q')x\big)\right]^{1/(1-q)}.
\label{STexp}
\end{equation}
For the displayed non-limit inverse, real-valued evaluation additionally requires $1+(1-q')x>0$ and $1+\frac{1-q}{1-q'}\ln(1+(1-q')x)>0$; the cases $q=1$ or $q'=1$ are defined by their continuous limits.

{The scaled Kaniadakis--Scarfone logarithm is exactly the symmetric-exponent slice $\log_{-|\kappa|,|\kappa|,\lambda}$ of the normalized Euler family \cite{kaniadakis2009,kaniadakis2017}. Its explicit form is}
\begin{align}
\log^{KS}_{\kappa,\lambda} (x) &= \frac{\lambda^{\kappa}(x^{\kappa}-1) - \lambda^{-\kappa}(x^{-\kappa}-1)}{\kappa (\lambda^{\kappa} + \lambda^{-\kappa})} \nonumber \\
& = \frac{2}{ \lambda^{\kappa} + \lambda^{-\kappa}} \left[ \log^K_{\kappa} (\lambda x)  - \log^K_{\kappa} (\lambda) \right], \quad x>0, \; \lambda > 0, \; 0<|\kappa|<1.
\label{defgenlogK}
\end{align}
{Equivalently,}
\begin{equation}
\log^{KS}_{\kappa,\lambda} (x) =  A \log_{-|\kappa|,|\kappa|} (\lambda x) + B,
\end{equation}
where this transformation involves three fundamental operations:
\begin{itemize}
 \item \textbf{Substitution} (linear scaling of the argument): $x \rightarrow \lambda x$.
 \item \textbf{Scaling} (multiplication by factor): $A = \frac{2}{\lambda^{\kappa} + \lambda^{-\kappa}}$.
 \item \textbf{Shift} (addition of constant): $B = \frac{\lambda^{-\kappa} - \lambda^{\kappa}}{\kappa(\lambda^{\kappa} + \lambda^{-\kappa})}$.
\end{itemize}

{The normalization factor also has the form}
\begin{equation}
\sqrt{1+\kappa^2 \bigl[\log^K_{\kappa}(\lambda)\bigr]^{2}} = \frac{\lambda^{\kappa} + \lambda^{-\kappa}}{2}.
\end{equation}
\rev{Hence we have:}
\begin{equation}
\log^{KS}_{\kappa,\lambda} (x) = \frac{\log^K_{\kappa} (\lambda x) - \log^K_{\kappa}(\lambda)}{\sqrt{1+\kappa^2 \bigl[\log^K_{\kappa}(\lambda)\bigr]^{2}}}.
\end{equation}

{Solving for the positive argument gives the inverse}
\begin{equation}
\exp^{KS}_{\kappa,\lambda} (x) = \frac{1}{\lambda} \exp^K_{\kappa} \left( x \sqrt{1+\kappa^2 \bigl[\log^K_{\kappa}(\lambda)\bigr]^{2}} + \log_{\kappa}^K (\lambda) \right).
\end{equation}
For $\lambda=1$ these functions reduce to the Kaniadakis pair. Since the scaled family is even in $\kappa$, take $\kappa>0$: as $\lambda\to\infty$, its logarithm tends to $(x^\kappa-1)/\kappa$ (Tsallis $q=1-\kappa$); as $\lambda\downarrow0$, it tends to $(1-x^{-\kappa})/\kappa$ (Tsallis $q=1+\kappa$). The corresponding inverse limits hold for arguments in the open range of the limiting logarithm. The case $\kappa=0$ is defined by continuity and gives $\ln x$.

{The two-parameter Tempesta logarithm has the form} \cite{Tempesta2015,tempesta2016}:
\begin{equation}
\log^{\text{Te}}_{\alpha,\kappa}(x) = \frac{1}{(1+\alpha)\kappa} \left[  \alpha x^{\kappa} -  x^{-\kappa} + 1- \alpha \right], \quad x>0, \; \alpha\ge0,\quad \kappa\ne0,
\label{entropyTe1}
\end{equation}
{For $\kappa>0$ and $\alpha>0$, setting $\lambda=\alpha^{1/(2\kappa)}$ gives $\log^{\rm Te}_{\alpha,\kappa}=\log_{-\kappa,\kappa,\lambda}$. Thus $\alpha$ changes the scale along the same symmetric-exponent slice. At $\alpha=0$ the link reduces to $(1-x^{-\kappa})/\kappa$. A second representation, valid also at this endpoint, is}
\begin{equation}
\log^{\text{Te}}_{\alpha,\kappa}(x) = \frac{2}{1+\alpha} \left[\log_{-|\kappa|,|\kappa|} (x) + \frac{\alpha-1}{2} \log_{0,\kappa} (x) \right].
\end{equation}

Setting $t=x^{\kappa}$ in the defining equation $y=\log^{\mathrm{Te}}_{\alpha,\kappa}(x)$ gives the quadratic relation $\alpha t^2-\widetilde y\,t-1=0$, where $\widetilde y=(1+\alpha)\kappa y+\alpha-1$. Hence, for $\alpha>0$ and $\kappa\neq0$, the positive inverse branch is
\begin{equation}
\exp^{\text{Te}}_{\alpha,\kappa}(y) = \left[ \frac{\widetilde y + \sqrt{\widetilde y^{2} + 4 \alpha}}{2\alpha} \right]^{1/\kappa}.
\label{eq:Tempesta-inverse-corrected}
\end{equation}
For $\alpha=0$, the continuous limiting inverse is $\exp^{\mathrm{Te}}_{0,\kappa}(y)=(1-\kappa y)^{-1/\kappa}$ on its real domain; for other parameter regimes, the appropriate real branch and discriminant conditions must be enforced explicitly.

\paragraph{The $\gamma$-exponential.} For $(a,b)=(-\gamma,2\gamma)$, $\gamma>0$, the substitution $t=x^\gamma$ turns $\log_{a,b}(x)=y$ into the cubic $t^3-3\gamma y\,t-1=0$, whose positive real root is given by Cardano's formula,
\begin{equation}
\exp_\gamma(y)=\Bigl[\Bigl(\tfrac{1+\sqrt{1-4\gamma^3y^3}}2\Bigr)^{1/3}+\Bigl(\tfrac{1-\sqrt{1-4\gamma^3y^3}}2\Bigr)^{1/3}\Bigr]^{1/\gamma},
\label{eq:gamma-exp}
\end{equation}
with real cube roots when $4\gamma^3y^3\le1$; for $4\gamma^3y^3>1$ the trigonometric form of Cardano's formula gives the unique positive root.

\section{Radial normalization and split updates}
\label{app:MMD-convergence}
\noindent\emph{Scope and notation.} This appendix is written for the full three-parameter scale-free link $\log_{a,b,\lambda}$ of Eq.~\eqref{logab}, with inverse $\exp_{a,b,\lambda}$, and the scale $\lambda>0$ is fixed throughout. As in the main text, $\nu$ denotes a normalization (KKT) multiplier; the letter $\lambda$ is never used for a multiplier here. All objects carry the full subscript: $F_{a,b,\lambda}=\sum_i\phi_{a,b,\lambda}(w_i)$ is the separable potential of Eq.~\eqref{eq:Euler-potential-vector}, $h_{a,b,\lambda}=\log_{a,b,\lambda}'$, $m_{a,b,\lambda}=1/h_{a,b,\lambda}$, $\omega_{a,b,\lambda}(x)=xh_{a,b,\lambda}(x)$, and $\bH_{a,b,\lambda}(\bw)=\nabla^2F_{a,b,\lambda}(\bw)=\diag\{h_{a,b,\lambda}(w_i)\}$. The arguments below use only monotonicity, the normalization $\log_{a,b,\lambda}(1)=0$, $h_{a,b,\lambda}(1)=1$, the full-range inverse for $a<0<b$, and curvature bounds, so every statement holds for each fixed $\lambda>0$. Setting $\lambda=1$ recovers the unscaled link $\log_{a,b}=\log_{a,b,1}$. Where an explicit constant depends on $\lambda$, it is given in terms of $\alpha_\lambda,\beta_\lambda$ of Eq.~\eqref{eq:scaled-power-mixture}.

Two scale identities are used repeatedly. Differentiating Eq.~\eqref{eq:scaled-Euler-identity}, with $\omega_0(x)=x\log_{a,b}'(x)$, gives
\begin{equation}
 h_{a,b,\lambda}(x)=\frac{\lambda\,\log_{a,b}'(\lambda x)}{\omega_0(\lambda)},\qquad
 \omega_{a,b,\lambda}(x)=\frac{\omega_0(\lambda x)}{\omega_0(\lambda)}
 =\alpha_\lambda x^a+\beta_\lambda x^b ,
 \label{eq:appG-scale-identities}
\end{equation}
and Eq.~\eqref{defexpEuler} expresses $\exp_{a,b,\lambda}$ through the unscaled inverse.

The main text always restores unit mass with a scalar multiplier inside the inverse, which gives the exact Euler step and the variational Euler softmax. A tempting shortcut, familiar from exponentiated gradient, is to compute an unconstrained proposal and divide it by its sum. For the ordinary logarithm the two procedures agree; for Euler potentials, at every scale $\lambda$, they do not. This appendix collects what is known about the radial alternative: the radial softmax (Section~\ref{app:radial-softmax}), the radially normalized mirror-descent update and its variational interpretation (Section~\ref{app:radial-md}), the difference between cumulative and recursive radial updates (Section~\ref{subsec:FTRL-radial}), and the convergence of a clipped variable-metric split update (Section~\ref{app:mmd-analysis}).

\subsection{The radial softmax}
\label{app:radial-softmax}
The radial map is $q_i^{\rm rad}=\exp_{a,b,\lambda}(s_i)/Z_{a,b,\lambda}$ with $Z_{a,b,\lambda}=\sum_j\exp_{a,b,\lambda}(s_j)$. In the full-range regime $a<0<b$, using $\exp_{a,b,\lambda}'=m_{a,b,\lambda}\circ\exp_{a,b,\lambda}$ (Proposition~\ref{prop:euler-tails}),
\begin{equation}
\frac{\partial q_i^{\rm rad}}{\partial s_j}
=(\delta_{ij}-q_i^{\rm rad})\frac{\exp_{a,b,\lambda}'(s_j)}{Z_{a,b,\lambda}}
=(\delta_{ij}-q_i^{\rm rad})\frac{m_{a,b,\lambda}\bigl(\exp_{a,b,\lambda}(s_j)\bigr)}{Z_{a,b,\lambda}}.
\label{eq:radial-softmax-Jacobian}
\end{equation}
This Jacobian is generally nonsymmetric, and the map is generally not invariant to a common logit shift. It should therefore be distinguished from the variational Euler softmax and its convex potential. At the ordinary exponential these two constructions agree.

\paragraph{What the scale does to the radial map.}
By Eq.~\eqref{defexpEuler}, $\exp_{a,b,\lambda}(s)=\lambda^{-1}\exp_{a,b}(\omega_0(\lambda)s+\log_{a,b}(\lambda))$. The common factor $\lambda^{-1}$ cancels in the ratio, so
\begin{equation}
 \bq^{\rm rad}_{a,b,\lambda}(\bs)
 =\bq^{\rm rad}_{a,b,1}\bigl(\omega_0(\lambda)\bs+\log_{a,b}(\lambda)\1\bigr).
 \label{eq:radial-softmax-scale}
\end{equation}
Thus, for the radial map, the scale acts as an inverse temperature $\omega_0(\lambda)$ together with a common logit offset $\log_{a,b}(\lambda)$. Because the unscaled radial map is not shift invariant, this offset changes the output and $\lambda$ cannot be absorbed into a temperature alone. In the logarithmic limit the offset is irrelevant and $\omega_0\equiv1$.

\subsection{Radially normalized Euler mirror descent}
\label{app:radial-md}
Throughout this subsection $L$ is a differentiable objective on the simplex and
$\widehat L(\bu)=L(\bu/(\1^T\bu))$ is its zero-homogeneous lift to positive vectors; its gradient is given in Eq.~\eqref{Grad-Lparc} below.

\medskip\noindent\textbf{Post-normalized Euler--Bregman/KL split update.}\ The radial update is
\begin{align}
\tilde{\bw}_{t+1}&=\exp_{a,b,\lambda}\!\left[\log_{a,b,\lambda}(\bw_t)-\eta_t\nabla\widehat L(\bw_t)\right],\label{eq:Euler-postnorm-unconstrained}\\
\bw_{t+1}^{\mathrm{rad}}&=\frac{\tilde{\bw}_{t+1}}{\1^T\tilde{\bw}_{t+1}},\label{eq:Euler-radial-split}\\
\text{or}\qquad \bw_{t+1}^{\mathrm{Euc}}&=\Pi_{\Delta_N}^{\mathrm{Euc}}(\tilde{\bw}_{t+1}),\label{eq:Euler-euclidean-projected}
\end{align}
with $\log_{a,b,\lambda}$ and $\exp_{a,b,\lambda}$ applied componentwise. For a general deformed exponential, $\exp_{a,b,\lambda}(s-\nu)$ is not a constant multiple of $\exp_{a,b,\lambda}(s)$. Therefore Eq.~\eqref{eq:Euler-radial-split} is \emph{not} the exact minimizer of the one-stage Euler--Bregman simplex problem \eqref{eq:exact-Euler-simplex-MD}. Nevertheless, a stronger statement is possible: Eqs.~\eqref{eq:Euler-postnorm-unconstrained}--\eqref{eq:Euler-radial-split} form an exact two-stage split/projective algorithm for a different, explicitly defined variational problem. The Euclidean projection in Eq.~\eqref{eq:Euler-euclidean-projected} remains a separate projected variant.

\begin{theorem}[Exact split and lifted variational interpretation of Eqs.~\eqref{eq:Euler-postnorm-unconstrained}--\eqref{eq:Euler-radial-split}]
\label{thm:Euler-postnorm-split}
Assume the full-range regime $a<0<b$, a fixed scale $\lambda>0$, and $\bw_t\in\Delta_N^{\circ}$. Let $\bg_t=\nabla\widehat L(\bw_t)$ and let $F_{a,b,\lambda}$ be the separable Euler potential in Eq.~\eqref{eq:Euler-potential-vector}, defined by integration when an exponent equals $-1$. Then:
\begin{enumerate}
\item Eq.~\eqref{eq:Euler-postnorm-unconstrained} is the unique unconstrained Euler--Bregman mirror step
\[
\tilde{\bw}_{t+1}=\operatorname*{arg\,min}_{\bu\in\mathbb R_{++}^{N}}
\left\{\eta_t\langle\bg_t,\bu\rangle+D_{F_{a,b,\lambda}}(\bu\Vert\bw_t)\right\}.
\]
\item For any positive proposal $\bu$, radial normalization $\mathcal N^{\rm rad}(\bu):=\bu/(\1^T\bu)$ (not to be confused with the exact simplex decision $\mathcal N_{a,b,\lambda}$ of Appendix~\ref{sec:notation-guide}) is the unique Kullback--Leibler information projection onto the simplex,
\[
\mathcal N^{\rm rad}(\bu)=\operatorname*{arg\,min}_{\bw\in\Delta_N}
D_{\mathrm{KL}}(\bw\Vert\bu),\qquad
D_{\mathrm{KL}}(\bw\Vert\bu)=\sum_i\left(w_i\ln\frac{w_i}{u_i}-w_i+u_i\right).
\]
Consequently Eq.~\eqref{eq:Euler-radial-split} is an exact KL projection of the Euler mirror proposal.
\item Equivalently, if $s_{t+1}=\1^T\tilde{\bw}_{t+1}$, the pair $(s_{t+1},\bw_{t+1}^{\mathrm{rad}})$ is the unique minimizer of the exact simplex-plus-scale problem
\[
\operatorname*{minimize}_{s>0,\;\bw\in\Delta_N^{\circ}}
\left\{\eta_t\langle\bg_t,s\bw\rangle+D_{F_{a,b,\lambda}}(s\bw\Vert\bw_t)\right\}.
\]
Thus Eqs.~\eqref{eq:Euler-postnorm-unconstrained}--\eqref{eq:Euler-radial-split} are exact for this lifted/projective problem, although they are not the exact one-stage simplex Euler mirror update \eqref{eq:exact-Euler-simplex-MD}.
\end{enumerate}
\end{theorem}
\begin{proof}
For the first stage, $\nabla F_{a,b,\lambda}=\log_{a,b,\lambda}$ componentwise, so strict convexity and stationarity give $\log_{a,b,\lambda}(\tilde\bw)=\log_{a,b,\lambda}(\bw_t)-\eta_t\bg_t$. The full-range inverse provides a positive solution for every finite dual input, so this stationary point is the unique global minimizer. For KL projection, the simplex KKT equations give $\ln(w_i/u_i)+\nu=0$ for a scalar multiplier $\nu$, hence $\bw=e^{-\nu}\bu=\bu/(\1^T\bu)$. Finally, the bijection $\bu=s\bw$, $s=\1^T\bu>0$, identifies the lifted minimizer with the first-stage minimizer. The lifted objective need not be jointly convex in $(s,\bw)$; its equivalence follows from the bijection. The two-stage construction alone gives no descent theorem for $L$; the relative-smoothness assumption below supplies it~\cite{Bauschke2017,Lu2018}.
\end{proof}

The lift $\widehat L(\bw)=L(\bw/\|\bw\|_1)$ is the auxiliary scale-invariant objective of \cite{Cichocki2024}. With $s=\1^T\bw$ and $\bx=\bw/s$, the exact chain rule is
\begin{equation}
\nabla_{\bw}\widehat L(\bw)=\frac{1}{s}\left[\nabla L(\bx)-(\bx^T\nabla L(\bx))\1\right].
\label{Grad-Lparc}
\end{equation}
At a normalized point ($s=1$, $\bx=\bw$), this becomes $\nabla L(\bw)-(\bw^T\nabla L(\bw))\1$. By contrast,
\begin{equation}
\Pi_{\1^{\perp}}^{\mathrm{Euc}}\nabla L(\bw)
=\nabla L(\bw)-\frac{\1^T\nabla L(\bw)}{N}\1
\label{eq:Euclidean-tangent-gradient}
\end{equation}
is the Euclidean projection of the gradient onto the hyperplane orthogonal to $\1$. The two coincide only when $\bw$ is the uniform vector or $\nabla L(\bw)$ is orthogonal to $\bw-\1/N$.

\begin{lemma}[Exact simplex flow versus radial split flow]\label{lem:audit-two-simplex-flows}
At an interior simplex point let $\bP=\bH_{a,b,\lambda}(\bw)^{-1}=\diag\{m_{a,b,\lambda}(w_i)\}$, $\bg=\nabla L(\bw)$, $\bB=\bI-\bw\1^T$, $\boldsymbol p=\bP\1$, and $s_P=\1^T\bP\1$. The first-order direction of exact constrained MD is
\begin{equation}\label{eq:audit-exact-simplex-flow}
\dot{\bw}=-\left(\bP-\frac{\boldsymbol p\boldsymbol p^T}{s_P}\right)\bg.
\end{equation}
The radial split using $\nabla\widehat L=\bB^T\bg$ instead has direction $-\bB\bP\bB^T\bg$, and
\begin{equation}\label{eq:audit-flow-difference}
\bB\bP\bB^T-\left(\bP-\frac{\boldsymbol p\boldsymbol p^T}{s_P}\right)
=s_P\left(\bw-\frac{\boldsymbol p}{s_P}\right)
\left(\bw-\frac{\boldsymbol p}{s_P}\right)^T.
\end{equation}
Thus their tangent preconditioners coincide at this state precisely when $\bP\1=s_P\bw$, that is, when $m_{a,b,\lambda}(w_i)\propto w_i$, equivalently when $\omega_{a,b,\lambda}(w_i)$ is the same for all $i$. This holds for ordinary entropy, and at the uniform point for every Euler pair, but not at every state for a generic Euler pair and scale.
\end{lemma}
\begin{proof}
Differentiate the exact KKT equation and enforce $\1^T\dot{\bw}=0$; the scalar multiplier removes $(\1^T\bP\bg)/(\1^T\bP\1)$ from the gradient. For the radial map, $D\mathcal N^{\rm rad}(\bw)=\bB$ and the lift gradient is $\bB^T\bg$. Expanding $\bB\bP\bB^T=\bP-\bw\boldsymbol p^T-\boldsymbol p\bw^T+s_P\bw\bw^T$ proves \eqref{eq:audit-flow-difference}; $s_P>0$ gives the equality criterion, and $m_{a,b,\lambda}(w_i)=c\,w_i$ is the same as $\omega_{a,b,\lambda}(w_i)=1/c$.
\end{proof}
The positive-semidefinite difference is not a convergence-speed ordering. It shows that radial normalization can change even the first-order constrained dynamics, although radial MD and radial MMD share a flow with each other (Theorem~\ref{thm:app-MD-MMD-equivalence}). An unconstrained Hessian flow used on the simplex likewise needs the multiplier in \eqref{eq:audit-exact-simplex-flow}.

\begin{assumption}[Relative smoothness along the split-MD steps]
\label{ass:Euler-postnorm-relative-smoothness}
On a set containing every pair $(\bw_t,\tilde{\bw}_{t+1})$, suppose that for some $L_F>0$,
\[
\widehat L(\bu)\le \widehat L(\bw)+\langle\nabla\widehat L(\bw),\bu-\bw\rangle+L_FD_{F_{a,b,\lambda}}(\bu\Vert\bw).
\]
This is the standard relative-smoothness/descent condition used in Bregman first-order analysis~\cite{Bauschke2017,Lu2018}. The constant $L_F$ depends on $(a,b,\lambda)$ through $F_{a,b,\lambda}$. The descent proof below only requires this inequality; convexity of the homogeneous lift $\widehat L$ is not assumed.
\end{assumption}

\begin{theorem}[Convergence and perturbation stability of the radially normalized MD update]
\label{thm:Euler-postnorm-convergence}
Let $L$ be continuously differentiable on a neighborhood of the relevant simplex states and let $\widehat L(\bu)=L(\bu/(\1^T\bu))$ on positive vectors. Assume the hypotheses of Theorem~\ref{thm:Euler-postnorm-split}, Assumption~\ref{ass:Euler-postnorm-relative-smoothness}, and $0<\eta_t<1/L_F$. Then the sequence generated by Eqs.~\eqref{eq:Euler-postnorm-unconstrained}--\eqref{eq:Euler-radial-split} satisfies
\[
\widehat L(\bw_{t+1})\le \widehat L(\bw_t)
-\left(\frac{1}{\eta_t}-L_F\right)D_{F_{a,b,\lambda}}(\tilde{\bw}_{t+1}\Vert\bw_t).
\]
Hence the lifted objective is monotonically nonincreasing.

Assume in addition a constant step $\eta\in(0,1/L_F)$, that $\widehat L$ is bounded below by $\widehat L_{\inf}$, and that all line segments joining $\bw_t$ to $\tilde{\bw}_{t+1}$ lie in a compact positive set on which
\[
\mu_F\bI\preceq\bH_{a,b,\lambda}(\bw)\preceq M_F\bI,
\qquad 0<\mu_F\le M_F<\infty.
\]
Then
\[
\min_{1\le t\le T}\|\nabla\widehat L(\bw_t)\|_2^2
\le
\frac{2M_F^2\,[\widehat L(\bw_1)-\widehat L_{\inf}]}
{\mu_F\eta(1-\eta L_F)T}.
\]
Summability of the squared gradients gives $\nabla\widehat L(\bw_t)\to\0$. The compact positive-set hypothesis keeps every cluster point in the simplex interior, where it satisfies the simplex KKT condition for $L$. If $L$ is convex on the simplex, every such cluster point is a global minimizer and $L(\bw_t)\to L^*$; if that minimizer is unique, the normalized sequence converges to it.

Finally, let the gradient used in one step be perturbed from $\bg$ to $\bg+\bolde$. If $h_{a,b,\lambda}=\log_{a,b,\lambda}'\ge\mu_F>0$ on every coordinate interval joining the two proposals (equivalently, on the inverse image of the intervening dual interval) and the two positive proposals have $\ell_1$ masses at least $s_{\min}>0$, then
\[
\left\|\mathcal N^{\rm rad}\!\left(\exp_{a,b,\lambda}[\log_{a,b,\lambda}(\bw)-\eta(\bg+\bolde)]\right)
-\mathcal N^{\rm rad}\!\left(\exp_{a,b,\lambda}[\log_{a,b,\lambda}(\bw)-\eta\bg]\right)\right\|_1
\le \frac{2\eta}{s_{\min}\mu_F}\|\bolde\|_1.
\]
Thus the update is locally Lipschitz-stable to gradient perturbations whenever the Euler metric is uniformly positive and the proposal mass is bounded away from zero.
\end{theorem}
\begin{proof}
Optimality of the first split stage, compared with the feasible point $\bw_t$, gives
$\eta_t\langle\bg_t,\tilde{\bw}_{t+1}-\bw_t\rangle+D_{F_{a,b,\lambda}}(\tilde{\bw}_{t+1}\Vert\bw_t)\le0$.
Combining this inequality with Assumption~\ref{ass:Euler-postnorm-relative-smoothness} and using scale invariance,
$\widehat L(\bw_{t+1})=\widehat L(\tilde{\bw}_{t+1})$, proves descent. Under the uniform metric bounds,
$D_{F_{a,b,\lambda}}(\tilde{\bw}_{t+1}\Vert\bw_t)\ge(\mu_F/2)\|\tilde{\bw}_{t+1}-\bw_t\|_2^2$. The mirror equation $\log_{a,b,\lambda}(\tilde w_{i,t+1})-\log_{a,b,\lambda}(w_{i,t})=-\eta g_{i,t}$ and the mean-value theorem imply
$\|\bg_t\|_2\le(M_F/\eta)\|\tilde{\bw}_{t+1}-\bw_t\|_2$. Hence each step decreases $\widehat L$ by at least $(1-\eta L_F)\mu_F\eta\|\bg_t\|_2^2/(2M_F^2)$; summing yields the stated $\mathcal O(1/T)$ stationarity bound and $\|\bg_t\|_2\to0$. At a normalized interior point, $\nabla\widehat L(\bw)=\0$ is equivalent to $\nabla L(\bw)=\nu\1$ with $\nu=\bw^T\nabla L(\bw)$, the simplex KKT equation. For stability, $\exp_{a,b,\lambda}'=1/(h_{a,b,\lambda}\circ\exp_{a,b,\lambda})\le1/\mu_F$ on the relevant interval, giving proposal sensitivity at most $(\eta/\mu_F)\|\bolde\|_1$; radial normalization obeys $\|\mathcal N^{\rm rad}(\bu)-\mathcal N^{\rm rad}(\bv)\|_1\le2\|\bu-\bv\|_1/s_{\min}$ when both masses are at least $s_{\min}$.
\end{proof}

The compact positive-set and relative-smoothness assumptions are substantive conditions on the trajectory; the update does not establish them automatically. In particular, this radial-MD stationarity theorem does not analyze convergence to a sparse boundary optimizer. Its perturbation estimate is a one-step sensitivity bound, not a contraction or a uniform-in-time noisy-trajectory guarantee.

The exponents choose the tail geometry, $\lambda$ places the transition between the two regimes, and $\eta$ sets the update scale. Select them against a held-out task metric with matched search budgets, using the interpretable controls in Section~\ref{subsec:two-parameter-gain}; Section~\ref{subsec:practical-online} gives evaluation criteria.

\paragraph{Special cases.}
In the logarithmic limit $(a,b)\to(0,0)$, for any fixed $\lambda$, the update reduces to the standard EG algorithm \eqref{EG}. For the Tsallis specialization $(a,b)=(\min(0,1-q),\max(0,1-q))$ the scale cancels, $\log_{a,b,\lambda}=\log_q^T$ for every $\lambda>0$, and a \emph{scalar} dual learning rate $\eta$ already yields the factor $w_{i,t}^{q-1}$ algebraically:
$\exp_q^T(\log_q^T w-\eta g)=w\,\exp_q^T(-\eta w^{q-1}g)$, since $1+(1-q)(\log_q^Tw-\eta g)=w^{1-q}[1-(1-q)\eta w^{q-1}g]$.
This identity holds on the inverse domain: for $q>1$ the power base must remain strictly positive, whereas for $q<1$ a positive-part extension permits zero proposals. No component-dependent learning-rate vector is needed. The resulting update, also obtained from alpha--beta-divergence arguments in \cite{Cichocki2024,Cia3}, is
\begin{align}
\tilde{\bw}_{t+1} &= \bw_{t} \odot \exp^T_{q}\left(-\eta \bw_t^{q-1} \odot \nabla_{\bw} \widehat{L}(\bw_{t})\right), \label{EGAB1}\\
\bw_{t+1} &= \frac{\tilde{\bw}_{t+1}}{\lVert \tilde{\bw}_{t+1} \rVert_1},\qquad \lVert\tilde{\bw}_{t+1}\rVert_1>0.
\label{EGAB2}
\end{align}
In the notation of the alpha--beta-divergence formulation~\cite{Cichocki2024}, its proposal has the form
\[
\tilde w_i=w_i\exp^T_{1-\beta}\!\left(-\eta w_i^{1-\alpha-\beta}g_i\right).
\]
The scalar-step Tsallis proposal above is recovered by $\alpha=1$ and $\beta=1-q$. More generally, the exponent $1-\alpha-\beta$ changes the coordinate scaling of the gradient. This identifies the precise intersection of the two constructions while retaining their distinct parameterizations. In the interior $a<0<b$ no such multiplicative factorization exists, and the scale $\lambda$ genuinely changes the proposal.

\paragraph{Clipped variable-metric (MMD) variant.}
Using the MMD/Hessian-preconditioned formula \eqref{diagMD}, we obtain the following additive gradient update:
\begin{empheq}[box=\fbox]{align}
\tilde{\bw}_{t+1} &= \left[\bw_t-\eta_t\bH_{a,b,\lambda}(\bw_t)^{-1}\nabla\widehat L(\bw_t)\right]_+,\label{diagMDL-prop}\\
\bw_{t+1} &= \frac{\tilde{\bw}_{t+1}}{\lVert \tilde{\bw}_{t+1} \rVert_1}, \quad \bw_t\in\Delta_N^\circ,\quad\lVert\tilde{\bw}_{t+1}\rVert_1>0,
\quad \forall t,
\label{diagMDL}
\end{empheq}
where $\bH_{a,b,\lambda}(\bw_t)^{-1}=\diag\{m_{a,b,\lambda}(w_{i,t})\}$ with $m_{a,b,\lambda}=1/\log_{a,b,\lambda}'$ and, by Eq.~\eqref{eq:scaled-power-mixture},
\begin{equation}
 m_{a,b,\lambda}(w)=\frac{1}{\alpha_\lambda w^{a-1}+\beta_\lambda w^{b-1}}
 =\frac{w}{\omega_{a,b,\lambda}(w)}.
 \label{eq:appG-mobility}
\end{equation}
This matrix is positive definite at strictly positive weights in the stated Euler regimes; for $a<0$ the Hessian diverges at a zero coordinate, while its reciprocal has limit zero. Thus the positive-definite interior metric assumptions do not extend to that boundary. By Eq.~\eqref{eq:appG-mobility} and the elasticity identity $\mathrm d\ln m_{a,b,\lambda}/\mathrm d\ln w=\varepsilon_{a,b,\lambda}(w)$ of Section~\ref{sec:logab-properties}, for $a<0<b$,
\[
\begin{gathered}
 m_{a,b,\lambda}(w)\sim\frac{w^{1-a}}{\alpha_\lambda}\ \ (w\downarrow0),\qquad
 m_{a,b,\lambda}(w)\sim\frac{w^{1-b}}{\beta_\lambda}\ \ (w\to\infty),\\
 m_{a,b,\lambda}(1)=1,\qquad
 m_{a,b,\lambda}'(1)=\varepsilon_{a,b,\lambda}(1)=1-\alpha_\lambda a-\beta_\lambda b .
\end{gathered}
\]
The exponents of the two power laws do not depend on $\lambda$, but their prefactors and the slope at $w=1$ do; at $\lambda=1$, $m_{a,b}'(1)=1-a-b$.

\medskip\noindent\textbf{Variational status of the clipped/normalized MMD update.}\
At an interior state, \rev{Eq.~\eqref{diagMDL-prop}} is the exact solution of a variable-metric projected-gradient subproblem on the nonnegative orthant, Eq.~\eqref{eq:app-MMD-QP}, and \rev{Eq.~\eqref{diagMDL}} is the KL projection of Theorem~\ref{thm:Euler-postnorm-split}. The two lines therefore define an exact variable-metric/KL split algorithm, which differs from the one-stage simplex update~\eqref{eq:exact-Euler-simplex-MD}. If clipping creates a zero, later unregularized Euler-Hessian steps are not finite-metric subproblems. Convergence, stability, the role of clipping, and the first-order relation to Eqs.~\eqref{eq:Euler-postnorm-unconstrained}--\eqref{eq:Euler-radial-split} are analyzed in Section~\ref{app:mmd-analysis}.

\subsection{Cumulative and recursive radial variants}
\label{subsec:FTRL-radial}
Exact Euler updates normalize through a scalar multiplier inside the inverse. A tempting shortcut is to apply the unconstrained update and then divide by the sum, as in ordinary EG. The next result shows that this shortcut changes the algorithm in a way that depends on implementation details: accumulating all gradients and normalizing once gives different weights from normalizing after every step, except for the ordinary logarithm. The free scale does not remove this discrepancy.

\begin{theorem}[Radial lazy versus radial greedy updates]\label{thm:lazy-greedy}
Let $N\ge2$, $\eta>0$, $\lambda>0$, and let $\log_{a,b,\lambda}$ be a full-range scaled Euler link ($a<0<b$) or the ordinary logarithm, with inverse $\exp_{a,b,\lambda}$. Set $\mathcal N^{\rm rad}(\bu)=\bu/(\1^T\bu)$ and initialize both sequences at the same $\bw_1\in\Delta_N^\circ$. Consider
\begin{empheq}[box=\fbox]{align}
\bw_{t+1}^{\rm greedy}&=\mathcal N^{\rm rad}\!\left(\exp_{a,b,\lambda}\Bigl(\log_{a,b,\lambda}(\bw_t^{\rm greedy})-\eta\bg_t\Bigr)\right),
\label{eq:radial-greedy}\\
\bw_{t+1}^{\rm lazy}&=\mathcal N^{\rm rad}\!\left(\exp_{a,b,\lambda}\Bigl(\log_{a,b,\lambda}(\bw_1)-\eta\sum_{s=1}^t\bg_s\Bigr)\right).
\label{eq:radial-lazy}
\end{empheq}
Without $\mathcal N^{\rm rad}$, recursive and cumulative updates coincide by telescoping. With radial normalization, they coincide for every finite gradient stream if and only if $\log_{a,b,\lambda}=\ln$. In particular, for $a<0<b$ they differ for every scale $\lambda>0$. These radial statements do not contradict Lemma~\ref{lem:audit-exact-ftrl}: exact FTRL uses a dual multiplier instead of radial normalization.
\end{theorem}
\begin{proof}
Sufficiency for $\ln$ follows from the ordinary exponential's translation-to-scaling identity $\exp(s-c)=e^{-c}\exp(s)$. For necessity, write $h_{a,b,\lambda}=\log_{a,b,\lambda}'>0$. A first gradient can produce any positive unnormalized proposal $\bu=s\bw$ with $s>0$ and $\bw\in\Delta_N^\circ$, since the link is full range. At $t=2$ the lazy iterate is $\mathcal N^{\rm rad}(\exp_{a,b,\lambda}(\log_{a,b,\lambda}(\bu)-\eta\bg_2))$ and the greedy iterate is $\mathcal N^{\rm rad}(\exp_{a,b,\lambda}(\log_{a,b,\lambda}(\bw)-\eta\bg_2))$; both equal $\bw$ at $\bg_2=\0$. Equality for every second gradient therefore requires equal derivatives with respect to $\bg_2$ at zero. Since $\exp_{a,b,\lambda}'(\log_{a,b,\lambda}(x))=1/h_{a,b,\lambda}(x)$, $D\mathcal N^{\rm rad}(\bu)=(\bI-\bw\1^T)/s$ and $D\mathcal N^{\rm rad}(\bw)=\bI-\bw\1^T$, equality gives
\begin{equation}\label{eq:audit-radial-necessity}
(\bI-\bw\1^T)\diag\!\left(\frac{1}{s\,h_{a,b,\lambda}(sw_i)}-\frac{1}{h_{a,b,\lambda}(w_i)}\right)=\boldsymbol0.
\end{equation}
Every column $\boldsymbol e_i-\bw$ is nonzero when $N\ge2$ and $\bw$ is interior. Thus $s\,h_{a,b,\lambda}(sw_i)=h_{a,b,\lambda}(w_i)$. Fix any component $w_i=c\in(0,1)$ and vary $s>0$: $x\,h_{a,b,\lambda}(x)=c\,h_{a,b,\lambda}(c)$ for every $x>0$, that is, $\omega_{a,b,\lambda}$ is constant. The normalization $h_{a,b,\lambda}(1)=1$, valid for every $\lambda$ by Eq.~\eqref{eq:scaled-Euler-identity}, yields $h_{a,b,\lambda}(x)=1/x$, and $\log_{a,b,\lambda}(1)=0$ gives $\log_{a,b,\lambda}=\ln$. For $a<0<b$ this is impossible at any scale: $\omega_{a,b,\lambda}(x)=\alpha_\lambda x^a+\beta_\lambda x^b$ with $\alpha_\lambda,\beta_\lambda>0$ is unbounded at both ends. For $N=1$ all normalized maps are trivial, so the dimension assumption is essential.
\end{proof}

\subsection{The clipped variable-metric split update}
\label{app:mmd-analysis}
This subsection collects the technical analysis of \rev{Eqs.~\eqref{diagMDL-prop}--\eqref{diagMDL}}. Let
$\bg_t=\nabla\widehat L(\bw_t)$, $\bH_t=\bH_{a,b,\lambda}(\bw_t)$, and $\bP_t=\bH_t^{-1}$. Because $\bH_t$ is diagonal and positive definite, the clipped MMD proposal is exactly
\begin{equation}
\tilde{\bw}_{t+1}=\operatorname*{arg\,min}_{\bu\in\mathbb R_+^N}
\left\{\langle\bg_t,\bu-\bw_t\rangle+
\frac{1}{2\eta_t}\|\bu-\bw_t\|_{\bH_t}^{2}\right\},
\label{eq:app-MMD-QP}
\end{equation}
where $\|\bx\|_{\bH}^{2}=\bx^T\bH\bx$. Its closed form is \rev{Eq.~\eqref{diagMDL-prop}}. This is a diagonal variable-metric projected-gradient step~\cite{Bertsekas1982}. If $\tilde{\bw}_{t+1}\ne\0$, the normalized vector is the exact KL projection described in Theorem~\ref{thm:Euler-postnorm-split}. Hence the complete MMD rule is an exact orthant-variable-metric/KL split method.

\paragraph{Boundary implementation is part of the algorithm.}
For $a<0$, $\phi_{a,b,\lambda}''(x)=h_{a,b,\lambda}(x)\sim\alpha_\lambda x^{a-1}\to+\infty$ as $x\downarrow0$. Therefore the finite, uniformly bounded metric assumptions below do not apply to repeated unregularized Euler updates after clipping. One rigorous choice is to enforce strictly positive proposals and retain uniform interior bounds. Another is to specify the modified metric
\begin{equation}\label{eq:audit-regularized-metric}
\bH_\varrho(\bw)=\diag\{\log_{a,b,\lambda}'(w_i+\varrho)\},\qquad\varrho>0.
\end{equation}
On the closed simplex this metric has finite positive upper and lower bounds. The clipped variable-metric results below apply to this explicitly regularized algorithm, but its geometry differs from the original Euler metric. Simply setting the reciprocal curvature to zero at zero creates absorbing coordinates and does not justify the same stationarity theorem. For example, with $L(w_1,w_2)=-w_1$, the vertex $(0,1)$ has centered gradient $(-1,0)$; a zero inverse-curvature extension leaves it fixed despite its failure of simplex KKT optimality.

Throughout this subsection let $\widehat L(\bu)=L(\bu/(\1^T\bu))$ on nonnegative nonzero vectors, with a continuously differentiable extension on a neighborhood of each relevant segment. All stated $\bH_t$ are finite positive diagonal matrices. They may be the unregularized Euler metric at interior states, or a specified boundary-regularized metric. With the exact centered gradient, the proposal cannot be the zero vector: $\bw^T\nabla\widehat L(\bw)=0$ implies some positive coordinate has nonpositive gradient and is not decreased by the diagonal proposal. Perturbed gradients still require an explicit mass check.

The same rule also has an exact simplex-plus-scale formulation: provided the minimizing proposal is nonzero,
\begin{equation}
(s_{t+1},\bw_{t+1})=\operatorname*{arg\,min}_{s>0,\;\bw\in\Delta_N}
\left\{\langle\bg_t,s\bw-\bw_t\rangle+
\frac{1}{2\eta_t}\|s\bw-\bw_t\|_{\bH_t}^{2}\right\},
\label{eq:app-MMD-joint}
\end{equation}
with $s_{t+1}=\1^T\tilde{\bw}_{t+1}>0$. This is again a different variational problem from the one-stage Euler--Bregman simplex problem \eqref{eq:exact-Euler-simplex-MD}.

\begin{theorem}[Descent and stationarity of clipped/normalized MMD]
\label{thm:app-MMD-descent}
Assume $\widehat L$ has an $L_g$-Lipschitz Euclidean gradient with a chosen bound $L_g>0$ on every segment joining $\bw_t$ and $\tilde{\bw}_{t+1}$, $\bH_t\succeq m_H\bI$ with $m_H>0$, $\1^T\tilde{\bw}_{t+1}>0$, and
$0<\eta_t<2m_H/L_g$. Then
\begin{equation}
\widehat L(\bw_{t+1})\le\widehat L(\bw_t)-
\left(\frac1{\eta_t}-\frac{L_g}{2m_H}\right)
\|\tilde{\bw}_{t+1}-\bw_t\|_{\bH_t}^{2}.
\label{eq:app-MMD-descent}
\end{equation}
For a constant admissible step $\eta$ and a lower bound $\widehat L_{\inf}$, define
\[
c_H:=\frac1\eta-\frac{L_g}{2m_H}>0.
\]
Then
\begin{equation}
\min_{1\le t\le T}\|\tilde{\bw}_{t+1}-\bw_t\|_{\bH_t}^{2}
\le \frac{\widehat L(\bw_1)-\widehat L_{\inf}}{c_H T}
=\mathcal O(T^{-1}).
\label{eq:app-MMD-residual}
\end{equation}
If, in addition, the normalized iterates and proposals remain in a compact subset of the nonnegative orthant excluding the origin, $\nabla\widehat L$ is continuous there, and $m_H\bI\preceq\bH_t\preceq M_H\bI$, every cluster point satisfies
\begin{equation}
\0\in\nabla\widehat L(\bw^*)+N_{\mathbb R_+^N}(\bw^*),
\label{eq:app-MMD-KKT}
\end{equation}
which, after radial normalization, is the corresponding simplex KKT condition for $L$. Therefore, for convex $L$, every such normalized cluster point is a global minimizer.
\end{theorem}
\begin{proof}
The variational inequality of Eq.~\eqref{eq:app-MMD-QP}, tested at the feasible point $\bw_t$, gives
$\langle\bg_t,\tilde{\bw}_{t+1}-\bw_t\rangle\le-\eta_t^{-1}\|\tilde{\bw}_{t+1}-\bw_t\|_{\bH_t}^{2}$. The Euclidean descent lemma and
$\|\bx\|_2^2\le m_H^{-1}\|\bx\|_{\bH_t}^{2}$ imply Eq.~\eqref{eq:app-MMD-descent} at the proposal. Scale invariance gives
$\widehat L(\bw_{t+1})=\widehat L(\tilde{\bw}_{t+1})$. For a constant admissible step, summing Eq.~\eqref{eq:app-MMD-descent} gives Eq.~\eqref{eq:app-MMD-residual} and hence $\|\tilde{\bw}_{t+1}-\bw_t\|_{\bH_t}\to0$. The optimality condition of Eq.~\eqref{eq:app-MMD-QP} is
$\0\in\bg_t+\eta^{-1}\bH_t(\tilde{\bw}_{t+1}-\bw_t)+N_{\mathbb R_+^N}(\tilde{\bw}_{t+1})$.
Along a convergent subsequence, the residual term vanishes under the uniform upper metric bound $\bH_t\preceq M_H\bI$; continuity of the gradient and closedness of the orthant normal-cone graph give Eq.~\eqref{eq:app-MMD-KKT}. Because the normalized limit satisfies $\1^T\bw^*=1$ and $\nabla\widehat L(\bw^*)=\nabla L(\bw^*)-(\bw^{*T}\nabla L(\bw^*))\1$, the orthant KKT relation is exactly the simplex KKT condition for $L$: equality holds on positive coordinates and the usual complementary inequality holds on zero coordinates.
\end{proof}

\begin{remark}[A certificate for the inexpensive clipped update]\label{rem:interpret-app-MMD-descent}
The variable-metric proposal is a quadratic projected step, allowing clipping to replace inverse-link evaluation. The theorem specifies a step range that yields descent and a computable residual that tends to zero. Under the compactness and metric bounds, limit points satisfy the constrained stationarity conditions, and convexity makes them global minimizers. The $\mathcal O(T^{-1})$ quantity is a squared update residual, not an objective gap. A finite positive boundary metric is part of implementing this result when zero coordinates occur. The constants $m_H,M_H$ depend on $(a,b,\lambda)$ through $h_{a,b,\lambda}$ on the region visited.
\end{remark}

\begin{theorem}[Perturbation stability of clipped/normalized MMD]
\label{thm:app-MMD-stability}
Fix $\bw$ and a positive diagonal matrix $\bH\succeq m_H\bI$. Let two MMD proposals be generated using gradients $\bg$ and $\bg+\bolde$, and assume both proposal masses are at least $s_{\min}>0$. Then
\begin{equation}
\|\bw^+(\bg+\bolde)-\bw^+(\bg)\|_1
\le\frac{2\eta}{s_{\min} m_H}\|\bolde\|_1.
\label{eq:app-MMD-stability}
\end{equation}
\end{theorem}
\begin{proof}
Componentwise clipping is nonexpansive. Since $\|\bH^{-1}\|_{1\to1}\le1/m_H$ for a positive diagonal $\bH\succeq m_H\bI$, the two proposals differ in $\ell_1$ norm by at most $(\eta/m_H)\|\bolde\|_1$. The same radial-normalization inequality used in Theorem~\ref{thm:Euler-postnorm-convergence} contributes the factor $2/s_{\min}$.
\end{proof}

\begin{remark}[How numerical or gradient errors enter one step]\label{rem:interpret-app-MMD-stability}
The constant separates three sources of sensitivity: step size, inverse metric curvature, and division by the proposal mass. It provides a practical reason to monitor the mass before normalization and to keep the metric uniformly positive. The estimate applies to two proposals from the same state and metric, including changes of active clipping coordinates. Uniform control of an entire noisy trajectory requires additional assumptions on how errors propagate between states.
\end{remark}

\begin{theorem}[First-order and local-linear relation between normalized MD and MMD]
\label{thm:app-MD-MMD-equivalence}
Fix $\lambda>0$. Assume clipping is inactive, the inverse mirror map $\exp_{a,b,\lambda}$ is twice continuously differentiable on the relevant compact interval, $\widehat L$ is twice continuously differentiable there, the gradients are bounded, and the proposal masses stay bounded away from zero. Then, for the same $\bw_t$ and $\bg_t$,
\begin{equation}
\bw_{t+1}^{\mathrm{MD}}-\bw_{t+1}^{\mathrm{MMD}}=\mathcal O(\eta_t^2),
\qquad \eta_t\downarrow0.
\label{eq:app-MD-MMD-O2}
\end{equation}
Both methods therefore have the same first-order normalized Hessian flow
\begin{equation}
\dot{\bw}=-(\bI-\bw\1^T)\bH_{a,b,\lambda}(\bw)^{-1}\nabla\widehat L(\bw).
\label{eq:app-common-normalized-flow}
\end{equation}
Moreover, at any common interior stationary point $\bw^*$ with $\nabla\widehat L(\bw^*)=\0$, the two discrete maps have exactly the same Jacobian. With $\bG_*:=\nabla^2\widehat L(\bw^*)$ and $\bP_*:=\bH_{a,b,\lambda}(\bw^*)^{-1}$, its action on simplex perturbations is
\begin{equation}
\bJ_*=(\bI-\bw^*\1^T)(\bI-\eta\bP_*\bG_*).
\label{eq:app-common-Jacobian}
\end{equation}
Consequently both fixed-step maps are locally linearly stable when the spectral radius of the tangent-space restriction of $\bJ_*$ is below one, and they have the same linearized spectral factor. Their nonlinear trajectories and observed asymptotic factors need not coincide. No universal necessity statement about equality of radially normalized maps follows from this first-order calculation. The Euler comparison here concerns the unregularized interior maps; a boundary-regularized MMD metric is a different construction.
\end{theorem}
\begin{proof}
Since $\exp_{a,b,\lambda}'(\log_{a,b,\lambda}(x))=m_{a,b,\lambda}(x)$, Taylor expansion of $\exp_{a,b,\lambda}(\log_{a,b,\lambda}(\bw_t)-\eta_t\bg_t)$ gives
$\bw_t-\eta_t\bH_{a,b,\lambda}(\bw_t)^{-1}\bg_t+\mathcal O(\eta_t^2)$, which is the unclipped MMD proposal to first order. Radial normalization is smooth on positive vectors with mass bounded away from zero, so the $\mathcal O(\eta_t^2)$ difference is preserved after normalization and yields Eq.~\eqref{eq:app-common-normalized-flow}. At a stationary point the terms involving derivatives of $\bH_{a,b,\lambda}^{-1}$ multiply $\bg(\bw^*)=\0$, so both unnormalized maps have differential $\bI-\eta\bH_{a,b,\lambda}(\bw^*)^{-1}\nabla\bg(\bw^*)$; applying the same differential of radial normalization proves equality of the normalized Jacobians. The Hessian-flow interpretation is consistent with the mirrorless-mirror-descent and Hessian-Riemannian literature~\cite{MMD2021,Alvarez2004}.
\end{proof}

\begin{remark}[A principled approximation and a shared local stability test]\label{rem:interpret-app-MD-MMD-equivalence}
Away from clipping and boundary singularities, MMD approximates the nonlinear mirror proposal to first order, with a one-step discrepancy of order $\eta^2$. At an interior stationary point their identical Jacobians give the same tangent-space linear stability criterion. This supports MMD as a less expensive local approximation and explains why the two methods can behave similarly near a solution. The agreement is local; it neither equates finite-step trajectories nor extends automatically to boundary-regularized metrics.
\end{remark}

The discussion above treats clipping as a possibility to be analyzed. It is
useful to know when it cannot occur at all, since in that regime the
clipped/normalized MMD rule coincides with the smooth variable-metric step and
the active-set caveats are vacuous. The relevant constant is the probability-domain floor
$c_{a,b,\lambda}=\min_{0<x\le1}\omega_{a,b,\lambda}(x)$ of Eq.~\eqref{eq:entropy-comparison-constant}.

\begin{theorem}[Explicit step size for which clipping is inactive]
\label{thm:clipping-inactive}
Let $a<0<b$, $\lambda>0$, and $w_{i,t}\in(0,1]$ for all $i$. Put $p=|a|$ and $r=(p/b)^{2/(p+b)}$. If
\begin{equation}
\eta_t\,\|\nabla\widehat L(\bw_t)\|_\infty\;<\;c_{a,b,\lambda}
:=\min_{0<w\le1} w\,\log_{a,b,\lambda}'(w)
=\begin{cases}
1, & \lambda\le r,\\[4pt]
\dfrac{p+b}{p\lambda^{a}+b\lambda^{b}}
\left(\dfrac{b}{p}\right)^{\frac{p-b}{p+b}}, & \lambda>r,
\end{cases}
\label{eq:clipping-inactive}
\end{equation}
then every component of the MMD proposal in \rev{Eq.~\eqref{diagMDL-prop}} is strictly
positive and the truncation $[\,\cdot\,]_+$ is never active. In particular, for
$\lambda\le r$ the classical condition $\eta_t\|\nabla\widehat L\|_\infty<1$
suffices. At $\lambda=1$ the condition $\lambda\le r$ reads $b\le|a|$, and for $b>|a|$ the threshold reduces to $c_{a,b,1}=(b/|a|)^{(|a|-b)/(|a|+b)}$.
\end{theorem}

\begin{proof}
Positivity of $w_{i,t}-\eta_tm_{a,b,\lambda}(w_{i,t})\,g_{i,t}$ is automatic when $g_{i,t}\le0$ and, by Eq.~\eqref{eq:appG-mobility}, is equivalent to
$\eta_t g_{i,t}<\omega_{a,b,\lambda}(w_{i,t})$ otherwise, so it suffices that
$\eta_t\|\bg_t\|_\infty<\min_{(0,1]}\omega_{a,b,\lambda}$. Now $\omega_{a,b,\lambda}(w)=\alpha_\lambda w^{a}+\beta_\lambda w^{b}$ and
\[
\omega_{a,b,\lambda}'(w)=w^{-1}\bigl(b\beta_\lambda w^{b}-p\,\alpha_\lambda w^{a}\bigr).
\]
Using $\beta_\lambda/\alpha_\lambda=(b/p)\lambda^{b-a}$, this vanishes only at
$w_\ast=(p/b)^{2/(b-a)}\lambda^{-1}=r/\lambda$, is negative for $w<w_\ast$ and positive
for $w>w_\ast$, so $w_\ast$ is the unique minimizer on $(0,\infty)$; it is the crossing point $x_m$ of Proposition~\ref{prop:elasticity}. If $\lambda\le r$, then $w_\ast\ge1$, $\omega_{a,b,\lambda}$ is
nonincreasing on $(0,1]$, and the minimum is $\omega_{a,b,\lambda}(1)=\alpha_\lambda+\beta_\lambda=1$. If $\lambda>r$, then $w_\ast<1$ and, by Eq.~\eqref{eq:appG-scale-identities},
\[
\omega_{a,b,\lambda}(w_\ast)=\frac{\omega_0(r)}{\omega_0(\lambda)},\qquad
\omega_0(x)=\frac{p\,x^{a}+b\,x^{b}}{p+b}.
\]
At $r$ one has $p\,r^{a}=(b^2/p)\,r^{b}$ and $b-a=p+b$, so
\[
p\,r^{a}+b\,r^{b}=\frac{b(p+b)}{p}\,r^{b}
=(p+b)\frac{b}{p}\left(\frac{p}{b}\right)^{\frac{2b}{p+b}}
=(p+b)\left(\frac{b}{p}\right)^{\frac{p-b}{p+b}},
\]
which gives Eq.~\eqref{eq:clipping-inactive}; it agrees with $K/D_\lambda$ in Eq.~\eqref{eq:entropy-comparison-constant}. At $\lambda=1$, $\omega_0(1)=1$ and $r\ge1$ exactly when $b\le p$.
\end{proof}

\begin{remark}[An explicit safeguard for positive proposals]\label{rem:interpret-clipping-inactive}
The constant converts positivity into a simple step-size check using the current maximum gradient magnitude. For example, with $0<\tau<1$ and $G_t=\|\nabla\widehat L(\bw_t)\|_\infty>0$, choosing $\eta_t\le\tau c_{a,b,\lambda}/G_t$ keeps clipping inactive. If $G_t=0$, the proposal is unchanged. This preserves the setting of the MD--MMD local comparison. Positivity and descent are separate requirements, so a descent-based step restriction should also be enforced when its hypotheses apply.
\end{remark}

\noindent
Three consequences are worth recording. First, $c_{a,b,\lambda}\le1$ for every scale, with equality exactly when $\lambda\le r=(|a|/b)^{2/(b-a)}$; for $\lambda>r$ it decreases strictly in $\lambda$ and behaves like $\lambda^{-b}$ as $\lambda\to\infty$. A small scale therefore admits the largest clipping-free step, although $\lambda$ also changes the metric and must be selected on the task as a whole. Second, in the unscaled case $\lambda=1$ the admissible step shrinks as $b$ grows relative to $|a|$, since $t\mapsto t^{(1-t)/(1+t)}$ decreases for $t=b/|a|>1$; for other scales this monotonicity in $b$ need not hold. Third, $c_{a,b,\lambda}\to0$ as $a\to0^-$ with fixed $b>0$ and $\lambda>0$: the bounded
branch $a=0$, recommended in Section~\ref{sec:euler-CE} for robust losses,
admits no positive, state-independent clipping-free threshold based only on a fixed gradient bound. A dual implementation still needs its finite-range check. Boundary iteration requires a specified metric extension such as \eqref{eq:audit-regularized-metric}, or a separately analyzed active-set method; the theorem does not define such a method implicitly.

\medskip\noindent\textbf{Role of clipping and normalization.}\ Clipping improves feasibility stability because the orthant projection is nonexpansive, but persistent clipping changes the active set and the dynamics are then governed by the orthant KKT system rather than the smooth interior Euler flow. No universal deterioration of the convergence rate should be asserted solely from the presence of clipping. Radial normalization is exact KL projection and leaves the scale-invariant objective $\widehat L$ unchanged, but it can amplify perturbations if the proposal mass approaches zero; the mass lower bound in Theorems~\ref{thm:Euler-postnorm-convergence} and \ref{thm:app-MMD-stability} makes this dependence explicit. Uniform interior bounds (or an explicit positive floor/active-set treatment) are therefore the natural conditions for stable Euler-metric dynamics.

\section*{Statement on AI usage}

I used \rev{OpenAI's} GPT-6.1 Sol and \rev{Anthropic's} Claude Opus 5.5 
to assist in a final stage with language editing, references, numerical checks, and, in part, the analysis of the proposed algorithms and theorems.  
However, all concepts presented in this manuscript are the author's own developments and ideas.
The author takes full responsibility for all content in the final version of the paper, including its results, proofs, and writing.

\end{document}